\documentclass{article} % For LaTeX2e
\usepackage{iclr2024_conference,times}
\usepackage{epsfig}
\usepackage{graphicx}
\usepackage{amsmath}
\usepackage{amssymb}
\usepackage{float}
\usepackage{booktabs}
\usepackage{caption}
\usepackage[font=small]{caption}
\usepackage[colorlinks,linkcolor=red]{hyperref}
\usepackage{amsmath,amsfonts,bm}

\def\eqref#1{equation~\ref{#1}}
\def\1{\bm{1}}

\DeclareMathAlphabet{\mathsfit}{\encodingdefault}{\sfdefault}{m}{sl}
\SetMathAlphabet{\mathsfit}{bold}{\encodingdefault}{\sfdefault}{bx}{n}

\newcommand{\bB}{\mathbf{B}}
\newcommand{\bc}{\mathbf{c}}\newcommand{\bC}{\mathbf{C}}
\newcommand{\bd}{\mathbf{d}}

\newcommand{\bF}{\mathbf{F}} %\bf already taken

\newcommand{\bI}{\mathbf{I}}

\newcommand{\bK}{\mathbf{K}}

\newcommand{\bR}{\mathbf{R}}

\newcommand{\bT}{\mathbf{T}}

\newcommand{\bx}{\mathbf{x}}\newcommand{\bX}{\mathbf{X}}

\newcommand{\btheta}{\boldsymbol{\theta}}

\newcommand{\bOmega}{\boldsymbol{\Omega}}

\newcommand{\cL}{\mathcal{L}}

\newcommand{\figrefer}[1]{Figure~\ref{#1}}
\newcommand{\eqnref}[1]{Eq.~\ref{#1}}
\newcommand{\tabnref}[1]{Table~\ref{#1}}

\usepackage{xspace}
\makeatletter
\DeclareRobustCommand\onedot{\futurelet\@let@token\@onedot}
\def\@onedot{\ifx\@let@token.\else.\null\fi\xspace}
\def\eg{{e.g}\onedot} 
\def\ie{{i.e}\onedot} 
 
\def\etc{{etc}\onedot}

\makeatother

\newcommand{\PAR}[1]{\vspace{0.1cm}\noindent{\bf #1} }

\newcommand{\norm}[1]{\left\lVert#1\right\rVert}

\usepackage{hyperref}
\usepackage{url}
\usepackage{wrapfig}

\title{USB-NeRF: Unrolling Shutter Bundle Adjusted Neural Radiance Fields}

\author{Moyang Li$^{1,2}$\footnotemark[1]\qquad Peng Wang$^{1,3}$\footnotemark[1]\qquad Lingzhe Zhao$^{1}$\qquad Bangyan Liao$^{1,3}$\qquad Peidong Liu$^{1}$\footnotemark[2]\\
$^{1}$Westlake University\qquad $^{2}$ETH Zürich \qquad $^{3}$Zhejiang University\\
\texttt{\tt\fontsize{8.25}{14}\selectfont \href{mailto:moyali@ethz.ch}{\textcolor{black}{moyali@ethz.ch}}, \{\href{mailto:wangpeng@westlake.edu.cn}{\textcolor{black}{wangpeng}}, \href{mailto:zhaolingzhe@westlake.edu.cn}{\textcolor{black}{zhaolingzhe}}, \href{mailto:liaobangyan@westlake.edu.cn}{\textcolor{black}{liaobangyan}}, \href{mailto:liupeidong@westlake.edu.cn}{\textcolor{black}{liupeidong}}\}@westlake.edu.cn}
}

\iclrfinalcopy % Uncomment for camera-ready version, but NOT for submission.
\begin{document}

\maketitle

\renewcommand{\thefootnote}{\fnsymbol{footnote}}
\footnotetext[1]{Equal contribution.}
\footnotetext[2]{Corresponding author.}

\vspace{-1.0em}
\begin{abstract}
\vspace{-1.0em}
% brief intro & motivation of our work
Neural Radiance Fields (NeRF) has received much attention recently due to its impressive capability to represent 3D scene and synthesize novel view images. Existing works usually assume that the input images are captured by a global shutter camera. Thus, rolling shutter (RS) images cannot be trivially applied to an off-the-shelf NeRF algorithm for novel view synthesis. Rolling shutter effect would also affect the accuracy of the camera pose estimation (\eg via COLMAP), which further prevents the success of NeRF algorithm with RS images.
% what we do & how we do
In this paper, we propose Unrolling Shutter Bundle Adjusted Neural Radiance Fields (USB-NeRF). USB-NeRF is able to correct rolling shutter distortions and recover accurate camera motion trajectory simultaneously under the framework of NeRF, by modeling the physical image formation process of a RS camera.
% what results do we deliver
Experimental results demonstrate that USB-NeRF achieves better performance compared to prior works, in terms of RS effect removal, novel view image synthesis as well as camera motion estimation. Furthermore, our algorithm can also be used to recover high-fidelity high frame-rate global shutter video from a sequence of RS images. Code and data are available at \url{https://github.com/WU-CVGL/USB-NeRF}.
\end{abstract}
%%%%% BODY TEXT
\section{Introduction}
\label{sec:intro}
%% Background (3D and NeRF)
% 3D->NeRF
Understanding and recovering 3D scenes from 2D images is a difficult but important problem in computer vision. Different from a 2D image which can be naturally formulated as an array of pixel values, there are many 3D representations to depict a 3D scene, such as 
the commonly used point clouds \citep{furukawa2009accurate}, height-map \citep{pollefeys2008detailed}, voxel grids \citep{niessner2013real, seitz1997photorealistic} and 3D triangular meshes \citep{Delaunoy2014CVPR}. Each has its own advantages and limitations. 
Recently, implicit neural representation by Neural Radiance Fields (NeRF) \citep{mildenhall2020nerf} has drawn great attention, due to its impressive 3D representation capability. NeRF represents the scene with a Multi-layer Perception (MLP) network. It takes a 5D vector (\ie the 3D position and 2D viewing direction of a query point) as input and outputs the corresponding radiance and volume density of the query point. The pixel intensity is then accumulated by differentiable volume rendering \citep{levoy1990efficient, max1995optical}. The parameters of the MLP network can be estimated by maximizing the photo-metric consistency across images captured from different viewpoints.

%% Motivation (reconstruct 3D from rs images)
To obtain a good representation of the 3D scene with NeRF, both high-quality images and corresponding accurate camera poses are usually necessary. However, it is usually difficult to acquire such perfect inputs in real-world scenarios, as real images can be easily degraded by motion blur, de-focus, rolling shutter (RS) effect \etc. Different from the commonly assumed global shutter camera model by NeRF, rolling shutter cameras capture images row by row sequentially, as illustrated in \figrefer{fig_Image_formation}. Different rows of the image are thus scanned at different timestamps, which would lead to rolling shutter distortions if it is captured by a moving camera. Neglecting these distortions usually can lead to performance degradation in 3D reconstruction, motion estimation as well as camera localization \etc., via rolling shutter images. 
A trivial solution to mitigate the effect of the rolling shutter distortions is to apply a state-of-the-art RS effect correction algorithm \citep{liu2020deepunroll,fan2021RSSR,fan2022CVR} to pre-process the images before they are fed into downstream tasks. However, those methods usually require to be pre-trained by a large dataset, which can be expensive to obtain in real-world scenarios. The generalization performance of those pre-trained networks is also limited as demonstrated in our experiments.
%  What we do
Therefore, we propose unrolling shutter bundle adjusted neural radiance fields in this paper. The proposed method is able to learn the 3D representation and recover the camera motion trajectory simultaneously. High quality un-distorted global shutter images can be further synthesized (\ie with RS effect removed) from the learned 3D scene representation. 

%% How do we do it in brief detail
In particular, we propose to represent the 3D scene with NeRF and model the camera motion trajectory with a differentiable continuous-time cubic B-Spline in the $\textbf{SE}(3)$ space. Given a sequence of rolling shutter images, we aim to optimize the camera motion trajectory (\ie estimate the parameters of the cubic B-Splines) and learn the implicit 3D representation simultaneously. The optimization is achieved by formulating the real physical image formation process of a RS camera, and maximizing the photometric consistency between the rendered and captured RS images. The method thus does not require any pre-training and would have better generalization performance compared to prior learning-based works as demonstrated in our experiments. Given the estimated continuous-time motion trajectory and the learned 3D scene representation, we can further recover the global shutter images in arbitrary desired frame-rate in high quality. We dub our method as {\textit{USB-NeRF}}, \ie, Unrolling Shutter Bundle Adjusted Neural Radiance Fields. 

\begin{figure}
	\vspace{-2.0em}
	\begin{center}
		\setlength\tabcolsep{1.2pt}
		\begin{tabular}{cccccc}
			\includegraphics[width=0.16\textwidth]{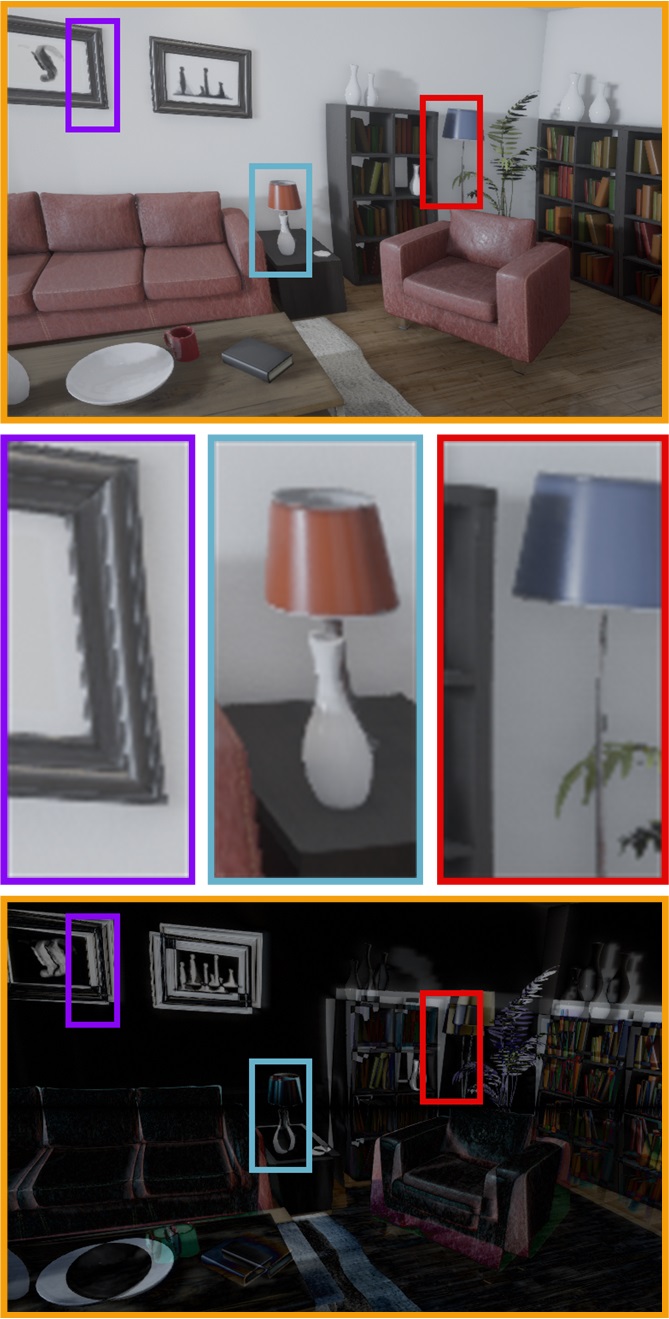} &
			\includegraphics[width=0.16\textwidth]{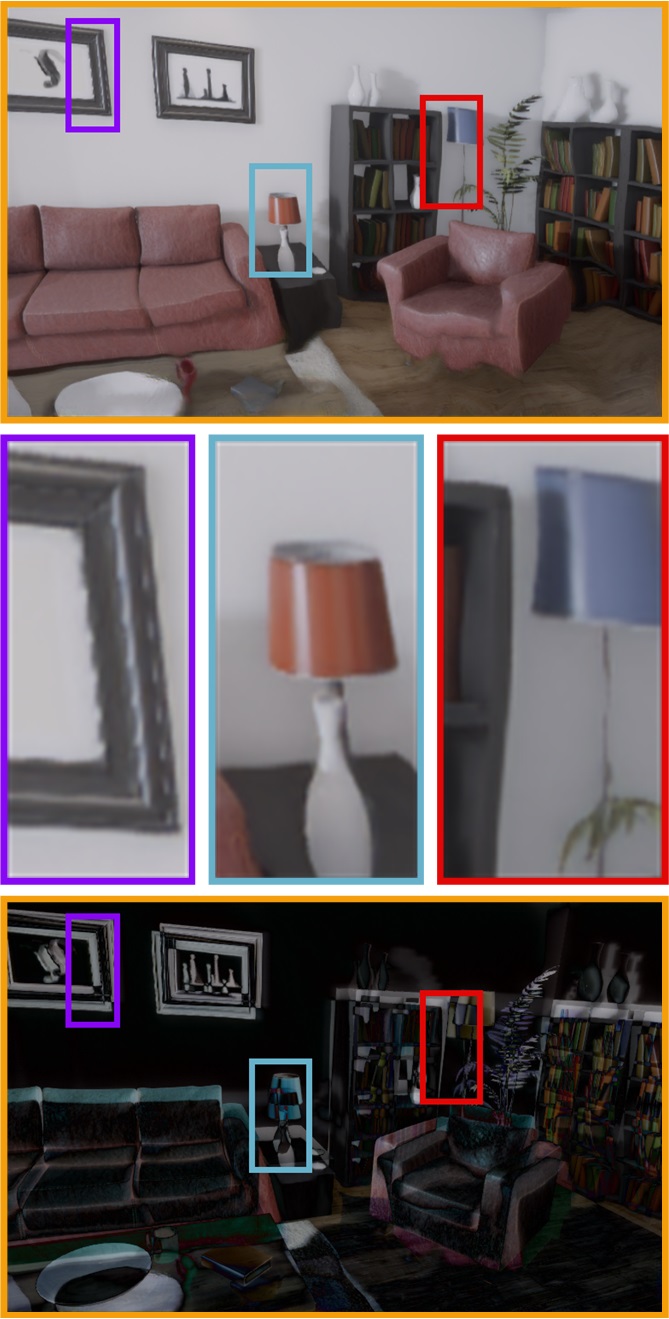} &
			\includegraphics[width=0.16\textwidth]{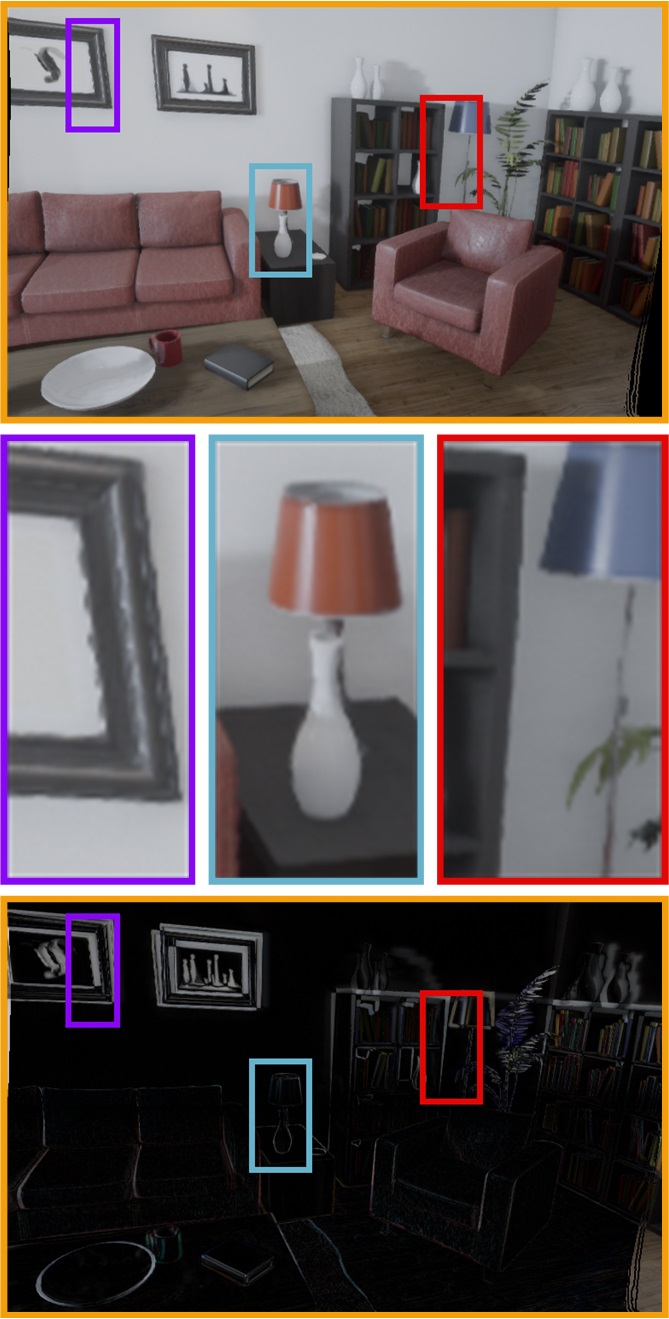} &
			\includegraphics[width=0.16\textwidth]{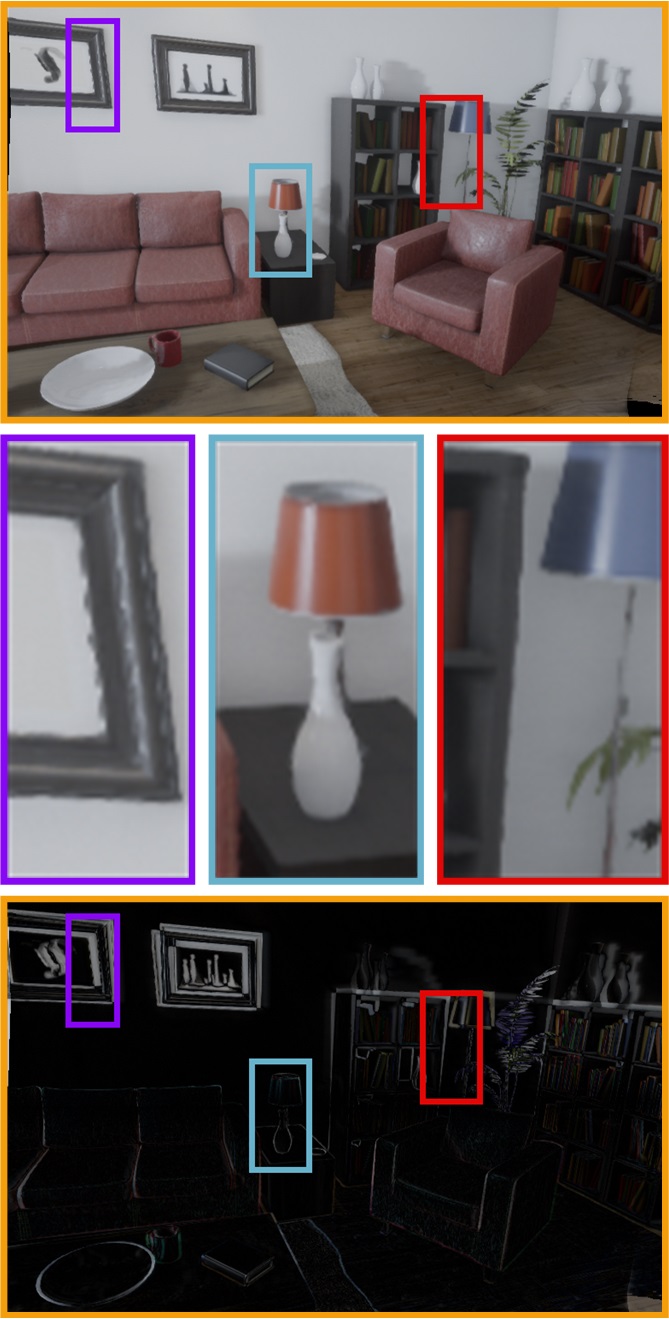} &
			\includegraphics[width=0.16\textwidth]{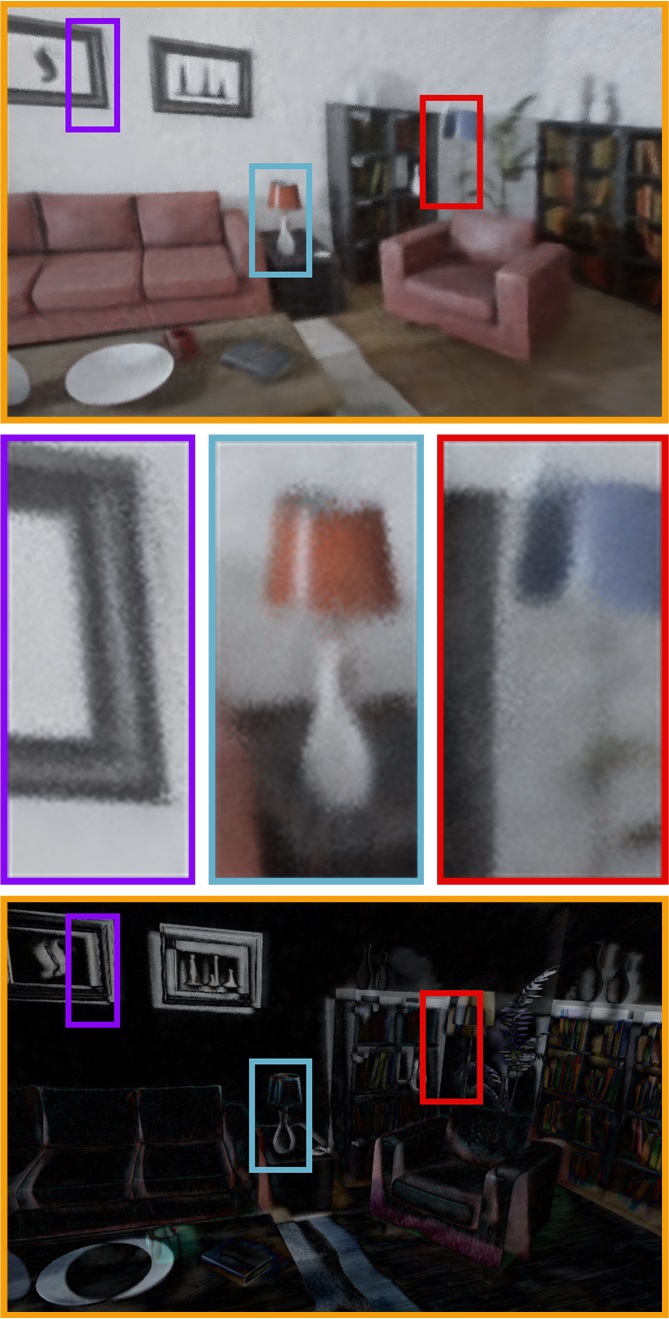} &
			\includegraphics[width=0.16\textwidth]{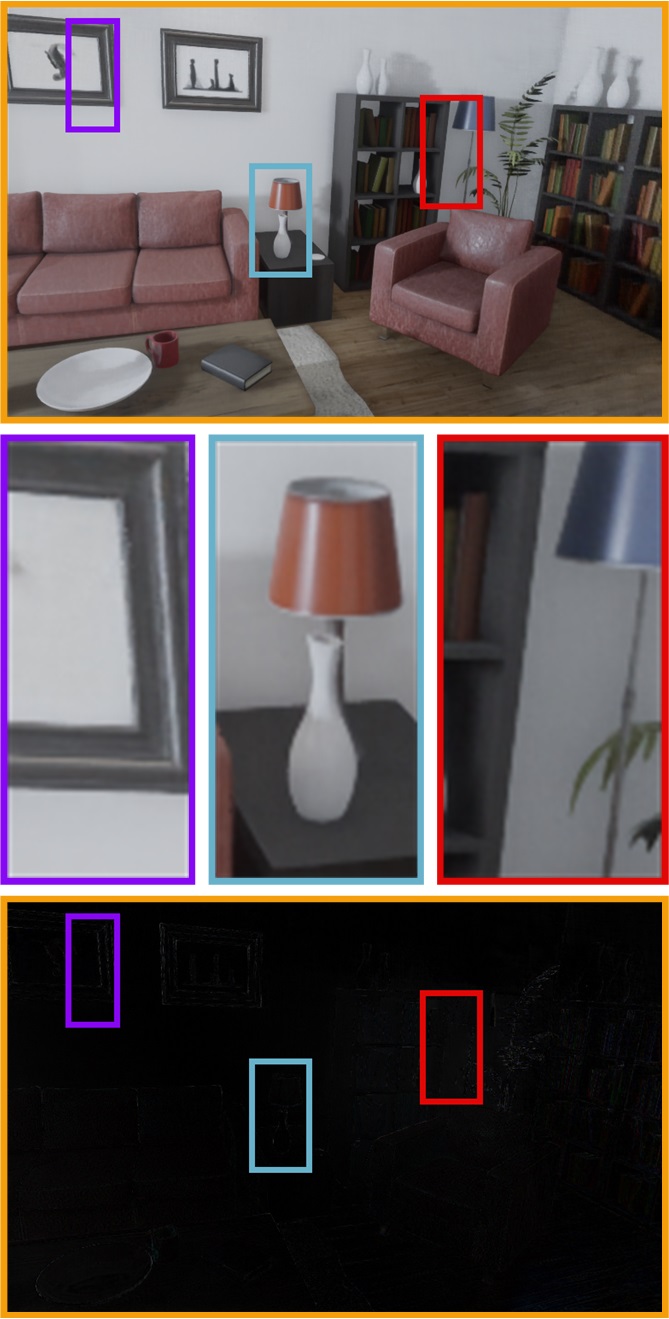} \\
%			Input RS image & DSUN \cite{liu2020deepunroll}  & RSSR \cite{fan2021RSSR} & CVR \cite{fan2022CVR} & BARF \cite{lin2021barf} & USB-NeRF (Ours)
			\scriptsize{Input RS image} & \scriptsize{DSUN}  & \scriptsize{RSSR} & \scriptsize{CVR} & \scriptsize{BARF} & \scriptsize{USB-NeRF (Ours)}
		\end{tabular}
		\vspace{-0.6em}
		\captionsetup {font={small,stretch=0.5}}
		\captionof{figure}{Given a sequence of rolling shutter images, our method is able to simultaneously learn the undistorted 3D scene representation and recover the continuous-time camera motion trajectory. Global shutter images with removed rolling shutter effect can then be rendered from the learned 3D representation. The {\bf third row} presents residual images (the darker the better) that are defined as the absolute difference between the corresponding images ({\bf first row}) and ground truth global shutter images. }
		\label{fig_teaser}
		\vspace{-2em}
	\end{center}
\end{figure}

%% What do we achieve and then summarize our main contributions
Extensive experimental evaluations are conducted with both synthetic and real datasets, to evaluate the performance of our method. The experimental results demonstrate that {\textit{USB-NeRF}} achieves superior performance compared to prior state-of-the-art methods (\eg as shown in \figrefer{fig_teaser}) in terms of rolling shutter effect removal, novel view image synthesis as well as camera motion estimation. 

%Our main {\bf{contributions}} are summarized as follows:
%\vspace{-0.3em}
%\begin{itemize}
%	\itemsep0em 
%	\item We propose a dense bundle adjustment (BA) formulation for rolling shutter images, under the framework of NeRF;
%%	\item We also provide theoretical analysis on the degeneracy issue of BA with sequential rolling shutter images, and then prove why the commonly applied constant velocity assumption can help mitigate such degeneracy, which has never been proved before;
%	\item Experimental results demonstrate the superior performance of our method in terms of RS effect correction, camera motion estimation and novel view image synthesis, compared to prior works;
%	\item Our method can also be used to restore high-fidelity global shutter video at an arbitrary frame rate from a sequence of low frame-rate RS images.
%\end{itemize}

\section{Related Work}
\label{sec:related}
We review the related works in two main areas: neural radiance fields and rolling shutter effect correction, which are the most related to our work. 

\noindent{\bf Neural Radiance Fields.} NeRF has received lots of attention recently due to its impressive capability to represent 3D scenes (\cite{mildenhall2020nerf}). 
Many extensions have been proposed to further improve its performance. For example, \cite{muller2022instant, yu2021plenoctrees, fridovich2022plenoxels, chen2022tensorf, garbin2021fastnerf} proposed approaches to accelerate its training and rendering efficiency. 
Other extensions also explore NeRF for dynamic scenes (\cite{pumarola2021d, gao2021dynamic, park2021nerfies, tretschk2021non}) and scene editing (\cite{li2022climatenerf, liu2021editing, sun2022fenerf, kania2022conerf}). 
Apart from this, there are also many variants have been proposed to address the training of NeRF with imperfect inputs, such as with unknown or inaccurate poses (\cite{wang2021nerf, lin2021barf, chen2022local, meng2021gnerf}), with degraded images (\eg blurry images \cite{ma2022deblur, wang2023bad}, dark/noisy images \cite{mildenhall2022nerf, pearl2022nan}, low dynamic range images \cite{huang2022hdr}), or with a limited number of input images \etc (\cite{niemeyer2022regnerf, yu2021pixelnerf, kim2022infonerf, xu2022sinnerf, deng2022depth}).

We will review those methods in detail which are the most related to our work as follows. To overcome the effect of inaccurate camera poses, NeRF-{}- (\cite{wang2021nerf}) sets the camera poses as learnable parameters and optimizes them with the weights of NeRF jointly by minimizing the photo-metric loss. GNeRF (\cite{meng2021gnerf}) further integrates an additional adversarial loss into the training of the whole pipeline to have a better camera pose estimation. BARF (\cite{lin2021barf}) and L2G-NeRF (\cite{chen2022local}) propose to gradually apply the positional encoding to achieve a coarse-to-fine training strategy, to better constrain the training of the network and camera pose estimation. 
Although these methods have achieved impressive results with imperfect poses, images with rolling shutter effect are still a problem for NeRF. Prior works usually assume a global shutter camera model and use a single transformation matrix to represent the pose of each view. They are thus not suitable for rolling shutter camera model, in which each row has different poses. We therefore parameterize the whole motion trajectory of the RS image sequence with a differentiable continuous-time cubic B-Spline parameterized in the $\textbf{SE}(3)$ space. We then formulate the image formation process of a rolling shutter camera into the joint training of NeRF and the parameter estimations of the cubic B-Splines. 

\noindent{\bf Rolling Shutter Effect Correction.}
RS effect removal is a challenging problem, and many related methods have been proposed over the last decades \cite{Forssen2010CVPR, Baker2011CVPR, Rengarajan2016CVPR, Purkait2017ICCV, Lao2018CVPR, Vasu2018CVPR} \etc. We will detail several recent state-of-the-art methods as follows. 
\cite{hedborg2012RSBA} propose to recover the camera poses and sparse 3D geometry from a sequence of rolling shutter images. They assume a piece-wise linear motion model for each frame and propose a sparse bundle adjustment solver for rolling shutter cameras.
\cite{grundmann2012calibration} present a mixture model of homographies to model rolling shutter distortions of video streams. 
\cite{zhuang2017sfm_RS} later develop an RS-aware differential Structure from Motion (SfM) algorithm to estimate the relative poses of two consecutive RS images and then rectify the distortions. 
As for unorganized RS images, RS effect correction has been shown to suffer from severe degeneracy (\cite{albl2016degeneracies}). To mitigate the degeneracy of rolling-shutter (RS) SfM, \cite{ito2017RS-sfm} propose to add a critical camera motion constraint; \cite{albl2020dual-RS} and \cite{zhong2022dual} propose to employ dual RS images with reversed directions to avoid the ambiguity. 
Deep-learning-based approaches have also been proposed to address RS effect correction recently. \cite{rengarajan2017unrolling} propose a convolutional neural network (CNN) to estimate the row-wise camera motion from a single RS image. \cite{liu2020deepunroll} and \cite{fan2021sunet} design special shutter unrolling networks to recover the global shutter image from two consecutive images. \cite{fan2021RSSR} and \cite{fan2022CVR} further developed Acceleration-Net and bilateral motion field approximation model to achieve RS temporal super-resolution. 
While those methods deliver state-of-the-art performance, they usually require a large dataset for network training. Those datasets are usually expensive to obtain in practice and further limit their generalization performance to images with different characteristics (as shown in our experiments). In contrast, our method does not require any pre-training with large datasets and would thus have no generalization issues.

%
%
%However, the above-mentioned methods face different limitations. Many approaches under the assumption of consecutive RS image frames only model the motion between two neighboring frames, while sometimes dropping the geometric information provided by multiple images. Learning-based methods require the supervision of plenty of GS images, thus leading to poor generalization into new datasets. Instead, we employ Bundle Adjustment, and scene representation with NeRF to optimize the camera motion trajectory and scene reconstruction simultaneously. Scene representation with NeRF employing the scene’s geometric constraint could better preserve the scene structure, thus improving the quality of retrieved GS images. And motion trajectory modeling could utilize the motion information to reconstruct high-fidelity GS image videos.

\section{Method}
\label{sec:method}

\begin{figure*}[!ht]
	\vspace{-2.0em}
	\small
	\begin{center}
		\setlength{\belowcaptionskip}{-1.6em}
		\includegraphics[width=0.9\textwidth]{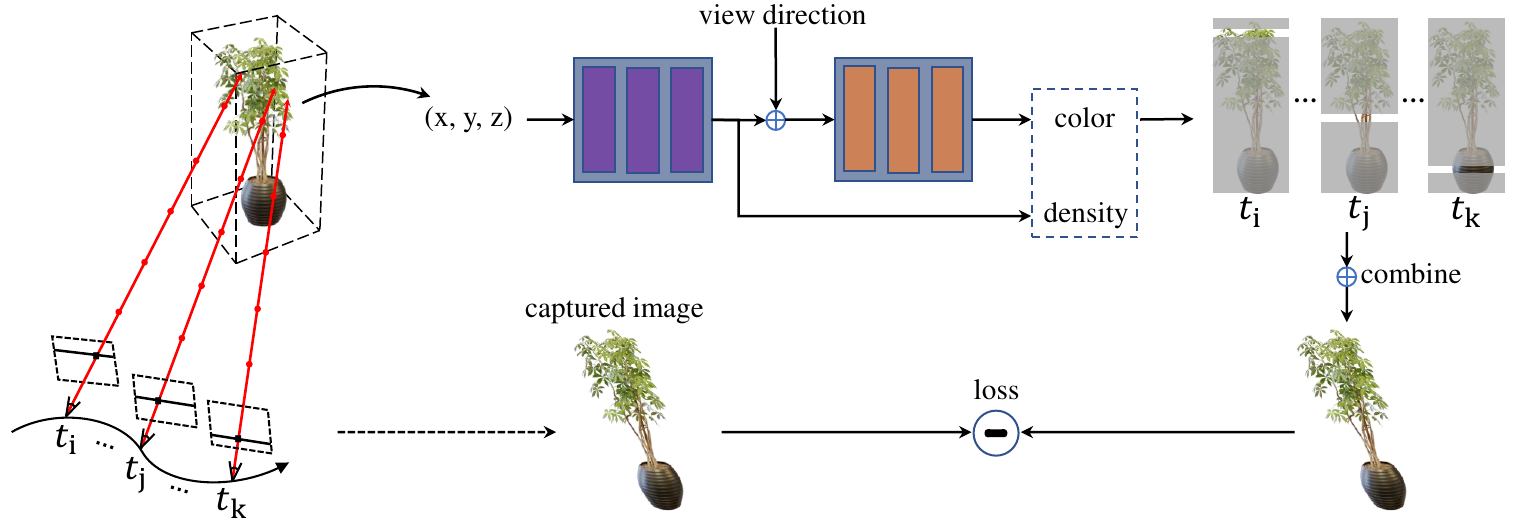}
		\vspace{-.8em}
		\captionsetup {font={small,stretch=0.5}}
		\caption{{\bf The pipeline of USB-NeRF.} Given a sequence of rolling shutter images, we train NeRF to learn the underlying undistorted 3D scene representations. We parameterize the camera motion trajectory of the image sequence by a continuous-time cubic B-Spline in $\textbf{SE}(3)$ space. Given the capturing time for each row of the rolling shutter image, we can interpolate its pose from the spline. Each rolling shutter image can then be synthesized by rendering all the image rows (\ie each with different poses) from NeRF. By maximizing the photo-metric consistency between the synthesized and captured RS images, we can learn the underlying 3D scene representation and recover the camera motion trajectory. Global shutter images can then be rendered from the learned 3D representation with known camera poses.}
		\label{fig_method}
	\end{center}
\end{figure*}

In this section, we present the details of our unrolling shutter bundle adjusted neural radiance fields (USB-NeRF). USB-NeRF takes a sequence of rolling shutter images as input. It then learns the underlying 3D scene representation and recovers the continuous camera motion trajectory simultaneously, by maximizing the photo-metric consistency between the rendered and captured RS images. The learned 3D representation is free of rolling shutter distortions and thus able to be used for arbitrary frame-rate global shutter image/video synthesis, provided the recovered continuous-time camera motion trajectory. The details of the method are shown in \figrefer{fig_method}. We will detail each component as follows. 

\subsection{Neural Radiance Fields}
We represent the 3D scene implicitly with a Multi-layer Perceptron (MLP) network. We adopt the original architecture of NeRF proposed by \citet{mildenhall2020nerf}. More advanced variants of NeRF, such as voxel-based NeRF representation with improved efficiency from \citet{yu2021plenoctrees} is also feasible to be used for our method. 

Given a camera view with known pose, we can render its corresponding image from the implicit 3D representation by using volume rendering. For convenience, we present the steps to render the intensity of a particular pixel to illustrate the concept. The rendering procedures of other pixels are the same. To render the pixel intensity $\bI(\bx)$ at pixel location $\bx$ for a particular image with pose $\bT_c^{w}$, we can query the radiance and volume density of each 3D point along the ray from camera center to the 3D space passing through $\bx$. The pixel intensity (\ie $\bI(\bx)$) can then be computed by accumulating the sampled radiance and volume densities along the ray. The whole procedure can be formally defined as follows. 

Assume the sampled 3D point along the ray has depth $\lambda$, its 3D position $\bX^{w}$ in the world coordinate frame can be computed by (assuming the camera has a pin-hole model): 
\begin{align}
	\bd^{c} &= \bK^{-1} \begin{bmatrix}
		\bx \\
		1
	\end{bmatrix}, \\
	\bX^{w} &= \bT_c^w \cdot \lambda \bd^{c},
\end{align}
where $\bd^{c}$ is the ray direction defined in the camera coordinate frame, $\bK$ is the camera intrinsic matrix, $\bx$ is its 2D pixel coordinate, and $\bT_c^w$ is the transformation matrix used to convert a 3D vector from the camera coordinate frame to world coordinate frame.
We can then query the MLP network $\bF_{\btheta}$ parameterized by $\btheta$ for the radiance $\bc$ and volume density $\sigma$ of the sampled 3D point $\bX^{w}$ by: 
\begin{equation} \label{eq_mlp_query}
	(\bc, \sigma) = \bF_{\btheta}(\gamma_{L_x}(\bX^w), \gamma_{L_d}(\bd^w)),
\end{equation}
where $\bd^w = \bR_c^{w} \cdot \bd^c$ is the viewing direction of the ray defined in the world coordinate frame, $\bR_c^{w}$ is the rotation matrix which transforms vectors from camera frame to world frame, and $\gamma_{*}$ represents positional encodings for $\bX^w$ and $\bd^w$ \citep{mildenhall2020nerf}. 
The final pixel intensity can then be computed from $N$ sampled 3D points along the ray via:
\begin{equation}
	\bI(\bx) = \sum_{i=1}^{N} T_i(1-{\rm exp}(-\sigma_i\delta_i))\bc_i,
\end{equation}
where $\bc_i$ and $\sigma_i$ are the predicted radiance and volume density of the $i^{th}$ point by \eqnref{eq_mlp_query} respectively, $\delta_i = \norm{\bX_{i+1}^w - \bX_i^w}_2$ is the distance between two adjacent points, and $T_i$ represents the  transmittance of the $i^{th}$ point and can be computed by:
\begin{equation}
	T_i = {\textrm{exp}} (-\sum_{k=1}^{i-1}\sigma_k\delta_k).
\end{equation}
According to the above derivations, we can see that $\bI(\bx)$ is also a function of the camera pose $\bT_c^w$. Since the whole rendering procedure is differentiable, $\bT_c^w$ can thus also be relaxed as a free parameter to be optimized during the training of the MLP network (\ie $\bF_{\btheta}$) \citep{lin2021barf}.

%
%-------------------------------------------------------------------------
%\begin{figure}[t]
%	\vspace{-2em}
%	\begin{center}
%	\small
%	\setlength{\belowcaptionskip}{-1.8em}
%	\begin{minipage}[t]{0.3\linewidth}
%		\centering
%		\includegraphics[width=\textwidth]{figures/Fig1-RS.jpg}
%		\centerline{(a) Rolling shutter}
%	\end{minipage}%
%	\begin{minipage}[t]{0.3\linewidth}
%		\centering
%		\includegraphics[width=\textwidth]{figures/Fig1-GS.jpg}
%		\centerline{(b) Global Shutter}
%	\end{minipage}
%	\vspace{-0.5em}
%	\captionsetup {font={small,stretch=0.5}}
%	\caption{{\bf Image formation models of a rolling shutter camera and a global shutter camera respectively.} It demonstrates that each row of a rolling shutter image is captured at different timestamps, and would thus lead to image distortions if the image is captured by a moving camera.}
%	\label{fig_Image_formation}
%	\end{center}
%\end{figure}
%

\subsection{Rolling Shutter Camera Model}
\begin{wrapfigure}{r}{0.5\textwidth}
	\vspace{-0.5cm}
	\begin{center}
		\begin{tabular}{cc}
			\includegraphics[width=0.23\textwidth]{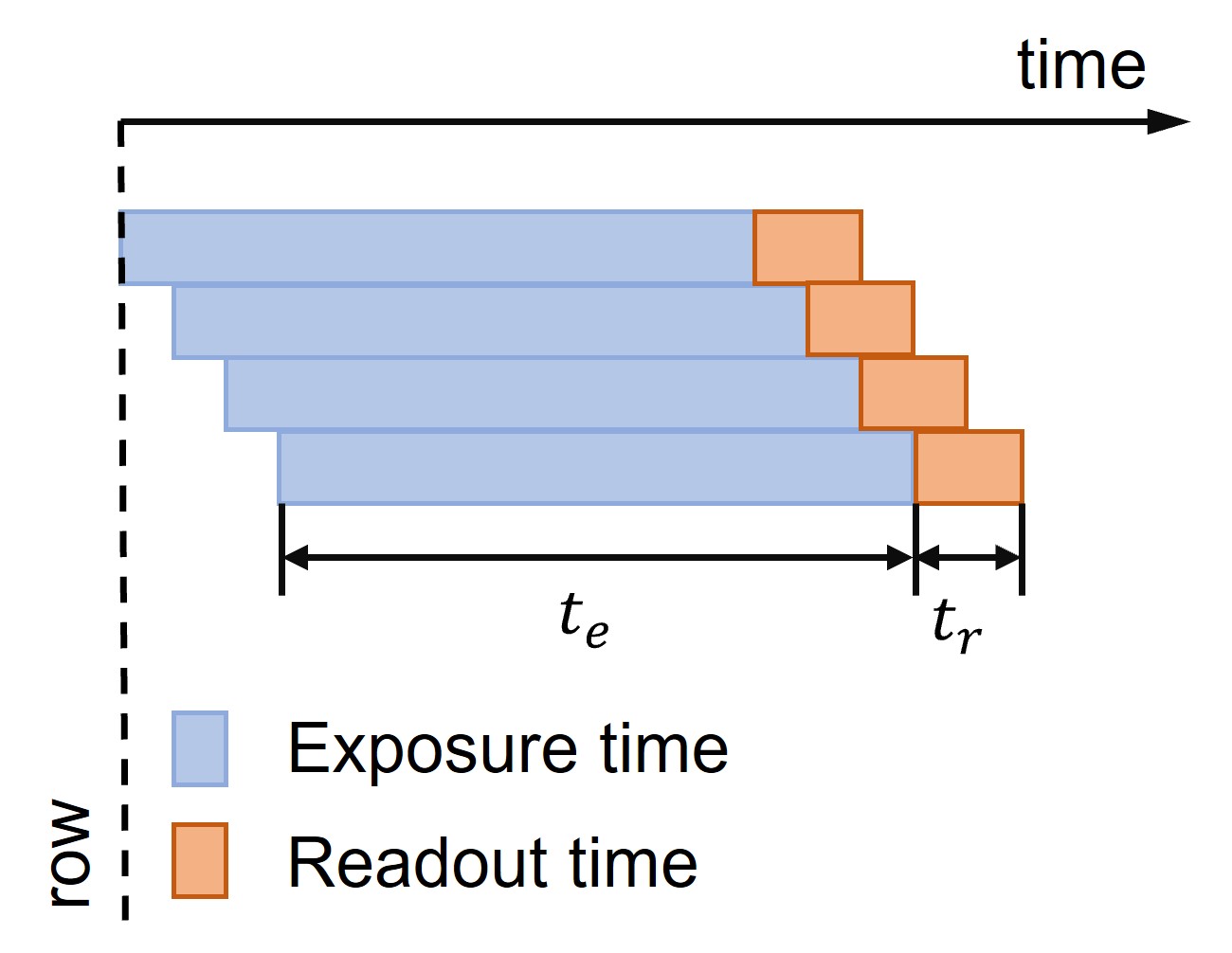} &
			\includegraphics[width=0.23\textwidth]{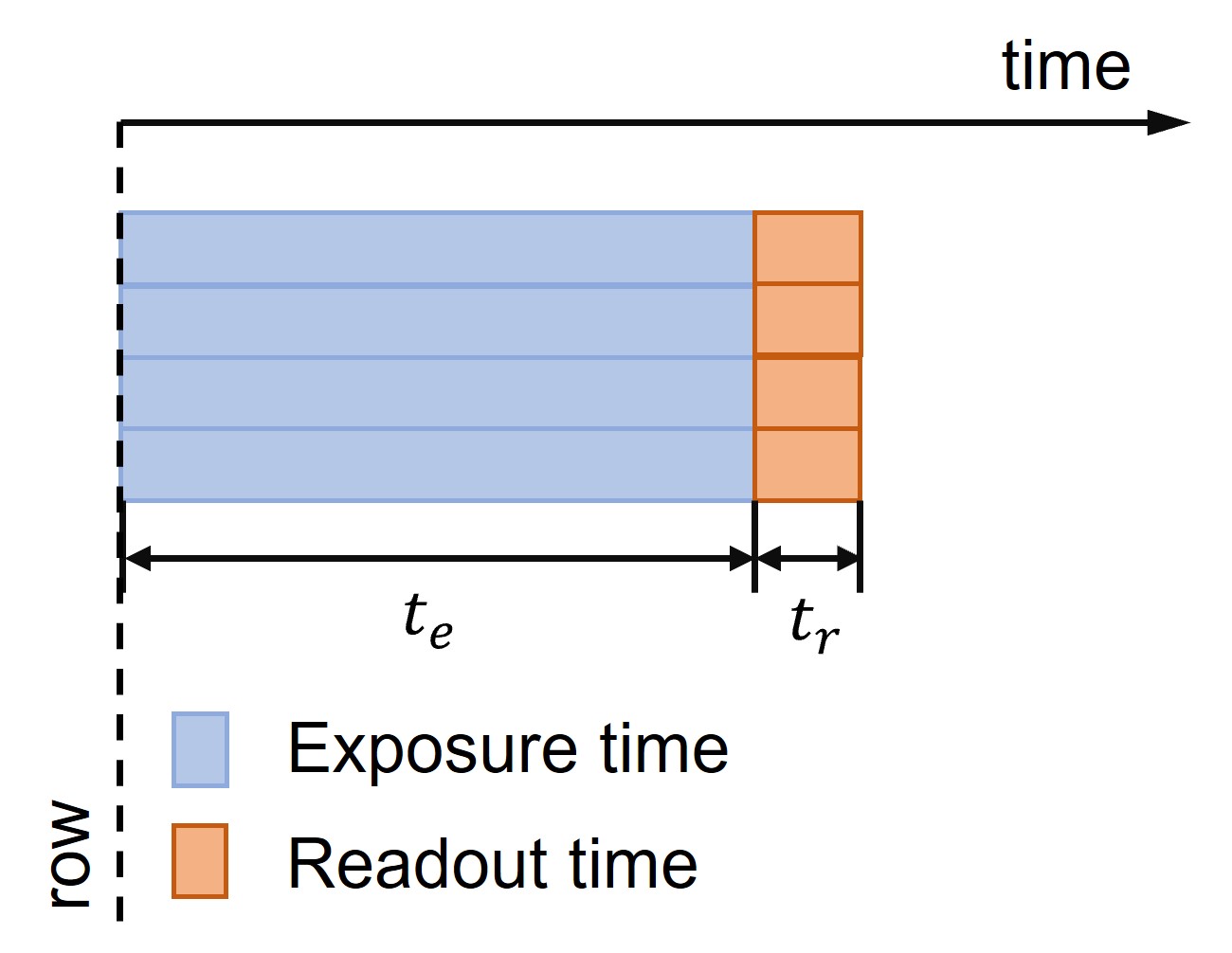} \\
			\footnotesize{(a) Rolling shutter} & \footnotesize{(b) Global shutter}
		\end{tabular}
	\end{center}
	\vspace{-0.4cm}
	\captionsetup {font={small,stretch=0.5}}
	\caption{{\bf Image formation models of a rolling shutter camera and a global shutter camera respectively.} It demonstrates that each row of a rolling shutter image is captured at different timestamps, and would thus lead to image distortions if the image is captured by a moving camera.}
	\label{fig_Image_formation}
	\vspace{-0.4cm}
\end{wrapfigure}

Different from global shutter cameras, each scanline/image row of the rolling shutter camera is captured at different timestamps. Without loss of generality, we assume the readout direction of RS camera is from top to bottom as shown in \figrefer{fig_Image_formation} in our formulation. This process can be mathematically modeled as (assuming infinitesimal exposure time):
\begin{equation}
	[\bI^r(\bx)]_i = [\bI^g_i(\bx)]_i,
\end{equation}
where $\bI^r(\bx)$ is the rolling shutter image, $[\bI(\bx)]_i$ denotes an operator which extracts the $i^{th}$ row from image $\bI(\bx)$, $\bI^g_i(\bx)$ is the global shutter image captured at the same pose as the $i^{th}$ row of $\bI^r(\bx)$. We denote the pose of the $i^{th}$ row of $\bI^r(\bx)$ as $\bT_{c_i}^w$. Thus, provided the 3D representation by NeRF and the known poses $\bT_{c_i}^w$ for $i=0,1,...,(H-1)$, where $H$ is the height of the image, we can easily render the corresponding rolling shutter image $\bI^r(\bx)$.

From the above derivations, we can see that $\bI^r(\bx)$ is a function of $\btheta$ (\ie the weight of the MLP network), and $\bT_{c_i}^w$ for $i=0,1,...,(H-1)$. Furthermore, we can also find that $\bI^r(\bx)$ is differentiable with respect to $\bT_{c_i}^w$ and $\btheta$. It thus lays the foundation for our bundle adjustment formulation with a sequence of rolling shutter images. 

\subsection{Camera Motion Trajectory Modeling}
To do bundle adjustment optimization with a sequence of rolling shutter images, we need to parameterize the pose of each row for each image as shown in the previous section. Commonly used parameterization is to assign a 6 DoF pose to the first row of each image and then do linear interpolation for subsequent rows \citep{hedborg2012RSBA}. Instead of using such kind of simple linear motion model, we propose to use cubic B-Splines parameterized in the $\textbf{SE}(3)$ space in this work, which can handle more realistic complex camera motions \citep{Lovegrove2013BMVC}. Experimental ablation studies also verify that cubic B-Splines formulation delivers better performance than the simple linear motion model for complex motions. 

We use a sequence of control knots with poses $\bT_{c_i}^w$ \footnote{Here we abuse the same notation as the previously defined transformation matrix of the $i^{th}$ row of the rolling shutter image.} for $i=0,1,...,n$, to represent the spline. For brevity, we denote $\bT_{c_i}^w$ with $\bT_i$ for subsequent derivations. We assume the control knots are sampled with a uniform time interval $\Delta t$ and the trajectory starts from $t_0$. Spline with a smaller $\Delta t$ can represent a smoother motion, with an expense of more parameters to optimize. Since four consecutive control knots determine the value of the spline curve at a particular timestamp, we can thus compute the starting index of the four control knots for time $t$ by:
\begin{equation}
	k = \lfloor \frac{t - t_0}{\Delta t} \rfloor,
\end{equation}
where $\lfloor * \rfloor$ is the floor operator. Then we can obtain the four control knots responsible for time $t$ as $\bT_k$, $\bT_{k+1}$, $\bT_{k+2}$ and $\bT_{k+3}$. We can further define $u = \frac{t - t_0}{\Delta t} - k$, where $u \in [0, 1)$ to transform $t$ into a uniform time representation. Using this time representation and based on the matrix representation for the De Boor-Cox formula \citep{Qin1998CGA}, we can write the matrix representation of a cumulative basis $\tilde{\bB}(u)$ as 
\begin{align}
	\tilde{\bB}(u) = \bC \begin{bmatrix}
		1 \\ u \\ u^2 \\ u^3
	\end{bmatrix},  \quad
    \bC = \frac{1}{6} \begin{bmatrix}
    	6 & 0 & 0 & 0 \\
    	5 & 3 & -3 & 1 \\
    	1 & 3 & 3 & -2 \\
    	0 & 0 & 0 & 1
    \end{bmatrix}.
\end{align}
The pose at time $t$ can then be computed as:
\begin{equation}
	\bT(u) = \bT_k \cdot \prod_{j=0}^2 \mathrm{exp}(\tilde{\bB}(u)_{j+1} \cdot \bOmega_{k+j}),
\end{equation}
where $\tilde{\bB}(u)_{j+1}$ denotes the $(j+1)^{th}$ element of the vector $\tilde{\bB}(u)$, $\bOmega_{k+j} = \mathrm{log}(\bT_{k+j}^{-1} \cdot \bT_{k+j+1})$. 

From the above derivations, we can see that the interpolated camera poses are functions of the poses of the control knots. They are also differentiable with respect to the poses of those control knots.

\subsection{Loss Function}
Given a sequence of rolling shutter images, we can then estimate the learnable parameters $\btheta$ of NeRF as well as the camera motion trajectory parameterized by cubic B-Spline (\ie $\bT_0, \bT_1, ...,$ and $\bT_n$), by minimizing the photo-metric loss:
\begin{equation} \label{eq_loss_function}
	\cL = \sum_{m=0}^{M-1} \norm{\bI^r_m(\bx) - \tilde{\bI}^r_m(\bx)}_2,
\end{equation}
where $M$ is the number of input images, $\bI^r_m(\bx)$ is the captured rolling shutter image, and $\tilde{\bI}^r_m(\bx)$ is the rendered rolling shutter image from NeRF, with known camera intrinsic parameters \etc. $\tilde{\bI}^r_m(\bx)$ is a function of $\btheta$ as well as part of the control knots' poses of the whole trajectory. We implement the above equations with PyTorch and exploit its automatic differentiation module to compute the Jacobian for back-propagation.

\section{Experiments}\label{sec:experiment}
\subsection{Datasets}
%In this section, we will give an overview of the datasets that are used for evaluating our method.

\noindent{\bf Synthetic datasets.}\quad
We use the scripts provided by \citet{liu2020deepunroll} to synthesize 4 datasets with the Unreal game engine (\ie Unreal-RS-BlueRoom, Unreal-RS-LivingRoom, Unreal-RS-WhiteRoom, Unreal-RS-Adornment) and 2 datasets with Blender (\ie Blender-RS-Factory, Blender-RS-Tanabata). We adopt the real motion trajectories from ETH3D \citep{ETH3D} (\ie the challenging shaky sequences) to synthesize the images. Since the ground truth pose of ETH3D dataset has only 100 Hz, we further interpolate the trajectories with cubic spline to have a continuous-time representation. We first capture GS images in the Unreal game engine and Blender at the frequency of 10,000 Hz, then synthesize RS images by simulating the physical image formation process of an RS camera. We configure the scanline readout time as 100 $\mu$s and the image resolution as 768$\times$480 pixels. We generate 40 RS images in total for Unreal-RS-BlueRoom, Unreal-RS-LivingRoom, Unreal-RS-WhiteRoom individually and 80 RS images for Unreal-RS-Adornment, Blender-RS-Factory, Blender-RS-Tanabata. We also evaluate our method on a public synthetic dataset (\ie Carla-RS\footnote{We find that COLMAP \citep{schonberger2016structure} is hardly to recover the poses of the Fastec-RS dataset \citet{liu2020deepunroll}, which is used to initialize our method. We did not evaluate on this dataset.} \citep{liu2020deepunroll}) for fair comparisons against other methods. 
% We evaluate our method on dataset-seq1, dataset-seq2 and dataset-seq6 of TUM-RS dataset with whole sequences as other whole sequences do not satisfy requirements of NeRF \citep{mildenhall2020nerf}.

\noindent{\bf Real datasets.}\quad We captured 5 sequences using GoPro HERO6 Black, Canon camera (EOS M3), and iPhone 14 Pro. All rolling shutter images are captured at the frequency of 30 Hz. The scanline readout time of aforementioned 3 cameras is approximately 13.89 $\mu$s, 18.52 $\mu$s and 3.70 $\mu$s, respectively. We also evaluate our method on the public real-world dataset TUM-RS \citep{schubert2019RS-VIO}. TUM-RS consists of 10 real challenging indoor sequences of rolling shutter images, which are originally used for RS visual-inertial odometry evaluations. It records RS images, groundtruth motion trajectories at the frequency of 20 Hz and 120 Hz, respectively. The scanline readout time of the RS camera is approximately 29.47 $\mu$s. As whole sequences are too long for NeRF to process, we choose a subset frames from each sequence. Details are presented in the Appendix \ref{sec:appendix_dataset}. Since there are no pixel-aligned RS-global shutter image pairs for this dataset, we only evaluate the accuracy of recovered camera motion trajectories. We also provide additional qualitative comparisons on the RS effect removal against its nearest neighbor global shutter images. 

\subsection{Baseline Methods and Evaluation Metrics}
\noindent{\bf Baselines.} We compare our method against learning-free method DiffSfM \citep{zhuang2017sfm_RS} and several learning-based state-of-the-art rolling shutter effect removal methods, \eg DSUN \citep{liu2020deepunroll}, SUNet \citep{fan2021sunet}, RSSR \citep{fan2021RSSR}, CVR \citep{fan2022CVR}. Those learning-based methods usually take two consecutive images as input and train a deep network to recover the corresponding global shutter image. The network training usually requires a large dataset, which would be expensive/difficult to obtain in real scenarios. For fair comparisons, we use the official pre-trained models (of those baseline methods) for evaluations with Carla-RS dataset. However, for the newly synthesized datasets (\eg Unreal-RS) as well as the real TUM-RS dataset, we are unable to fine-tune those methods due to a limited number of images. We, therefore, use the official pre-trained models for evaluations. 
Additionally, we also compare against the performance of the original NeRF \citep{mildenhall2020nerf} and BARF \citep{lin2021barf} assuming the inputs are global shutter images with poses computed by COLMAP \citep{schonberger2016structure}.

\noindent{\bf Evaluation metrics.}\quad We evaluate the performance regarding rolling shutter effect removal and novel view image synthesis with the commonly used metrics, \eg PSNR, SSIM and LPIPS  \citep{zhang2018LPIPS}, between the recovered global shutter images and the ground truth global shutter images. We also compute the absolute trajectory error (ATE) against that estimated by COLMAP \citep{schonberger2016structure}, BARF \citep{lin2021barf} RSBA \citep{hedborg2012RSBA} and NW-RSBA \citep{liao2023NW_RSBA}, which is the most related to ours. The ATE error metric is commonly used for trajectory estimation evaluations in the visual odometry community.

%\begin{table}
%	\setlength{\belowcaptionskip}{-9pt}
%	\setlength\tabcolsep{2pt}
%    \footnotesize
%	% \resizebox{\linewidth}{!}{
	%		\begin{tabular}{c|c c c c}
		%			\toprule
		%			&  COLMAP \citep{schonberger2016structure} & BARF \citep{lin2021barf} & USB-NeRF-linear & USB-NeRF-cubic \\
		%			\midrule
		%			1 &{.0313}$\pm${.0240} &.0376$\pm$.0184 &.0071$\pm$.0034 &{\bf.0065}$\pm$.0020\\
		%			2 &{.1550}$\pm${.0710} &.1831$\pm$.0625 &{\bf.0894}$\pm$.0444 &.0954$\pm$.0461\\
		%			3 &.0575$\pm${.0330} &.0781$\pm$.0441 &.0210$\pm$.0131 &{\bf.0181}$\pm$.0144\\
		%			4 &.0183$\pm${.0052} &.0310$\pm$.0129 &{\bf.0052}$\pm$.0017 &.0053$\pm$.0019\\
		%			5 &.0674$\pm${.0240} &.0995$\pm$.0392 &.0124$\pm$.0046 &{\bf.0119}$\pm$.0039\\
		%			6 &.0408$\pm${.0225} &.0769$\pm$.0467 &.0205$\pm$.0120 &{\bf.0198}$\pm$.0119\\
		%			7 &.0196$\pm${.0065} &.0188$\pm$.0065 &.0041$\pm$.0020 &{\bf.0033}$\pm$.0016\\
		%			8 &.0416$\pm${.0198} &.0667$\pm$.0360 &{\bf.0124}$\pm$.0088 &.0125$\pm$.0088\\
		%			9 &.0517$\pm${.0174} &.1500$\pm$.0846 &.0291$\pm$.0213 &{\bf.0135}$\pm$.0082\\
		%			10 &.0451$\pm${.0139} &.1445$\pm$.0476 &.0194$\pm$.0134 &{\bf.0185}$\pm$.0111\\
		%			\bottomrule
		%	    \end{tabular}
	%    % }
%	\vspace{-0.5em}
%	\caption{{\textbf{Quantitative comparisons on TUM-RS datasets \citep{schubert2019RS-VIO} in terms of the accuracy of trajectory estimation.}} The experimental results demonstrate that our method performs much better than prior works in terms of the accuracy of motion trajectory estimation.} 
%	\label{TUM_ATE}
%	\vspace{-1em}
%\end{table}

\subsection{Experimental Results}
\PAR{Ablation studies} We evaluate four different camera pose interpolation strategies, to justify the advantage to use cubic B-Spline for whole trajectory parameterization. Besides cubic B-Spline, we also explore the linear interpolation strategy used in \citet{hedborg2012RSBA}, which assigns the first row of each RS image a pose parameter (\ie $\bT_i$) and then do linear interpolations for subsequent rows by using two neighboring poses (\ie $\bT_i$ and $\bT_{i+1}$). These two strategies bring in additional constraints between neighboring frames, \ie the pose of a particular row in the $i^{th}$ frame would also depends on the pose of the $(i+1)^{th}$ frame. To relax this constraint, we also experiment with two additional strategies. We parameterize the camera trajectory for each RS image independently, \ie assigning four controls knots for each RS image for cubic B-Spline parameterization, and two control poses (\ie $\bT_{start}$ and $\bT_{end}$) for the linear interpolation case. 

We conduct experiments with the synthetic Carla-RS and the Unreal-RS datasets, by evaluating the rolling shutter effect removal performance in terms of PSNR, SSIM and LPIPS metrics. The experimental results are presented in \tabnref{tabel_ablation_study}. The results demonstrate that the optimization cannot be properly constrained if there is no pose dependency between neighboring frames. It can be explained by the degeneracy analysis for rolling shutter structure from motion (SfM) done by \citet{albl2016degeneracies}, which states that near parallel readout directions of RS images would lead to degenerate solutions for SfM. The pose dependency by the other two parameterizations would bring in additional constraints to avoid the degenerate solutions, which are verified by the experimental results. 

The results also reveal that cubic B-Spline interpolation performs similarly to linear interpolation if the camera moves at a constant velocity (\eg Carla-RS dataset). However, it performs much better than linear interpolation, if the camera has realistic complex motions (\eg Unreal-RS dataset). Therefore, we present experimental results for subsequent evaluations with the cubic B-Spline interpolation unless explicitly stated. 

\begin{table}[t]
	\begin{center}
		\vspace{-2.0em}
		\captionsetup {font={small,stretch=0.5}}
		\caption{{\textbf{Ablation studies for motion trajectory parameterization.} {\bf USB-NeRF-lin-nodep} denotes the trajectory is parameterized with linear interpolation and there is no dependency between neighboring frames. {\bf USB-NeRF-cub-nodep} denotes the trajectory is parameterized with cubic B-Spline and there is no dependency between neighboring frames. The experimental results demonstrate that cubic B-Spline parameterization performs better than linear interpolation in general, and the pose dependency between frames is also necessary.} }
		\label{tabel_ablation_study}
		\vspace{-0.5em}
		\setlength{\belowcaptionskip}{-5pt}
		\setlength\tabcolsep{2pt}
		%	\normalsize
		\scriptsize
		\resizebox{\linewidth}{!}{
			\begin{tabular}{c|ccc|ccc|ccc|ccc}
				\toprule
				&  \multicolumn{3}{c}{Carla} & \multicolumn{3}{|c}{Blue Room} & \multicolumn{3}{|c}{Living Room} & \multicolumn{3}{|c}{White Room} \\
				& PSNR$\uparrow$ & SSIM$\uparrow$ & LPIPS$\downarrow$ & PSNR$\uparrow$ & SSIM$\uparrow$ & LPIPS$\downarrow$ & PSNR$\uparrow$ & SSIM$\uparrow$ & LPIPS$\downarrow$ & PSNR$\uparrow$ & SSIM$\uparrow$ & LPIPS$\downarrow$ \\
				\midrule
				USB-NeRF-lin-nodep &16.78 &0.551 &0.2864 &19.16 &0.590 &0.1546 &16.42 &0.580 &0.3598 &15.71 &0.469 &0.3327\\
				USB-NeRF-cub-nodep &17.52 &0.569 &0.2501 &17.77 &0.544 &0.2005 &16.73 &0.582 &0.3375 &15.89 &0.471 &0.2976\\
				USB-NeRF-linear &{\bf32.15} & {\bf 0.892} &0.0704 &27.74 &0.847 &0.0928 &29.01 &0.858 &0.1080 &26.18 &0.806 &0.1040\\
				USB-NeRF-cubic & 31.90 & 0.889 &{\bf 0.0701} &{\bf 31.85} &{\bf 0.909} &{\bf 0.0573} &{\bf 34.89} &{\bf 0.939} &{\bf 0.0415} &{\bf 30.57} &{\bf 0.892} &{\bf 0.0576}\\
				\bottomrule
			\end{tabular}
		}
		\vspace{-2em}
	\end{center}
\end{table}
\begin{table}[t]
	\vspace{-2.0em}
	\begin{center}
		\captionsetup {font={small,stretch=0.5}}
		\caption{{\textbf{Quantitative comparisons on the synthetic datasets in terms of rolling shutter effect removal.}} Experimental results demonstrate that our method performs similarly to other learning-based methods on the Carla-RS dataset, on which those networks have been properly trained. However, our method performs much better on the Unreal-RS dataset, due to the poor generalization performance of other methods. USB-NeRF also performs better than the original NeRF and BARF, since they did not model the rolling shutter effect in their formulation.} 
		\label{table_our_synthesis}
		\vspace{-0.5em}
		\setlength\tabcolsep{2.7pt}
		\setlength{\belowcaptionskip}{-12pt}
		%    \normalsize
		\scriptsize
		\resizebox{\linewidth}{!}{
			\begin{tabular}{c|ccc|ccc|ccc|ccc}
				\toprule
				&  \multicolumn{3}{c}{Carla}  &  \multicolumn{3}{|c}{Blue Room}  &  \multicolumn{3}{|c}{Living Room}  &  \multicolumn{3}{|c}{White Room} \\
				& PSNR$\uparrow$ & SSIM$\uparrow$ & LPIPS$\downarrow$ & PSNR$\uparrow$ & SSIM$\uparrow$ & LPIPS$\downarrow$ & PSNR$\uparrow$ & SSIM$\uparrow$ & LPIPS$\downarrow$ & PSNR$\uparrow$ & SSIM$\uparrow$ & LPIPS$\downarrow$\\
				\midrule
				%			diffSfM &26.46 &0.807 &0.0703 &23.27 &0.497 &0.496 &26.07728411	&0.600554526 &0.600554526 &23.13402044 &0.469696939 &0.469696939\\
				
				%			DSUN \citep{liu2020deepunroll} &26.46 &0.807 &0.0703 &20.89 &0.660 &0.1154 &22.58 &0.742 &0.1138 &19.64 &0.618 &0.1500\\
				%			SUNet \citep{fan2021sunet} &29.18 &0.850 &0.0658 &16.58 &0.513 &0.2498 &18.64 &0.618 &0.2424 &15.84 &0.463 &0.3679\\
				%			RSSR \citep{fan2021RSSR} &24.78 &0.867 &0.0695 &22.20 &0.748 &0.0751 &22.09 &0.779 &0.0835 &19.94 &0.713 &0.0983\\
				%			CVR \citep{fan2022CVR} &31.74 &{\bf 0.929} &{\bf 0.0368} &24.05 &0.809 &0.0598 &23.48 &0.825 &0.0672 &20.89 &0.732 &0.0782\\
				
				DiffSfM\citep{zhuang2017sfm_RS} &24.20 &0.775 &0.1322 &17.10 &0.497 &0.2073 &18.68 &0.601 &0.2236 &14.94 &0.469 &0.2218 \\
				%            DiffSfM\citep{zhuang2017sfm_RS} &24.20 &0.775 &0.1322 &23.27 &0.497 &0.2073 &26.08 &0.601 &0.2236 &23.13 &0.470 &0.2218\\
				DSUN \citep{liu2020deepunroll} &26.46 &0.807 &0.0703 &21.25 &0.682 &0.1746 &23.22 &0.762 &0.1507 &20.45 &0.643 &0.1852\\
				SUNet \citep{fan2021sunet} &29.18 &0.850 &0.0658 &23.32 &0.721 &0.1513 &26.64 &0.7957 &0.1167 &22.37 &0.6866 &0.1422\\
				RSSR \citep{fan2021RSSR} &24.78 &0.867 &0.0695 &22.15 &0.731 &0.1369 &22.14 &0.770 &0.1258 &19.19 &0.697 &0.1392\\
				CVR \citep{fan2022CVR} &31.74 &{\bf 0.929} &{\bf 0.0368} &23.25 &0.745 &0.1268 &23.14 &0.785 &0.1141 &20.85 &0.715 &0.1212\\
				NeRF \citep{mildenhall2020nerf} &20.85 &0.620 &0.1734 &19.12 &0.569 &0.3289 &21.44 &0.682 &0.3495 &18.29 &0.509 &0.4104\\
				BARF \citep{lin2021barf} &20.95 &0.664 &0.1845 &19.14 &0.576 &0.3446 &21.48 &0.690 &0.3341 &18.43 &0.545 &0.3837\\
				%			USB-NeRF (ours) 
				%			&{\bf 32.15} &0.892 &0.0704 &27.74 &0.847 &0.0928 &29.01 &0.858 &0.1080 &26.18 &0.806 &0.1040 &25.04 &0.763 &0.1174\\
				\specialrule{0.08em}{1pt}{1pt}
				USB-NeRF (ours) &{\bf 31.90} &0.889 &0.0701 &{\bf 31.85} &{\bf 0.909} &{\bf 0.0573} &{\bf 34.89} &{\bf 0.939} &{\bf 0.0415} &{\bf 30.57} &{\bf 0.892} &{\bf 0.0576}\\
				%			USB-NeRF** &- &- &- &33.53 &0.936 &0.0336 &37.92 &0.970 &0.0139 &32.90 &0.940 &0.0266 &27.52 &0.845 &0.0613\\
				\specialrule{0.08em}{1pt}{1pt}
			\end{tabular}
		}
		\vspace{-0.85em}
	\end{center}
\end{table}
\begin{table}[t]
	\begin{center}
		\vspace{-1em}
		\captionsetup {font={small,stretch=0.5}}
		\caption{{\textbf{Quantitative comparisons on both synthetic and real datasets in terms of the accuracy of trajectory estimation for translation error (m).}} The experimental results demonstrate that rolling shutter distortions affect the accuracy of motion trajectory estimations. Due to proper modeling, our method performs much better than state-of-the-art methods. It also demonstrates that cubic B-Spline interpolation is superior to linear interpolation. The ATE metrics of the TUM-RS \citep{schubert2019RS-VIO} dataset are averaged over 10 sequences and details of every sequence are presented in the Appendix \ref{sec:appendix_experiment} (\tabnref{TUM_ATE}). x denotes method failed on the corresponding sequence.}
		\label{Syn_ATE}
		\vspace{-0.5em}
		\scriptsize
		\setlength{\tabcolsep}{2.5mm}
		\setlength{\belowcaptionskip}{-10pt}
		\resizebox{\linewidth}{!}{
			\begin{tabular}{c|c c c c c c}
				\toprule
				\makebox[0.08\textwidth][c]{} & \makebox[0.08\textwidth][c]{\scriptsize COLMAP} & \makebox[0.08\textwidth][c]{\scriptsize BARF} & \makebox[0.08\textwidth][c]{\scriptsize RSBA} & \makebox[0.08\textwidth][c]{\scriptsize NW-RSBA} & \makebox[0.08\textwidth][c]{\scriptsize USB-NeRF-linear} &\makebox[0.08\textwidth][c]{\scriptsize USB-NeRF-cubic} \\
				%			&{\scriptsize \qquad COLMAP \qquad} &{\scriptsize BARF} &{\scriptsize \qquad RSBA \qquad} & {\scriptsize NW-RSBA} & {\scriptsize \qquad USB-NeRF-linear \qquad} & {\scriptsize \qquad USB-NeRF-cubic \qquad} \\
				\midrule
				Carla 		&{.1931}$\pm${.1090} 	&{.2245}$\pm${.1293} 	&{.1923}$\pm${.0959} 	&{.2720}$\pm${.1404} &{.0570}$\pm${.0335} &{\bf.0530}$\pm${.0342} \\
				Blue Room 	&{.1593}$\pm${.0949} 	&{.1446}$\pm${.0819} 	&{.0640}$\pm${.0308} 	& x 					&.0062$\pm${.0035} &{\bf.0013}$\pm${.0013}\\
				Living Room &{.0967}$\pm${.0422} 	&{.0919}$\pm${.0400} 	&{.1010}$\pm${.0400} 	& x 					&.0144$\pm${.0089} &{\bf.0035}$\pm${.0025}\\
				White Room 	&{.1097}$\pm${.0422} 	&{.1191}$\pm${.0521} 	&{.1210}$\pm${.0410} 	& x 					&{.0115}$\pm${.0067} &{\bf.0044}$\pm${.0033}\\
				Adornment 	&{.3269}$\pm${.1952} 	&{.3918}$\pm${.2189} 	&x  					&x  					&{.0536}$\pm${.0834} &{\bf.0162}$\pm${.0122}\\
				Factory 	&{.2443}$\pm${.1003} 	&{.2149}$\pm${.1326} 	&x 						&x 						&{.0123}$\pm${.0076} &{\bf .0072}$\pm${.0052}\\
				Tanabata 	&{.1397}$\pm${.0745} 	&{.1957}$\pm${.1020} 	&x 						&x 						&{.0154}$\pm${.0085} &{\bf.0130}$\pm${.0077}\\
				\midrule
				TUM-RS &{.0486}$\pm${.0228} &{.0873}$\pm${.0391} &{.0688}$\pm${.0322} &{.1374}$\pm${.0664} &{.0150}$\pm${.0104} &{\bf.0136}$\pm${.0090}\\
				\bottomrule
			\end{tabular}
		}
		\vspace{-3.5em}
	\end{center}
\end{table}

\PAR{Quantitative evaluation results.}
We evaluate the performance of our method against other state-of-the-art methods in terms of rolling shutter effect removal and the accuracy of trajectory estimation, with both synthetic and real datasets.
\tabnref{table_our_synthesis} presents the rolling shutter effect removal comparisons. It demonstrates that our method performs similarly to prior learning-based methods if those methods are trained on the respective dataset (\eg Carla-RS dataset). However, those learning-based methods exhibit poor generalization performance when no fine-tuning on new dataset is performed (\eg Unreal-RS dataset), while our method delivers good performance consistently, as our method does not rely on pre-training. Our method also performs better than learning-free method DiffSfM \citep{zhuang2017sfm_RS} for all datasets. Additional experimental results in terms of rolling shutter effect removal on other synthetic sequences are presented in Appendix \ref{sec:appendix_experiment} (\tabnref{table_our_synthesis_longer}).

The results shown in \tabnref{table_our_synthesis} also reveal that original NeRF \citep{mildenhall2020nerf} and BARF \citep{lin2021barf} cannot deliver satisfying results if the rolling shutter image formation process is not explicitly considered. Even BARF \citep{lin2021barf} also optimizes the camera poses to eliminate the effect of inaccurate poses, it still cannot learn the correct underlying 3D representations, which delivers poor recovered global shutter images. 

\tabnref{Syn_ATE} presents the camera motion trajectory estimation results with both synthetic and real datasets. The results demonstrate that both COLMAP \citep{schonberger2016structure} and BARF \citep{lin2021barf} suffer from the rolling shutter effect. The introduced distortions would affect camera pose estimations if they are not properly handled. On the contrary, our method does not have such limitations, since we formulate the physical image formation process of RS camera into the training of NeRF. Our method also performs better than SOTA RS-aware bundle adjustment methods (i.e. RSBA \citep{hedborg2012RSBA}, NW-RSBA \citep{liao2023NW_RSBA}) in terms of the ATE metric. The results also reveal that cubic B-Spline interpolation performs better than linear interpolation for both synthetic and real datasets, in terms of the accuracy of the recovered motion trajectories. Due to the limited space, we present detailed quantitative experimental results about trajectory estimation with real datasets in Appendix \ref{sec:appendix_experiment} (\tabnref{TUM_ATE}).

More quantitative evaluation results (e.g. on novel view image synthesis, with un-ordered image sequences) are presented in Appendix \ref{sec:appendix_experiment} (\tabnref{table_novel_view} and \tabnref{table_unorganized}). They also demonstrate the better performance of our method over prior state-of-the-art methods.

%The visualization of the estimated trajectories can be found in the appendix \ref{sec:trajectory_visualization} due to limited space.

%\begin{table}
%	\setlength\tabcolsep{2pt}
%	\setlength{\belowcaptionskip}{-15pt}
%	\resizebox{\linewidth}{!}{
%		\begin{tabular}{c|c c c c c}
%			\toprule
%			Seq. &  COLMAP \citep{schonberger2016structure} & BARF \citep{lin2021barf} &NW-RSBA &USB-NeRF-linear & USB-NeRF-cubic \\
%			\midrule
%			Carla &{.1931}$\pm${.1090} &{.2245}$\pm${.1293} &{.2720}$\pm${.1404} &{.0570}$\pm${.0335} &{\bf.0530}$\pm${.0342} \\
%			Blue Room &{.1593}$\pm${.0949} &{.1446}$\pm${.0819} &x &.0062$\pm${.0035} &{\bf.0013}$\pm${.0013}\\
%			Living Room &{.0967}$\pm${.0422} &{.0919}$\pm${.0400} &x &.0144$\pm${.0089} &{\bf.0035}$\pm${.0025}\\
%			White Room &{.1097}$\pm${.0422} &{.1191}$\pm${.0521} &x &{.0115}$\pm${.0067} &{\bf.0044}$\pm${.0033}\\
%			\bottomrule
%	\end{tabular}
%    }
%	\vspace{-0.5em}
%	\caption{{\textbf{Quantitative comparisons on synthetic datasets in terms of the accuracy of trajectory estimation.}} The experimental results demonstrate that rolling shutter distortions affect the accuracy of motion trajectory estimations. Due to proper modeling, our method performs much better than state-of-the-art methods. It also demonstrates that cubic B-Spline interpolation is superior to linear interpolation.} 
%	\label{Syn_ATE}
%\end{table}

%
%
\begin{figure}[!ht]
	\vspace{-2em}
	\setlength\tabcolsep{1.pt}
	\centering
	\small
	\begin{tabular}{ccccccc}
		\raisebox{.7in}{\rotatebox[origin=t]{90}{\normalsize Carla-RS \citep{liu2020deepunroll}}}
		&\includegraphics[width=0.15\textwidth]{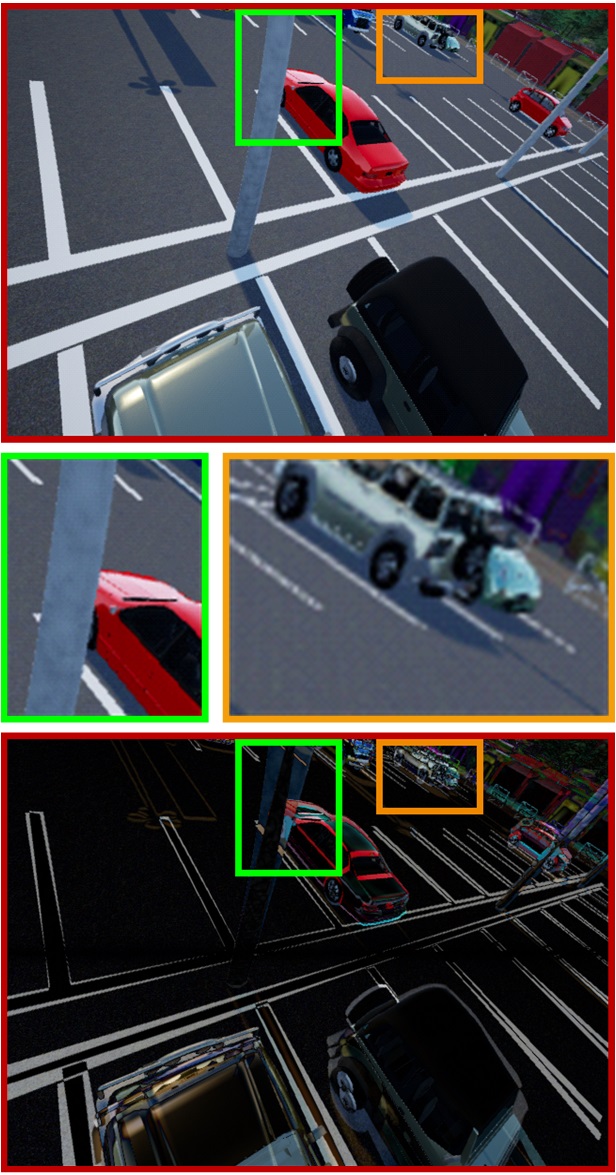} &
		\includegraphics[width=0.15\textwidth]{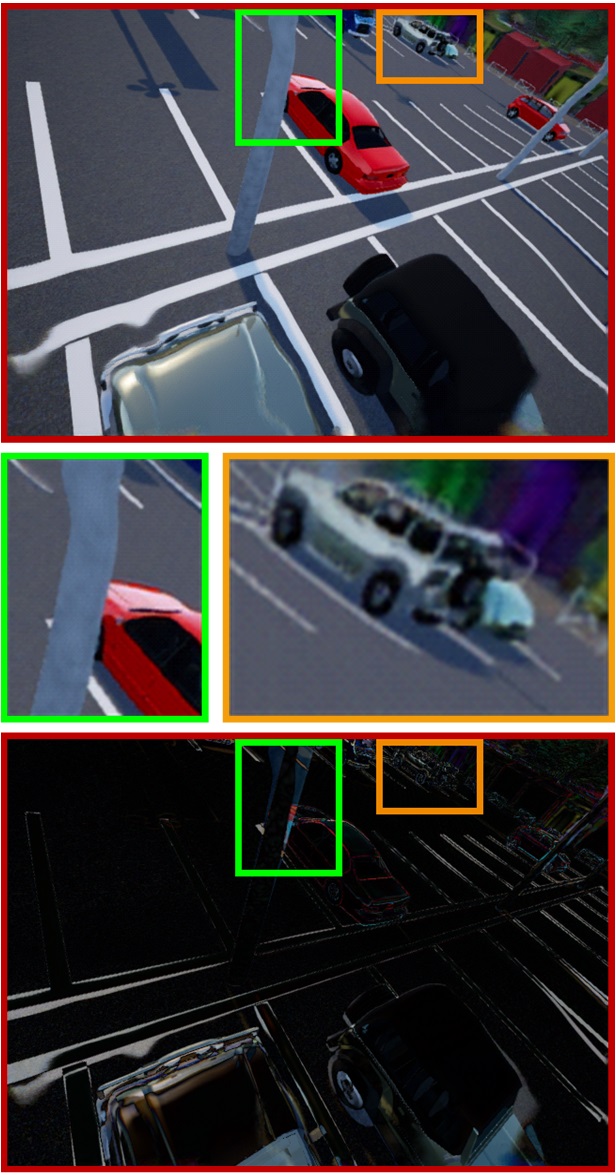} &
		\includegraphics[width=0.15\textwidth]{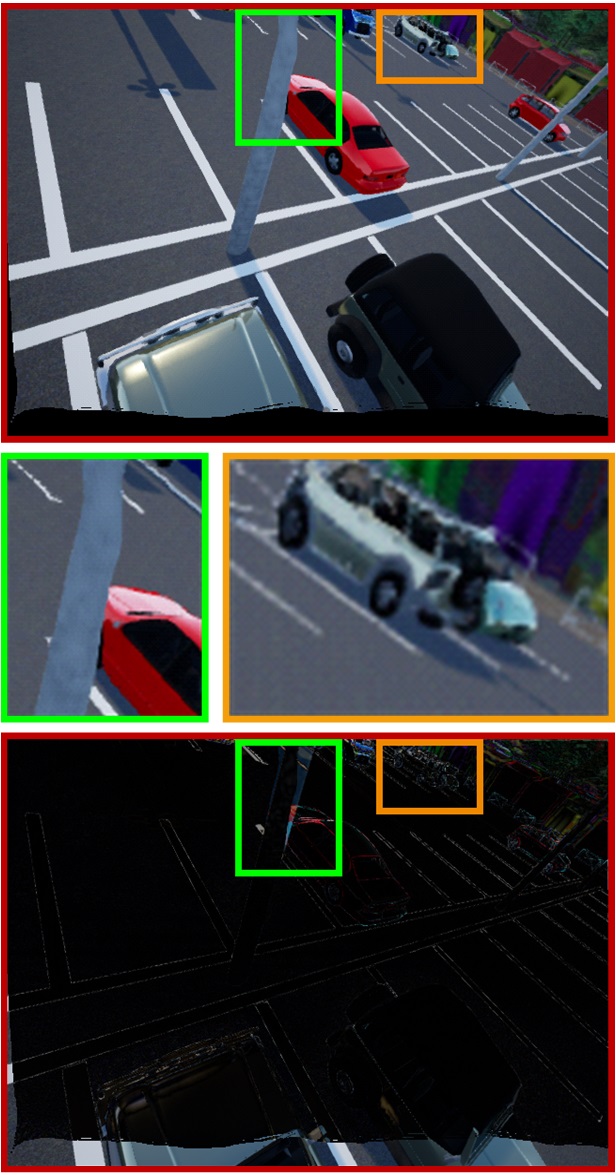} &
		\includegraphics[width=0.15\textwidth]{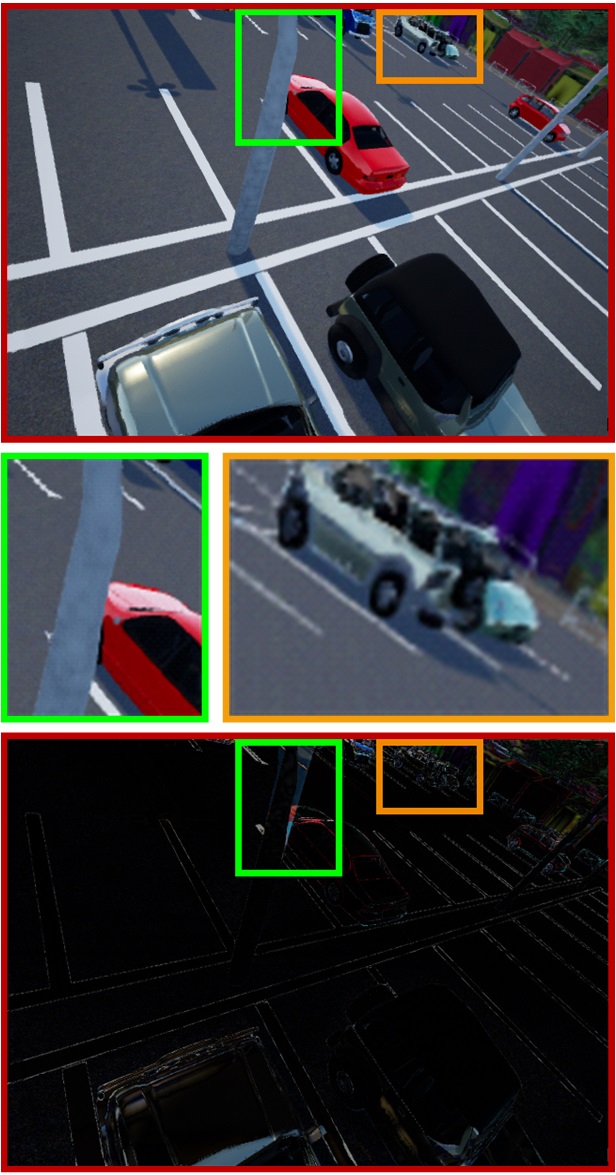} &
		\includegraphics[width=0.15\textwidth]{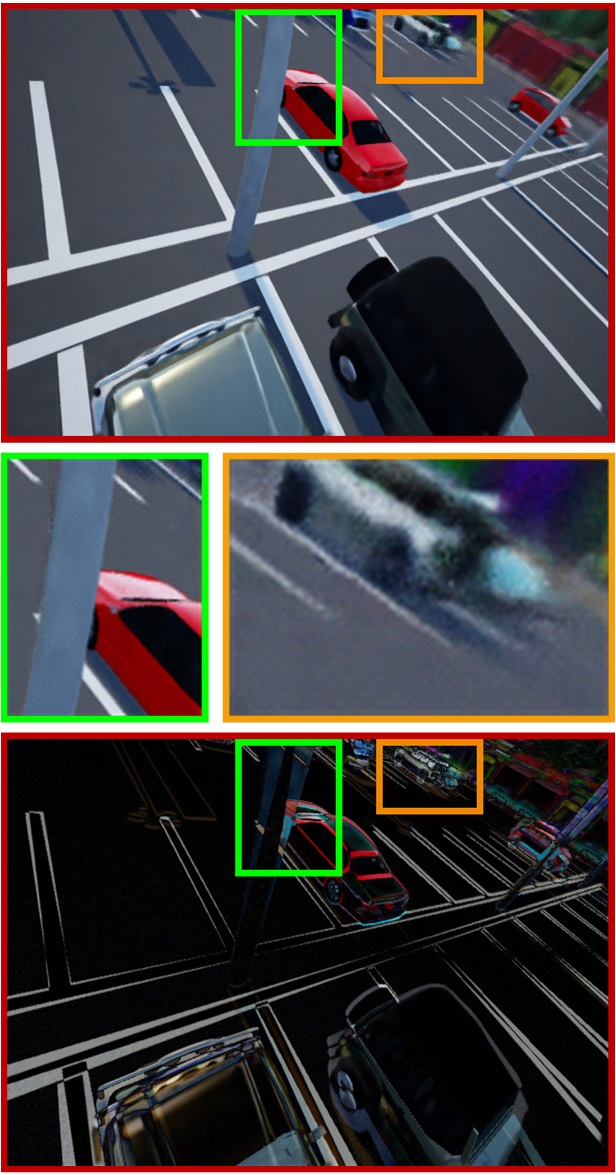} & 
		\includegraphics[width=0.15\textwidth]{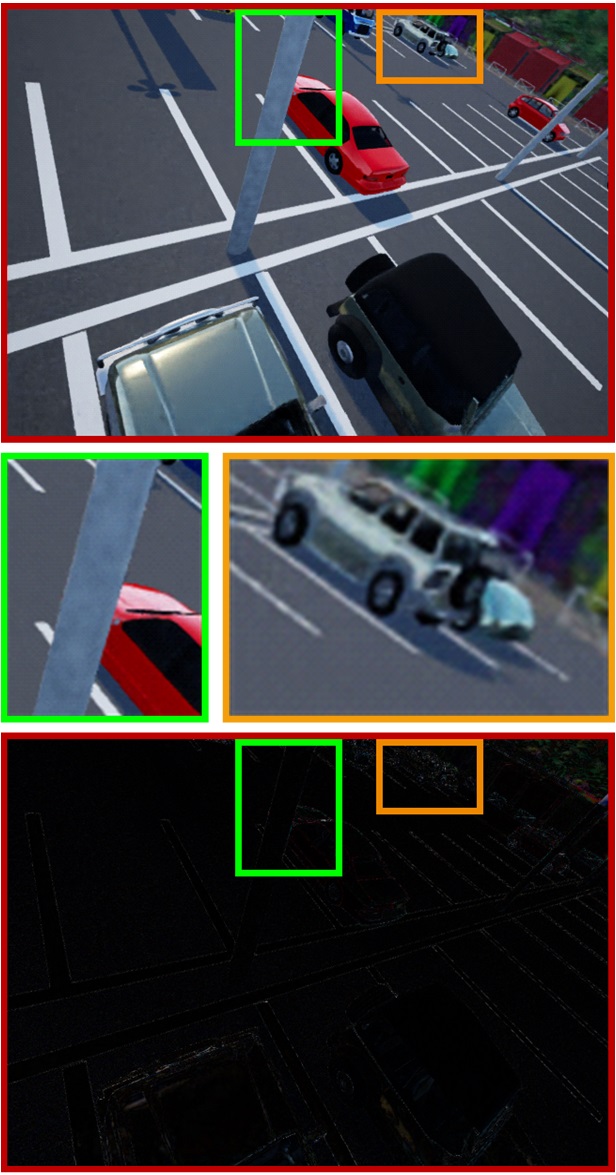} \\
		\specialrule{-0.25em}{0.05pt}{0.05pt}
		\raisebox{.7in}{\rotatebox[origin=t]{90}{\normalsize Blue Room}} &\includegraphics[width=0.15\textwidth]{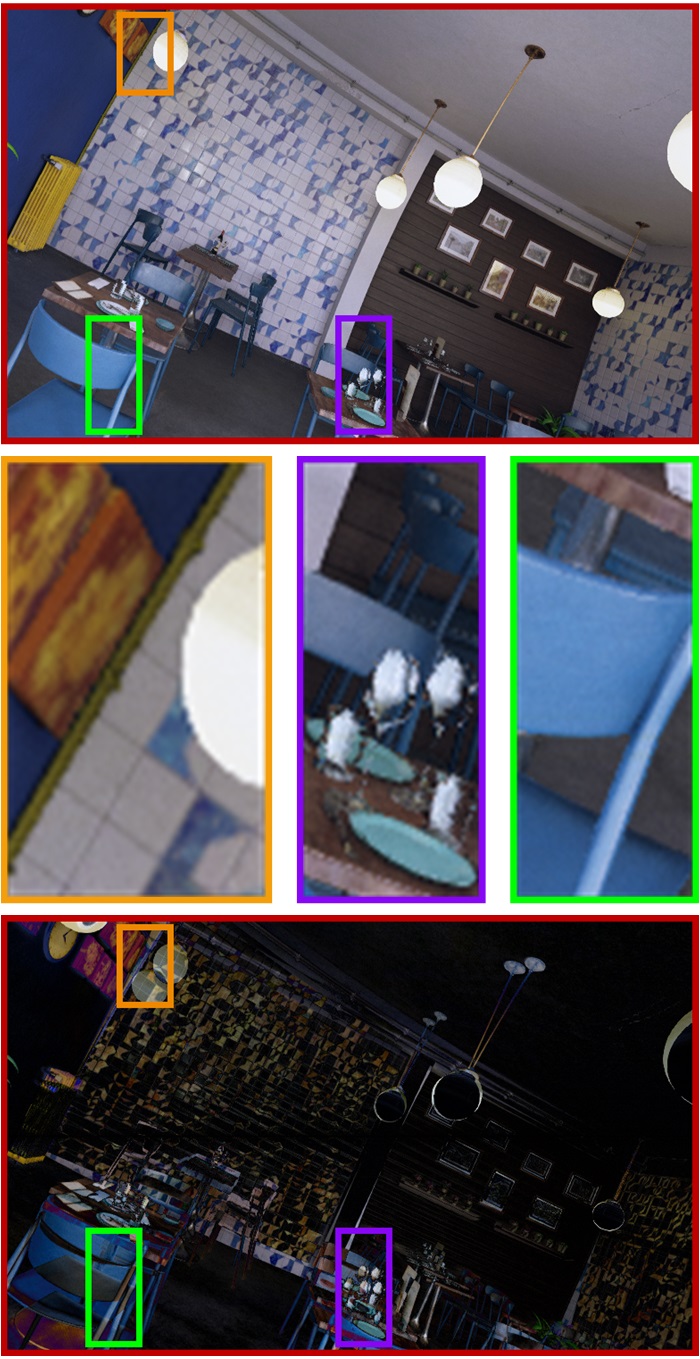} &
		\includegraphics[width=0.15\textwidth]{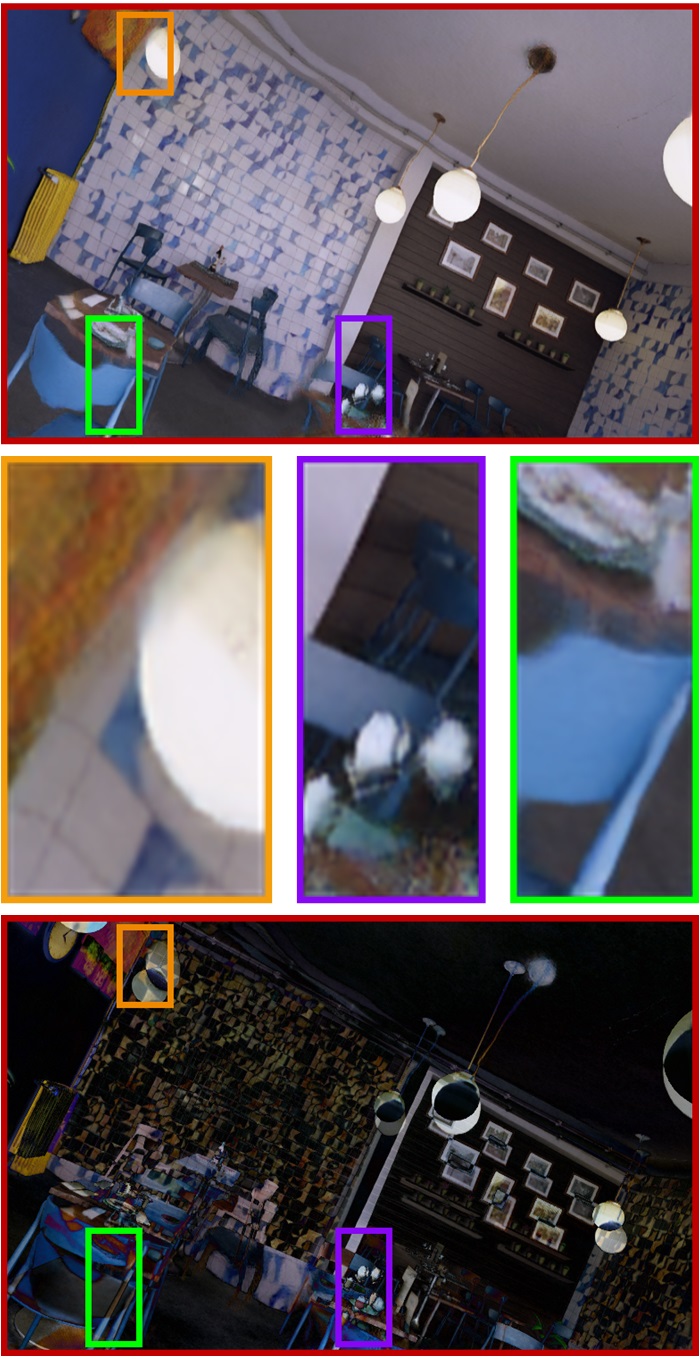} &
		\includegraphics[width=0.15\textwidth]{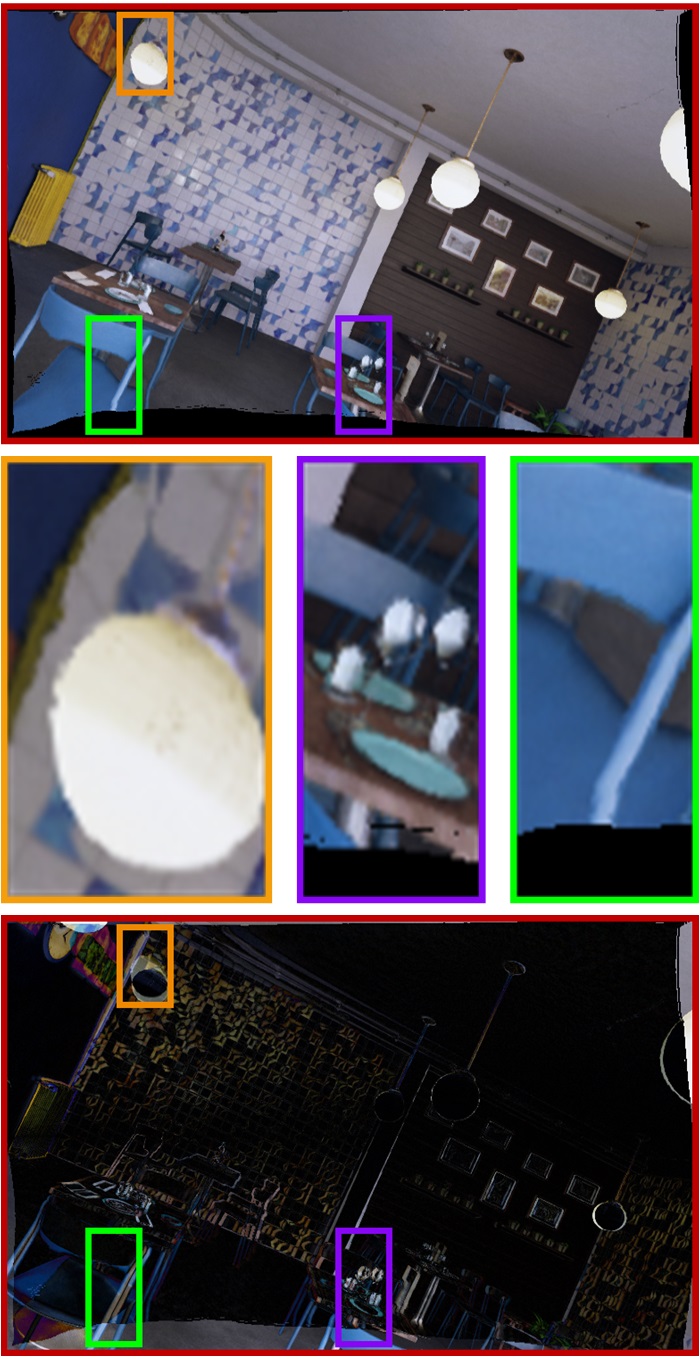} &
		\includegraphics[width=0.15\textwidth]{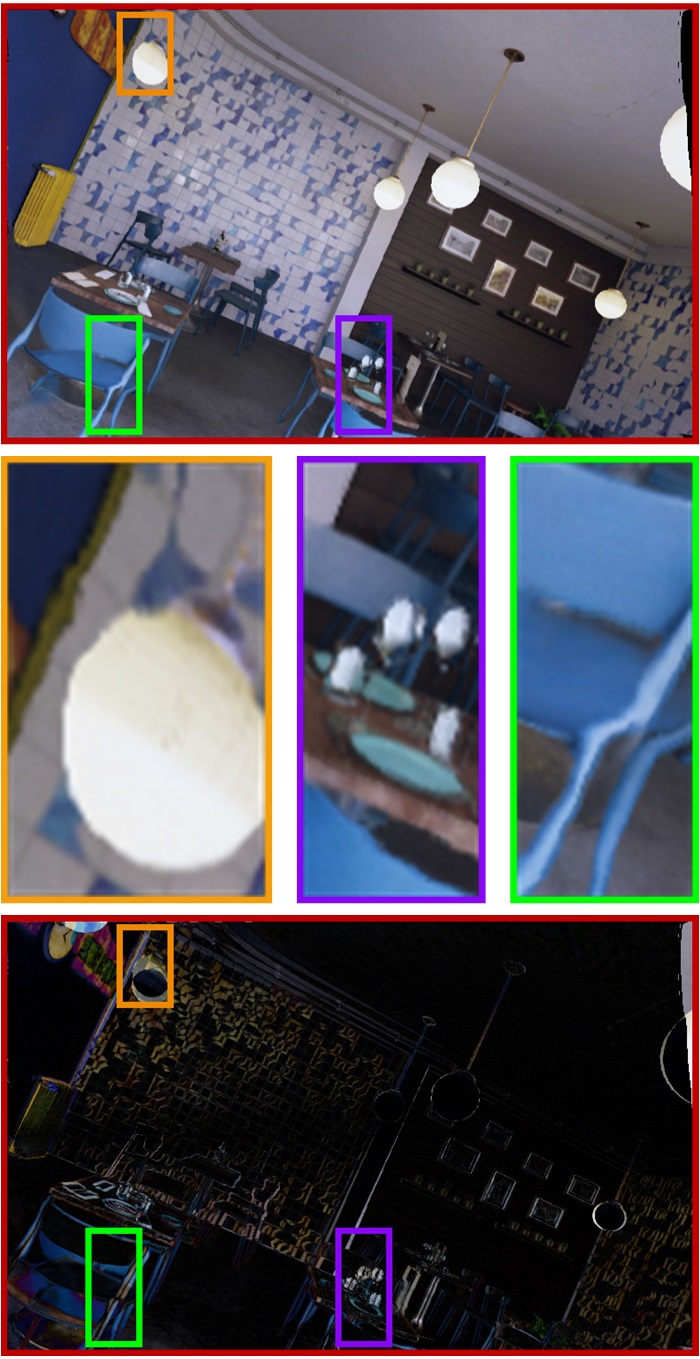} &
		\includegraphics[width=0.15\textwidth]{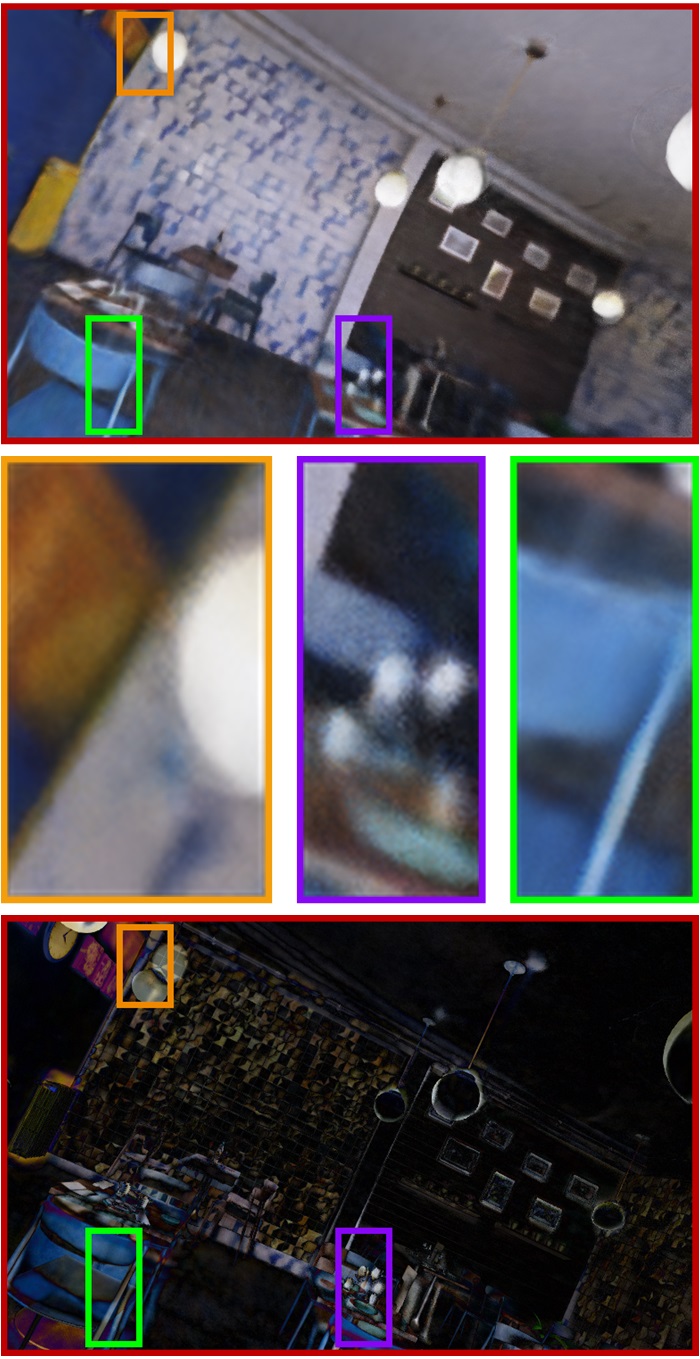} &
		\includegraphics[width=0.15\textwidth]{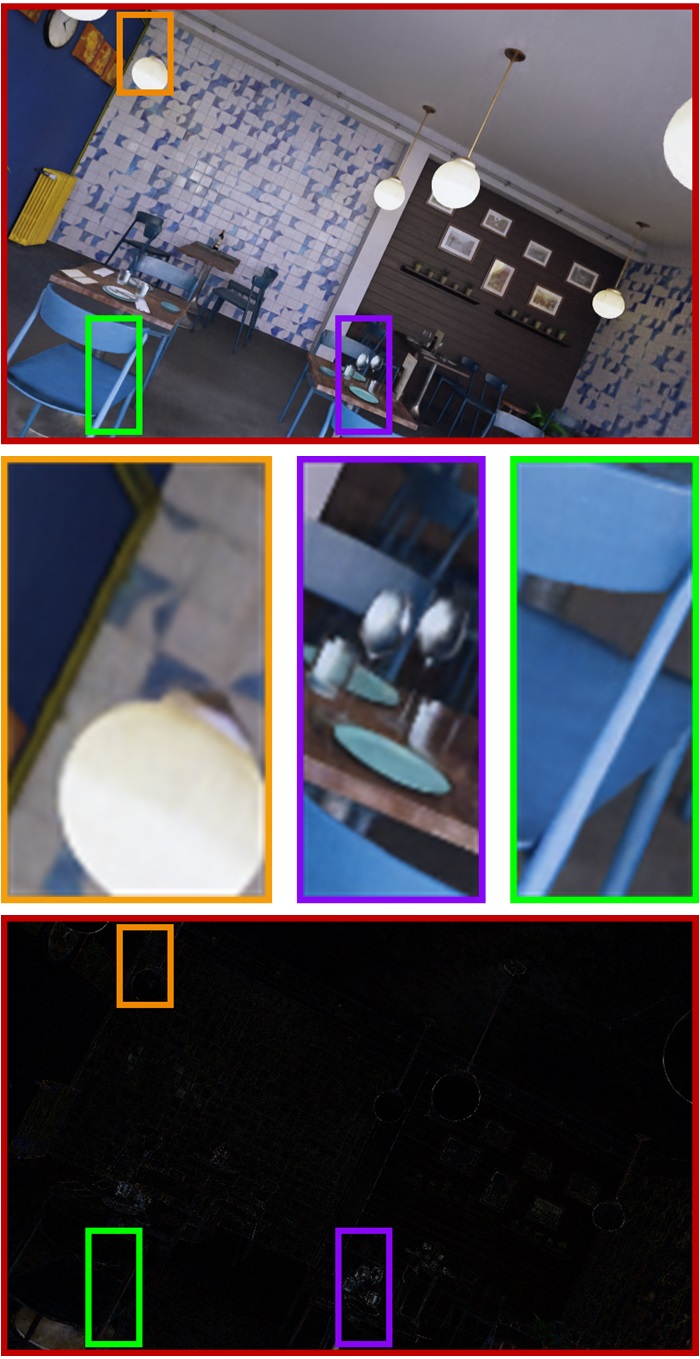} \\
		\specialrule{0em}{0.05pt}{0.05pt}
%		& Rolling shutter & DSUN \citep{liu2020deepunroll} & RSSR \citep{fan2021RSSR} & CVR \citep{fan2022CVR} & BARF \citep{lin2021barf} & USB-NeRF
		& Rolling shutter & DSUN & RSSR & CVR & BARF & USB-NeRF
	\end{tabular}
    \vspace{-0.5em}
    \captionsetup {font={small,stretch=0.5}}
    \caption{{\bf{Qualitative comparisons with Carla-RS datasets \citep{liu2020deepunroll} and Unreal-RS datasets.}} The experimental results demonstrate that our method achieves better performance compared to prior works. The darker the $3^{rd}$ and the $6^{th}$ rows, the performance is better.}
	\label{fig_carla}
	\vspace{-0.em}
\end{figure}

\begin{figure}[!ht]
	\vspace{-0.5em}
	\setlength{\belowcaptionskip}{-6pt}
	\setlength\tabcolsep{1pt}
	\centering
    \small
	\begin{tabular}{ccccccc}
        \raisebox{.9in}{\rotatebox[origin=t]{90}{\,}}
		\includegraphics[width=0.15\textwidth]{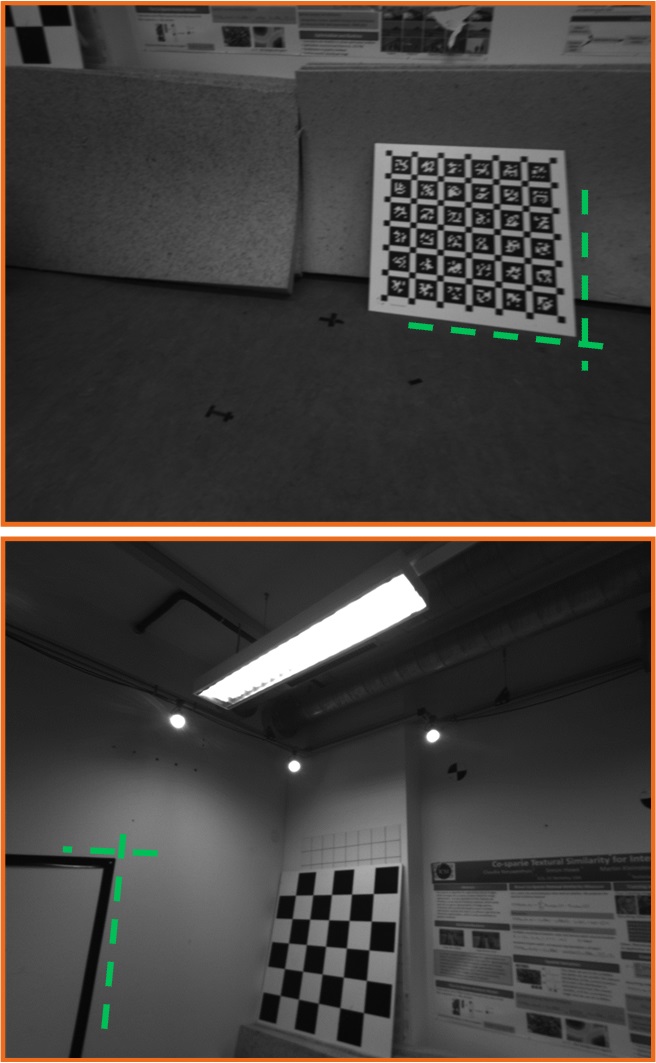} &
		\includegraphics[width=0.15\textwidth]{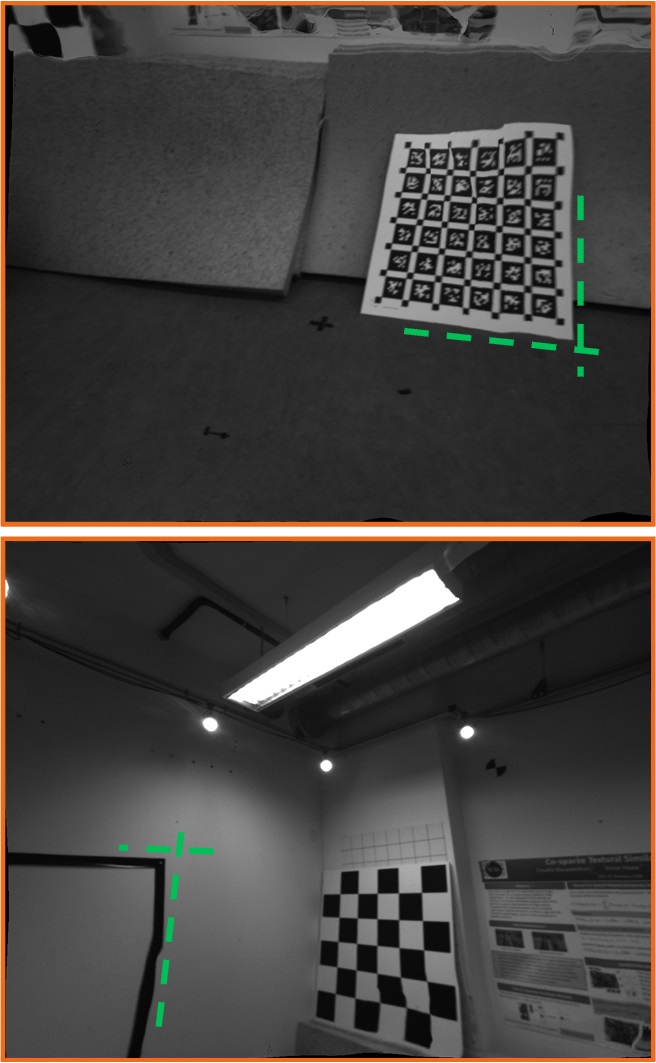} &
		\includegraphics[width=0.15\textwidth]{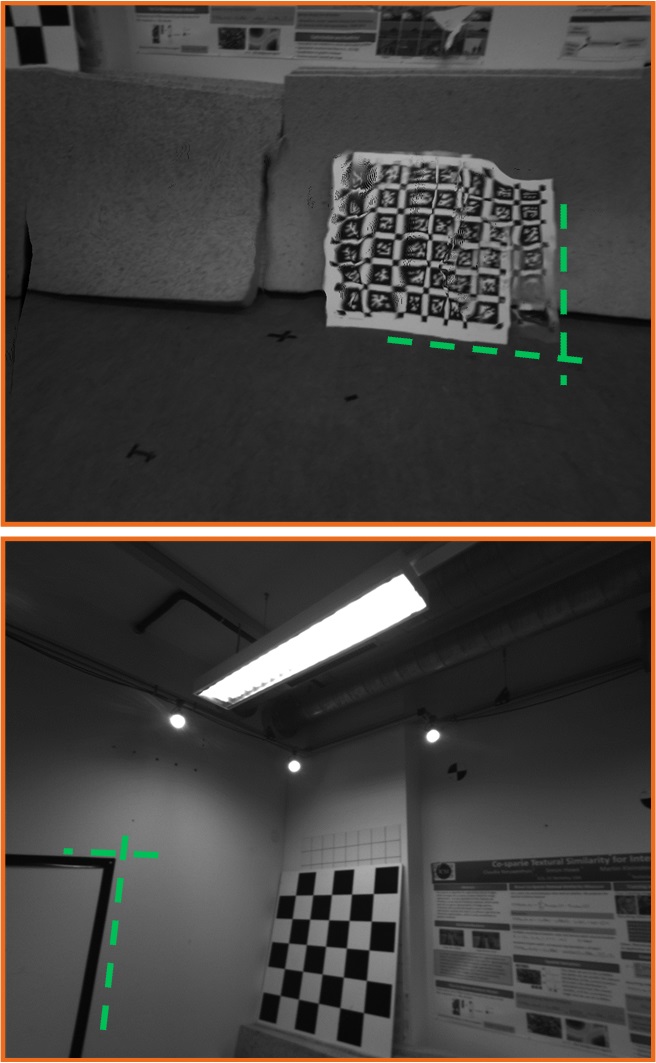} &
		\includegraphics[width=0.15\textwidth]{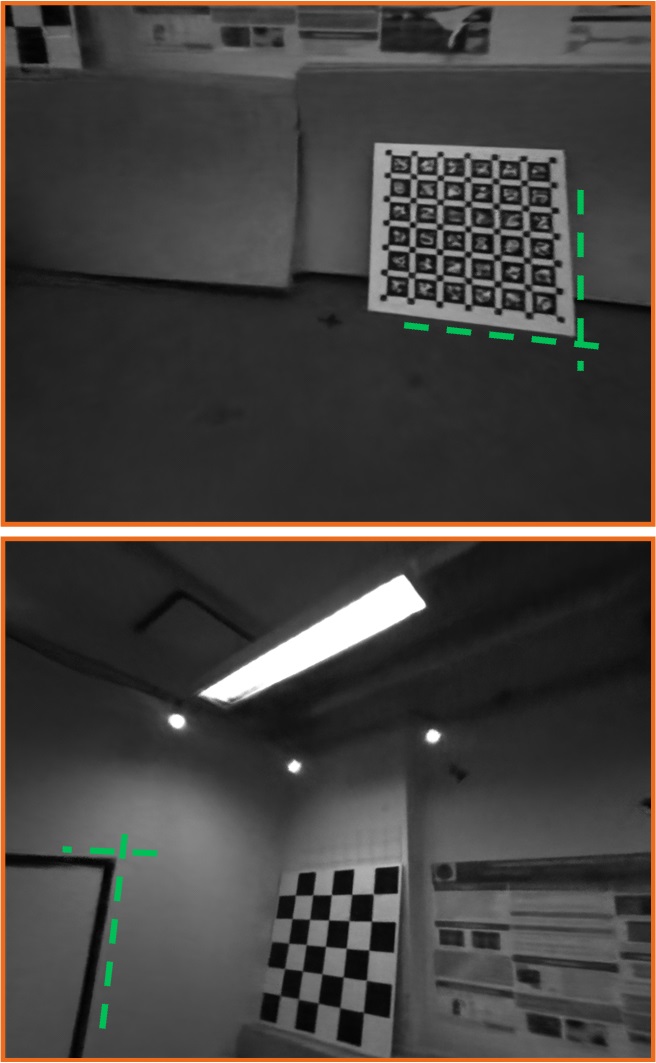} &
		\includegraphics[width=0.15\textwidth]{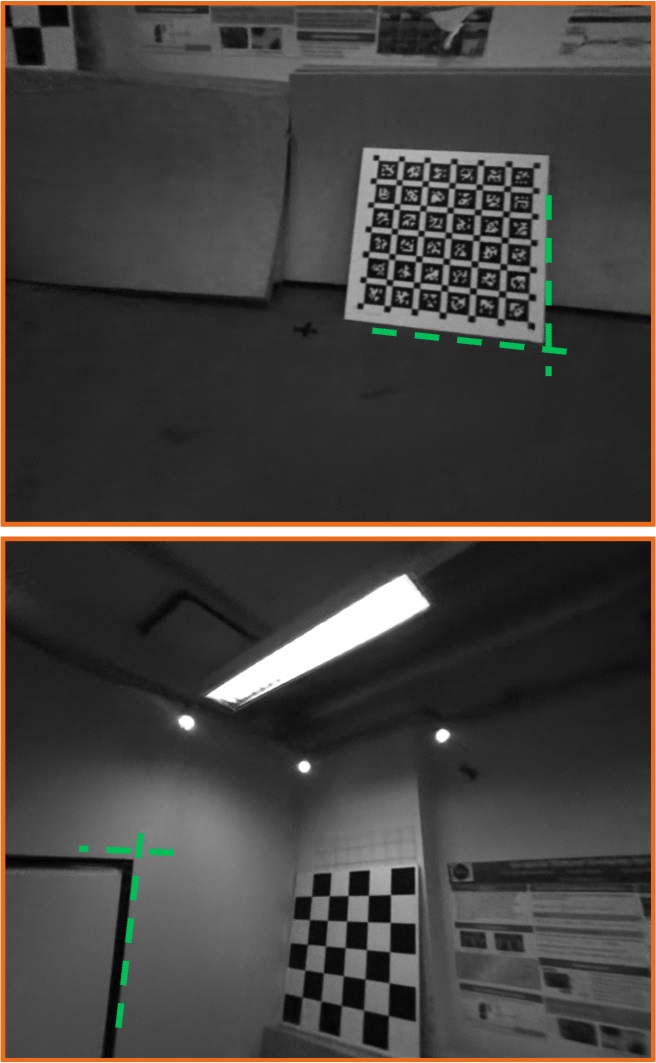} &
		\includegraphics[width=0.15\textwidth]{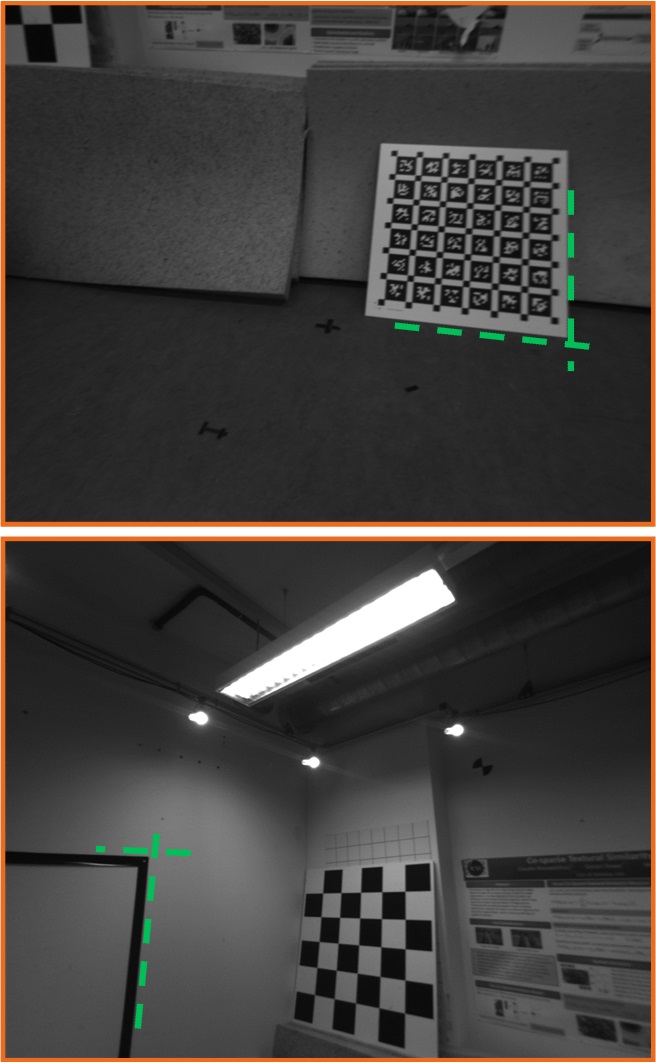} \\
		\specialrule{0em}{0.05pt}{0.05pt}
%		Rolling shutter & RSSR \citep{fan2021RSSR} & CVR \citep{fan2022CVR} & BARF \citep{lin2021barf} & USB-NeRF & Global shutter
		Rolling shutter & RSSR & CVR & BARF & USB-NeRF & Global shutter
	\end{tabular}
    \vspace{-0.5em}
    \captionsetup {font={small,stretch=0.5}}
	\caption{{\bf{Qualitative comparisons with real TUM-RS datasets \citep{schubert2019RS-VIO}.}} Since the dataset does not have pixel-aligned rolling-global shutter image pairs, we choose the nearest neighbor global shutter images for comparisons. The experimental results demonstrate that RSSR and CVR fail to correct the RS effect due to their poor generalization performance. BARF also fails since it does not consider the rolling shutter camera model, while our method successfully removes the RS effect.}
	\label{fig_TUM}
	\vspace{-1em}
\end{figure}

\PAR{Qualitative evaluation results.} We also evaluate the qualitative performance of our method against the other baseline methods. \figrefer{fig_carla} presents the comparisons for both Carla-RS and Unreal-RS datasets. \figrefer{fig_TUM} presents the results with the TUM-RS dataset \citep{schubert2019RS-VIO}. Since the TUM-RS dataset does not provide pixel-aligned rolling-global shutter image pairs, we choose the nearest neighbor global shutter image (captured by another global shutter camera) for comparison. 

The experimental results demonstrate that our method can better exploit multi-view information for rolling shutter effect removal, compared to DSUN \citep{liu2020deepunroll}, RSSR \citep{fan2021RSSR} and CVR \citep{fan2022CVR} in \figrefer{fig_carla} (Carla-RS), even they are properly trained on the corresponding dataset. \figrefer{fig_carla} (Blue Room) and \figrefer{fig_TUM} demonstrate that both RSSR \citep{fan2021RSSR} and CVR \citep{fan2022CVR} have a poor generalization performance if they are not fine-tuned on the respective datasets, which is common for practical applications. On the contrary, our method does not have such limitation and performs better than those learning-based RS effect removal methods consistently. The results also reveal that BARF \citep{lin2021barf} fails to learn the underlying undistorted 3D scene representation even though it optimizes the camera poses. It proves the necessity to properly model the physical image formation process of RS camera into the training of NeRF for better 3D scene reconstruction. 
More qualitative results can also be found in Appendix \ref{sec:appendix_experiment} (e.g. novel view image synthesis, rolling shutter effect removal, trajectory estimation etc.) and supplementary video. They also demonstrate the superior performance of our method over prior works.

\section{Conclusion}

In this paper, we presented unrolling shutter bundle-adjusted neural radiance fields. The method takes advantage of the powerful representation ability of NeRF and a continuous-time trajectory representation with cubic B-Spline. Given a sequence of rolling shutter images, our method successfully learns the true underlying 3D representations and recovers the motion trajectory accurately. Experimental results demonstrate the superior performance of our method against prior state-of-the-art works, in terms of camera motion estimation, rolling shutter effect removal and novel view image synthesis \etc.

% High framerate video extraction?

%In this paper, we present a NeRF-based unrolling shutter network with bundle adjustment. USB-NeRF employ the inherent property of NeRF representation to help recover more accurate GS images with the help of geometric constraint. Given images with RS distortions and inaccurate camera poses, our approach could learn accurate scene representation and obtain precise motion trajectory estimation simultaneously. Our proposed framework also shows the possibilities to exploit the time-relevant information embedded in RS images to retrieve a sequence of high-fidelity GS images.
%
%Now our method could only handle with static scenes, we hope to expand USB-NeRF to RS effect removal of dynamic scenes, which will be more meaningful for real world. Besides, our method relies on the initialization with poses estimated through COLMAP \cite{schonberger2016structure}, we plan to overcome this limitation in the future.
\subsubsection*{Acknowledgments}
This work was supported in part by NSFC under Grant 62202389, in part by a grant from the Westlake University-Muyuan Joint Research Institute, and
in part by the Westlake Education Foundation.

\bibliography{egbib,bibography_peidong}

\begin{thebibliography}{117}
\providecommand{\natexlab}[1]{#1}
\providecommand{\url}[1]{\texttt{#1}}
\expandafter\ifx\csname urlstyle\endcsname\relax
  \providecommand{\doi}[1]{doi: #1}\else
  \providecommand{\doi}{doi: \begingroup \urlstyle{rm}\Url}\fi

\bibitem[Albl et~al.(2016)Albl, Sugimoto, and Pajdla]{albl2016degeneracies}
Cenek Albl, Akihiro Sugimoto, and Tomas Pajdla.
\newblock {Degeneracies in Rolling Shutter SfM}.
\newblock In \emph{European Conference on Computer Vision (ECCV)}, pp.\  36--51. Springer, 2016.

\bibitem[Albl et~al.(2020)Albl, Kukelova, Larsson, Polic, Pajdla, and Schindler]{albl2020dual-RS}
Cenek Albl, Zuzana Kukelova, Viktor Larsson, Michal Polic, Tomas Pajdla, and Konrad Schindler.
\newblock {From Two Rolling Shutters to One Global Shutter}.
\newblock In \emph{Computer Vision and Pattern Recognition (CVPR)}, pp.\  2505--2513, 2020.

\bibitem[Athar et~al.(2022)Athar, Xu, Research, Sunkavalli, Shechtman, and Shu]{Athar2022}
Shahrukh Athar, Zexiang Xu, Adobe Research, Kalyan Sunkavalli, Eli Shechtman, and Zhixin Shu.
\newblock {RigNeRF: Fully Controllable Neural 3D Portraits}.
\newblock In \emph{Computer Vision and Pattern Recognition (CVPR)}, pp.\  20364--20373, 2022.

\bibitem[Baker et~al.(2010)Baker, Bennett, Kang, and Szeliski]{Baker2011CVPR}
Simon Baker, Eric Bennett, Sing~Bing Kang, and Richard Szeliski.
\newblock Removing rolling shutter wobble.
\newblock In \emph{CVPR}, 2010.

\bibitem[Bi et~al.(2020)Bi, Xu, Srinivasan, Mildenhall, Sunkavalli, Ha{\v{s}}an, Hold-Geoffroy, Kriegman, and Ramamoorthi]{Bi2020}
Sai Bi, Zexiang Xu, Pratul Srinivasan, Ben Mildenhall, Kalyan Sunkavalli, Milo{\v{s}} Ha{\v{s}}an, Yannick Hold-Geoffroy, David Kriegman, and Ravi Ramamoorthi.
\newblock {Neural Reflectance Fields for Appearance Acquisition}.
\newblock \emph{arXiv preprint arXiv:2008.03824}, 2020.

\bibitem[Boss et~al.(2021)Boss, Braun, Jampani, Barron, Liu, and Lensch]{Boss2021}
Mark Boss, Raphael Braun, Varun Jampani, Jonathan~T. Barron, Ce~Liu, and Hendrik P.~A. Lensch.
\newblock {NeRD: Neural Reflectance Decomposition from Image Collections}.
\newblock In \emph{International Conference on Computer Vision (ICCV)}, 2021.
\newblock URL \url{http://arxiv.org/abs/2012.03918}.

\bibitem[Chen et~al.(2022{\natexlab{a}})Chen, Xu, Geiger, Yu, and Su]{chen2022tensorf}
Anpei Chen, Zexiang Xu, Andreas Geiger, Jingyi Yu, and Hao Su.
\newblock {TensoRF: Tensorial Radiance Fields}.
\newblock In \emph{European Conference on Computer Vision (ECCV)}, pp.\  333--350. Springer, 2022{\natexlab{a}}.

\bibitem[Chen et~al.(2022{\natexlab{b}})Chen, Chen, Wang, Zhang, Guo, Shan, and Wang]{chen2022local}
Yue Chen, Xingyu Chen, Xuan Wang, Qi~Zhang, Yu~Guo, Ying Shan, and Fei Wang.
\newblock {Local-to-Global Registration for Bundle-Adjusting Neural Radiance Fields}.
\newblock \emph{arXiv preprint arXiv:2211.11505}, 2022{\natexlab{b}}.

\bibitem[Cho \& Lee(2009)Cho and Lee]{Cho2009ToG}
Sunghyun Cho and Seungyong Lee.
\newblock {Fast Motion Deblurring}.
\newblock \emph{ACM Transactions on Graphics (ToG)}, 28\penalty0 (5), 2009.

\bibitem[Cornelis et~al.(2008)Cornelis, Leibe, Cornelis, and Van~Gool]{Cornelis2008IJCV}
N.~Cornelis, B.~Leibe, K.~Cornelis, and L.~J. Van~Gool.
\newblock {3D} urban scene modeling integrating recognition and reconstruction.
\newblock \emph{International Journal of Computer Vision (IJCV)}, 78\penalty0 (2-3):\penalty0 121--141, July 2008.

\bibitem[Delaunoy \& Pollefeys(2014)Delaunoy and Pollefeys]{Delaunoy2014CVPR}
Ama{\"{e}}l Delaunoy and Marc Pollefeys.
\newblock {Photometric Bundle Adjustment for Dense Multi-view 3D Modeling}.
\newblock In \emph{Computer Vision and Pattern Recognition (CVPR)}, 2014.

\bibitem[Deng et~al.(2022{\natexlab{a}})Deng, Liu, Zhu, and Ramanan]{deng2022depth}
Kangle Deng, Andrew Liu, Jun-Yan Zhu, and Deva Ramanan.
\newblock {Depth-supervised NeRF: Fewer Views and Faster Training for Free}.
\newblock In \emph{Computer Vision and Pattern Recognition (CVPR)}, pp.\  12882--12891, 2022{\natexlab{a}}.

\bibitem[Deng et~al.(2022{\natexlab{b}})Deng, Yang, Xiang, and Tong]{Deng2022}
Yu~Deng, Jiaolong Yang, Jianfeng Xiang, and Xin Tong.
\newblock {GRAM: Generative Radiance Manifolds for 3D-Aware Image Generation}.
\newblock In \emph{Computer Vision and Pattern Recognition (CVPR)}, volume~i, 2022{\natexlab{b}}.
\newblock URL \url{http://arxiv.org/abs/2112.08867}.

\bibitem[Fan \& Dai(2021)Fan and Dai]{fan2021RSSR}
Bin Fan and Yuchao Dai.
\newblock {Inverting a Rolling Shutter Camera: Bring Rolling Shutter Images to High Framerate Global Shutter Video}.
\newblock In \emph{International Conference on Computer Vision (ICCV)}, pp.\  4228--4237, 2021.

\bibitem[Fan et~al.(2021)Fan, Dai, and He]{fan2021sunet}
Bin Fan, Yuchao Dai, and Mingyi He.
\newblock {SUNet: Symmetric Undistortion Network for Rolling Shutter Correction}.
\newblock In \emph{International Conference on Computer Vision (ICCV)}, pp.\  4541--4550, 2021.

\bibitem[Fan et~al.(2022)Fan, Dai, Zhang, Liu, and He]{fan2022CVR}
Bin Fan, Yuchao Dai, Zhiyuan Zhang, Qi~Liu, and Mingyi He.
\newblock {Context-Aware Video Reconstruction for Rolling Shutter Cameras}.
\newblock In \emph{Computer Vision and Pattern Recognition (CVPR)}, pp.\  17572--17582, 2022.

\bibitem[Fergus et~al.(2006)Fergus, Singh, Hertzmann, Roweis, and Freeman]{Fergus2006SIGGGRAPH}
Rob Fergus, Barun Singh, Aaron Hertzmann, Sam~T. Roweis, and William~T. Freeman.
\newblock {Removing camera shake from a single photograph}.
\newblock In \emph{ACM Transactions on Graphics (ToG)}, 2006.

\bibitem[Forssen \& Ringaby(2010)Forssen and Ringaby]{Forssen2010CVPR}
Per~Erik Forssen and Erik Ringaby.
\newblock Rectifying rolling shutter video from hand-held devices.
\newblock In \emph{CVPR}, 2010.

\bibitem[Fridovich-Keil et~al.(2022)Fridovich-Keil, Yu, Tancik, Chen, Recht, and Kanazawa]{fridovich2022plenoxels}
Sara Fridovich-Keil, Alex Yu, Matthew Tancik, Qinhong Chen, Benjamin Recht, and Angjoo Kanazawa.
\newblock {Plenoxels: Radiance Fields without Neural Networks}.
\newblock In \emph{Computer Vision and Pattern Recognition (CVPR)}, pp.\  5501--5510, 2022.

\bibitem[Furukawa \& Ponce(2009)Furukawa and Ponce]{furukawa2009accurate}
Yasutaka Furukawa and Jean Ponce.
\newblock {Accurate, Dense, and Robust Multiview Stereopsis}.
\newblock \emph{IEEE Transactions on Pattern Analysis and Machine Intelligence (PAMI)}, 32\penalty0 (8):\penalty0 1362--1376, 2009.

\bibitem[Furukawa \& Ponce(2010)Furukawa and Ponce]{Furukawa2010PAMI}
Yasutaka Furukawa and Jean Ponce.
\newblock {Accurate, Dense, and Robust Multi-View Stereopsis}.
\newblock \emph{IEEE Transactions on Pattern Analysis and Machine Intelligence (PAMI)}, 32\penalty0 (8):\penalty0 1362--1376, 2010.

\bibitem[Gafni et~al.(2021)Gafni, Thies, Zollhofer, and Niesner]{Gafni2021}
Guy Gafni, Justus Thies, Michael Zollhofer, and Matthias Niesner.
\newblock {Dynamic Neural Radiance Fields for Monocular 4D Facial Avatar Reconstruction}.
\newblock In \emph{Computer Vision and Pattern Recognition (CVPR)}. IEEE Computer Society, 2021.

\bibitem[Gallup et~al.(2010)Gallup, Pollefeys, and Frahm]{Gallup2010DAGM}
David Gallup, Marc Pollefeys, and Jan-Michael Frahm.
\newblock {3D Reconstruction Using an n-Layer Heightmap}.
\newblock In \emph{DAGM German Conference on Pattern Recognition (DAGM GCPR)}, 2010.

\bibitem[Gao et~al.(2021)Gao, Saraf, Kopf, and Huang]{gao2021dynamic}
Chen Gao, Ayush Saraf, Johannes Kopf, and Jia-Bin Huang.
\newblock {Dynamic View Synthesis From Dynamic Monocular Video}.
\newblock In \emph{International Conference on Computer Vision (ICCV)}, pp.\  5712--5721, 2021.

\bibitem[Garbin et~al.(2021{\natexlab{a}})Garbin, Kowalski, Johnson, Shotton, and Valentin]{Garbin2021}
Stephan~J. Garbin, Marek Kowalski, Matthew Johnson, Jamie Shotton, and Julien Valentin.
\newblock {FastNeRF: High-Fidelity Neural Rendering at 200FPS}.
\newblock In \emph{International Conference on Computer Vision (ICCV)}, 2021{\natexlab{a}}.
\newblock URL \url{http://arxiv.org/abs/2103.10380}.

\bibitem[Garbin et~al.(2021{\natexlab{b}})Garbin, Kowalski, Johnson, Shotton, and Valentin]{garbin2021fastnerf}
Stephan~J Garbin, Marek Kowalski, Matthew Johnson, Jamie Shotton, and Julien Valentin.
\newblock {FastNeRF: High-Fidelity Neural Rendering at 200FPS}.
\newblock In \emph{International Conference on Computer Vision (ICCV)}, pp.\  14326--14335. IEEE, 2021{\natexlab{b}}.

\bibitem[Grundmann et~al.(2012)Grundmann, Kwatra, Castro, and Essa]{grundmann2012calibration}
Matthias Grundmann, Vivek Kwatra, Daniel Castro, and Irfan Essa.
\newblock {Calibration-Free Rolling Shutter Removal}.
\newblock In \emph{IEEE International Conference on Computational Photography (ICCP)}, pp.\  1--8. IEEE, 2012.

\bibitem[Gu et~al.(2022)Gu, Liu, Wang, and Theobalt]{Gu2022}
Jiatao Gu, Lingjie Liu, Peng Wang, and Christian Theobalt.
\newblock {StyleNeRF: A Style-Based 3D Aware Generator for High Resolution Image Synthesis}.
\newblock In \emph{International Conference on Learning Representations (ICLR)}, pp.\  1--20, 2022.

\bibitem[Hedborg et~al.(2012)Hedborg, Forss{\'e}n, Felsberg, and Ringaby]{hedborg2012RSBA}
Johan Hedborg, Per-Erik Forss{\'e}n, Michael Felsberg, and Erik Ringaby.
\newblock {Rolling Shutter Bundle Adjustment}.
\newblock In \emph{Computer Vision and Pattern Recognition (CVPR)}, pp.\  1434--1441. IEEE, 2012.

\bibitem[Huang et~al.(2022{\natexlab{a}})Huang, Zhang, Feng, Li, Wang, and Wang]{Huang2022}
Xin Huang, Qi~Zhang, Ying Feng, Hongdong Li, Xuan Wang, and Qing Wang.
\newblock {HDR-NeRF: High Dynamic Range Neural Radiance Fields}.
\newblock In \emph{Computer Vision and Pattern Recognition (CVPR)}, pp.\  18398--18408, 2022{\natexlab{a}}.
\newblock URL \url{http://arxiv.org/abs/2111.14451}.

\bibitem[Huang et~al.(2022{\natexlab{b}})Huang, Zhang, Feng, Li, Wang, and Wang]{huang2022hdr}
Xin Huang, Qi~Zhang, Ying Feng, Hongdong Li, Xuan Wang, and Qing Wang.
\newblock {HDR-NeRF: High Dynamic Range Neural Radiance Fields}.
\newblock In \emph{Computer Vision and Pattern Recognition (CVPR)}, pp.\  18398--18408, 2022{\natexlab{b}}.

\bibitem[Ito \& Okatani(2017)Ito and Okatani]{ito2017RS-sfm}
Eisuke Ito and Takayuki Okatani.
\newblock {Self-Calibration-Based Approach to Critical Motion Sequences of Rolling-Shutter Structure from Motion}.
\newblock In \emph{Computer Vision and Pattern Recognition (CVPR)}, pp.\  801--809, 2017.

\bibitem[Jeong et~al.(2021)Jeong, Ahn, Choy, Anandkumar, Cho, and Park]{Jeong2021}
Yoonwoo Jeong, Seokjun Ahn, Christopher Choy, Animashree Anandkumar, Minsu Cho, and Jaesik Park.
\newblock {Self-Calibrating Neural Radiance Fields}.
\newblock \emph{International Conference on Computer Vision (ICCV)}, 2021.
\newblock URL \url{http://arxiv.org/abs/2108.13826}.

\bibitem[Kania et~al.(2022)Kania, Yi, Kowalski, Trzci{\'n}ski, and Tagliasacchi]{kania2022conerf}
Kacper Kania, Kwang~Moo Yi, Marek Kowalski, Tomasz Trzci{\'n}ski, and Andrea Tagliasacchi.
\newblock {CoNeRF: Controllable Neural Radiance Fields}.
\newblock In \emph{Computer Vision and Pattern Recognition (CVPR)}, pp.\  18623--18632, 2022.

\bibitem[Kim et~al.(2022)Kim, Seo, and Han]{kim2022infonerf}
Mijeong Kim, Seonguk Seo, and Bohyung Han.
\newblock {InfoNeRF: Ray Entropy Minimization for Few-Shot Neural Volume Rendering}.
\newblock In \emph{Computer Vision and Pattern Recognition (CVPR)}, pp.\  12912--12921, 2022.

\bibitem[Kingma \& Ba(2014)Kingma and Ba]{kingma2014adam}
Diederik~P Kingma and Jimmy Ba.
\newblock {Adam: A Method for Stochastic Optimization}.
\newblock \emph{arXiv preprint arXiv:1412.6980}, 2014.

\bibitem[Krishnan \& Fergus(2009)Krishnan and Fergus]{Krishnan2009NIPS}
Dilip Krishnan and Rob Fergus.
\newblock {Fast image deconvolution using Hyper-Laplacian priors}.
\newblock In \emph{Advances in Neural Information Processing Systems (NIPS)}, 2009.

\bibitem[Kupyn et~al.(2019)Kupyn, Martyniuk, Wu, and Wang]{Kupyn2019ICCV}
Orest Kupyn, Tetiana Martyniuk, Junru Wu, and Zhangyang Wang.
\newblock {DeblurGAN-v2: Deblurring (Orders-of-Magnitude) Faster and Better}.
\newblock In \emph{International Conference on Computer Vision (ICCV)}, 2019.

\bibitem[Lao \& Ait-Aider(2018)Lao and Ait-Aider]{Lao2018CVPR}
Yizhen Lao and Omar Ait-Aider.
\newblock A robust method for strong rolling shutter effects correction using lines with automatic feature selection.
\newblock In \emph{CVPR}, 2018.

\bibitem[Lassner \& Zollhofer(2021)Lassner and Zollhofer]{Lassner2021}
Christoph Lassner and Michael Zollhofer.
\newblock {Pulsar: Efficient sphere-based neural rendering}.
\newblock In \emph{Computer Vision and Pattern Recognition (CVPR)}, 2021.
\newblock URL \url{http://openaccess.thecvf.com/content/CVPR2021/html/Lassner_Pulsar_Efficient_Sphere-Based_Neural_Rendering_CVPR_2021_paper.html}.

\bibitem[Levin et~al.(2009)Levin, Weiss, Durand, and Freeman]{Levin2009CVPR}
Anat Levin, Yair Weiss, Fredo Durand, and William~T. Freeman.
\newblock {Understanding and evaluating blind deconvolution algorithm}.
\newblock In \emph{Computer Vision and Pattern Recognition (CVPR)}, 2009.

\bibitem[Levoy(1990)]{levoy1990efficient}
Marc Levoy.
\newblock {Efficient Ray Tracing of Volume Data}.
\newblock \emph{ACM Transactions on Graphics (ToG)}, 9\penalty0 (3):\penalty0 245--261, 1990.

\bibitem[Li et~al.(2022)Li, Lin, Forsyth, Huang, and Wang]{li2022climatenerf}
Yuan Li, Zhi-Hao Lin, David Forsyth, Jia-Bin Huang, and Shenlong Wang.
\newblock {ClimateNeRF: Physically-based Neural Rendering for Extreme Climate Synthesis}.
\newblock \emph{arXiv preprint arXiv:2211.13226}, 2022.

\bibitem[Liao et~al.(2023)Liao, Qu, Xue, Zhang, and Lao]{liao2023NW_RSBA}
Bangyan Liao, Delin Qu, Yifei Xue, Huiqing Zhang, and Yizhen Lao.
\newblock Revisiting rolling shutter bundle adjustment: Toward accurate and fast solution.
\newblock In \emph{Proceedings of the IEEE/CVF Conference on Computer Vision and Pattern Recognition}, pp.\  4863--4871, 2023.

\bibitem[Lin et~al.(2021)Lin, Ma, Torralba, and Lucey]{lin2021barf}
Chen-Hsuan Lin, Wei-Chiu Ma, Antonio Torralba, and Simon Lucey.
\newblock {BARF: Bundle-Adjusting Neural Radiance Fields}.
\newblock In \emph{International Conference on Computer Vision (ICCV)}, pp.\  5741--5751, 2021.

\bibitem[Liu et~al.(2020)Liu, Cui, Larsson, and Pollefeys]{liu2020deepunroll}
Peidong Liu, Zhaopeng Cui, Viktor Larsson, and Marc Pollefeys.
\newblock {Deep Shutter Unrolling Network}.
\newblock In \emph{Computer Vision and Pattern Recognition (CVPR)}, pp.\  5941--5949, 2020.

\bibitem[Liu et~al.(2021{\natexlab{a}})Liu, Zhang, Zhang, Zhang, Zhu, and Russell]{Liu2021}
Steven Liu, Xiuming Zhang, Zhoutong Zhang, Richard Zhang, Jun-Yan Zhu, and Bryan Russell.
\newblock {Editing Conditional Radiance Fields}.
\newblock In \emph{International Conference on Computer Vision (ICCV)}, 2021{\natexlab{a}}.
\newblock URL \url{http://arxiv.org/abs/2105.06466}.

\bibitem[Liu et~al.(2021{\natexlab{b}})Liu, Zhang, Zhang, Zhang, Zhu, and Russell]{liu2021editing}
Steven Liu, Xiuming Zhang, Zhoutong Zhang, Richard Zhang, Jun-Yan Zhu, and Bryan Russell.
\newblock {Editing Conditional Radiance Fields}.
\newblock In \emph{International Conference on Computer Vision (ICCV)}, pp.\  5773--5783, 2021{\natexlab{b}}.

\bibitem[Lovegrove et~al.(2013)Lovegrove, Patron-Perez, and Sibley]{Lovegrove2013BMVC}
Steven Lovegrove, Alonso Patron-Perez, and Gabe Sibley.
\newblock {Spline Fusion: A Continuous-time representation for visual-inertial fusion with application to rolling shutter cameras}.
\newblock In \emph{British Machine Vision Conference (BMVC)}, 2013.

\bibitem[Ma et~al.(2022)Ma, Li, Liao, Zhang, Wang, Wang, and Sander]{ma2022deblur}
Li~Ma, Xiaoyu Li, Jing Liao, Qi~Zhang, Xuan Wang, Jue Wang, and Pedro~V Sander.
\newblock {Deblur-NeRF: Neural Radiance Fields from Blurry Images}.
\newblock In \emph{Computer Vision and Pattern Recognition (CVPR)}, pp.\  12861--12870, 2022.

\bibitem[Martin-Brualla et~al.(2021)Martin-Brualla, Radwan, Sajjadi, Barron, Dosovitskiy, and Duckworth]{MartinBrualla2021}
Ricardo Martin-Brualla, Noha Radwan, Mehdi~S.M. Sajjadi, Jonathan~T. Barron, Alexey Dosovitskiy, and Daniel Duckworth.
\newblock {NeRF in the Wild: Neural Radiance Fields for Unconstrained Photo Collections}.
\newblock In \emph{Computer Vision and Pattern Recognition (CVPR)}, pp.\  7206--7215, 2021.
\newblock ISBN 9781665445092.
\newblock \doi{10.1109/CVPR46437.2021.00713}.

\bibitem[Max(1995)]{max1995optical}
Nelson Max.
\newblock {Optical Models for Direct Volume Rendering}.
\newblock \emph{IEEE Transactions on Visualization and Computer Graphics (TVCG)}, 1\penalty0 (2):\penalty0 99--108, 1995.

\bibitem[Meng et~al.(2021)Meng, Chen, Luo, Wu, Su, Xu, He, and Yu]{meng2021gnerf}
Quan Meng, Anpei Chen, Haimin Luo, Minye Wu, Hao Su, Lan Xu, Xuming He, and Jingyi Yu.
\newblock {GNeRF: GAN-based Neural Radiance Field without Posed Camera}.
\newblock In \emph{International Conference on Computer Vision (ICCV)}, pp.\  6351--6361, 2021.

\bibitem[Mildenhall et~al.(2019)Mildenhall, Srinivasan, Ortiz-Cayon, Kalantari, Ramamoorthi, Ng, and Kar]{mildenhall2019llff}
Ben Mildenhall, Pratul~P Srinivasan, Rodrigo Ortiz-Cayon, Nima~Khademi Kalantari, Ravi Ramamoorthi, Ren Ng, and Abhishek Kar.
\newblock Local light field fusion: Practical view synthesis with prescriptive sampling guidelines.
\newblock \emph{ACM Transactions on Graphics (TOG)}, 38\penalty0 (4):\penalty0 1--14, 2019.

\bibitem[Mildenhall et~al.(2020)Mildenhall, Srinivasan, Tancik, Barron, Ramamoorthi, and Ng]{mildenhall2020nerf}
Ben Mildenhall, Pratul~P. Srinivasan, Matthew Tancik, Jonathan~T. Barron, Ravi Ramamoorthi, and Ren Ng.
\newblock {NeRF: Representing Scenes as Neural Radiance Fields for View Synthesis}.
\newblock In \emph{European Conference on Computer Vision (ECCV)}, 2020.

\bibitem[Mildenhall et~al.(2022{\natexlab{a}})Mildenhall, Hedman, Martin-Brualla, Srinivasan, and Barron]{Mildenhall2022}
Ben Mildenhall, Peter Hedman, Ricardo Martin-Brualla, Pratul~P Srinivasan, and Jonathan~T Barron.
\newblock {NeRF in the Dark: High Dynamic Range View Synthesis from Noisy Raw Images}.
\newblock In \emph{Computer Vision and Pattern Recognition (CVPR)}, pp.\  16190--16199, 2022{\natexlab{a}}.

\bibitem[Mildenhall et~al.(2022{\natexlab{b}})Mildenhall, Hedman, Martin-Brualla, Srinivasan, and Barron]{mildenhall2022nerf}
Ben Mildenhall, Peter Hedman, Ricardo Martin-Brualla, Pratul~P Srinivasan, and Jonathan~T Barron.
\newblock {NeRF in the Dark: High Dynamic Range View Synthesis from Noisy Raw Images}.
\newblock In \emph{Computer Vision and Pattern Recognition (CVPR)}, pp.\  16190--16199, 2022{\natexlab{b}}.

\bibitem[M\"uller et~al.(2022)M\"uller, Evans, Schied, and Keller]{mueller2022instant}
Thomas M\"uller, Alex Evans, Christoph Schied, and Alexander Keller.
\newblock {Instant Neural Graphics Primitives with a Multiresolution Hash Encoding}.
\newblock \emph{ACM Transactions on Graphics (ToG)}, 41\penalty0 (4):\penalty0 102:1--102:15, July 2022.
\newblock \doi{10.1145/3528223.3530127}.
\newblock URL \url{https://doi.org/10.1145/3528223.3530127}.

\bibitem[M{\"u}ller et~al.(2022)M{\"u}ller, Evans, Schied, and Keller]{muller2022instant}
Thomas M{\"u}ller, Alex Evans, Christoph Schied, and Alexander Keller.
\newblock {Instant Neural Graphics Primitives with a Multiresolution Hash Encoding}.
\newblock \emph{ACM Transactions on Graphics (ToG)}, 41\penalty0 (4):\penalty0 1--15, 2022.

\bibitem[Nah et~al.(2017)Nah, Kim, and Lee]{Nah2017CVPR}
Seungjun Nah, Tae~Hyun Kim, and Kyoung~Mu Lee.
\newblock {Deep Multi-Scale Convolutional Neural Network for Dynamic Scene Deblurring}.
\newblock In \emph{Computer Vision and Pattern Recognition (CVPR)}, July 2017.

\bibitem[Niemeyer \& Geiger(2021)Niemeyer and Geiger]{Niemeyer2021}
Michael Niemeyer and Andreas Geiger.
\newblock {GIRAFFE: Representing Scenes as Compositional Generative Neural Feature Fields}.
\newblock In \emph{Computer Vision and Pattern Recognition (CVPR)}, pp.\  11448--11459, 2021.
\newblock \doi{10.1109/cvpr46437.2021.01129}.

\bibitem[Niemeyer et~al.(2022)Niemeyer, Barron, Mildenhall, Sajjadi, Geiger, and Radwan]{niemeyer2022regnerf}
Michael Niemeyer, Jonathan~T Barron, Ben Mildenhall, Mehdi~SM Sajjadi, Andreas Geiger, and Noha Radwan.
\newblock {RegNeRF: Regularizing Neural Radiance Fields for View Synthesis from Sparse Inputs}.
\newblock In \emph{Computer Vision and Pattern Recognition (CVPR)}, pp.\  5480--5490, 2022.

\bibitem[Nie{\ss}ner et~al.(2013{\natexlab{a}})Nie{\ss}ner, Zollh\"ofer, Izadi, and Stamminger]{Niesner2013SIGGRAPH}
M.~Nie{\ss}ner, M.~Zollh\"ofer, S.~Izadi, and M.~Stamminger.
\newblock {Real-time 3D Reconstruction at Scale using Voxel Hashing}.
\newblock In \emph{ACM Transactions on Graphics (ToG)}, 2013{\natexlab{a}}.

\bibitem[Nie{\ss}ner et~al.(2013{\natexlab{b}})Nie{\ss}ner, Zollh{\"o}fer, Izadi, and Stamminger]{niessner2013real}
Matthias Nie{\ss}ner, Michael Zollh{\"o}fer, Shahram Izadi, and Marc Stamminger.
\newblock {Real-time 3D Reconstruction at Scale using Voxel Hashing}.
\newblock \emph{ACM Transactions on Graphics (ToG)}, 32\penalty0 (6):\penalty0 1--11, 2013{\natexlab{b}}.

\bibitem[Or-El et~al.(2022)Or-El, Luo, Shan, Shechtman, Park, and Kemelmacher-Shlizerman]{OrEl2021}
Roy Or-El, Xuan Luo, Mengyi Shan, Eli Shechtman, Jeong~Joon Park, and Ira Kemelmacher-Shlizerman.
\newblock {StyleSDF: High-Resolution 3D-Consistent Image and Geometry Generation}.
\newblock In \emph{Computer Vision and Pattern Recognition (CVPR)}, pp.\  13503--13513, 2022.
\newblock URL \url{http://arxiv.org/abs/2112.11427}.

\bibitem[Park \& Lee(2017)Park and Lee]{Park2017}
Haesol Park and Kyoung~Mu Lee.
\newblock {Joint Estimation of Camera Pose, Depth, Deblurring, and Super-Resolution from a Blurred Image Sequence}.
\newblock In \emph{International Conference on Computer Vision (ICCV)}, volume 2017-Octob, pp.\  4623--4631, 2017.
\newblock ISBN 9781538610329.
\newblock \doi{10.1109/ICCV.2017.494}.

\bibitem[Park et~al.(2020)Park, Sinha, Barron, Bouaziz, Goldman, Seitz, and Martin-Brualla]{Park2020}
Keunhong Park, Utkarsh Sinha, Jonathan~T. Barron, Sofien Bouaziz, Dan~B Goldman, Steven~M. Seitz, and Ricardo Martin-Brualla.
\newblock {Nerfies: Deformable Neural Radiance Fields}.
\newblock In \emph{International Conference on Computer Vision (ICCV)}, 2020.
\newblock URL \url{http://arxiv.org/abs/2011.12948}.

\bibitem[Park et~al.(2021)Park, Sinha, Barron, Bouaziz, Goldman, Seitz, and Martin-Brualla]{park2021nerfies}
Keunhong Park, Utkarsh Sinha, Jonathan~T Barron, Sofien Bouaziz, Dan~B Goldman, Steven~M Seitz, and Ricardo Martin-Brualla.
\newblock {Nerfies: Deformable Neural Radiance Fields}.
\newblock In \emph{International Conference on Computer Vision (ICCV)}, pp.\  5865--5874, 2021.

\bibitem[Pearl et~al.(2022)Pearl, Treibitz, and Korman]{pearl2022nan}
Naama Pearl, Tali Treibitz, and Simon Korman.
\newblock {NAN: Noise-Aware NeRFs for Burst-Denoising}.
\newblock In \emph{Computer Vision and Pattern Recognition (CVPR)}, pp.\  12672--12681, 2022.

\bibitem[Peng et~al.(2021{\natexlab{a}})Peng, Dong, Wang, Zhang, Shuai, Zhou, and Bao]{Peng2021a}
Sida Peng, Junting Dong, Qianqian Wang, Shangzhan Zhang, Qing Shuai, Xiaowei Zhou, and Hujun Bao.
\newblock {Animatable neural radiance fields for modeling dynamic human bodies}.
\newblock In \emph{International Conference on Computer Vision (ICCV)}, 2021{\natexlab{a}}.
\newblock URL \url{http://openaccess.thecvf.com/content/ICCV2021/html/Peng_Animatable_Neural_Radiance_Fields_for_Modeling_Dynamic_Human_Bodies_ICCV_2021_paper.html}.

\bibitem[Peng et~al.(2021{\natexlab{b}})Peng, Zhang, Xu, Wang, Shuai, Bao, and Zhou]{Peng2021}
Sida Peng, Yuanqing Zhang, Yinghao Xu, Qianqian Wang, Qing Shuai, Hujun Bao, and Xiaowei Zhou.
\newblock {Neural Body: Implicit Neural Representations with Structured Latent Codes for Novel View Synthesis of Dynamic Humans}.
\newblock In \emph{Computer Vision and Pattern Recognition (CVPR)}, pp.\  9050--9059, 2021{\natexlab{b}}.
\newblock ISBN 9781665445092.
\newblock \doi{10.1109/CVPR46437.2021.00894}.
\newblock URL \url{http://openaccess.thecvf.com/content/CVPR2021/html/Peng_Neural_Body_Implicit_Neural_Representations_With_Structured_Latent_Codes_for_CVPR_2021_paper.html}.

\bibitem[Piala \& Clark(2021)Piala and Clark]{Piala2022}
Martin Piala and Ronald Clark.
\newblock {TermiNeRF: Ray Termination Prediction for Efficient Neural Rendering}.
\newblock In \emph{International Conference on 3D Vision (3DV)}, 2021.

\bibitem[Pollefeys et~al.(2008)Pollefeys, Nist{\'e}r, Frahm, Akbarzadeh, Mordohai, Clipp, Engels, Gallup, Kim, Merrell, et~al.]{pollefeys2008detailed}
Marc Pollefeys, David Nist{\'e}r, J-M Frahm, Amir Akbarzadeh, Philippos Mordohai, Brian Clipp, Chris Engels, David Gallup, S-J Kim, Paul Merrell, et~al.
\newblock {Detailed Real-Time Urban 3D Reconstruction from Video}.
\newblock \emph{International Journal of Computer Vision (IJCV)}, 78:\penalty0 143--167, 2008.

\bibitem[Pumarola et~al.(2021{\natexlab{a}})Pumarola, Corona, Pons-Moll, and Moreno-Noguer]{Pumarola2021}
Albert Pumarola, Enric Corona, Gerard Pons-Moll, and Francesc Moreno-Noguer.
\newblock {D-NeRF: Neural Radiance Fields for Dynamic Scenes}.
\newblock In \emph{Computer Vision and Pattern Recognition (CVPR)}, pp.\  10313--10322, 2021{\natexlab{a}}.
\newblock ISBN 9781665445092.
\newblock \doi{10.1109/CVPR46437.2021.01018}.
\newblock URL \url{http://openaccess.thecvf.com/content/CVPR2021/html/Pumarola_D-NeRF_Neural_Radiance_Fields_for_Dynamic_Scenes_CVPR_2021_paper.html}.

\bibitem[Pumarola et~al.(2021{\natexlab{b}})Pumarola, Corona, Pons-Moll, and Moreno-Noguer]{pumarola2021d}
Albert Pumarola, Enric Corona, Gerard Pons-Moll, and Francesc Moreno-Noguer.
\newblock {D-NeRF: Neural Radiance Fields for Dynamic Scenes}.
\newblock In \emph{Computer Vision and Pattern Recognition (CVPR)}, pp.\  10318--10327, 2021{\natexlab{b}}.

\bibitem[Purkait et~al.(2017)Purkait, Zach, and Leonardis]{Purkait2017ICCV}
Pulak Purkait, Christopher Zach, and Ales Leonardis.
\newblock Rolling shutter correction in manhattan world.
\newblock In \emph{ICCV}, 2017.

\bibitem[Qin(1998)]{Qin1998CGA}
Kaihuai Qin.
\newblock {General Matrix Representations for B-Splines}.
\newblock In \emph{Sixth Pacific Conference on Computer Graphics and Applications}, 1998.

\bibitem[Rebain et~al.(2021)Rebain, Jiang, Yazdani, Li, Yi, and Tagliasacchi]{Rebain2021}
Daniel Rebain, Wei Jiang, Soroosh Yazdani, Ke~Li, Kwang~Moo Yi, and Andrea Tagliasacchi.
\newblock {Derf: Decomposed radiance fields}.
\newblock In \emph{Computer Vision and Pattern Recognition (CVPR)}, pp.\  14148--14156, 2021.
\newblock ISBN 9781665445092.
\newblock \doi{10.1109/CVPR46437.2021.01393}.

\bibitem[Reiser et~al.(2021)Reiser, Peng, Liao, and Geiger]{Reiser2021}
Christian Reiser, Songyou Peng, Yiyi Liao, and Andreas Geiger.
\newblock {KiloNeRF: Speeding up Neural Radiance Fields with Thousands of Tiny MLPs}.
\newblock In \emph{International Conference on Computer Vision (ICCV)}, 2021.
\newblock URL \url{http://arxiv.org/abs/2103.13744}.

\bibitem[Rengarajan et~al.(2016)Rengarajan, Rajagopalan, and Aravind]{Rengarajan2016CVPR}
Vijay Rengarajan, A.N. Rajagopalan, and R.~Aravind.
\newblock From bows to arrows: rolling shutter rectification of urban scenes.
\newblock In \emph{CVPR}, 2016.

\bibitem[Rengarajan et~al.(2017)Rengarajan, Balaji, and Rajagopalan]{rengarajan2017unrolling}
Vijay Rengarajan, Yogesh Balaji, and AN~Rajagopalan.
\newblock {Unrolling the Shutter: CNN to Correct Motion Distortions}.
\newblock In \emph{Computer Vision and Pattern Recognition (CVPR)}, pp.\  2291--2299, 2017.

\bibitem[Schonberger \& Frahm(2016)Schonberger and Frahm]{schonberger2016structure}
Johannes~L Schonberger and Jan-Michael Frahm.
\newblock {Structure-from-Motion Revisited}.
\newblock In \emph{Computer Vision and Pattern Recognition (CVPR)}, pp.\  4104--4113, 2016.

\bibitem[Schops et~al.(2019)Schops, Sattler, and Pollefeys]{ETH3D}
Thomas Schops, Torsten Sattler, and Marc Pollefeys.
\newblock {BAD SLAM: Bundle Adjusted Direct RGB-D SLAM}.
\newblock In \emph{Computer Vision and Pattern Recognition (CVPR)}, pp.\  134--144, 2019.

\bibitem[Schubert et~al.(2019)Schubert, Demmel, von Stumberg, Usenko, and Cremers]{schubert2019RS-VIO}
David Schubert, Nikolaus Demmel, Lukas von Stumberg, Vladyslav Usenko, and Daniel Cremers.
\newblock {Rolling-Shutter Modelling for Direct Visual-Inertial Odometry}.
\newblock In \emph{International Conference on Intelligent Robots and Systems (IROS)}, pp.\  2462--2469. IEEE, 2019.

\bibitem[Seitz \& Dyer(1997)Seitz and Dyer]{seitz1997photorealistic}
Steven~M Seitz and Charles~R Dyer.
\newblock {Photorealistic Scene Reconstruction by Voxel Coloring}.
\newblock In \emph{Computer Vision and Pattern Recognition (CVPR)}, pp.\  1067--1073, 1997.

\bibitem[Shan et~al.(2008)Shan, Jia, and A.]{Shan2008ToG}
Q~Shan, Jiaya Jia, and Agarwala A.
\newblock {High-quality Motion Deblurring from a Single Image}.
\newblock \emph{ACM Transactions on Graphics (ToG)}, 27\penalty0 (3), 2008.

\bibitem[Srinivasan et~al.(2021)Srinivasan, Deng, Zhang, Tancik, Mildenhall, and Barron]{Srinivasan2021}
Pratul~P. Srinivasan, Boyang Deng, Xiuming Zhang, Matthew Tancik, Ben Mildenhall, and Jonathan~T. Barron.
\newblock {NeRV: Neural Reflectance and Visibility Fields for Relighting and View Synthesis}.
\newblock In \emph{Computer Vision and Pattern Recognition (CVPR)}, pp.\  7491--7500, 2021.
\newblock ISBN 9781665445092.
\newblock \doi{10.1109/CVPR46437.2021.00741}.

\bibitem[Steinbruecker et~al.(2014)Steinbruecker, Sturm, and Cremers]{Steinbruecker2014ICRA}
F.~Steinbruecker, J.~Sturm, and D.~Cremers.
\newblock {Volumetric 3D Mapping in Real-Time on a CPU}.
\newblock In \emph{International Conference on Robotics and Automation (ICRA)}, 2014.

\bibitem[Su et~al.(2017)Su, Delbracio, and Wang]{Su2017CVPR}
Shuochen Su, Mauricio Delbracio, and Jue Wang.
\newblock {Deep Video Deblurring for Hand-held Cameras}.
\newblock In \emph{Computer Vision and Pattern Recognition (CVPR)}, 2017.

\bibitem[Sun et~al.(2022)Sun, Wang, Zhang, Li, Zhang, Liu, and Wang]{sun2022fenerf}
Jingxiang Sun, Xuan Wang, Yong Zhang, Xiaoyu Li, Qi~Zhang, Yebin Liu, and Jue Wang.
\newblock {FENeRF: Face Editing in Neural Radiance Fields}.
\newblock In \emph{Computer Vision and Pattern Recognition (CVPR)}, pp.\  7672--7682, 2022.

\bibitem[Tancik et~al.(2022)Tancik, Casser, Yan, Pradhan, Mildenhall, Srinivasan, Barron, and Kretzschmar]{Tancik2022}
Matthew Tancik, Vincent Casser, Xinchen Yan, Sabeek Pradhan, Ben Mildenhall, Pratul~P. Srinivasan, Jonathan~T. Barron, and Henrik Kretzschmar.
\newblock {Block-NeRF: Scalable Large Scene Neural View Synthesis}.
\newblock In \emph{Computer Vision and Pattern Recognition (CVPR)}, 2022.
\newblock URL \url{http://arxiv.org/abs/2202.05263}.

\bibitem[Tao et~al.(2018)Tao, Gao, Shen, Wang, and Jia]{Tao2018CVPR}
Xin Tao, Hongyun Gao, Xiaoyong Shen, Jue Wang, and Jiaya Jia.
\newblock {Scale-recurrent network for deep image deblurring}.
\newblock In \emph{Computer Vision and Pattern Recognition (CVPR)}, 2018.

\bibitem[Tretschk et~al.(2021)Tretschk, Tewari, Golyanik, Zollh{\"o}fer, Lassner, and Theobalt]{tretschk2021non}
Edgar Tretschk, Ayush Tewari, Vladislav Golyanik, Michael Zollh{\"o}fer, Christoph Lassner, and Christian Theobalt.
\newblock {Non-Rigid Neural Radiance Fields: Reconstruction and Novel View Synthesis of a Dynamic Scene From Monocular Video}.
\newblock In \emph{International Conference on Computer Vision (ICCV)}, pp.\  12959--12970, 2021.

\bibitem[Turki et~al.(2022)Turki, Ramanan, and Satyanarayanan]{Turki2022}
Haithem Turki, Deva Ramanan, and Mahadev Satyanarayanan.
\newblock {Mega-NeRF: Scalable Construction of Large-Scale NeRFs for Virtual Fly-Throughs}.
\newblock In \emph{Computer Vision and Pattern Recognition (CVPR)}, pp.\  12922--12931, 2022.
\newblock URL \url{http://arxiv.org/abs/2112.10703}.

\bibitem[Vasu et~al.(2018)Vasu, Mohan, and Rajagopalan]{Vasu2018CVPR}
Subeesh Vasu, Mahesh Mohan, and A.N. Rajagopalan.
\newblock Occlusion aware rolling shutter rectification of 3d scenes.
\newblock In \emph{CVPR}, 2018.

\bibitem[Vaswani et~al.(2017)Vaswani, Shazeer, Parmar, Uszkoreit, Jones, Gomez, Kaiser, and Polosukhin]{vaswani2017attention}
Ashish Vaswani, Noam Shazeer, Niki Parmar, Jakob Uszkoreit, Llion Jones, Aidan~N Gomez, {\L}ukasz Kaiser, and Illia Polosukhin.
\newblock {Attention Is All You Need}.
\newblock \emph{Advances in Neural Information Processing Systems (NIPS)}, 30, 2017.

\bibitem[Wang et~al.(2022)Wang, Chai, He, Chen, and Liao]{Wang2022}
Can Wang, Menglei Chai, Mingming He, Dongdong Chen, and Jing Liao.
\newblock {CLIP-NeRF: Text-and-Image Driven Manipulation of Neural Radiance Fields}.
\newblock In \emph{Computer Vision and Pattern Recognition (CVPR)}, pp.\  1--10, 2022.
\newblock URL \url{http://arxiv.org/abs/2112.05139}.

\bibitem[Wang et~al.(2023)Wang, Zhao, Ma, and Liu]{wang2023bad}
Peng Wang, Lingzhe Zhao, Ruijie Ma, and Peidong Liu.
\newblock {BAD-NeRF: Bundle Adjusted Deblur Neural Radiance Fields}.
\newblock In \emph{Computer Vision and Pattern Recognition (CVPR)}, 2023.

\bibitem[Wang et~al.(2021)Wang, Wu, Xie, Chen, and Prisacariu]{wang2021nerf}
Zirui Wang, Shangzhe Wu, Weidi Xie, Min Chen, and Victor~Adrian Prisacariu.
\newblock {NeRF-{}-: Neural radiance fields without known camera parameters}.
\newblock \emph{arXiv preprint arXiv:2102.07064}, 2021.

\bibitem[Weng et~al.(2022)Weng, Curless, Srinivasan, Barron, and Kemelmacher-Shlizerman]{Weng2022}
Chung-Yi Weng, Brian Curless, Pratul~P. Srinivasan, Jonathan~T. Barron, and Ira Kemelmacher-Shlizerman.
\newblock {HumanNeRF: Free-viewpoint Rendering of Moving People from Monocular Video}.
\newblock In \emph{Computer Vision and Pattern Recognition (CVPR)}, pp.\  16210--16220, 2022.
\newblock URL \url{http://arxiv.org/abs/2201.04127}.

\bibitem[Whelan et~al.(2015)Whelan, Leutenegger, Salas{-}Moreno, Glocker, and Davison]{Whelan2015RSS}
Thomas Whelan, Stefan Leutenegger, Renato~F. Salas{-}Moreno, Ben Glocker, and Andrew~J. Davison.
\newblock {ElasticFusion: Dense {SLAM} Without {A} Pose Graph}.
\newblock In \emph{Robotics: Science and Systems (RSS)}, 2015.

\bibitem[Wizadwongsa et~al.(2021)Wizadwongsa, Phongthawee, Yenphraphai, and Suwajanakorn]{Wizadwongsa2021}
Suttisak Wizadwongsa, Pakkapon Phongthawee, Jiraphon Yenphraphai, and Supasorn Suwajanakorn.
\newblock {NeX: Real-time View Synthesis with Neural Basis Expansion}.
\newblock In \emph{Computer Vision and Pattern Recognition (CVPR)}, pp.\  8530--8539, 2021.
\newblock ISBN 9781665445092.
\newblock \doi{10.1109/CVPR46437.2021.00843}.
\newblock URL \url{http://openaccess.thecvf.com/content/CVPR2021/html/Wizadwongsa_NeX_Real-Time_View_Synthesis_With_Neural_Basis_Expansion_CVPR_2021_paper.html}.

\bibitem[Xian et~al.(2021)Xian, Huang, Kopf, and Kim]{Xian2021}
Wenqi Xian, Jia~Bin Huang, Johannes Kopf, and Changil Kim.
\newblock {Space-time Neural Irradiance Fields for Free-Viewpoint Video}.
\newblock In \emph{Computer Vision and Pattern Recognition (CVPR)}, pp.\  9416--9426, 2021.
\newblock ISBN 9781665445092.
\newblock \doi{10.1109/CVPR46437.2021.00930}.
\newblock URL \url{http://openaccess.thecvf.com/content/CVPR2021/html/Xian_Space-Time_Neural_Irradiance_Fields_for_Free-Viewpoint_Video_CVPR_2021_paper.html}.

\bibitem[Xiangli et~al.(2022)Xiangli, Xu, Pan, Zhao, Rao, Theobalt, Dai, and Lin]{Xiangli2021}
Yuanbo Xiangli, Linning Xu, Xingang Pan, Nanxuan Zhao, Anyi Rao, Christian Theobalt, Bo~Dai, and Dahua Lin.
\newblock {CityNeRF: Building NeRF at City Scale}.
\newblock In \emph{European Conference on Computer Vision (ECCV)}, 2022.

\bibitem[Xu et~al.(2022)Xu, Jiang, Wang, Fan, Shi, and Wang]{xu2022sinnerf}
Dejia Xu, Yifan Jiang, Peihao Wang, Zhiwen Fan, Humphrey Shi, and Zhangyang Wang.
\newblock {SinNeRF: Training Neural Radiance Fields on Complex Scenes from a Single Image}.
\newblock In \emph{European Conference on Computer Vision (ECCV)}, pp.\  736--753. Springer, 2022.

\bibitem[Xu \& Jia(2010)Xu and Jia]{Xu2010ECCV}
Li~Xu and Jiaya Jia.
\newblock {Two-Phase Kernel Estimation for robust motion deblurring}.
\newblock In \emph{European Conference on Computer Vision (ECCV)}, 2010.

\bibitem[Yang et~al.(2021)Yang, Zhang, Xu, Li, Zhou, Bao, Zhang, and Cui]{Yang2021}
Bangbang Yang, Yinda Zhang, Yinghao Xu, Yijin Li, Han Zhou, Hujun Bao, Guofeng Zhang, and Zhaopeng Cui.
\newblock {Learning Object-Compositional Neural Radiance Field for Editable Scene Rendering}.
\newblock In \emph{International Conference on Computer Vision (ICCV)}, pp.\  13779--13788, 2021.
\newblock URL \url{http://arxiv.org/abs/2109.01847}.

\bibitem[Yang et~al.(2022)Yang, Chen, Chen, Chen, and Wong]{Yang2022}
Wenqi Yang, Guanying Chen, Chaofeng Chen, Zhenfang Chen, and Kwan-Yee~K. Wong.
\newblock {PS-NeRF: Neural Inverse Rendering for Multi-view Photometric Stereo}.
\newblock In \emph{European Conference on Computer Vision (ECCV)}, 2022.
\newblock URL \url{http://arxiv.org/abs/2207.11406}.

\bibitem[Yu et~al.(2021{\natexlab{a}})Yu, Li, Tancik, Li, Ng, and Kanazawa]{Yu2021}
Alex Yu, Ruilong Li, Matthew Tancik, Hao Li, Ren Ng, and Angjoo Kanazawa.
\newblock {PlenOctrees for Real-time Rendering of Neural Radiance Fields}.
\newblock In \emph{International Conference on Computer Vision (ICCV)}, 2021{\natexlab{a}}.

\bibitem[Yu et~al.(2021{\natexlab{b}})Yu, Li, Tancik, Li, Ng, and Kanazawa]{yu2021plenoctrees}
Alex Yu, Ruilong Li, Matthew Tancik, Hao Li, Ren Ng, and Angjoo Kanazawa.
\newblock {PlenOctrees for Real-time Rendering of Neural Radiance Fields}.
\newblock In \emph{International Conference on Computer Vision (ICCV)}, pp.\  5752--5761, 2021{\natexlab{b}}.

\bibitem[Yu et~al.(2021{\natexlab{c}})Yu, Ye, Tancik, and Kanazawa]{yu2021pixelnerf}
Alex Yu, Vickie Ye, Matthew Tancik, and Angjoo Kanazawa.
\newblock {pixelNeRF: Neural Radiance Fields from One or Few Images}.
\newblock In \emph{Computer Vision and Pattern Recognition (CVPR)}, pp.\  4578--4587, 2021{\natexlab{c}}.

\bibitem[Yuan et~al.(2022)Yuan, Sun, Lai, Ma, Jia, and Gao]{Yuan2022}
Yu-jie Yuan, Yang-tian Sun, Yu-kun Lai, Yuewen Ma, Rongfei Jia, and Lin Gao.
\newblock {NeRF-Editing: Geometry Editing of Neural Radiance Fields}.
\newblock In \emph{Computer Vision and Pattern Recognition (CVPR)}, pp.\  18353--18364, 2022.

\bibitem[Zhang et~al.(2021{\natexlab{a}})Zhang, Yang, Tulsiani, and Ramanan]{Zhang2021a}
Jason~Y. Zhang, Gengshan Yang, Shubham Tulsiani, and Deva Ramanan.
\newblock {NeRS: Neural Reflectance Surfaces for Sparse-view 3D Reconstruction in the Wild}.
\newblock In \emph{Advances in Neural Information Processing Systems (NeurIPS)}, pp.\  1--13, 2021{\natexlab{a}}.
\newblock URL \url{http://arxiv.org/abs/2110.07604}.

\bibitem[Zhang et~al.(2018)Zhang, Isola, Efros, Shechtman, and Wang]{zhang2018LPIPS}
Richard Zhang, Phillip Isola, Alexei~A Efros, Eli Shechtman, and Oliver Wang.
\newblock {The Unreasonable Effectiveness of Deep Features as a Perceptual Metric}.
\newblock In \emph{Computer Vision and Pattern Recognition (CVPR)}, pp.\  586--595, 2018.

\bibitem[Zhang et~al.(2021{\natexlab{b}})Zhang, Srinivasan, Deng, Debevec, Freeman, and Barron]{Zhang2021}
Xiuming Zhang, Pratul~P. Srinivasan, Boyang Deng, Paul Debevec, William~T. Freeman, and Jonathan~T. Barron.
\newblock {NeRFactor: Neural Factorization of Shape and Reflectance Under an Unknown Illumination}.
\newblock \emph{ACM Transactions on Graphics (ToG)}, 40\penalty0 (6):\penalty0 1--18, dec 2021{\natexlab{b}}.
\newblock ISSN 0730-0301.
\newblock \doi{10.1145/3478513.3480496}.

\bibitem[Zhong et~al.(2022)Zhong, Cao, Sun, Wu, Zhou, Zheng, Lin, and Sato]{zhong2022dual}
Zhihang Zhong, Mingdeng Cao, Xiao Sun, Zhirong Wu, Zhongyi Zhou, Yinqiang Zheng, Stephen Lin, and Imari Sato.
\newblock {Bringing Rolling Shutter Images Alive with Dual Reversed Distortion}.
\newblock In \emph{European Conference on Computer Vision (ECCV)}, pp.\  233--249. Springer, 2022.

\bibitem[Zhuang et~al.(2017)Zhuang, Cheong, and Hee~Lee]{zhuang2017sfm_RS}
Bingbing Zhuang, Loong-Fah Cheong, and Gim Hee~Lee.
\newblock {Rolling-Shutter-Aware Differential SfM and Image Rectification}.
\newblock In \emph{International Conference on Computer Vision (ICCV)}, pp.\  948--956, 2017.

\end{thebibliography}
\bibliographystyle{iclr2024_conference}

\appendix
\section{Appendix}

In the appendix, we present the details on method implementation and training, frame selections from TUM-RS \citep{schubert2019RS-VIO} datasets, more experimental results on image quality of synthesized novel views and visualization of estimated motion trajectories. The rendered novel view high frame-rate global shutter video is presented in the supplementary video. We will present each part as follows. 

\subsection{Implementation and Training Details}
We implement our method in PyTorch. We use Adam \citep{kingma2014adam} optimizer to estimate the weights of the MLP network and the pose parameters of the spline. The optimizer is configured with ${\beta}_1=0.9$ and ${\beta}_2=0.999$ for NeRF and pose optimizations. We set the learning rates to be $5 \times 10^{-4}$ and  $1 \times 10^{-3}$ for the NeRF and pose optimizers respectively. Both learning rates gradually decay to $5 \times 10^{-5}$ and $1 \times 10^{-5}$ respectively. During each training step, we randomly select 7200 pixels from all training images to minimize the loss function presented in \eqnref{eq_loss_function} and we run a total of 200K steps on an NVIDIA RTX 3090 GPU. We adopt the linear adjustment of the positional encoding starting from steps 20K to 100K to achieve the coarse-to-fine training strategy as in BARF \citep{lin2021barf}. We select the pose corresponding to the first row of each image as the control knots of the spline, and they are initialized with the poses computed via COLMAP \citep{schonberger2016structure} from the sequence of rolling shutter images. 
%%%%% BODY TEXT
\subsection{Details on frame selections from TUM-RS datasets}
\label{sec:appendix_dataset}

Since the total number of frames of each TUM-RS \citep{schubert2019RS-VIO} sequence is too long to be processed by NeRF \citep{mildenhall2020nerf} and our method, we choose sub-sequence frames to evaluate our method. The details are listed in \tabnref{TUM_frames}.

\begin{table}[htbp]
	\begin{center}
	\captionsetup {font={small,stretch=0.5}}
	\caption{{\textbf{Selected frames from each sequence of TUM-RS datasets \citep{schubert2019RS-VIO}}.}} 
	\label{TUM_frames}
	\setlength{\belowcaptionskip}{-9pt}
	\setlength\tabcolsep{10pt}
	\scriptsize
	% \resizebox{\linewidth}{!}{
		\begin{tabular}{c|c|c}
			\toprule
			Seq. &  Start frame timestamp & End frame timestamp \\
			\midrule
			1 &1548685771426550686 &1548685773576617686\\
			2 &1548685824354495627 &1548685826504562627\\
			3 &1548685846189303526 &1548685848339371526\\
			4 &1548685907331049323 &1548685909481116323\\
			5 &1548686030960692141 &1548686033110759141\\
			6 &1548689768046781352 &1548689770196849352\\
			7 &1548689850274927799 &1548689852424994799\\
			8 &1548689861503747264 &1548689863653814264\\
			9 &1548689939497068095 &1548689941647135095\\
			10 &1548689997870607830 &1548690000020674830\\
			\bottomrule
		\end{tabular}
		% }
%	\vspace{0.5em}
%	\vspace{-1em}
	\end{center}
\end{table}

\subsection{Additional experimental results}
\label{sec:appendix_experiment}

To evaluate the performance of our method for novel view image synthesis , we select 6 additional views from each scene of Unreal-RS datasets. Since DiffSfM \citep{zhuang2017sfm_RS}, DSUN \citep{liu2020deepunroll}, SUNet \citep{fan2021sunet}, RSSR \citep{fan2021RSSR} and CVR\citep{fan2022CVR} cannot synthesize novel view image, we first apply these approaches to restore global shutter images, and then train the original NeRF \citep{mildenhall2020nerf}. \tabnref{table_novel_view} and \figrefer{fig_novel_view} demonstrate that our method outperforms prior methods in terms of novel view image synthesis. The poor performances of DSUN \citep{liu2020deepunroll}, SUNet \citep{fan2021sunet}, RSSR \citep{fan2021RSSR} and CVR\citep{fan2022CVR}, are caused by their poor generalization capabilities on domain-shifted datasets. Learning-free method DiffSfM \citep{zhuang2017sfm_RS} similarly shows poor performance on all datasets as it could not realize perfect rolling shutter effect removal even with bundle adjustment. The experimental results also reveal that both NeRF \citep{mildenhall2020nerf} and BARF \citep{lin2021barf} cannot perform well either, due to their ignorance of the rolling shutter effect presented in the training images. Since the Carla-RS \citep{liu2020deepunroll} does not provide additional images for novel view image synthesis evaluation, we only present the qualitative results in \figrefer{fig_carla_novel_view}. It also demonstrates that our method could take advantage of multi-view information instead of only two views, and thus performs better than existing methods in terms of novel view image synthesis.

\begin{table}
	\captionsetup {font={small,stretch=0.5}}
	\caption{{\textbf{Quantitative comparisons on the synthetic datasets in terms of novel view synthesis.}} Experimental results demonstrate that our method is able to synthesize high-quality novel view global shutter images.} 
	\label{table_novel_view}
	\begin{center}
	\setlength\tabcolsep{5pt}
	\setlength{\belowcaptionskip}{0pt}
	\scriptsize
	% \resizebox{\linewidth}{!}{
		\begin{tabular}{c|ccc|ccc|ccc}
			\toprule
			&  \multicolumn{3}{|c}{Blue Room}  &  \multicolumn{3}{|c}{Living Room}  &  \multicolumn{3}{|c}{White Room} \\
			& PSNR$\uparrow$ & SSIM$\uparrow$ & LPIPS$\downarrow$ & PSNR$\uparrow$ & SSIM$\uparrow$ & LPIPS$\downarrow$ & PSNR$\uparrow$ & SSIM$\uparrow$ & LPIPS$\downarrow$\\
			\midrule
			NeRF+DiffSfM\citep{zhuang2017sfm_RS} &13.87 &0.420 &0.7271 &13.35 &0.398 &0.6415 &14.11 &0.467 &0.5926 \\
			NeRF+DSUN \citep{liu2020deepunroll} &18.08 &0.550 &0.4391 &17.47 &0.574 &0.5005 &18.96 &0.592 &0.3821\\
			NeRF+SUNet \citep{fan2021sunet} &19.52 &0.641 &0.3732 &18.69 &0.662 &0.4031 &14.87 &0.412 &0.5885\\
			NeRF+RSSR \citep{fan2021RSSR} &17.86 &0.566 &0.3175 &18.73 &0.651 &0.3628 &11.83 &0.296 &0.6422\\
			NeRF+CVR \citep{fan2022CVR} &15.94 &0.464 &0.4968 &15.88 &0.503 &0.5753 &20.67 &0.703 &0.2506\\
			NeRF \citep{mildenhall2020nerf} &18.56 &0.562 &0.3820 &15.62 &0.446 &0.6098 &15.60 &0.426 &0.5616\\
			BARF \citep{lin2021barf} &19.13 & 0.540 & 0.4592 & 16.03 & 0.544 & 0.5296 & 12.94 & 0.389 & 0.7600 \\
			\specialrule{0.08em}{1pt}{1pt}
			USB-NeRF (ours) &{\bf 28.99} &{\bf 0.886} &{\bf 0.0757} &{\bf 33.33} &{\bf 0.936} &{\bf 0.0468} &{\bf 29.11} &{\bf 0.889} &{\bf 0.0598}\\
			\specialrule{0.08em}{1pt}{1pt}
		\end{tabular}
		\vspace{1em}
		% }
	
	\end{center}
\end{table}

\begin{figure}[!ht]
	\setlength\tabcolsep{1.pt}
	\centering
	\begin{tabular}{rcccccc}
		%		\raisebox{.24in}{\rotatebox[origin=t]{90}{\normalsize DSUN \citep{liu2020deepunroll}}}
		\raisebox{.22in}{\rotatebox[origin=t]{90}{\scriptsize DiffSfM}}
		&\includegraphics[width=0.16\textwidth]{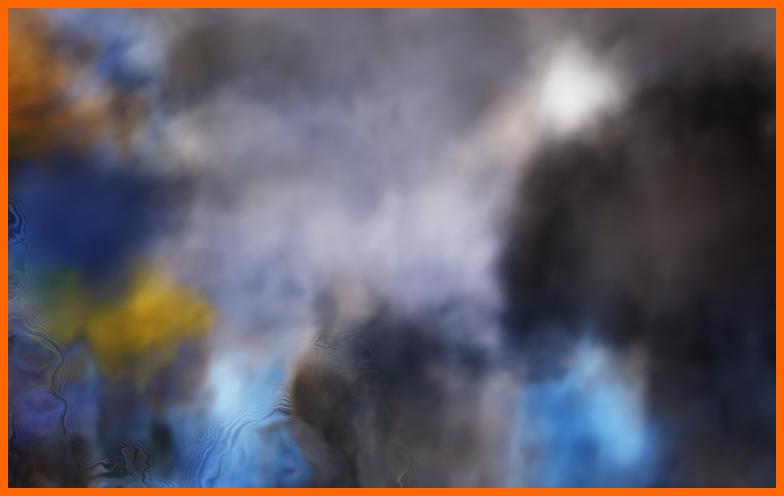} &
		\includegraphics[width=0.16\textwidth]{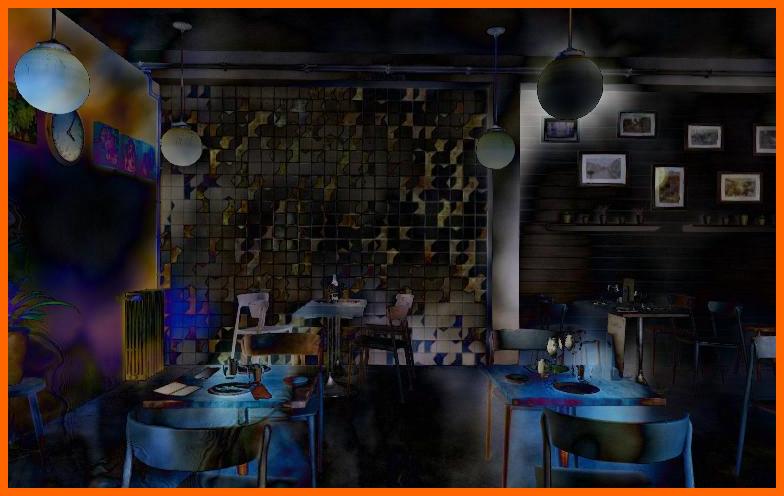} &
		\includegraphics[width=0.16\textwidth]{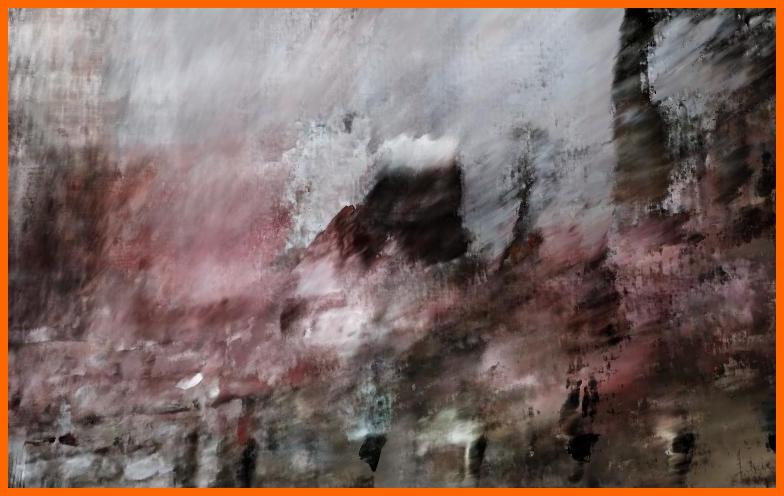} &
		\includegraphics[width=0.16\textwidth]{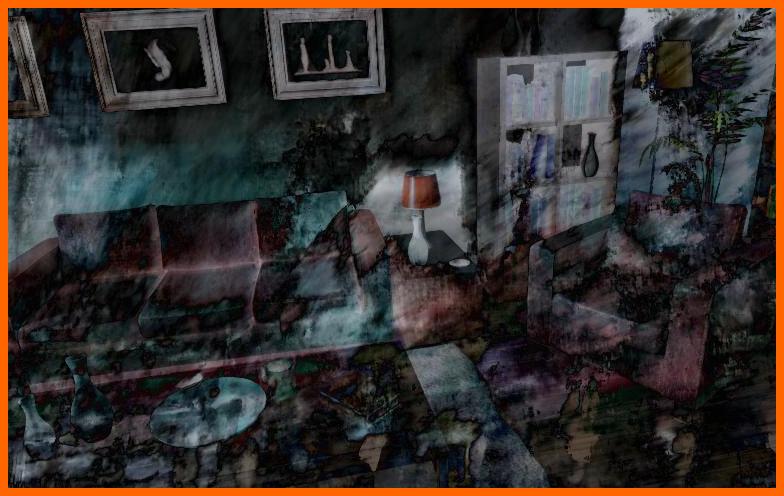} &
		\includegraphics[width=0.16\textwidth]{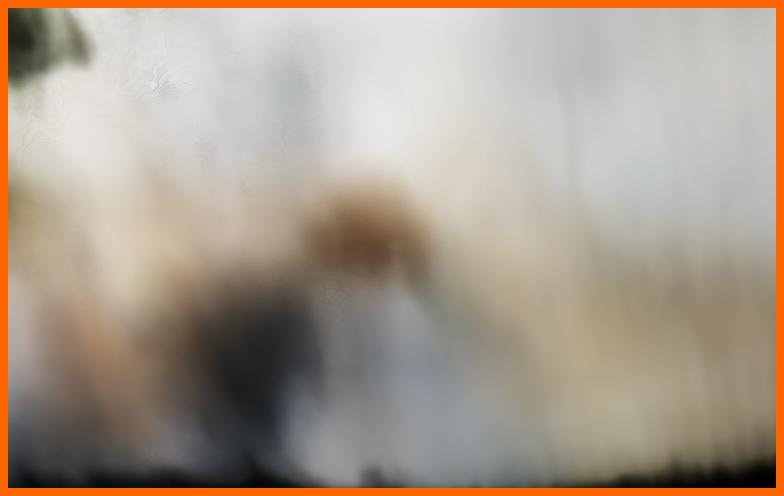} &
		\includegraphics[width=0.16\textwidth]{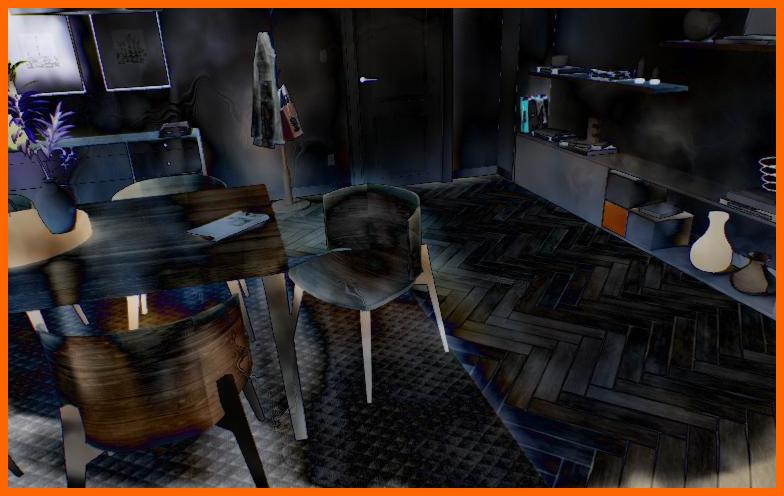} \\
		\specialrule{-0.1em}{0.05pt}{0.05pt}
		\raisebox{.24in}{\rotatebox[origin=t]{90}{\scriptsize DSUN}}
		&\includegraphics[width=0.16\textwidth]{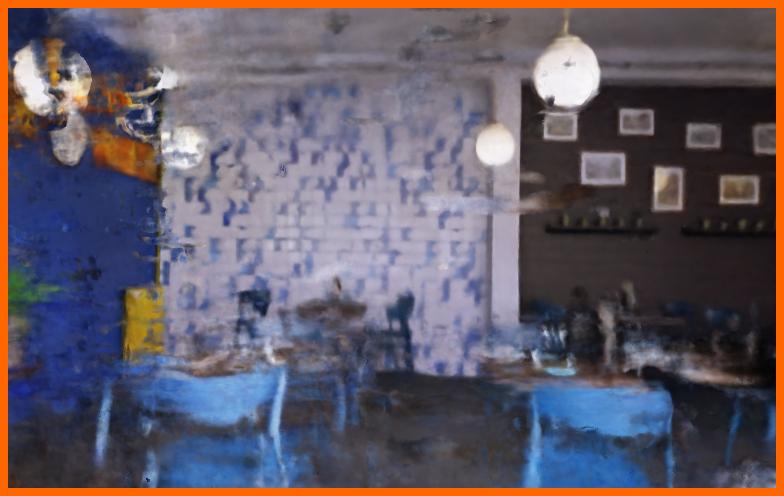} &
		\includegraphics[width=0.16\textwidth]{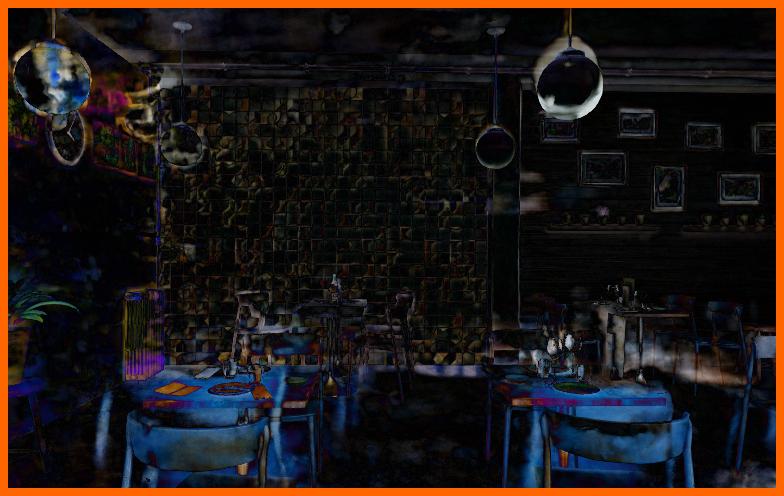} &
		\includegraphics[width=0.16\textwidth]{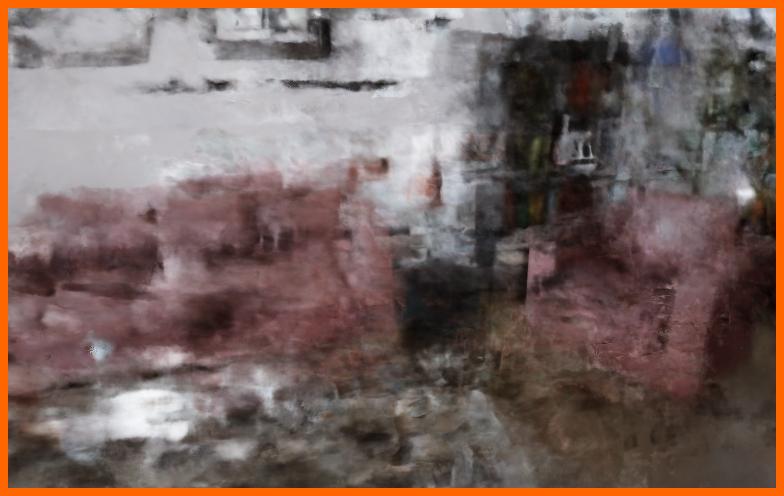} &
		\includegraphics[width=0.16\textwidth]{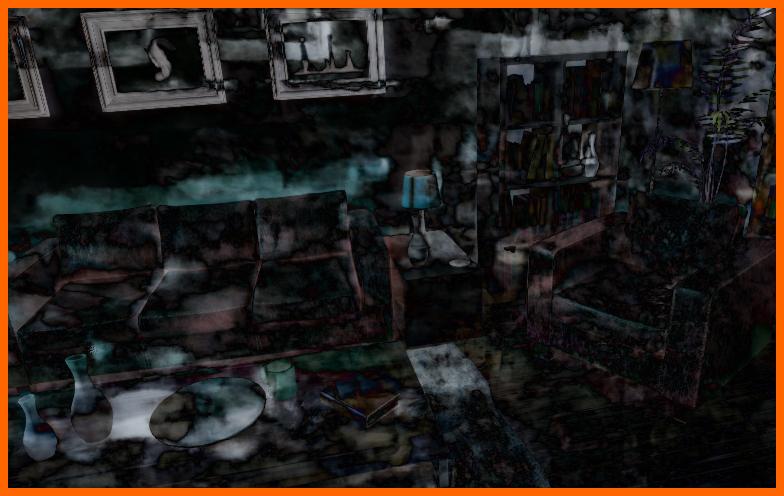} &
		\includegraphics[width=0.16\textwidth]{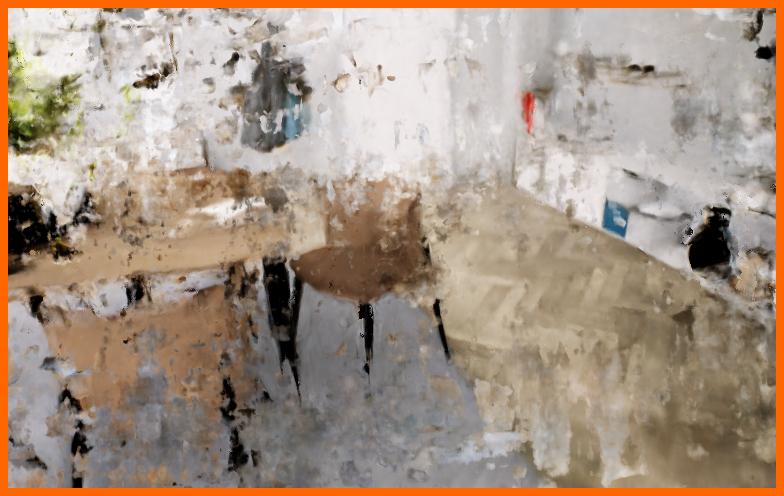} &
		\includegraphics[width=0.16\textwidth]{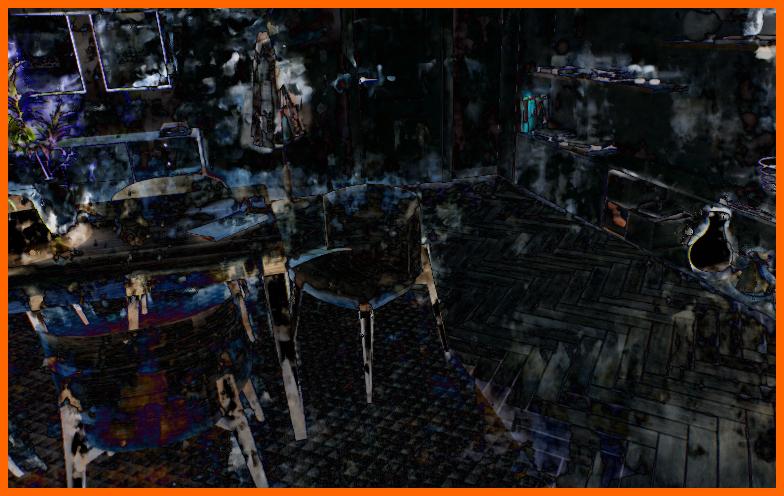} \\
		\specialrule{-0.1em}{0.05pt}{0.05pt}
		%		\raisebox{.24in}{\rotatebox[origin=t]{90}{\normalsize SUNet \citep{fan2021sunet}}}
		\raisebox{.24in}{\rotatebox[origin=t]{90}{\scriptsize SUNet}}
		&\includegraphics[width=0.16\textwidth]{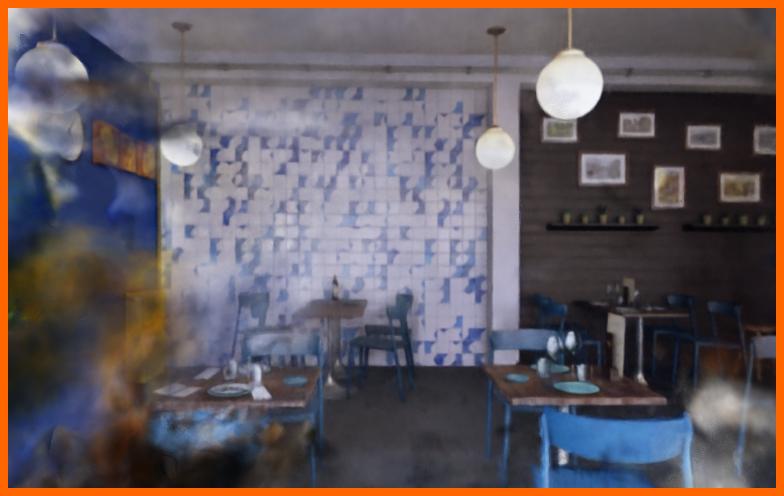} &
		\includegraphics[width=0.16\textwidth]{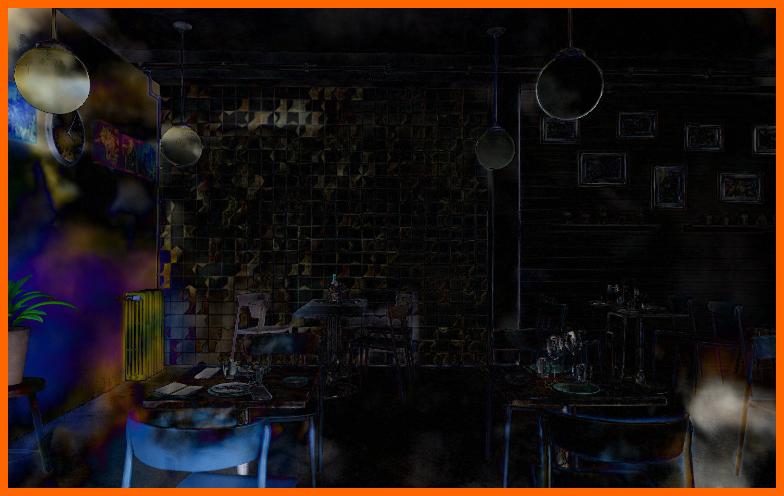} &
		\includegraphics[width=0.16\textwidth]{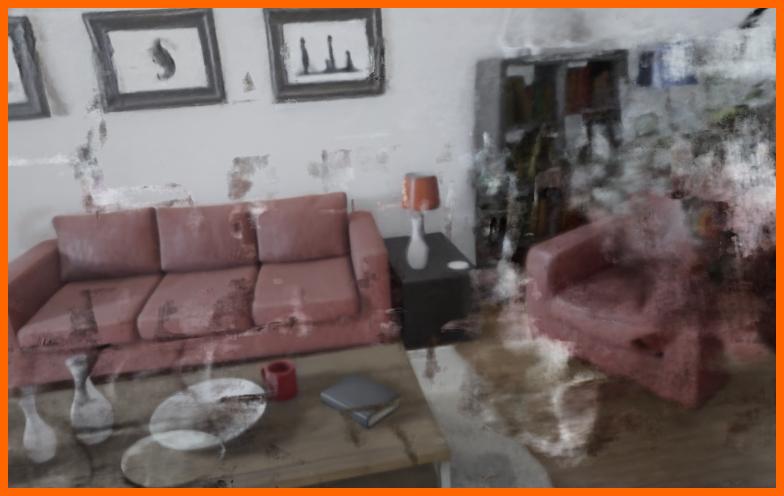} &
		\includegraphics[width=0.16\textwidth]{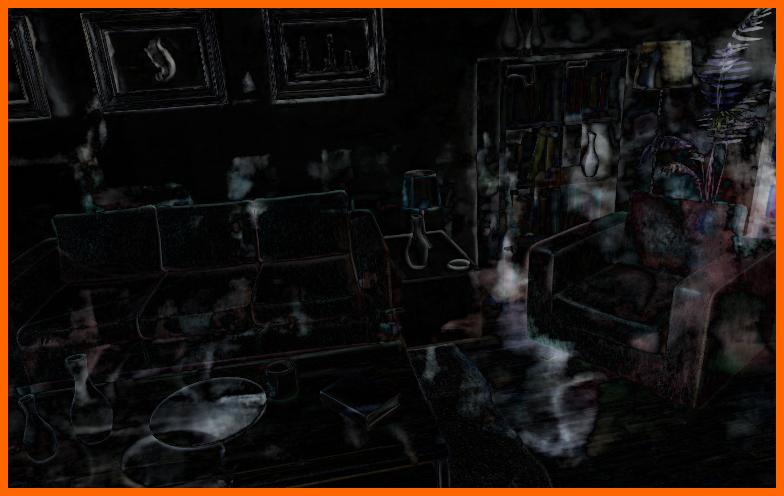} &
		\includegraphics[width=0.16\textwidth]{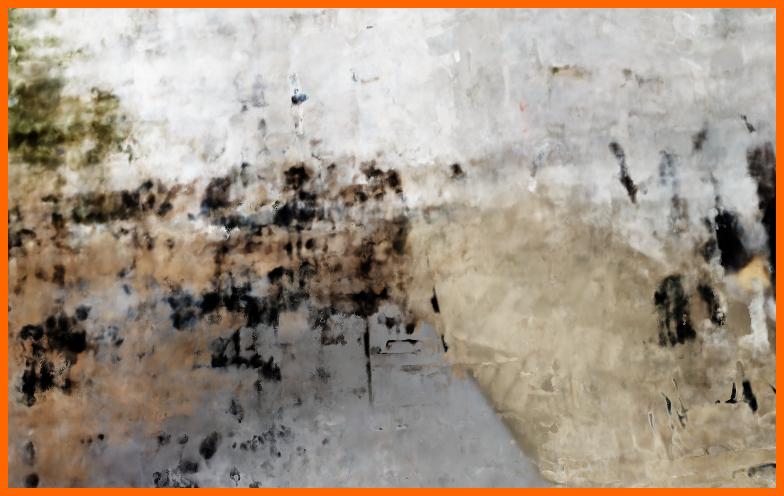} &
		\includegraphics[width=0.16\textwidth]{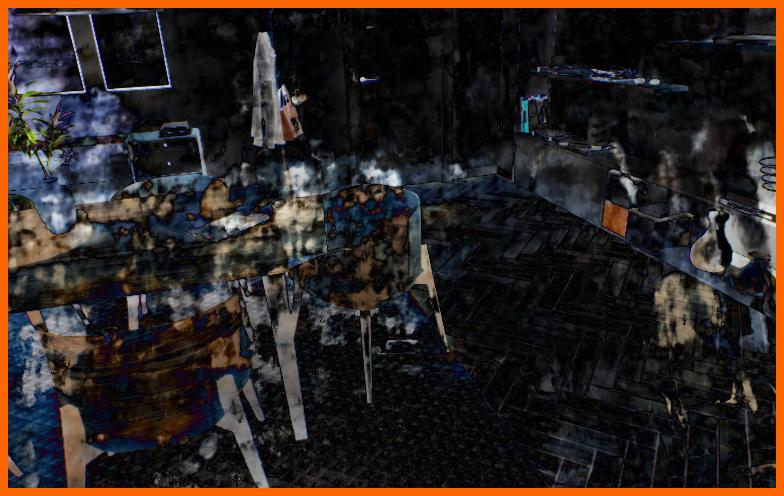} \\
		\specialrule{-0.1em}{0.05pt}{0.05pt}
		%		\raisebox{.25in}{\rotatebox[origin=t]{90}{\normalsize RSSR \citep{fan2021RSSR}}}
		\raisebox{.25in}{\rotatebox[origin=t]{90}{\scriptsize RSSR}}
		&\includegraphics[width=0.16\textwidth]{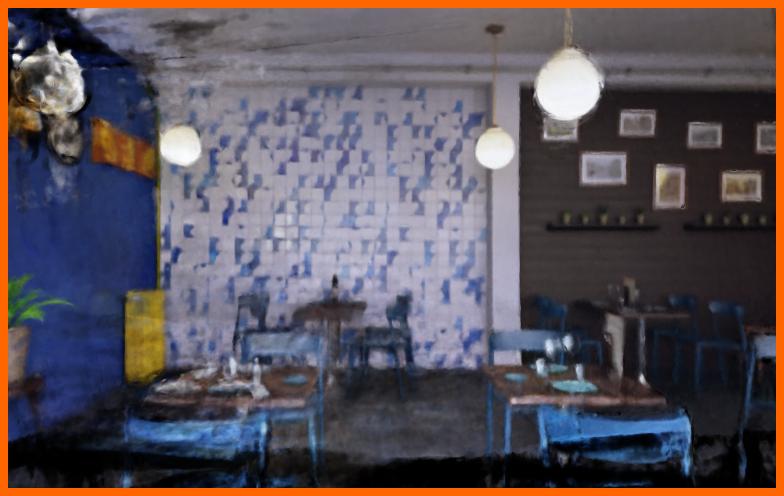} &
		\includegraphics[width=0.16\textwidth]{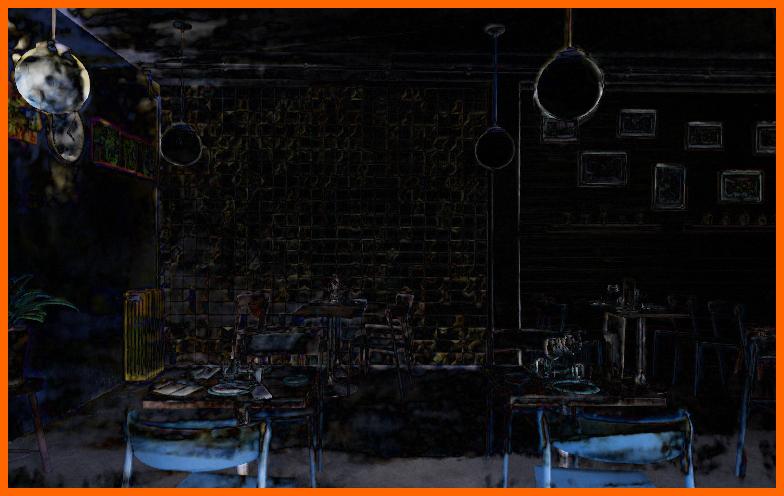} &
		\includegraphics[width=0.16\textwidth]{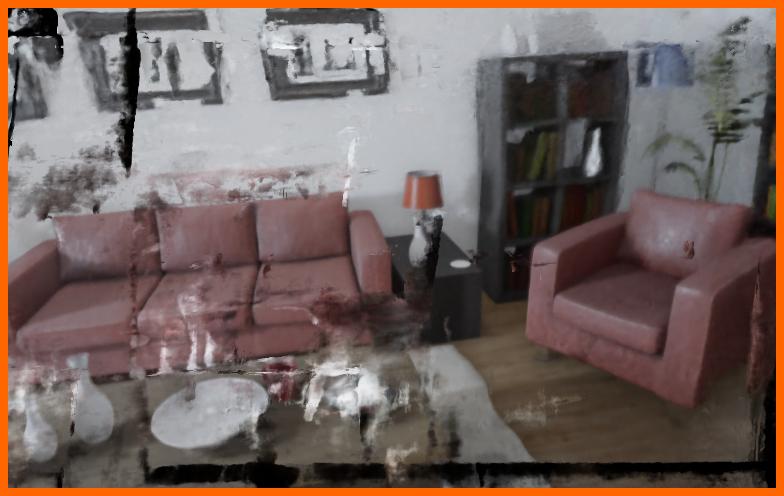} &
		\includegraphics[width=0.16\textwidth]{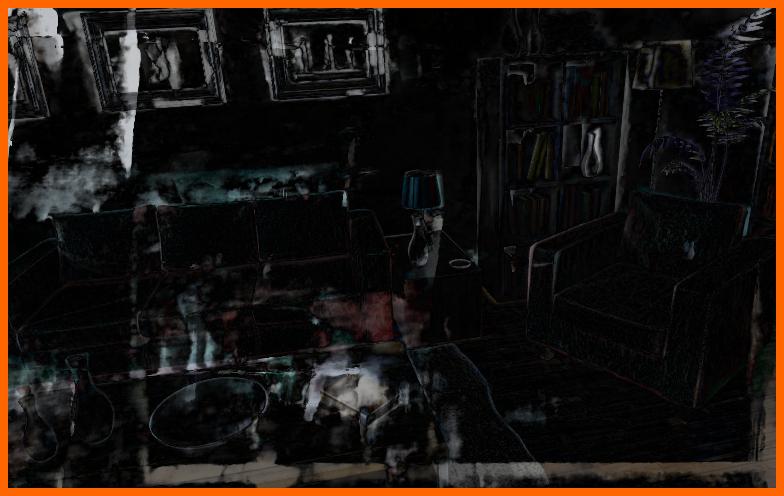} &
		\includegraphics[width=0.16\textwidth]{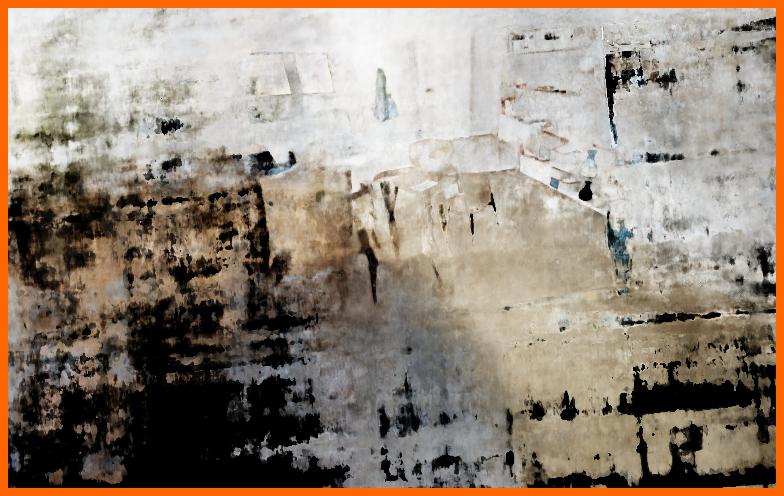} &
		\includegraphics[width=0.16\textwidth]{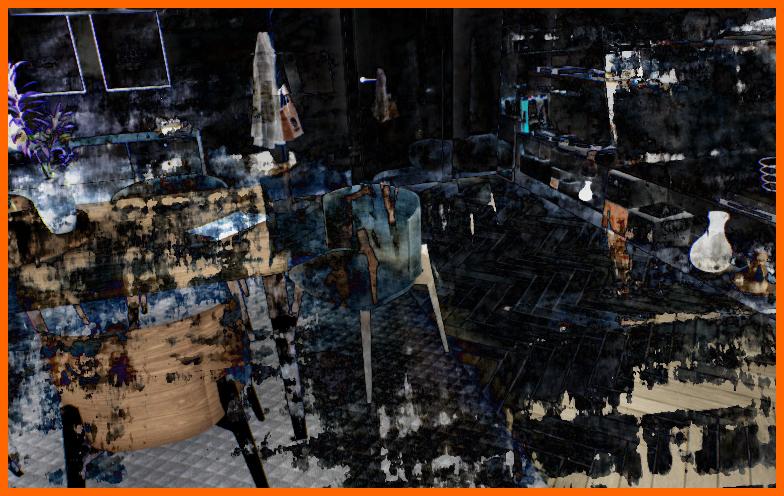} \\
		\specialrule{-0.1em}{0.05pt}{0.05pt}
		%		\raisebox{.23in}{\rotatebox[origin=t]{90}{\normalsize CVR\citep{fan2022CVR}}}
		\raisebox{.23in}{\rotatebox[origin=t]{90}{\scriptsize CVR}}
		&\includegraphics[width=0.16\textwidth]{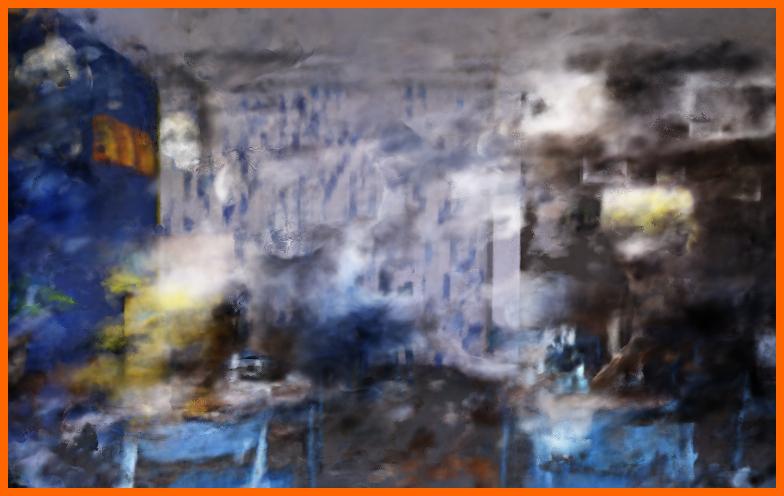} &
		\includegraphics[width=0.16\textwidth]{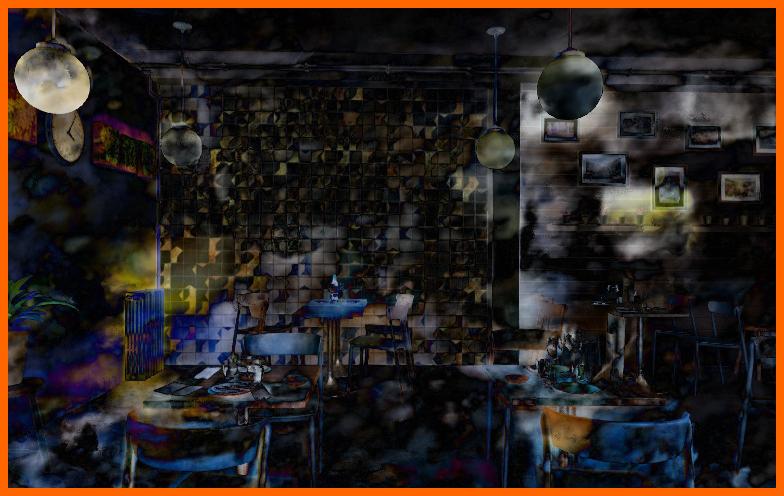} &
		\includegraphics[width=0.16\textwidth]{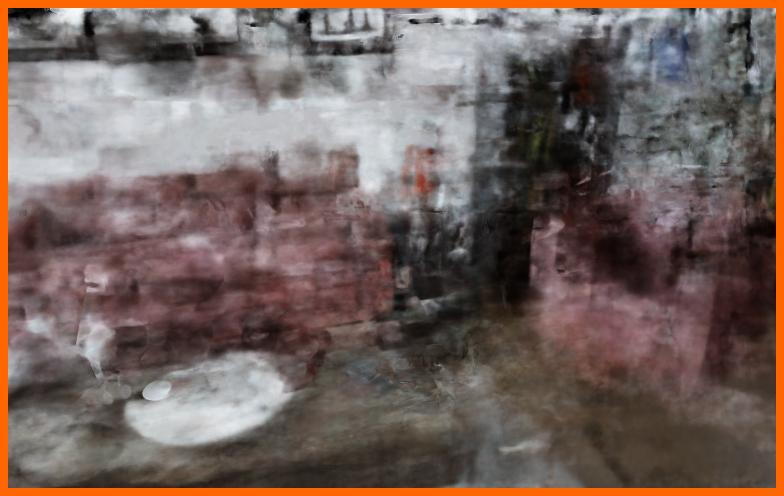} &
		\includegraphics[width=0.16\textwidth]{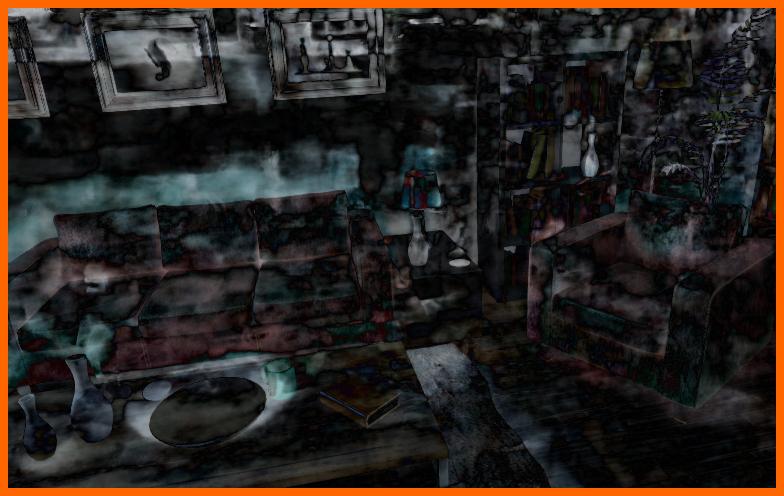} &
		\includegraphics[width=0.16\textwidth]{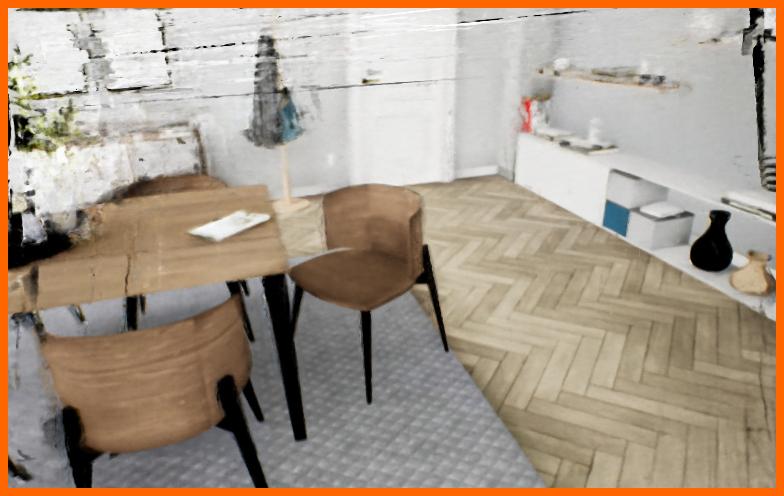} &
		\includegraphics[width=0.16\textwidth]{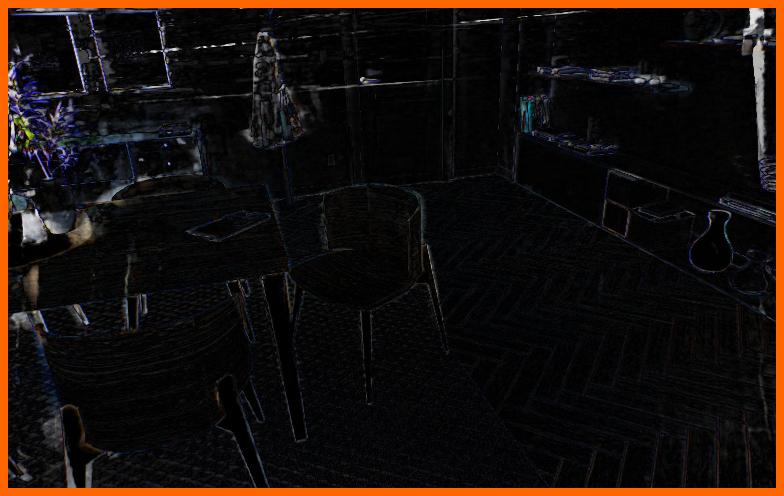} \\
		\specialrule{-0.1em}{0.05pt}{0.05pt}
		%		\raisebox{.25in}{\rotatebox[origin=t]{90}{\normalsize NeRF \citep{mildenhall2020nerf}}}
		\raisebox{.25in}{\rotatebox[origin=t]{90}{\scriptsize NeRF}}
		&\includegraphics[width=0.16\textwidth]{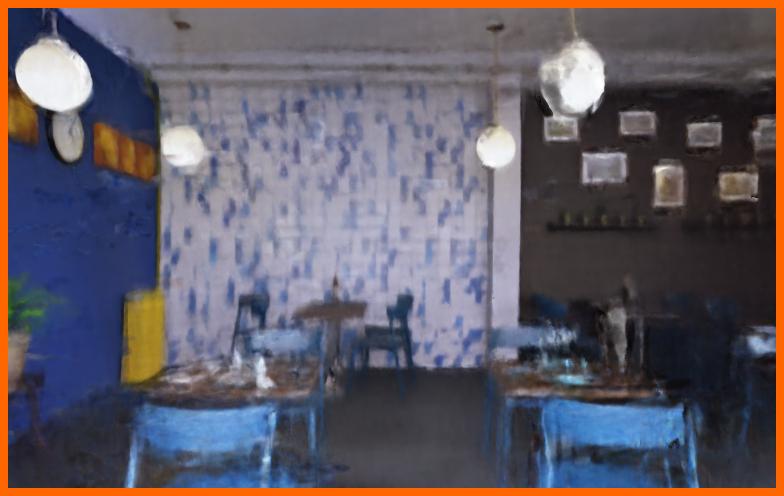} &
		\includegraphics[width=0.16\textwidth]{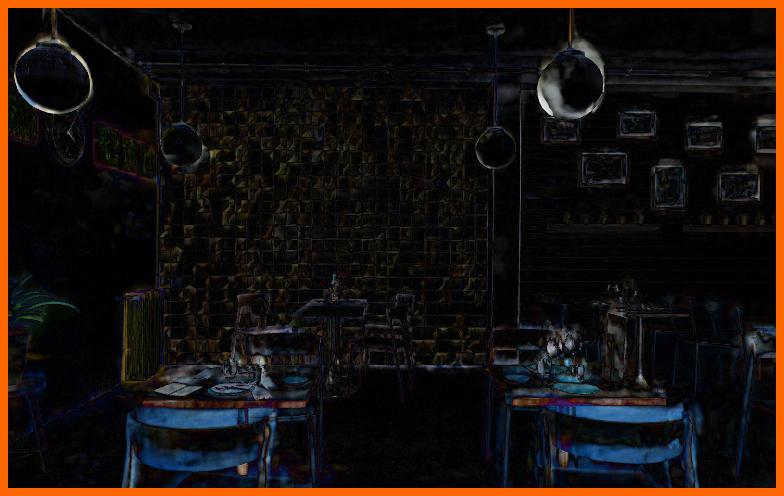} &
		\includegraphics[width=0.16\textwidth]{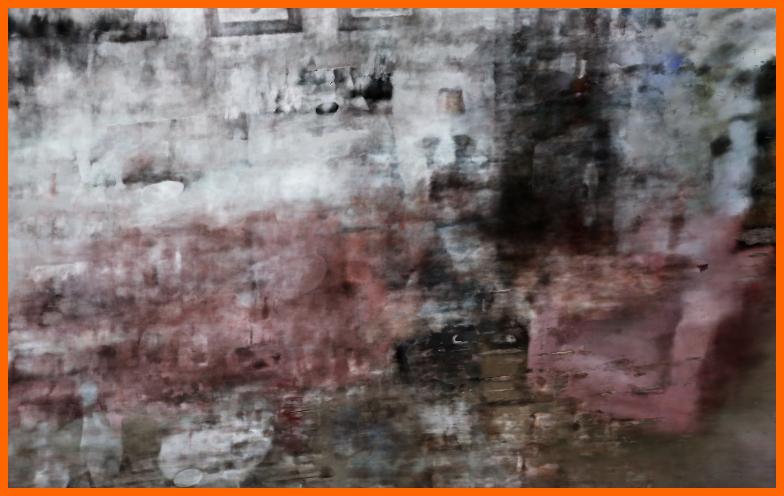} &
		\includegraphics[width=0.16\textwidth]{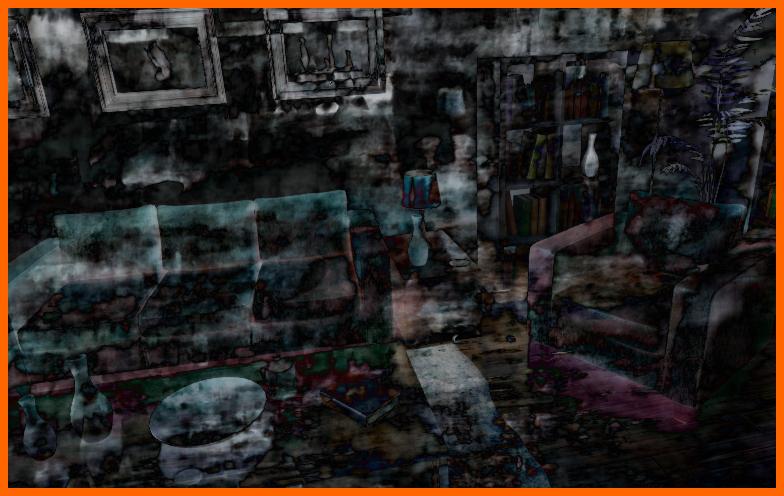} &
		\includegraphics[width=0.16\textwidth]{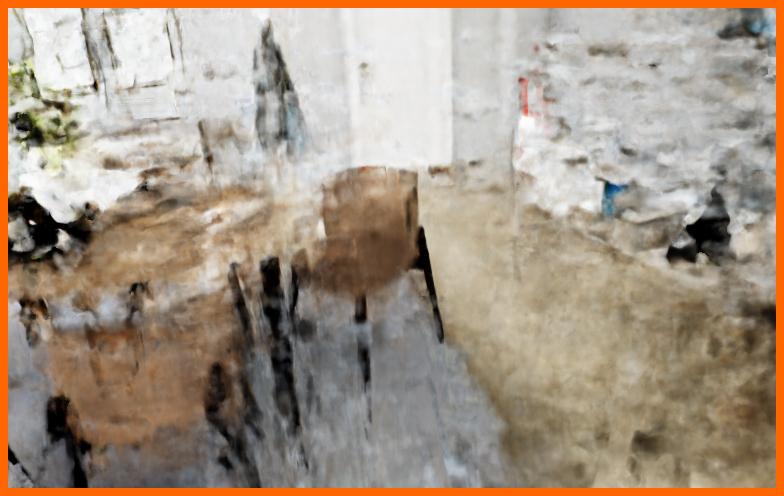} &
		\includegraphics[width=0.16\textwidth]{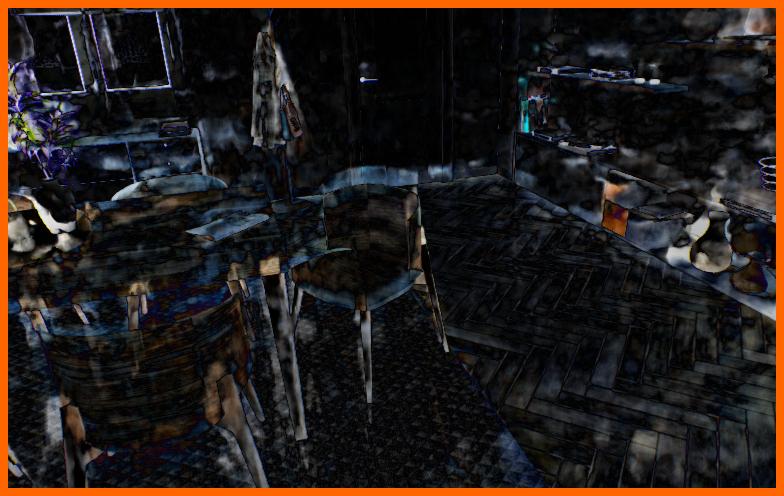} \\
		\specialrule{-0.1em}{0.05pt}{0.05pt}
		%		\raisebox{.25in}{\rotatebox[origin=t]{90}{\normalsize BARF \citep{lin2021barf}}}
		\raisebox{.25in}{\rotatebox[origin=t]{90}{\scriptsize BARF}}
		&\includegraphics[width=0.16\textwidth]{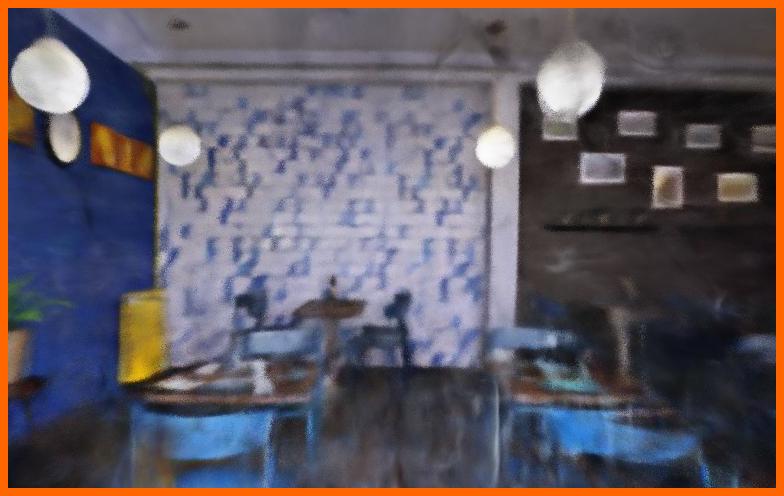} &
		\includegraphics[width=0.16\textwidth]{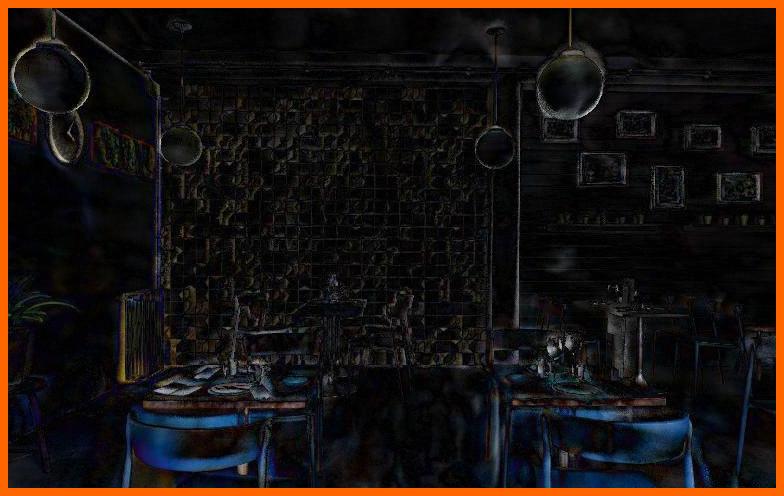} &
		\includegraphics[width=0.16\textwidth]{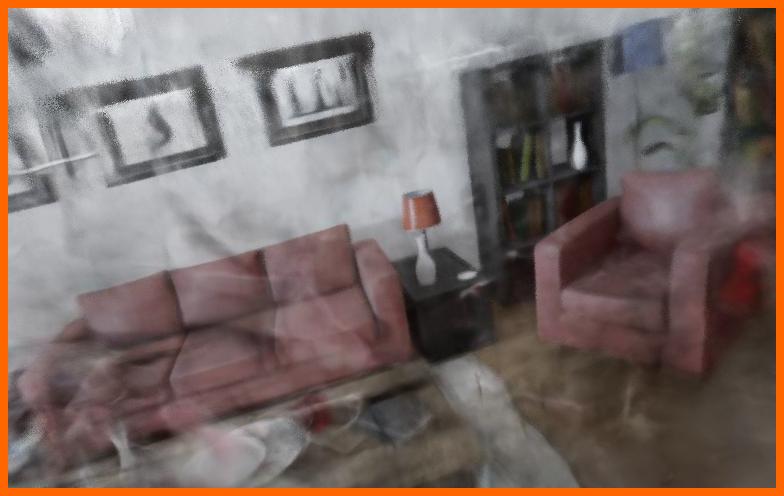} &
		\includegraphics[width=0.16\textwidth]{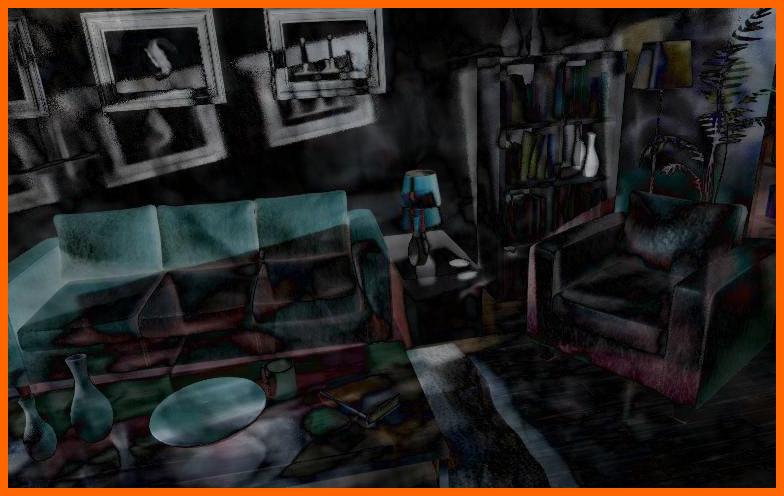} &
		\includegraphics[width=0.16\textwidth]{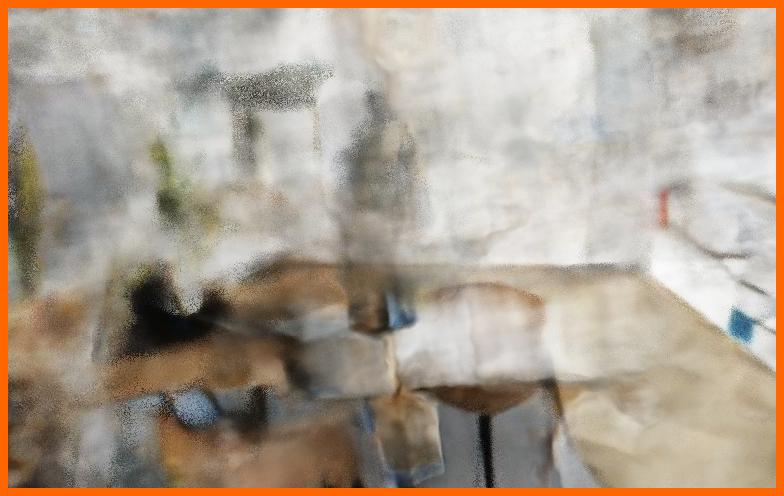} &
		\includegraphics[width=0.16\textwidth]{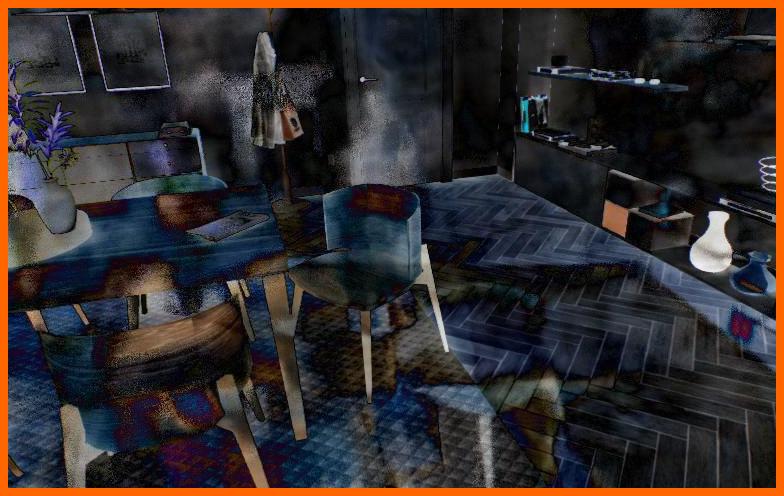} \\
		\specialrule{-0.1em}{0.05pt}{0.05pt}
		\raisebox{.25in}{\rotatebox[origin=t]{90}{\scriptsize USB-NeRF}}
		&\includegraphics[width=0.16\textwidth]{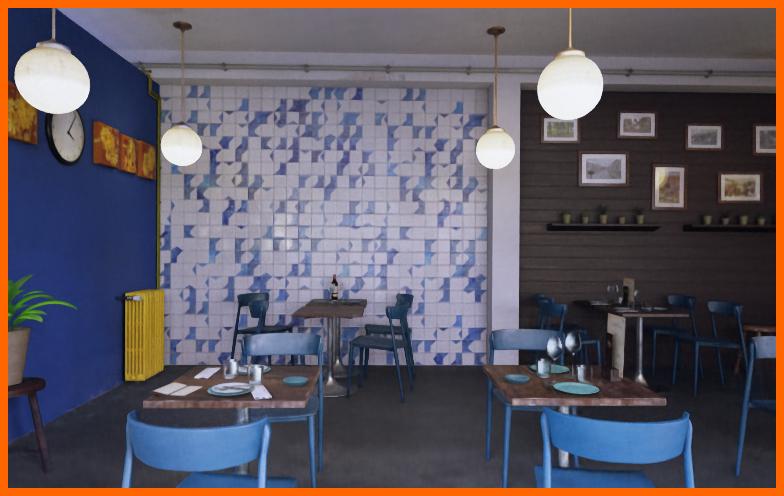} &
		\includegraphics[width=0.16\textwidth]{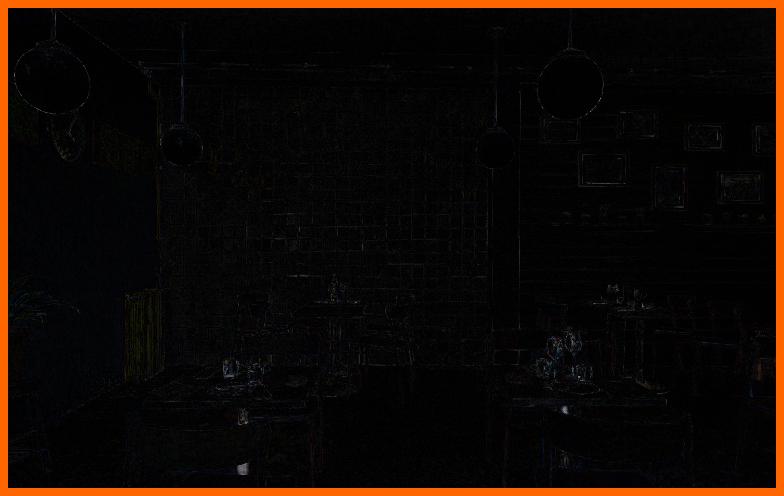} &
		\includegraphics[width=0.16\textwidth]{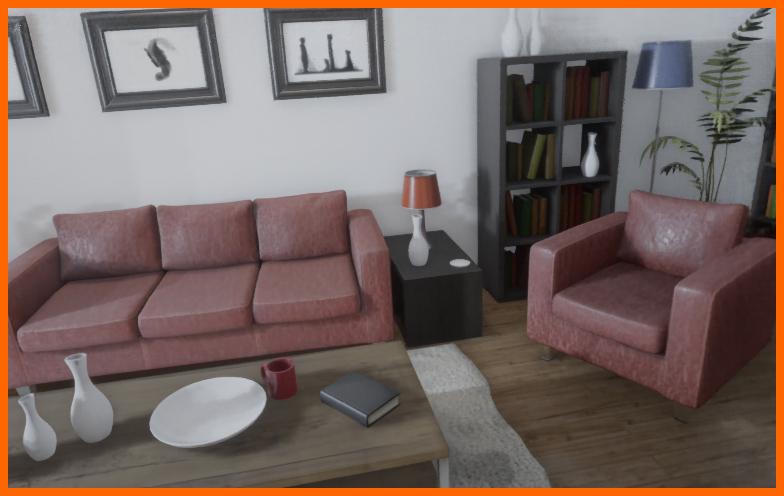} &
		\includegraphics[width=0.16\textwidth]{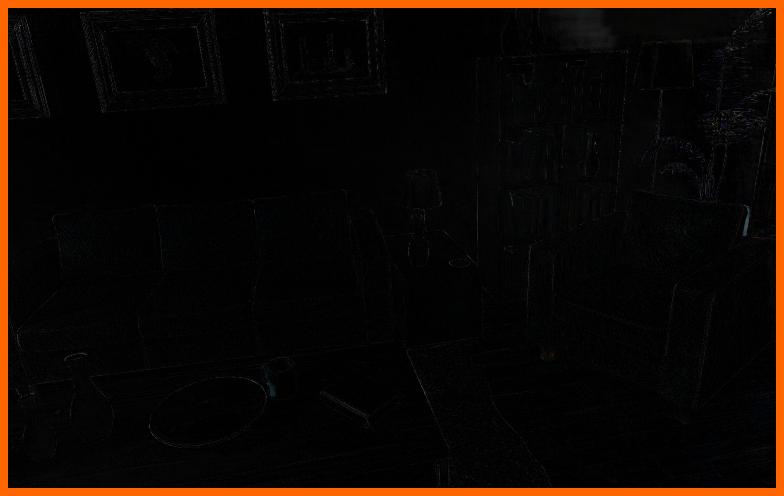} &
		\includegraphics[width=0.16\textwidth]{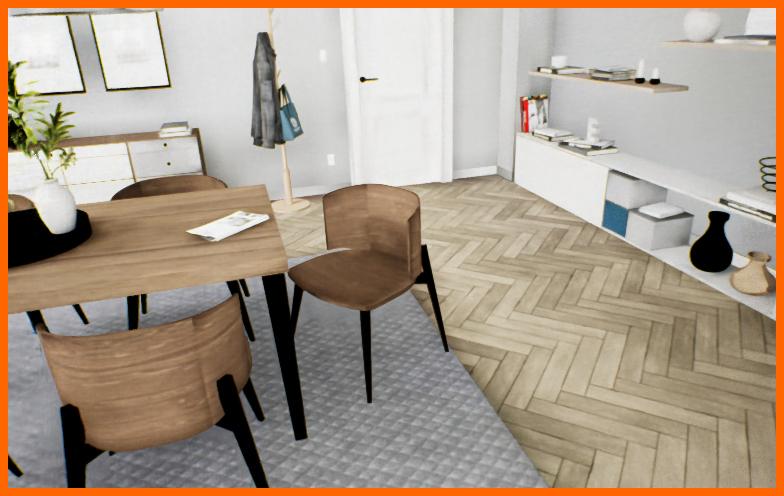} &
		\includegraphics[width=0.16\textwidth]{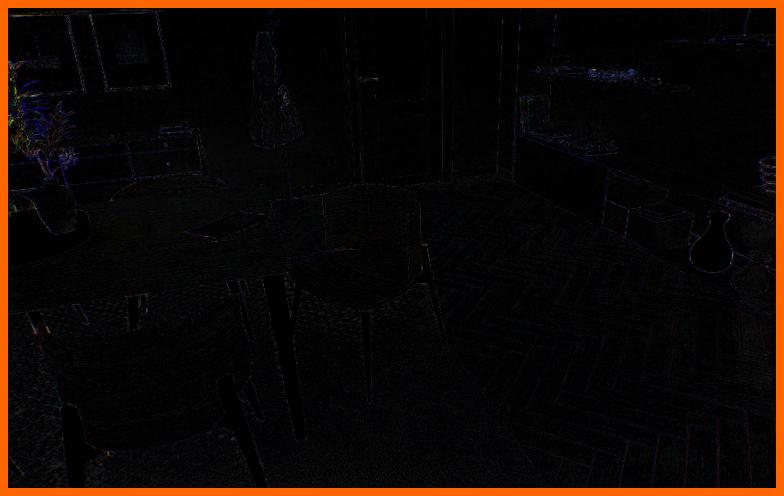} \\
		\specialrule{0em}{0.05pt}{0.05pt}
	\end{tabular}
%	\vspace{-0.5em}
	\caption{{\bf{Qualitative comparisons with synthetic datasets for novel view synthesis.}} The experimental results demonstrate the ability of our approach to synthesize novel view images in good quality. The even columns present the error residual images between the rendered and ground-truth global shutter images. The darker the better. }
	\label{fig_novel_view}
%	\vspace{-0.6em}
\end{figure}

\begin{figure}[!ht]
	\setlength\tabcolsep{1.pt}
	\centering
	\begin{tabular}{ccccc}
		\includegraphics[width=0.19\textwidth]{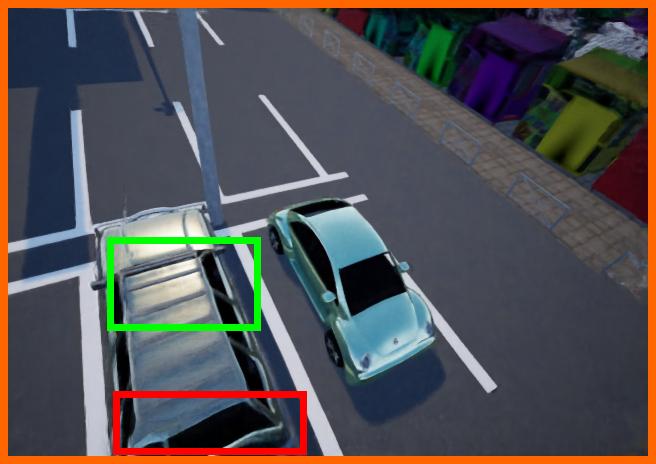} &
		\includegraphics[width=0.19\textwidth]{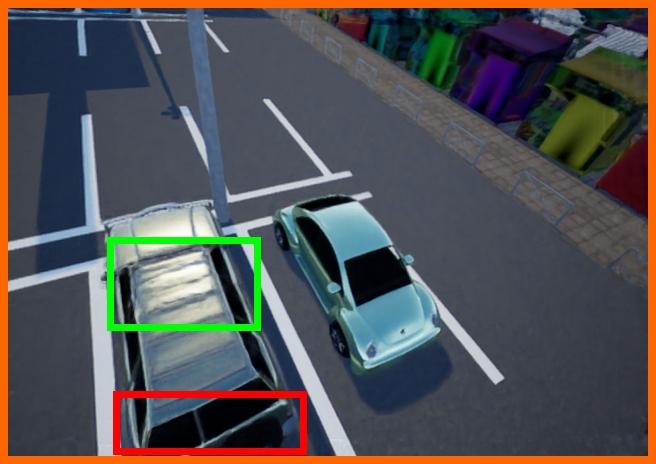} &
		\includegraphics[width=0.19\textwidth]{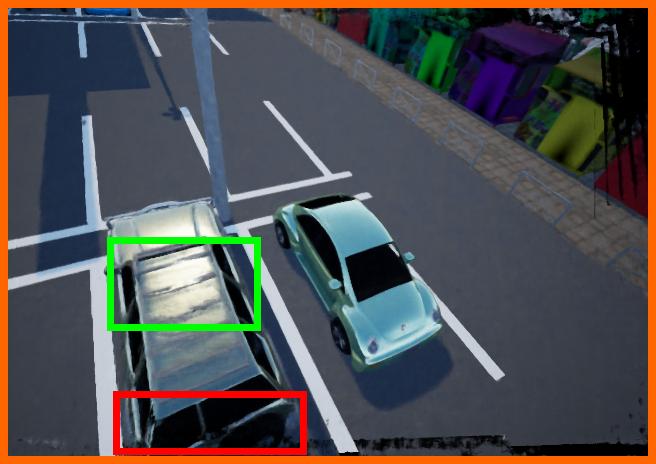} &
		\includegraphics[width=0.19\textwidth]{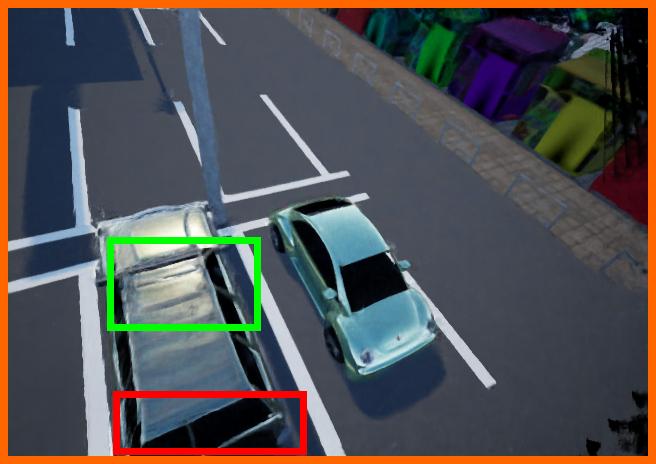} &
		\includegraphics[width=0.19\textwidth]{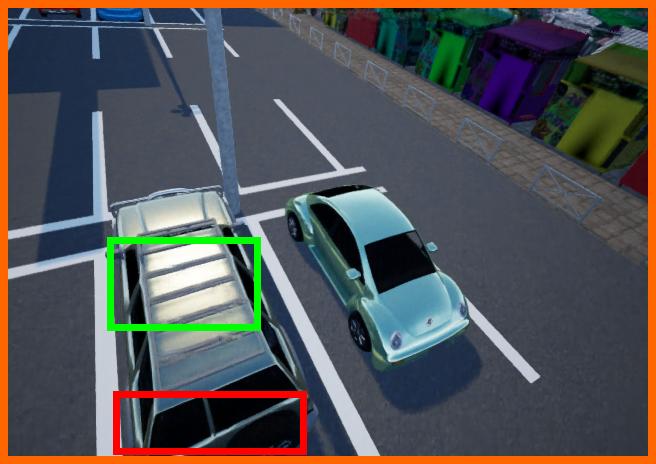} \\
		\specialrule{-0.1em}{0.05pt}{0.05pt}
		\includegraphics[width=0.19\textwidth]{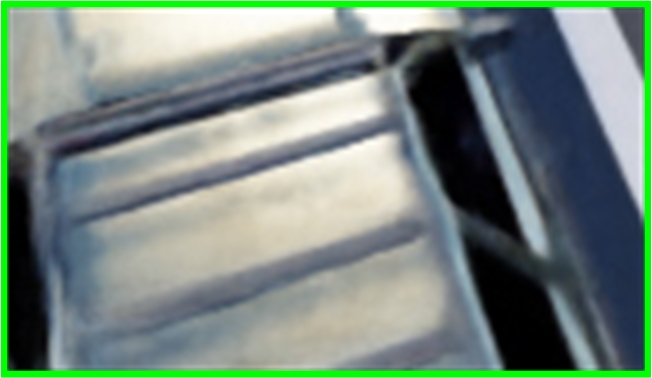} &
		\includegraphics[width=0.19\textwidth]{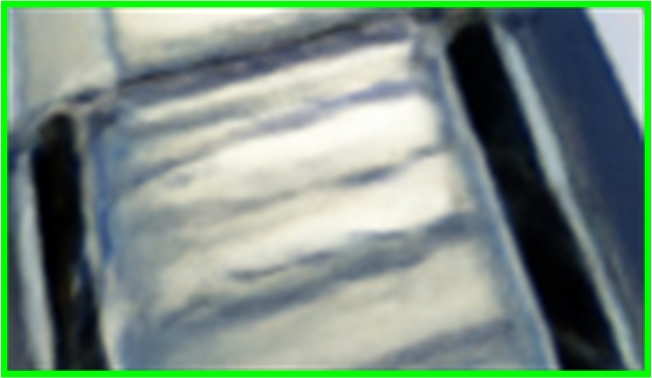} &
		\includegraphics[width=0.19\textwidth]{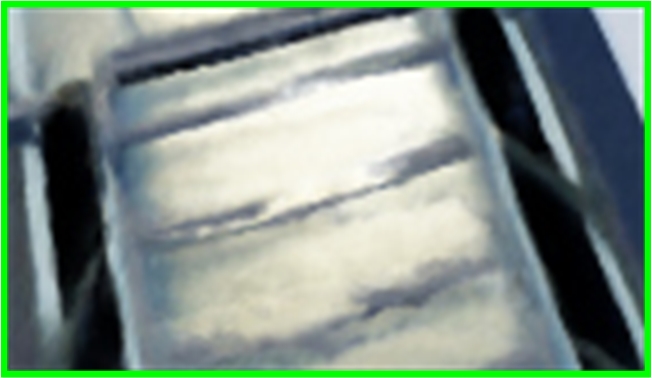} &
		\includegraphics[width=0.19\textwidth]{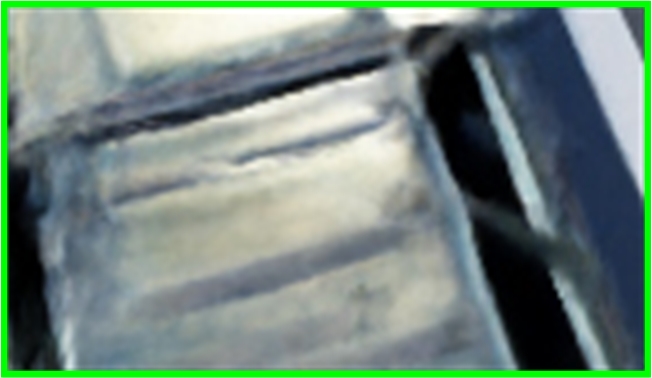} &
		\includegraphics[width=0.19\textwidth]{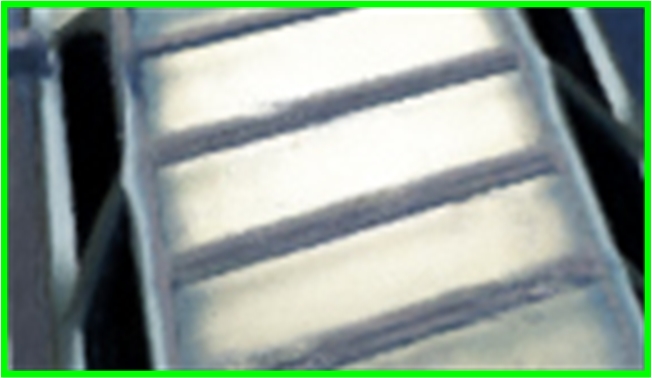} \\
		\specialrule{-0.1em}{0.05pt}{0.05pt}
		\includegraphics[width=0.19\textwidth]{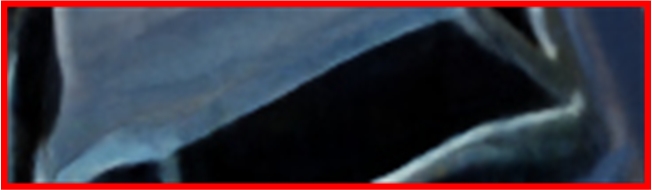} &
		\includegraphics[width=0.19\textwidth]{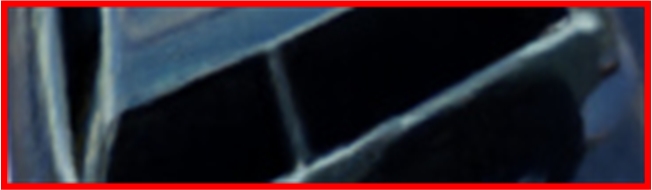} &
		\includegraphics[width=0.19\textwidth]{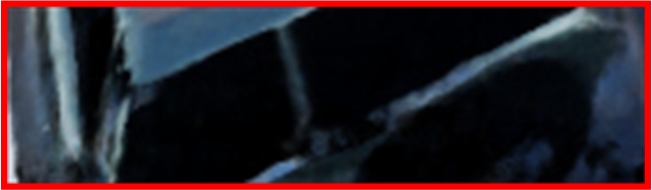} &
		\includegraphics[width=0.19\textwidth]{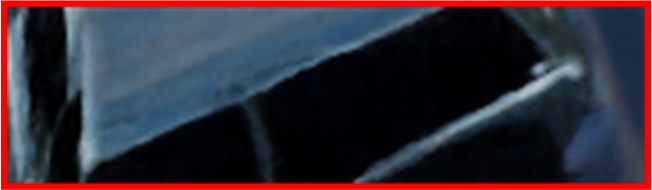} &
		\includegraphics[width=0.19\textwidth]{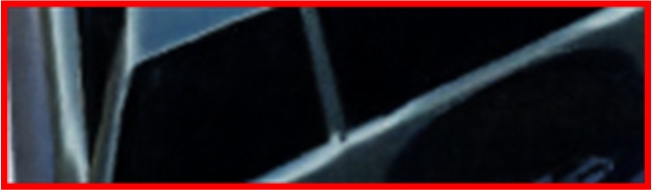} \\
		\specialrule{0em}{0.05pt}{0.05pt}
		%		NeRF+DSUN \citep{liu2020deepunroll} &  NeRF+SUNet \citep{fan2021sunet} & NeRF+RSSR \citep{fan2021RSSR} & NeRF+CVR\citep{fan2022CVR} & USB-NeRF
		NeRF+DSUN &  NeRF+SUNet & NeRF+RSSR & NeRF+CVR & USB-NeRF
	\end{tabular}
%	\vspace{-0.5em}
	\captionsetup {font={small,stretch=0.5}}
	\caption{{\bf{Qualitative comparisons with Carla-RS datasets for novel view synthesis.}} The experimental results demonstrate the ability of our approach to synthesize novel view images in good quality, by taking advantage of multi-view images.}
	\label{fig_carla_novel_view}
\end{figure}

\begin{table}
	\vspace{1em}
	\caption{{\textbf{Quantitative comparisons on the unorganized synthetic datasets.}} The experimental results demonstrate that our method also performs better than prior methods with un-ordered rolling shutter images, in terms of rolling shutter effect correction. }
	\label{table_unorganized}
%	\vspace{-0.5em}
	\setlength{\belowcaptionskip}{0pt}
	\scriptsize
	% \resizebox{\linewidth}{!}{
		\begin{tabular}{c|ccc|ccc|ccc}
			\toprule
			&  \multicolumn{3}{|c}{Blue Room}  &  \multicolumn{3}{|c}{Living Room}  &  \multicolumn{3}{|c}{Roof}\\
			& PSNR$\uparrow$ & SSIM$\uparrow$ & LPIPS$\downarrow$ & PSNR$\uparrow$ & SSIM$\uparrow$ & LPIPS$\downarrow$ & PSNR$\uparrow$ & SSIM$\uparrow$ & LPIPS$\downarrow$\\
			\midrule
			NeRF \citep{mildenhall2020nerf} &20.10 &0.601 &0.3177 &24.98 &0.784 &0.2364 &19.10 &0.489 &0.5200\\		
			BARF \citep{lin2021barf} &20.28 &0.591 &0.3608 &24.58 &0.776 &0.2533 &19.14 &0.493 &0.5234\\
			USB-NeRF &{\bf 31.13} & {\bf0.900} & {\bf0.0615} & {\bf32.46} & {\bf0.923} & {\bf0.0364} & {\bf27.26} & {\bf0.747} & {\bf0.1469}\\
%			USB-NeRF* &31.08 &{\bf 0.905} &{\bf 0.0540} &{\bf 35.18} &{\bf 0.950} &{\bf 0.0321} &25.46 &0.798 &0.1289 &{\bf 27.97} &{\bf 0.816} &{\bf 0.0805}\\
			%			USB-NeRF** &- &- &- &33.53 &0.936 &0.0336 &37.92 &0.970 &0.0139 &32.90 &0.940 &0.0266 &27.52 &0.845 &0.0613\\
			\specialrule{0.08em}{1pt}{1pt}
		\end{tabular}
		\vspace{-0.0em}
		% }
\end{table}

\begin{figure}[!ht]
	\vspace{1.em}
	\setlength\tabcolsep{1.pt}
	\centering
	\begin{tabular}{cccccc}
		\includegraphics[width=0.16\textwidth]{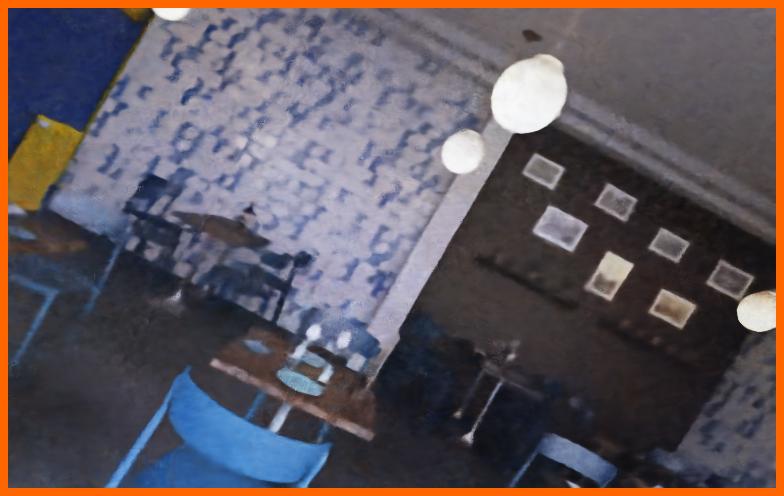} &
		\includegraphics[width=0.16\textwidth]{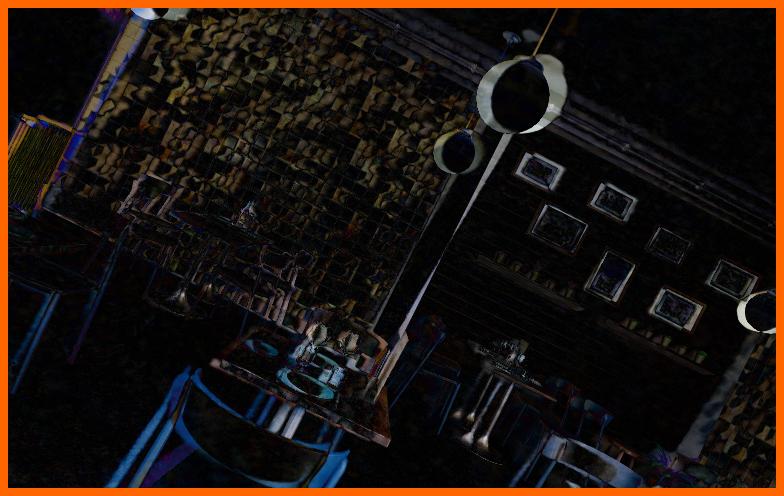} &
		\includegraphics[width=0.16\textwidth]{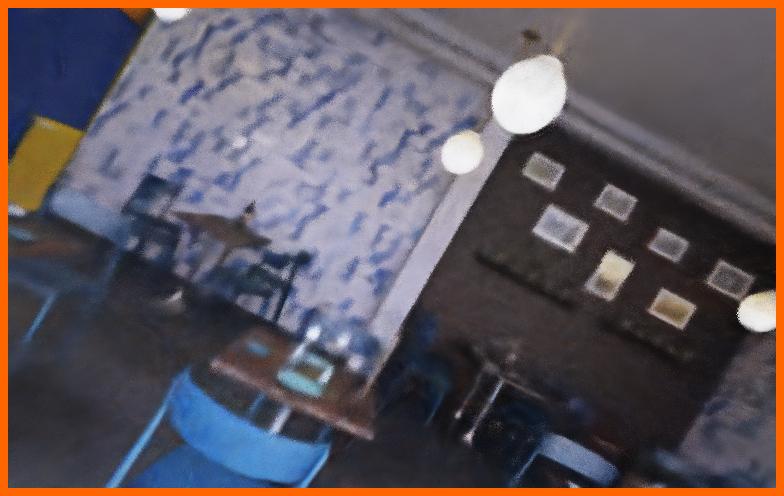} &
		\includegraphics[width=0.16\textwidth]{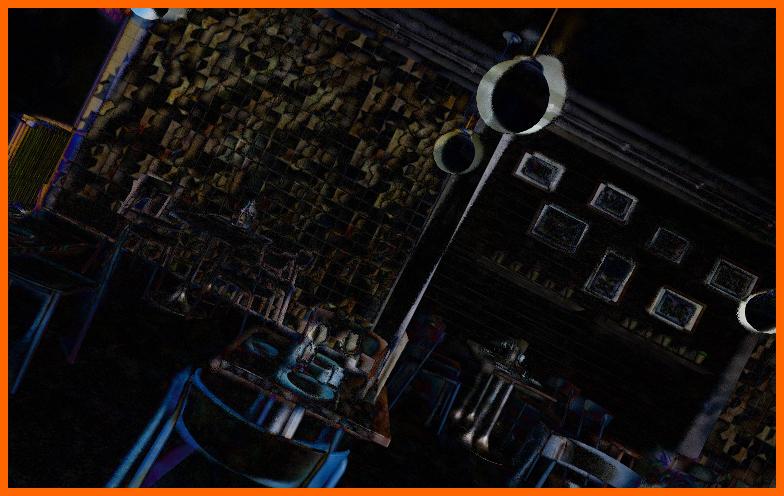} &
		\includegraphics[width=0.16\textwidth]{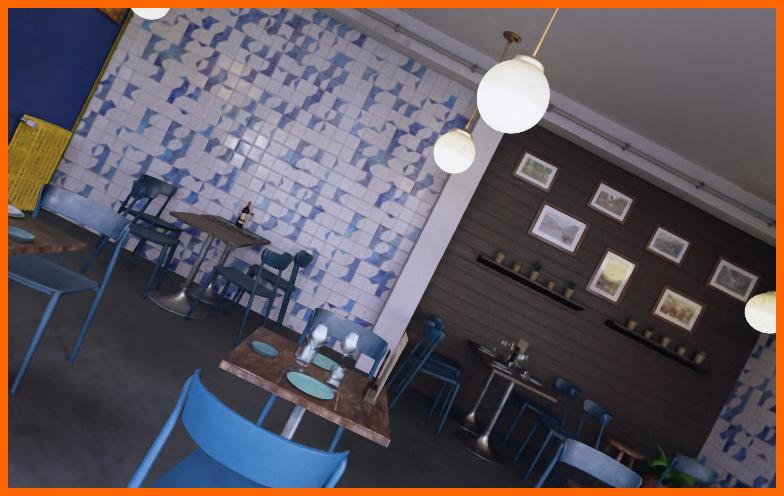} &
		\includegraphics[width=0.16\textwidth]{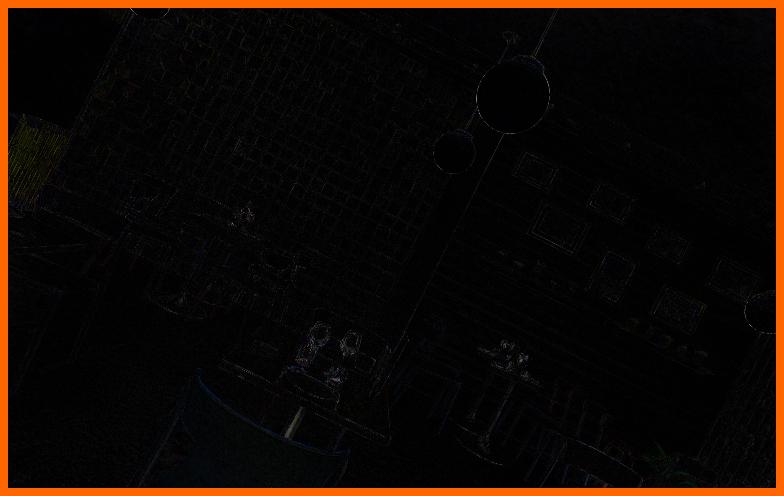} \\
		\specialrule{-0.1em}{0.05pt}{0.05pt}
		\includegraphics[width=0.16\textwidth]{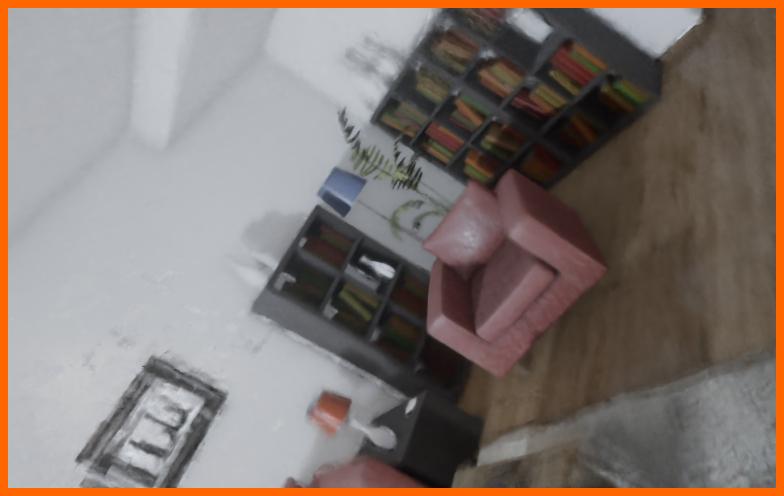} &
		\includegraphics[width=0.16\textwidth]{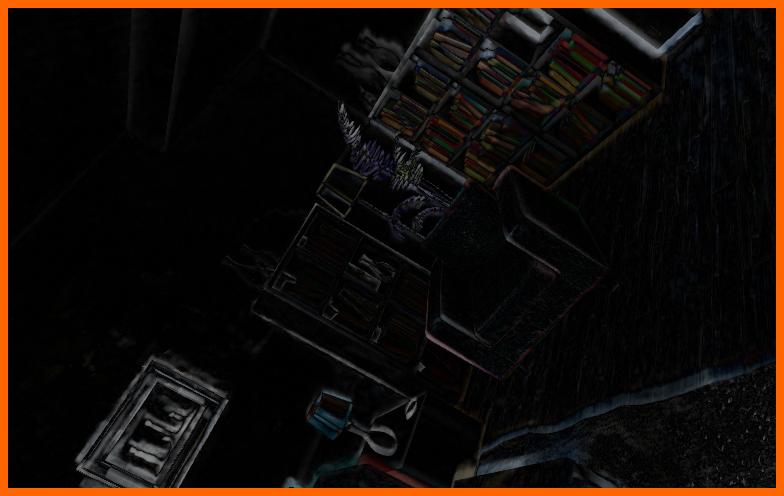} &
		\includegraphics[width=0.16\textwidth]{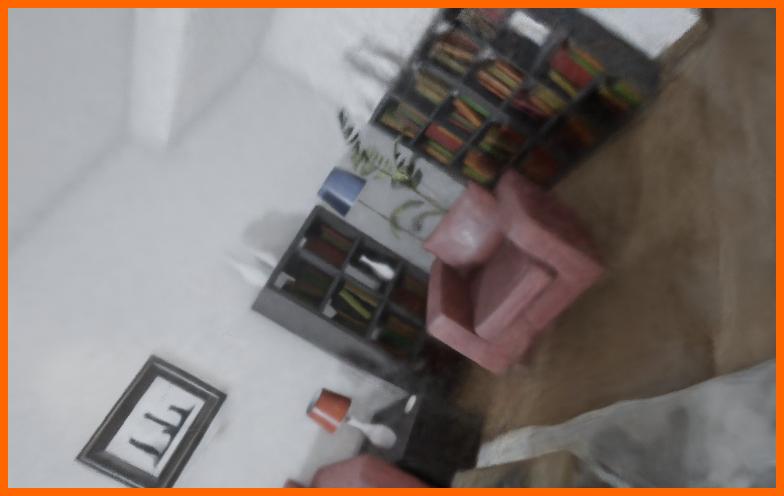} &
		\includegraphics[width=0.16\textwidth]{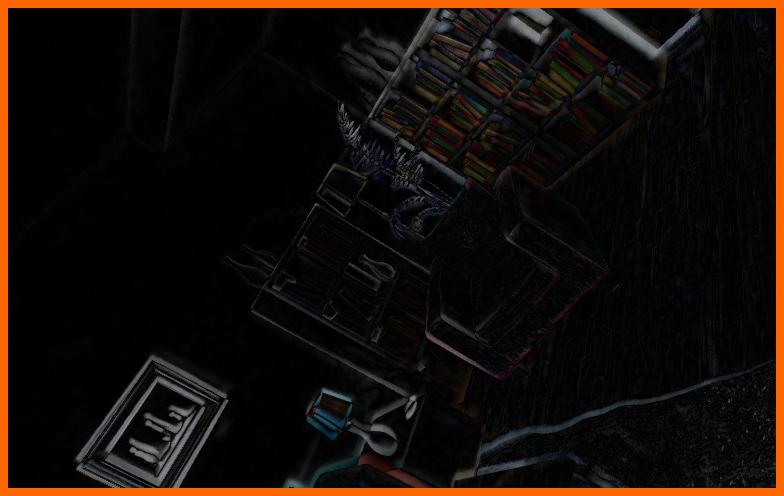} &
		\includegraphics[width=0.16\textwidth]{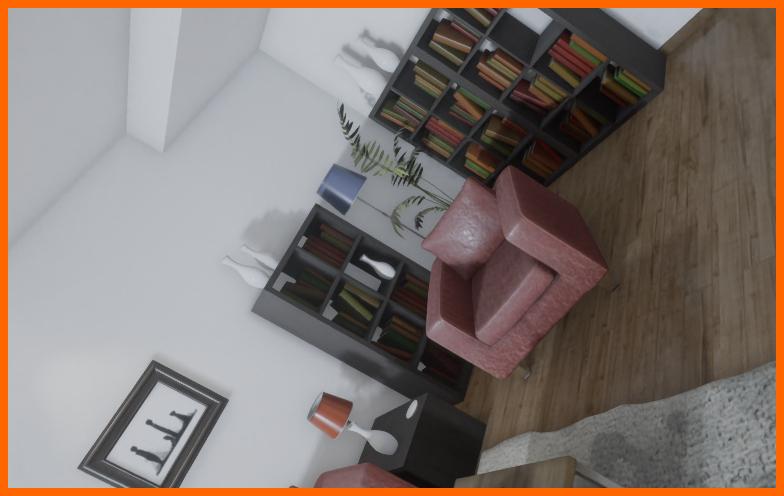} &
		\includegraphics[width=0.16\textwidth]{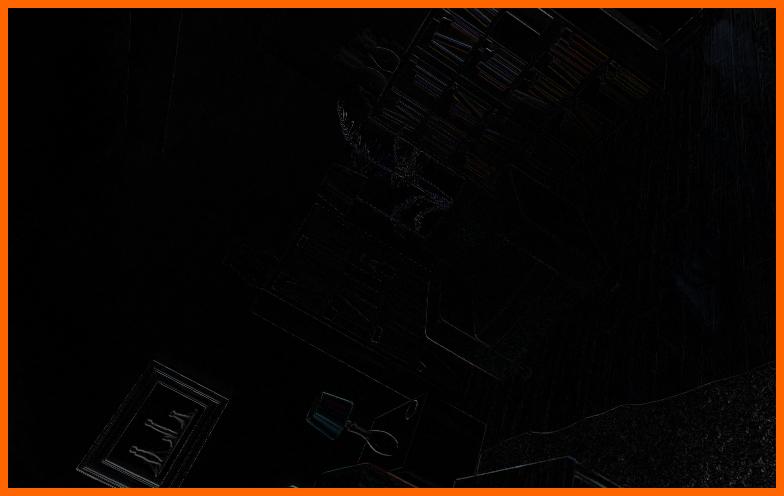} \\
		\specialrule{-0.1em}{0.05pt}{0.05pt}
%		\includegraphics[width=0.16\textwidth]{supplementary_figure/unordered_data/PinkCastle/NeRF.jpg} &
%		\includegraphics[width=0.16\textwidth]{supplementary_figure/unordered_data/PinkCastle/NeRF-diff.jpg} &
%		\includegraphics[width=0.16\textwidth]{supplementary_figure/unordered_data/PinkCastle/BARF.jpg} &
%		\includegraphics[width=0.16\textwidth]{supplementary_figure/unordered_data/PinkCastle/BARF-diff.jpg} &
%		\includegraphics[width=0.16\textwidth]{supplementary_figure/unordered_data/PinkCastle/USB-NeRF.jpg} &
%		\includegraphics[width=0.16\textwidth]{supplementary_figure/unordered_data/PinkCastle/USB-NeRF-diff.jpg} \\
%		\specialrule{-0.1em}{0.05pt}{0.05pt}
		\includegraphics[width=0.16\textwidth]{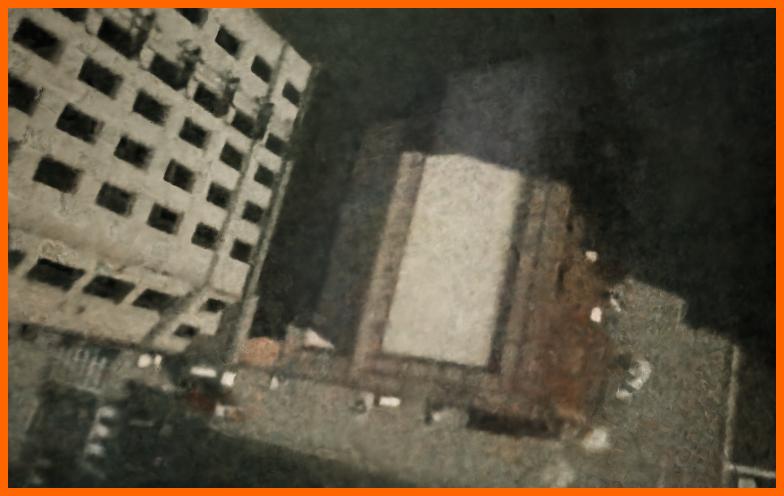} &
		\includegraphics[width=0.16\textwidth]{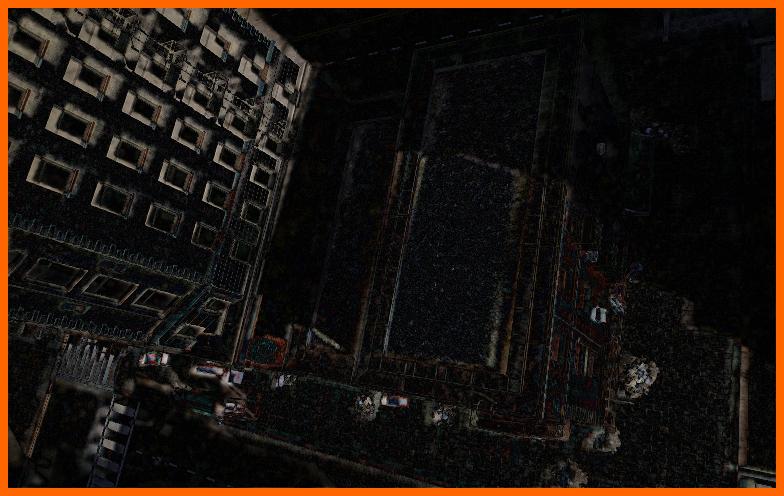} &
		\includegraphics[width=0.16\textwidth]{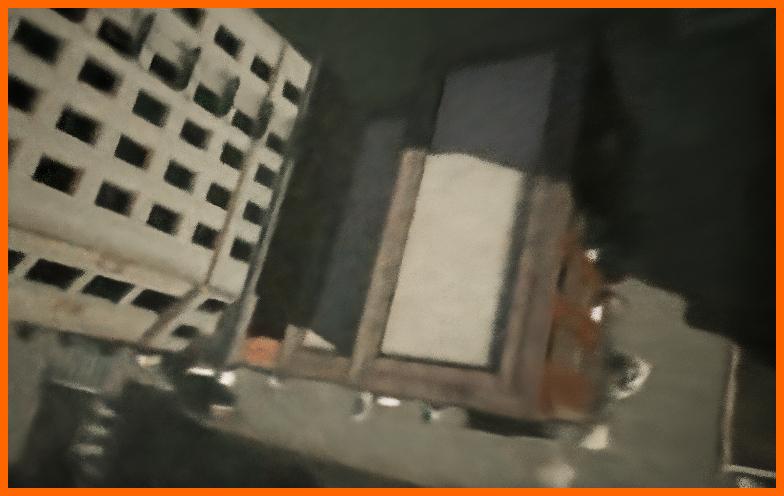} &
		\includegraphics[width=0.16\textwidth]{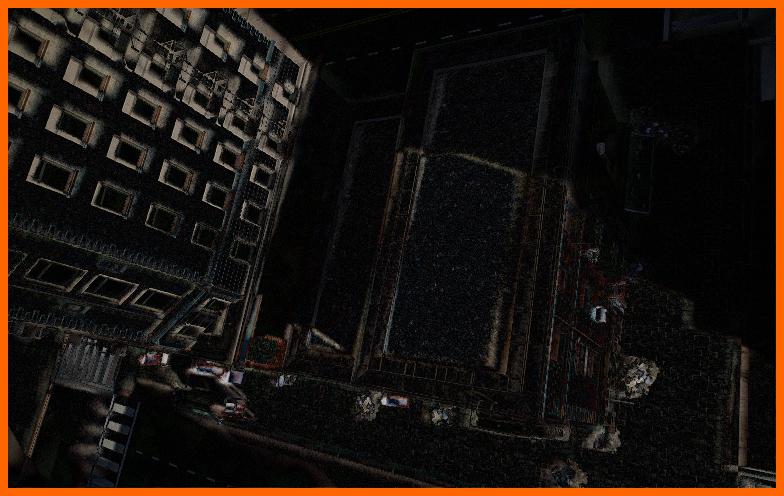} &
		\includegraphics[width=0.16\textwidth]{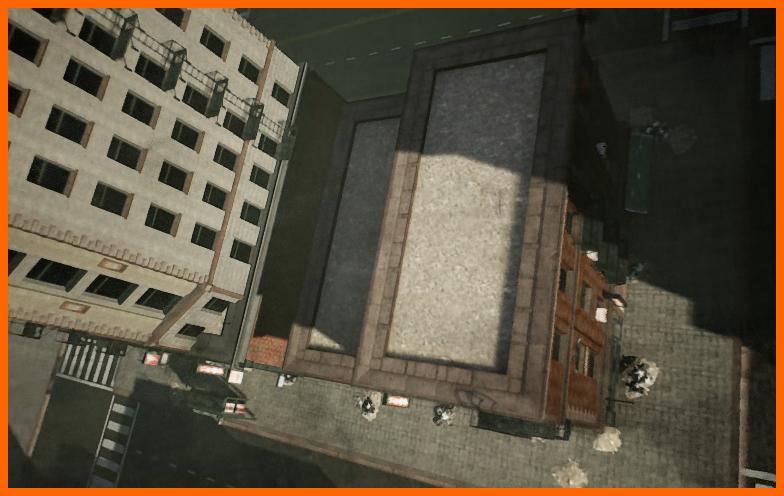} &
		\includegraphics[width=0.16\textwidth]{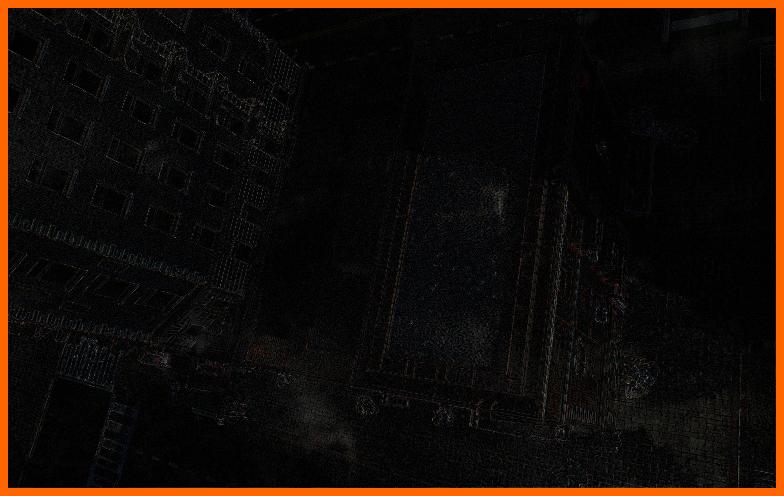} \\
		\specialrule{-0.1em}{0.05pt}{0.05pt}
		\specialrule{0em}{0.05pt}{0.05pt}
		\multicolumn{2}{c}{NeRF \citep{mildenhall2020nerf}}  & \multicolumn{2}{c}{BARF \citep{lin2021barf}}  & \multicolumn{2}{c}{USB-NeRF}
	\end{tabular}
%	\vspace{-0.5em}
	\captionsetup {font={small,stretch=0.5}}
	\caption{{\bf{Qualitative comparisons with unordered Unreal-RS datasets.}} The experimental results demonstrate that our approach could also generate high-quality global shutter images with a set of un-ordered rolling shutter images. The even columns present the error residual images between the rendered and ground-truth global shutter images. The darker the better.}
	\label{fig_unordered}
\end{figure}

\begin{table}[!ht]
%	\vspace{-0.5em}
	\begin{center}
		\captionsetup {font={small,stretch=0.5}}
		\caption{{\textbf{Quantitative comparisons on the synthetic datasets in terms of rolling shutter effect removal.}} Experimental results demonstrate that our method also achieves better performance compared to previous methods on these additional sequences.}
%		\vspace{-0.5em}
		\label{table_our_synthesis_longer}
		\setlength\tabcolsep{2.7pt}
		\setlength{\belowcaptionskip}{-12pt}
		%	\normalsize
		\scriptsize
		\resizebox{\linewidth}{!}{
			\begin{tabular}{c|ccc|ccc|ccc}
				\toprule
				&  \multicolumn{3}{c}{Adornment}  &  \multicolumn{3}{|c}{Factory}  &  \multicolumn{3}{|c}{Tanabata} \\
				& PSNR$\uparrow$ & SSIM$\uparrow$ & LPIPS$\downarrow$ & PSNR$\uparrow$ & SSIM$\uparrow$ & LPIPS$\downarrow$ & PSNR$\uparrow$ & SSIM$\uparrow$ & LPIPS$\downarrow$\\
				\midrule
				DiffSfM\citep{zhuang2017sfm_RS}	&14.91 &0.395	&0.3012 &15.21 &0.406	&0.2310 &11.63 &0.337	&0.3174 \\
				%			DiffSfM\citep{zhuang2017sfm_RS}	&20.57 &0.395 &0.3012 &18.47 &0.397 &0.3303 &22.23 &0.406 &0.2310 &18.87 &0.337 &0.3174\\
				DSUN\citep{liu2020deepunroll}    &17.84 &0.534 &0.2603 &19.75 &0.565 &0.1835 &16.06 &0.493 &0.2135\\
				SUNet\citep{fan2021sunet}	&19.11 &0.563 &0.2273 &22.84 &0.647 &0.1411 &17.79 &0.558 &0.1832\\
				RSSR\citep{fan2021RSSR}	&18.56 &0.602 &0.1892 &19.34 &0.631 &0.1438 &16.06 &0.566 &0.1660\\
				CVR\citep{fan2022CVR}   &19.70 &0.620 &0.1774 &20.89 &0.649 &0.1292 &17.68 &0.581 &0.1480\\
				
				NeRF \citep{mildenhall2020nerf} &17.28 &0.473 &0.6193 &18.02 &0.459 &0.4621 &15.04 &0.369 &0.5828\\
				BARF \citep{lin2021barf} &17.46 &0.485 &0.5614 &17.94 &0.459 &0.436 &15.28 &0.351 &0.6742\\
				\specialrule{0.08em}{1pt}{1pt}
				%			USB-NeRF-linear	&25.18 &0.758 &0.1779 &21.71 &0.695 &0.2035 &26.89 &0.790 &0.1350 &21.62 &0.667 &0.2407\\
				USB-NeRF (ours) &{\bf 29.97} &{\bf 0.876} &{\bf 0.0892} &{\bf 32.67} &{\bf 0.898} &{\bf 0.0819} &{\bf 23.84} &{\bf 0.750} &{\bf 0.1947}\\			
				\specialrule{0.08em}{1pt}{1pt}
			\end{tabular}
		}
	\end{center}
\end{table}

\begin{figure}[!ht]
	\vspace{1em}
	\setlength\tabcolsep{1.pt}
	\centering
	\begin{tabular}{rcccccc}
		%		\raisebox{.24in}{\rotatebox[origin=t]{90}{\normalsize DSUN \citep{liu2020deepunroll}}}
		\raisebox{.22in}{\rotatebox[origin=t]{90}{\scriptsize DiffSfM}} &\includegraphics[width=0.16\textwidth]{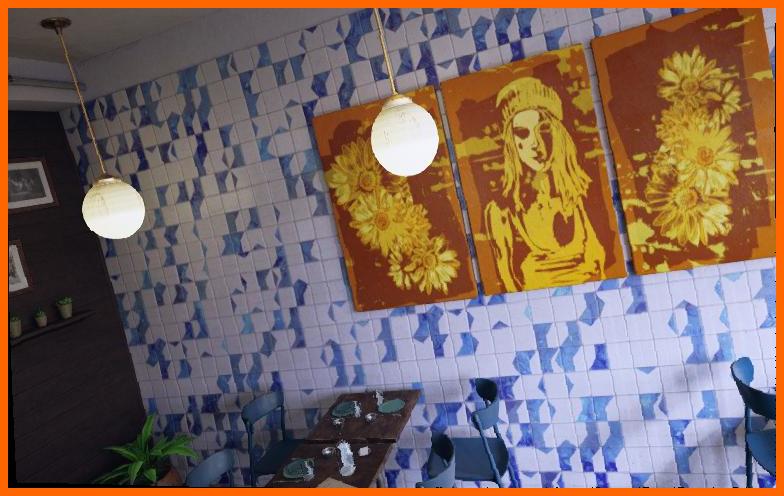} &
		\includegraphics[width=0.16\textwidth]{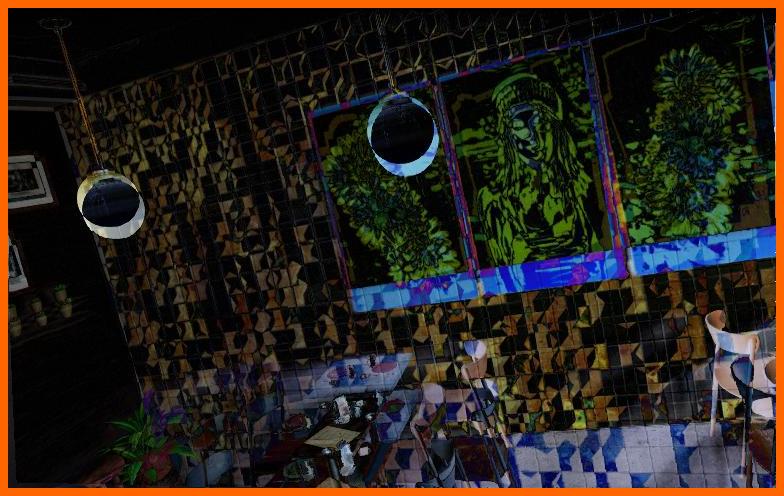} &
		\includegraphics[width=0.16\textwidth]{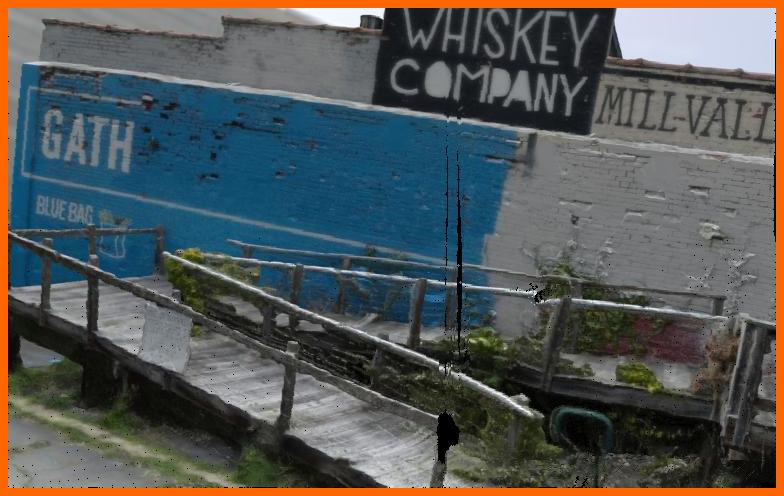} &
		\includegraphics[width=0.16\textwidth]{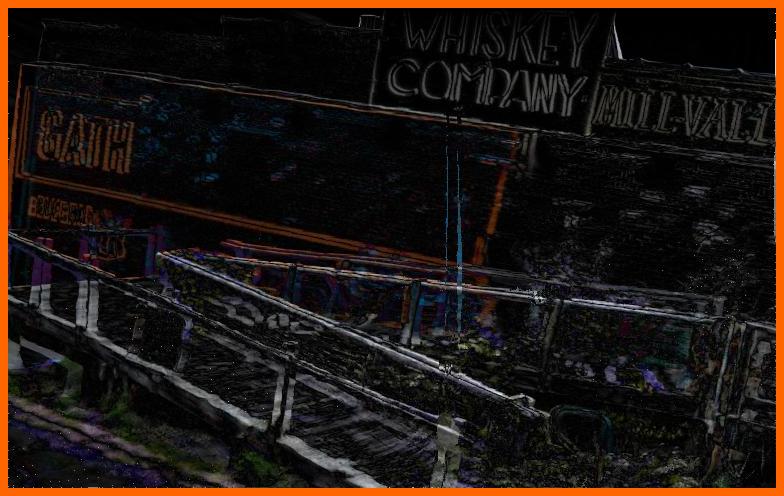} &
		\includegraphics[width=0.16\textwidth]{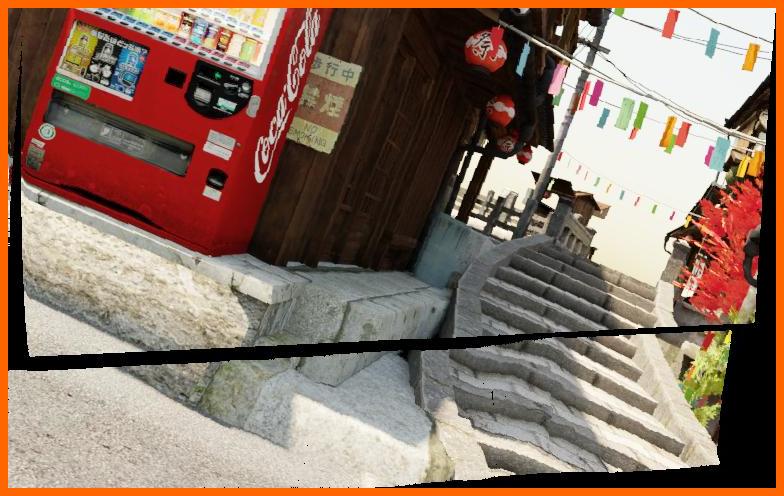} &
		\includegraphics[width=0.16\textwidth]{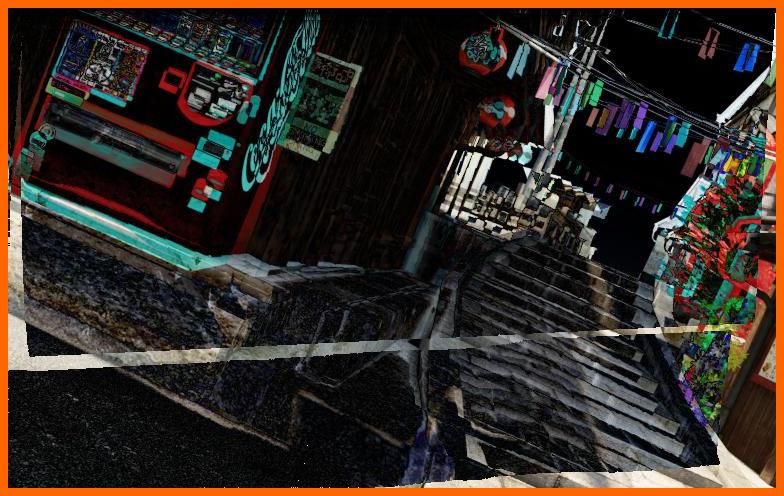}\\
		
		\specialrule{-0.1em}{0.05pt}{0.05pt}
		\raisebox{.24in}{\rotatebox[origin=t]{90}{\scriptsize DSUN}}
		&\includegraphics[width=0.16\textwidth]{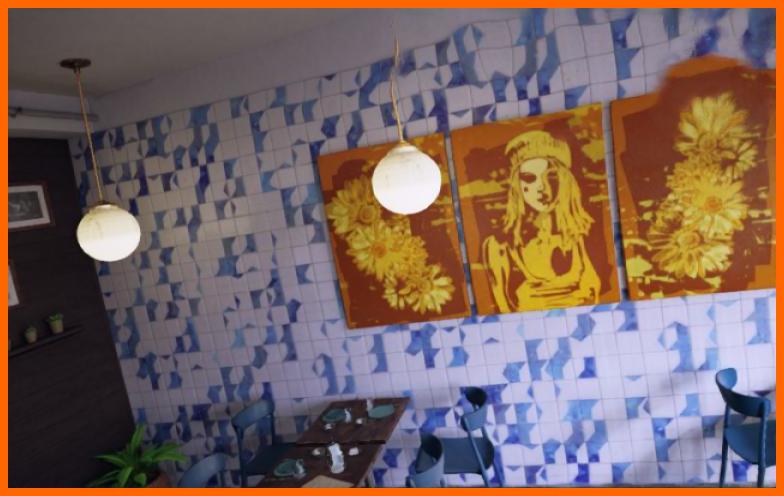} &
		\includegraphics[width=0.16\textwidth]{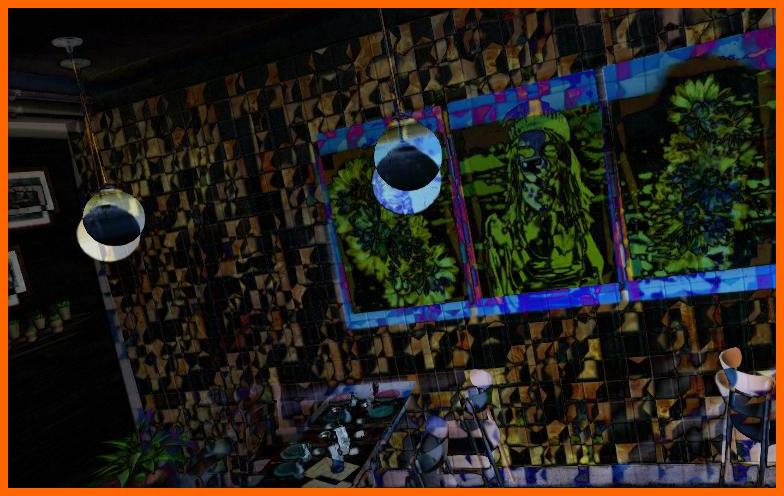} &
		\includegraphics[width=0.16\textwidth]{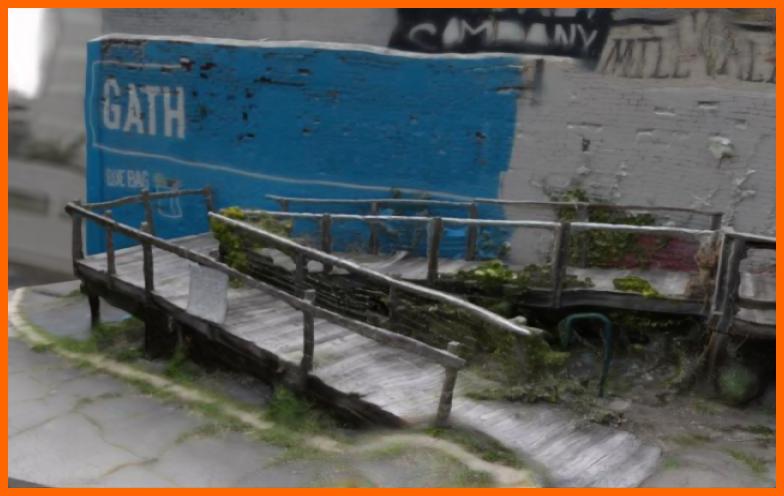} &
		\includegraphics[width=0.16\textwidth]{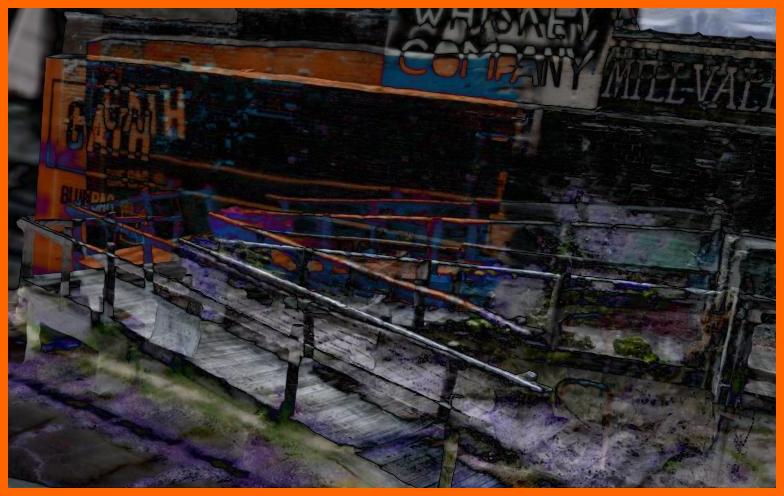} &
		\includegraphics[width=0.16\textwidth]{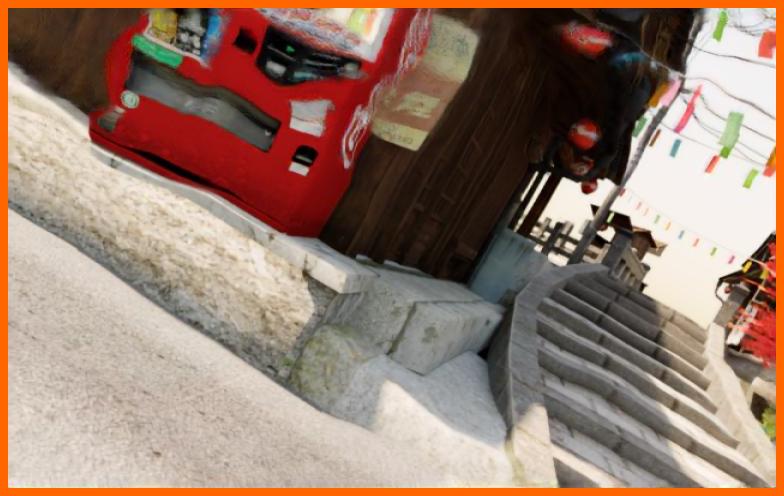} &
		\includegraphics[width=0.16\textwidth]{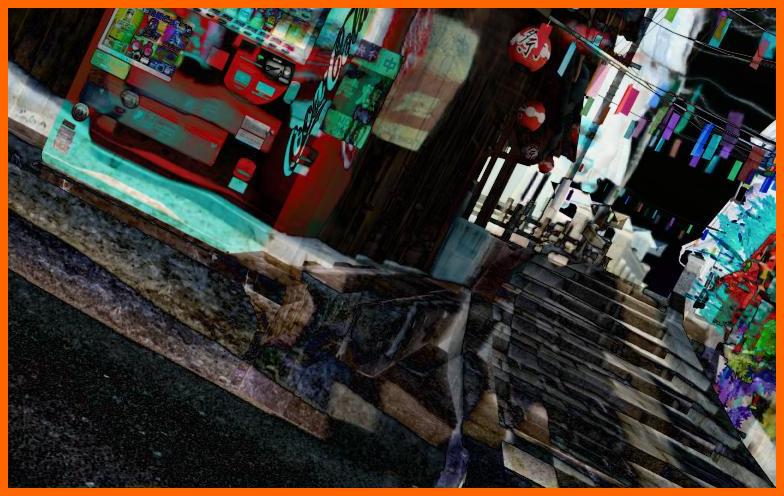}\\
		
		\specialrule{-0.1em}{0.05pt}{0.05pt}
		\raisebox{.24in}{\rotatebox[origin=t]{90}{\scriptsize SUNet}}
		&\includegraphics[width=0.16\textwidth]{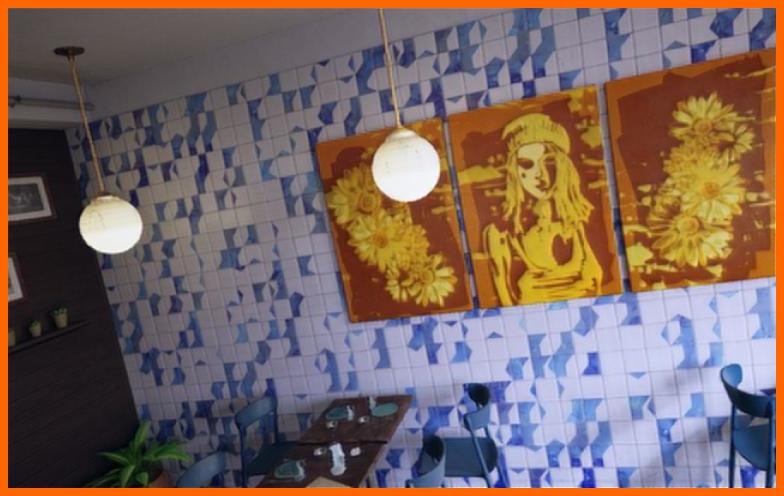} &
		\includegraphics[width=0.16\textwidth]{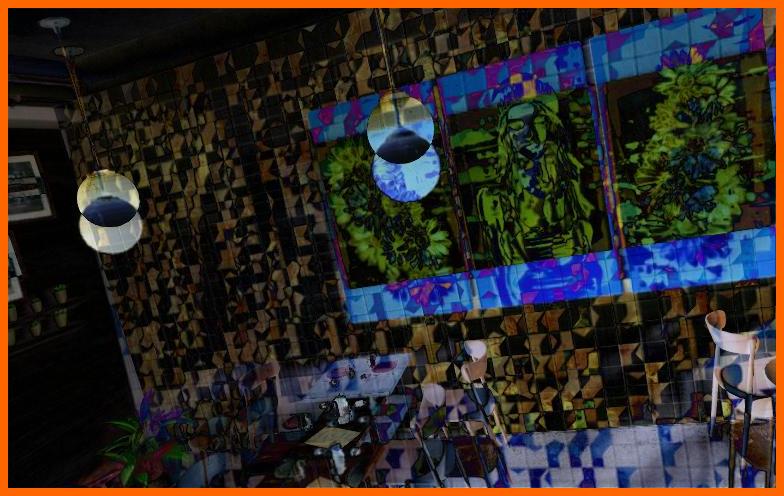} &
		\includegraphics[width=0.16\textwidth]{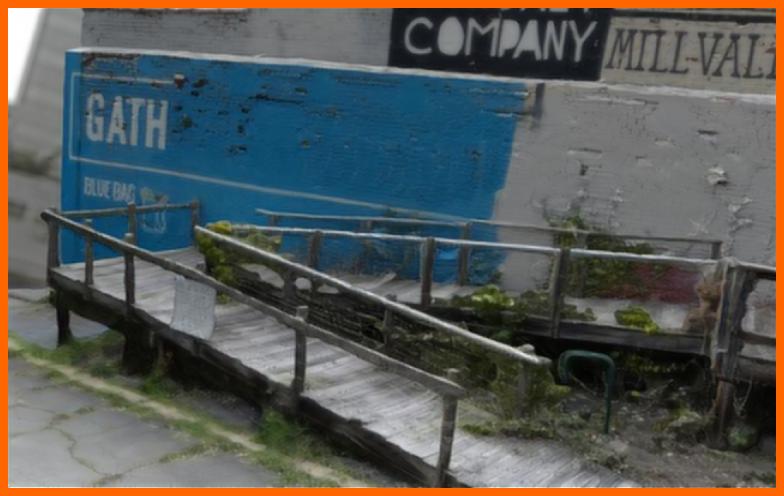} &
		\includegraphics[width=0.16\textwidth]{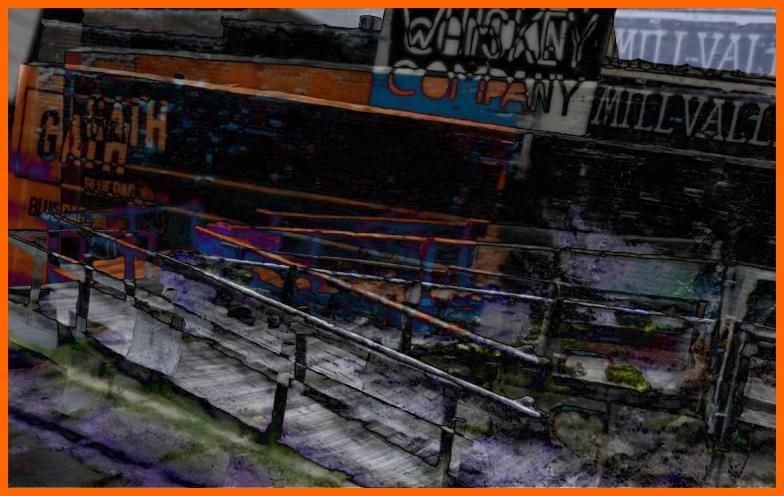} &
		\includegraphics[width=0.16\textwidth]{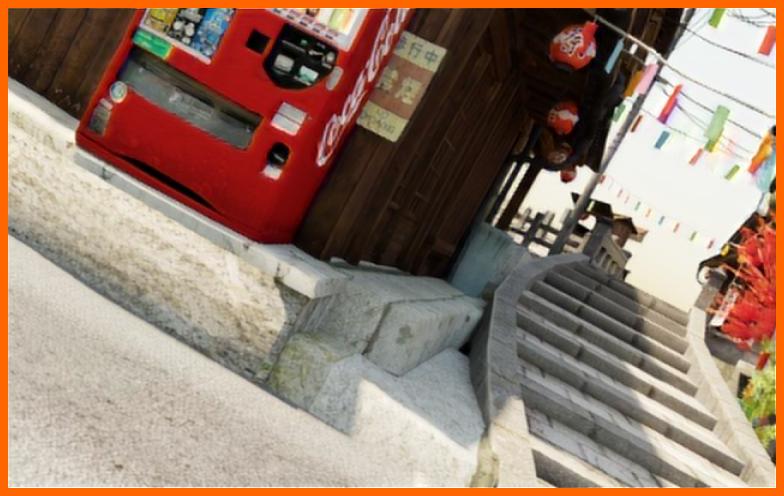} &
		\includegraphics[width=0.16\textwidth]{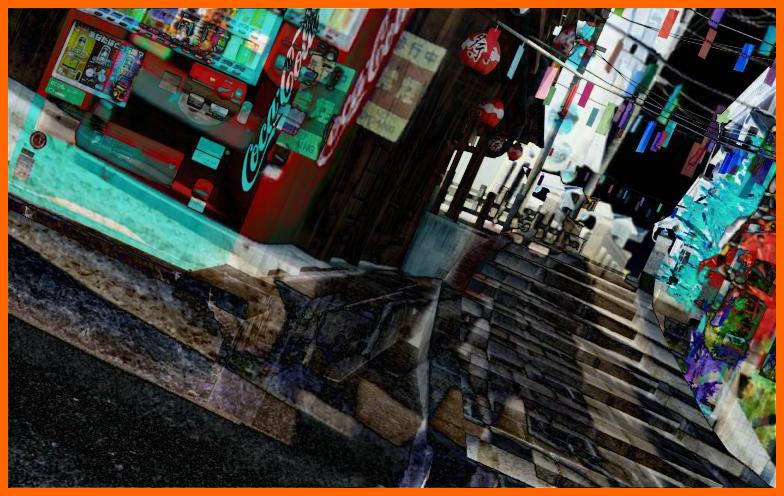}\\
		
		\specialrule{-0.1em}{0.05pt}{0.05pt}
		\raisebox{.25in}{\rotatebox[origin=t]{90}{\scriptsize RSSR}}
		&\includegraphics[width=0.16\textwidth]{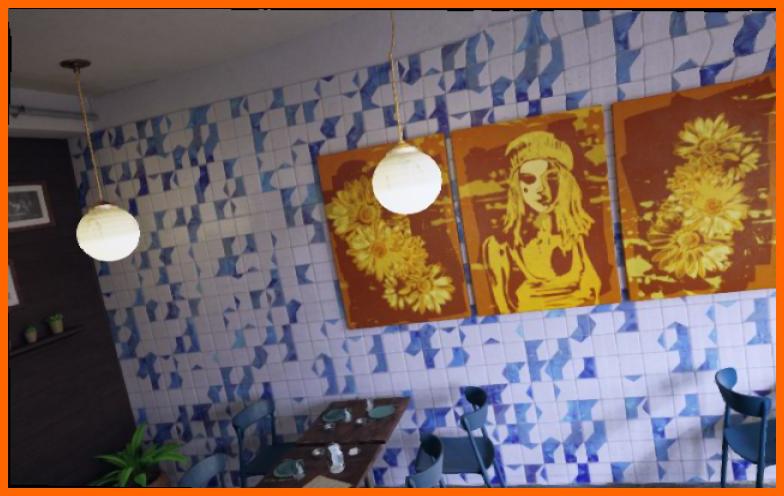} &
		\includegraphics[width=0.16\textwidth]{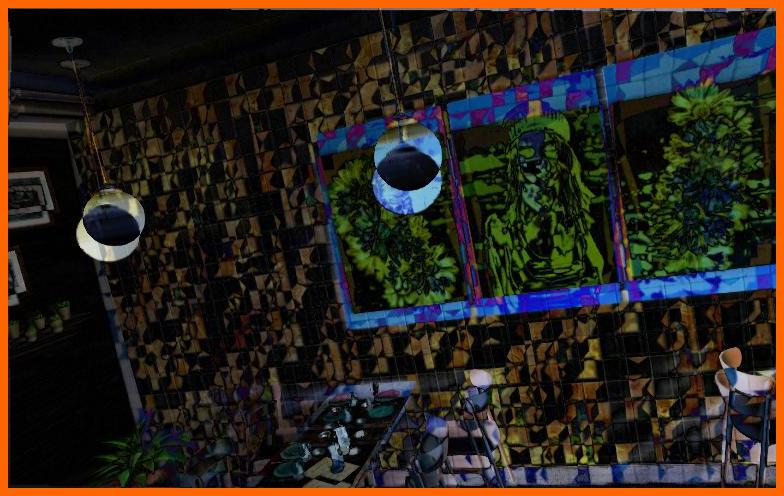} &
		\includegraphics[width=0.16\textwidth]{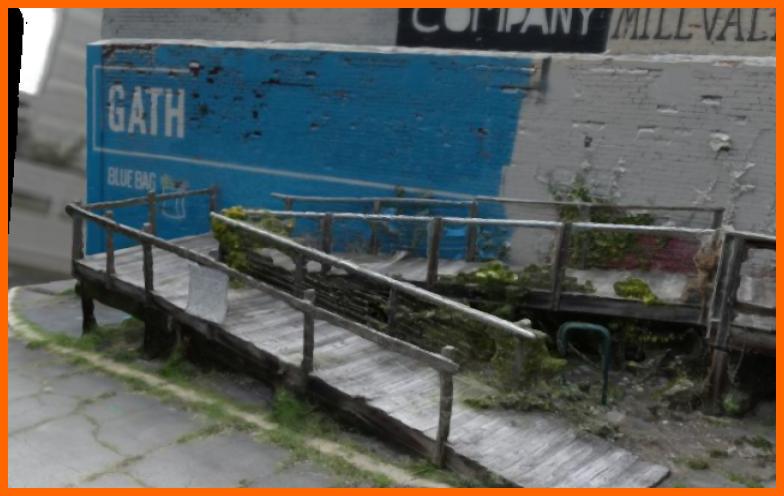} &
		\includegraphics[width=0.16\textwidth]{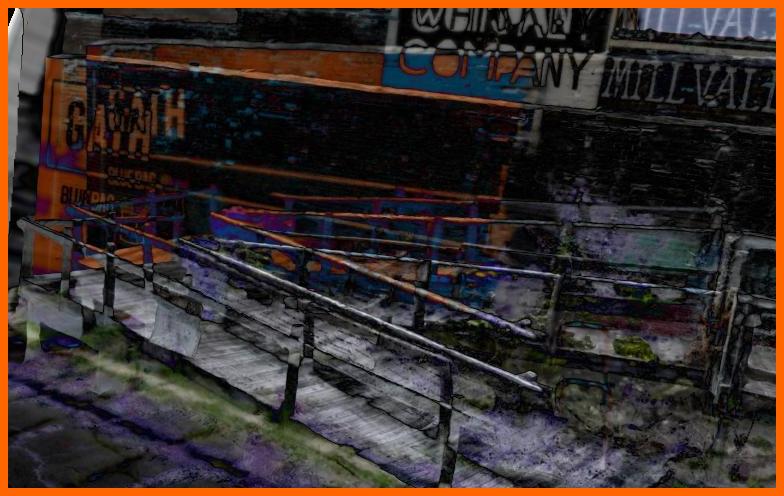} &
		\includegraphics[width=0.16\textwidth]{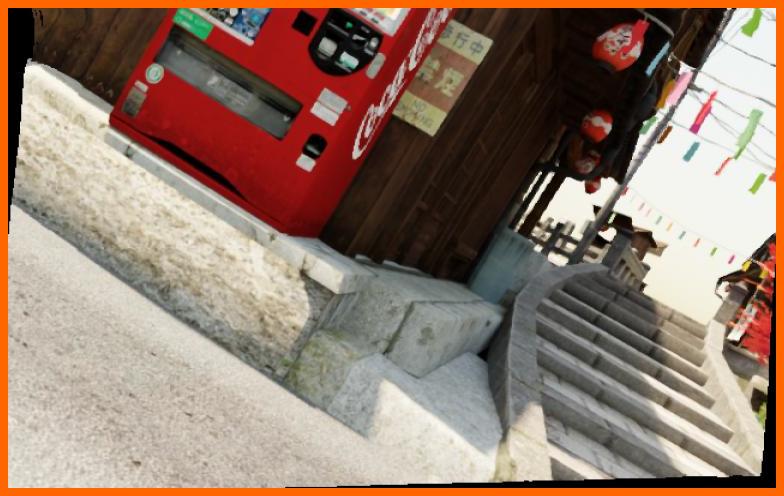} &
		\includegraphics[width=0.16\textwidth]{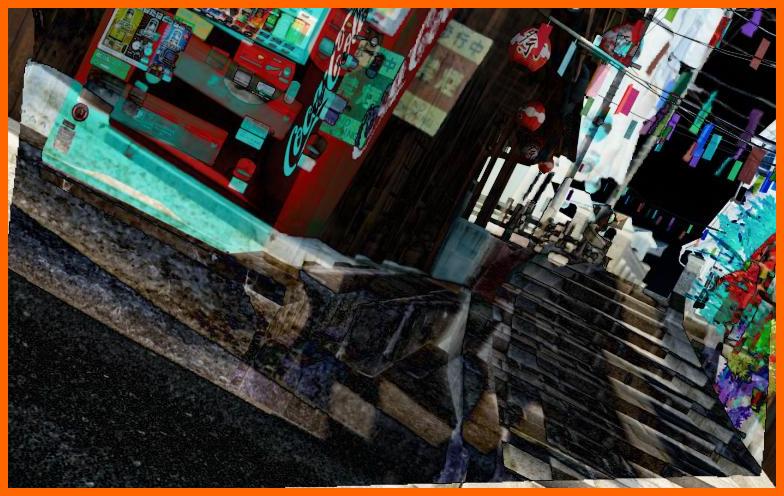}\\
		
		\specialrule{-0.1em}{0.05pt}{0.05pt}
		\raisebox{.23in}{\rotatebox[origin=t]{90}{\scriptsize CVR}}
		&\includegraphics[width=0.16\textwidth]{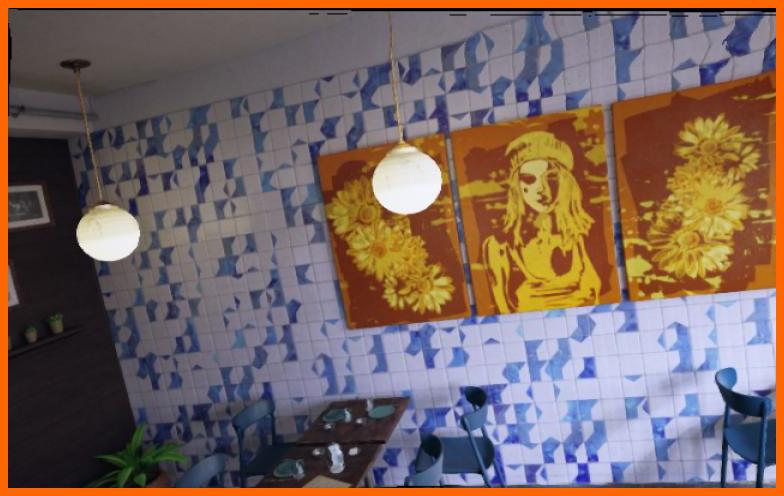} &
		\includegraphics[width=0.16\textwidth]{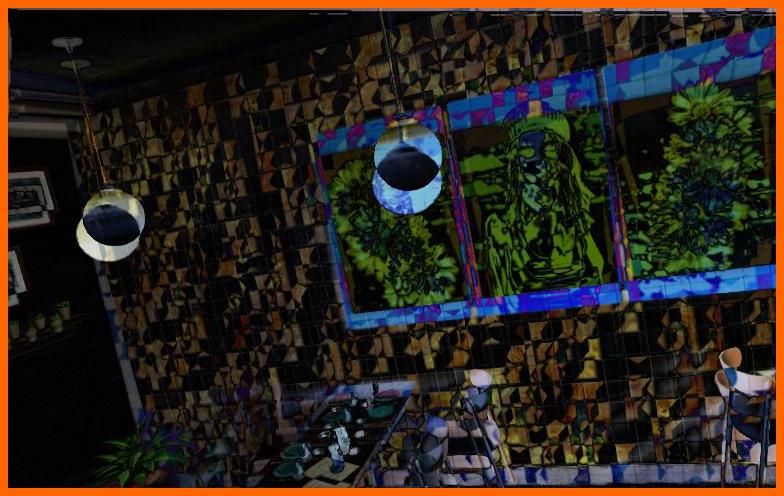} &
		\includegraphics[width=0.16\textwidth]{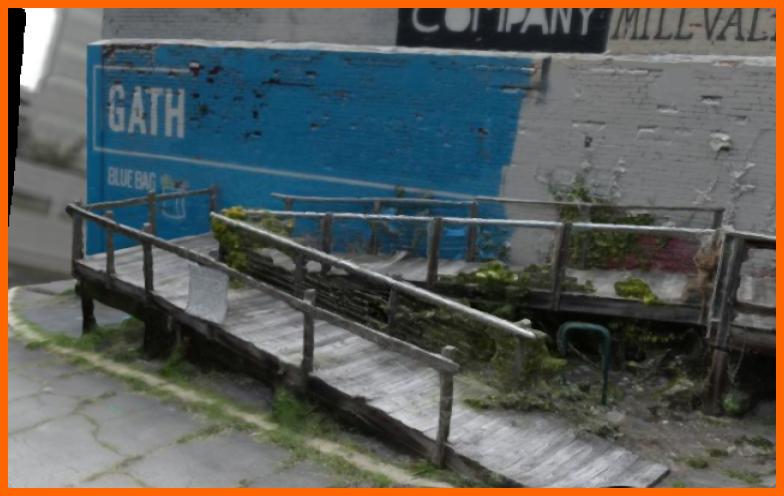} &
		\includegraphics[width=0.16\textwidth]{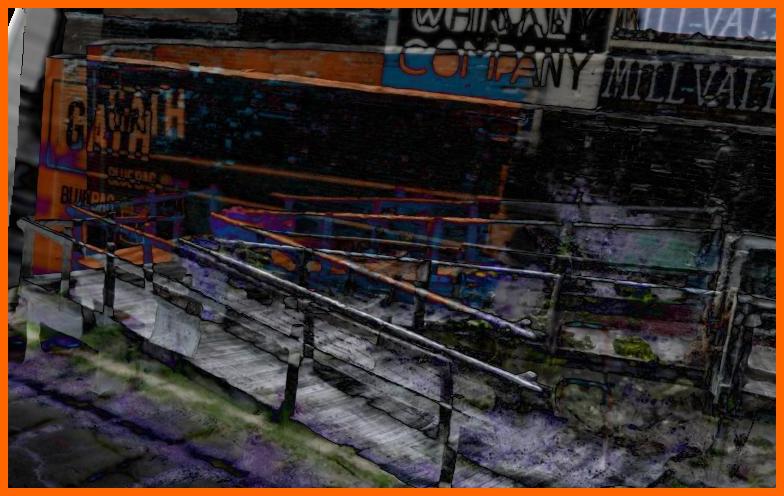} &
		\includegraphics[width=0.16\textwidth]{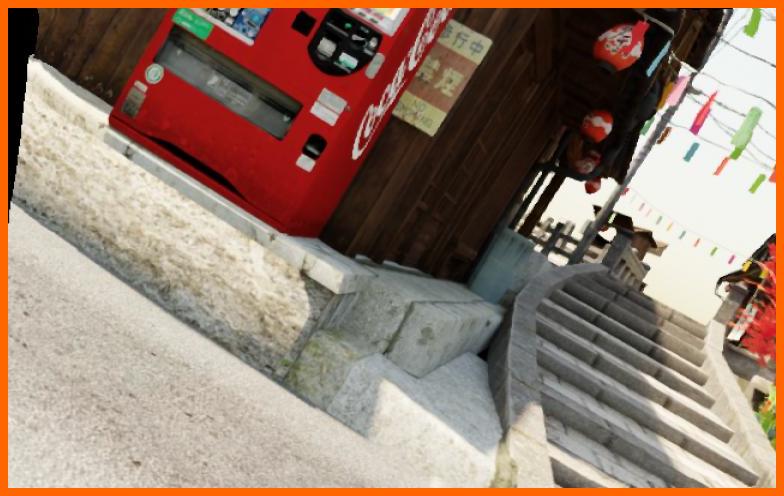} &
		\includegraphics[width=0.16\textwidth]{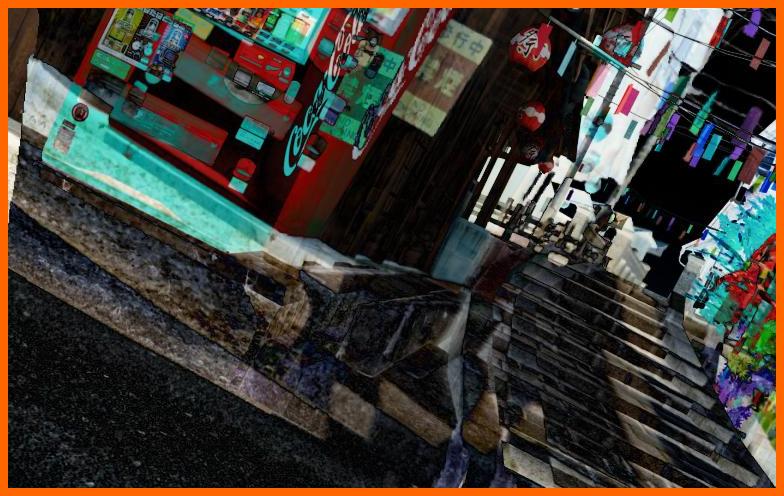}\\
		
		\specialrule{-0.1em}{0.05pt}{0.05pt}
		\raisebox{.25in}{\rotatebox[origin=t]{90}{\scriptsize NeRF}}
		&\includegraphics[width=0.16\textwidth]{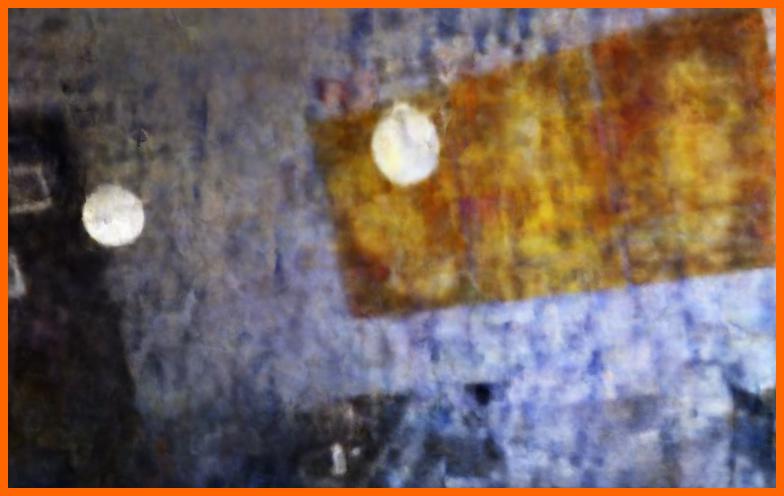} &
		\includegraphics[width=0.16\textwidth]{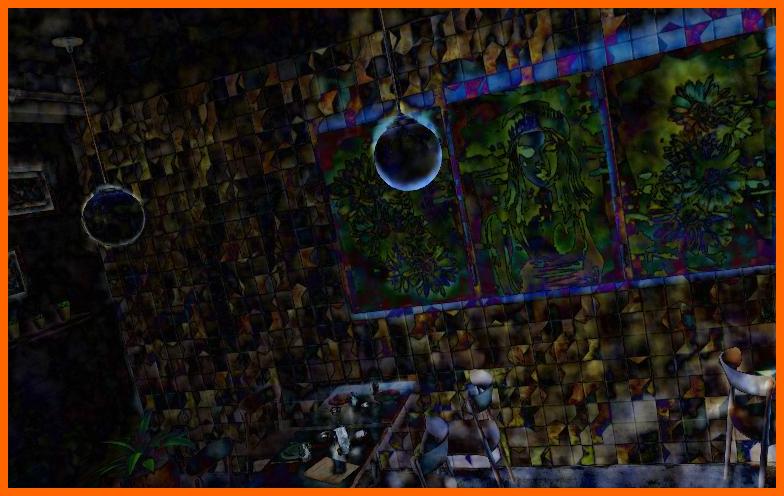} &
		\includegraphics[width=0.16\textwidth]{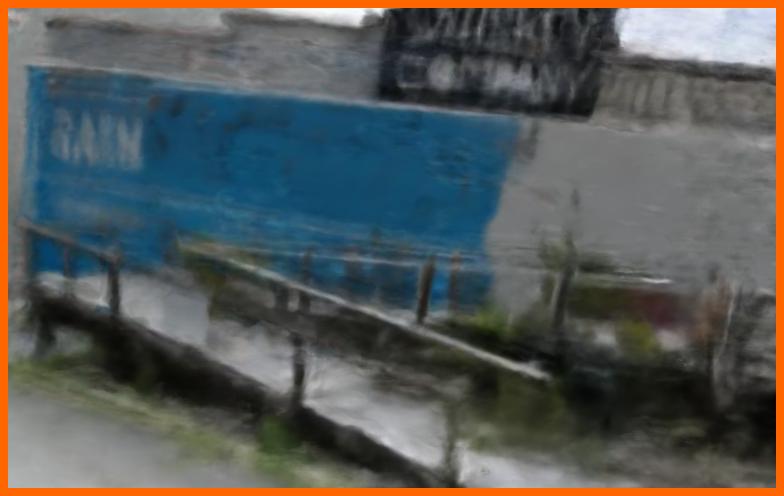} &
		\includegraphics[width=0.16\textwidth]{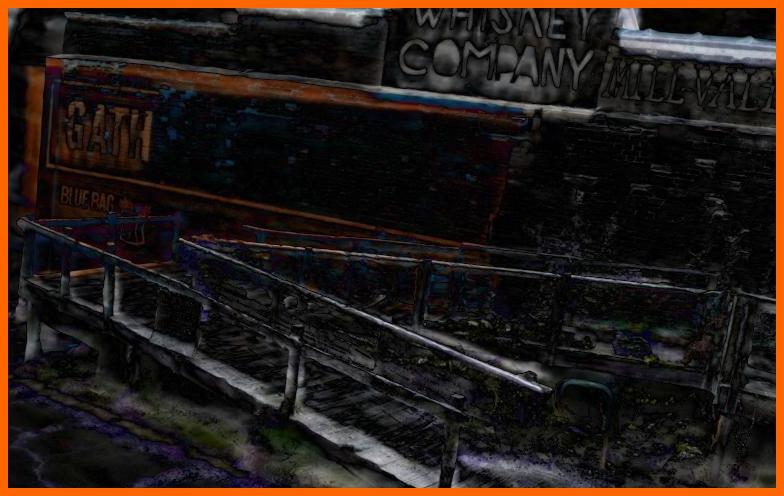} &
		\includegraphics[width=0.16\textwidth]{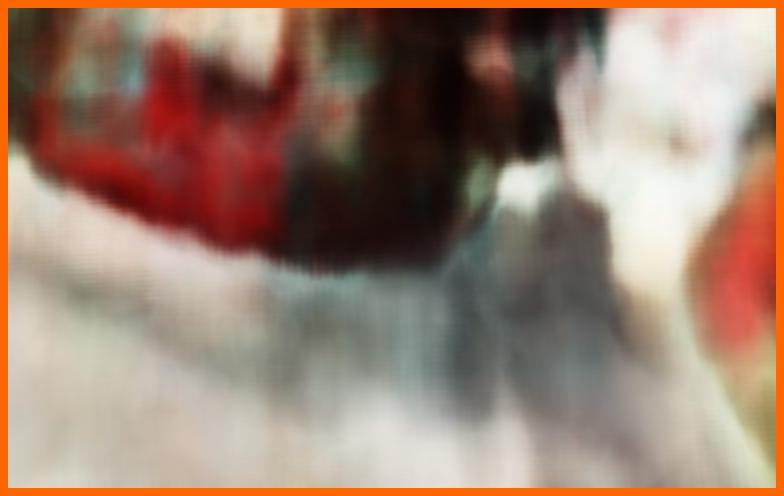} &
		\includegraphics[width=0.16\textwidth]{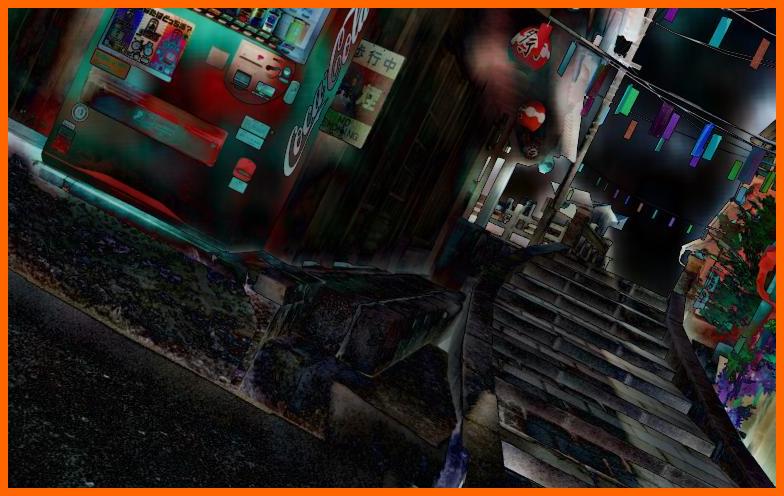}\\
		
		\specialrule{-0.1em}{0.05pt}{0.05pt}
		\raisebox{.25in}{\rotatebox[origin=t]{90}{\scriptsize BARF}}
		&\includegraphics[width=0.16\textwidth]{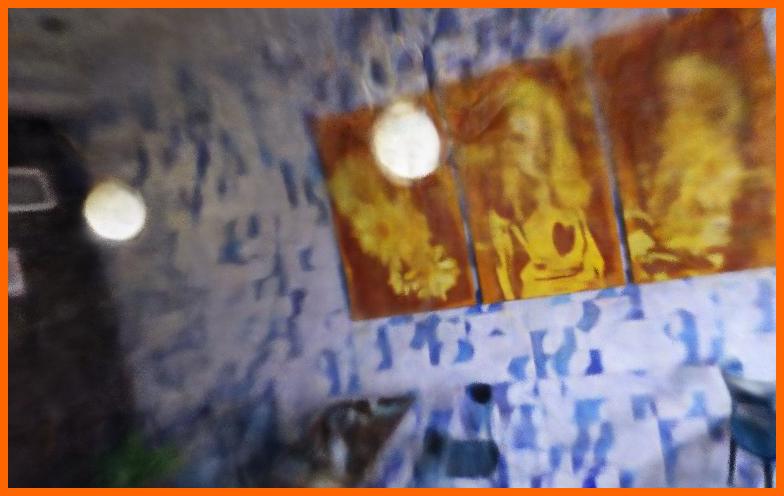} &
		\includegraphics[width=0.16\textwidth]{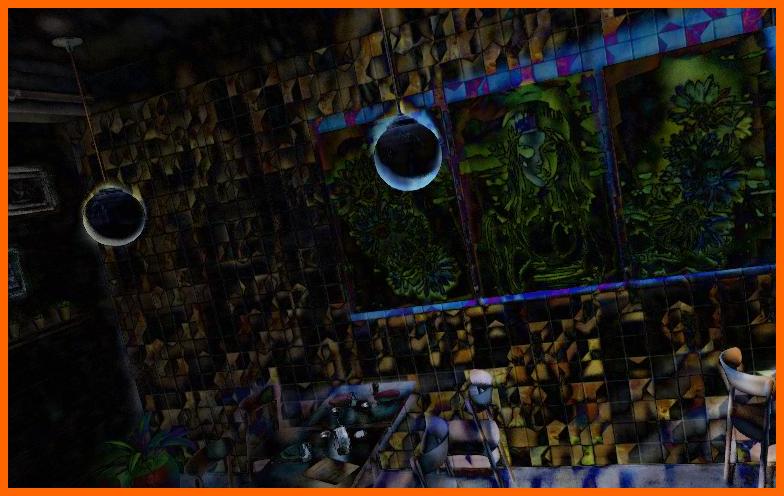} &
		\includegraphics[width=0.16\textwidth]{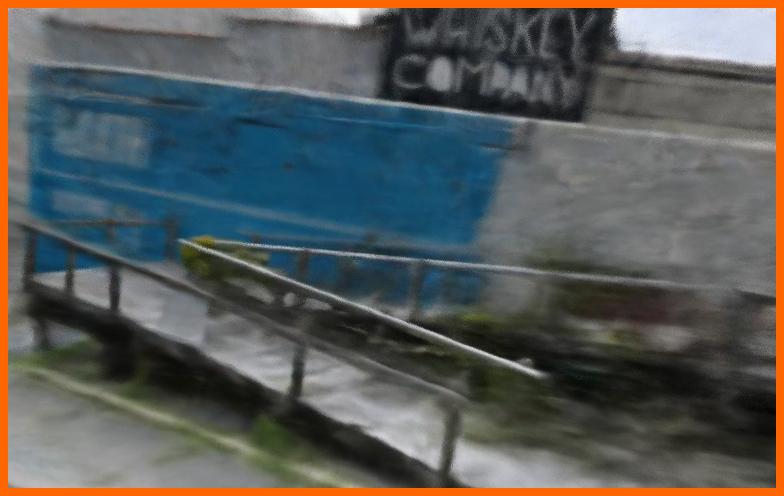} &
		\includegraphics[width=0.16\textwidth]{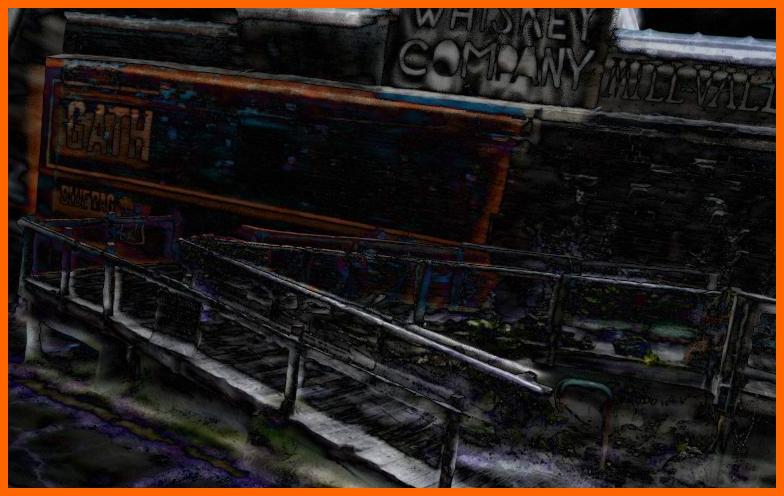} &
		\includegraphics[width=0.16\textwidth]{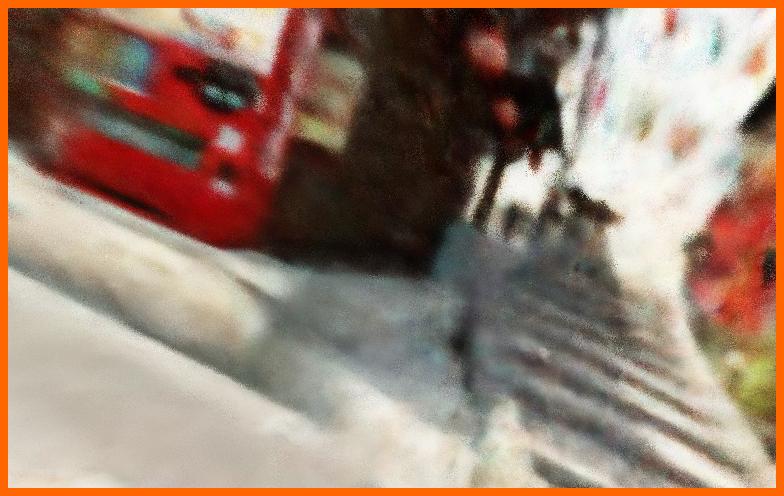} &
		\includegraphics[width=0.16\textwidth]{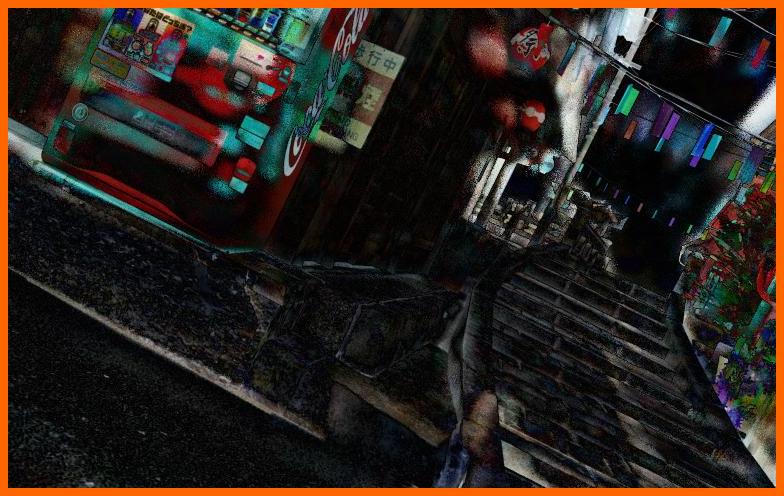}\\
		
		\specialrule{-0.1em}{0.05pt}{0.05pt}
		\raisebox{.25in}{\rotatebox[origin=t]{90}{\scriptsize USB-NeRF}}
		&\includegraphics[width=0.16\textwidth]{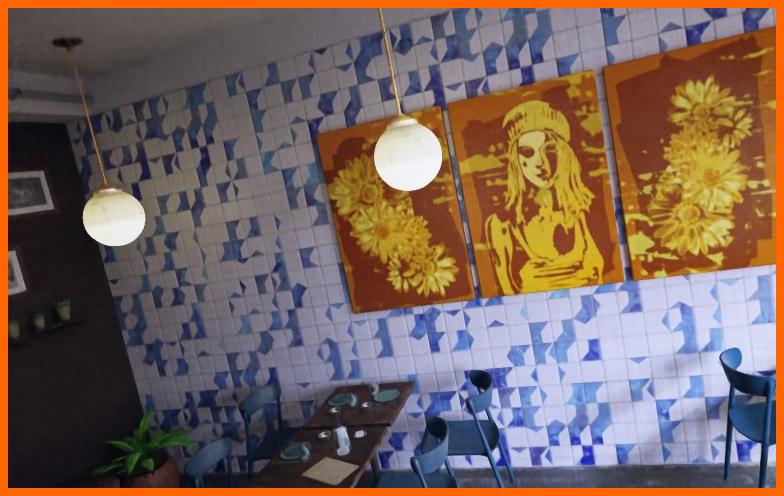} &
		\includegraphics[width=0.16\textwidth]{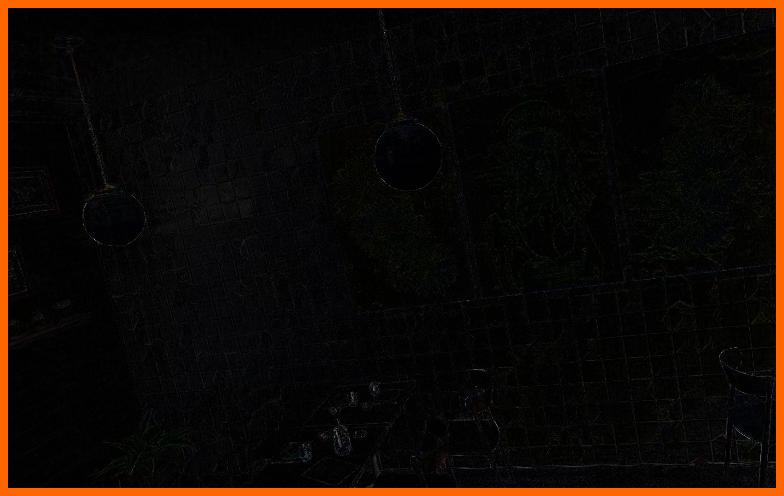} &
		\includegraphics[width=0.16\textwidth]{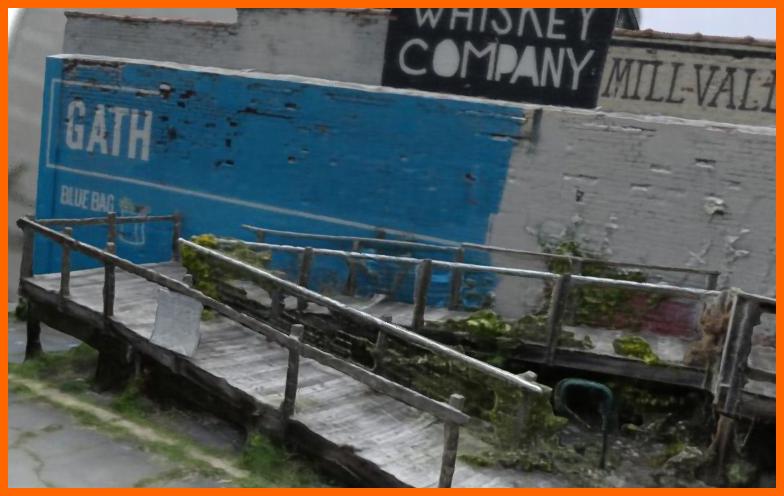} &
		\includegraphics[width=0.16\textwidth]{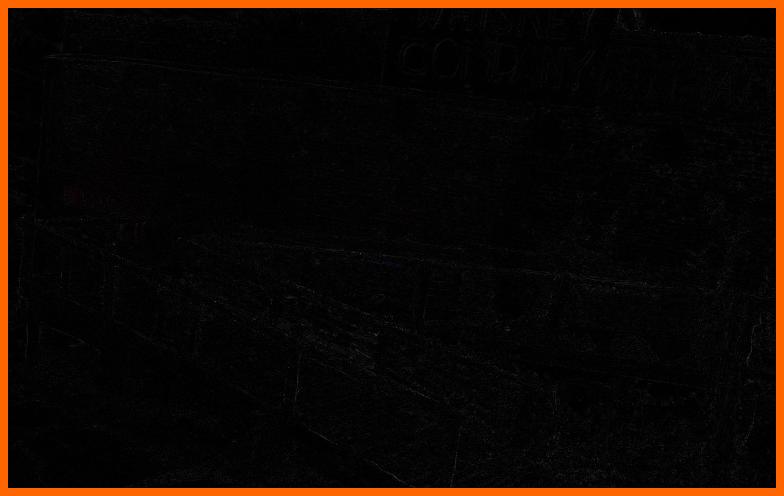} &
		\includegraphics[width=0.16\textwidth]{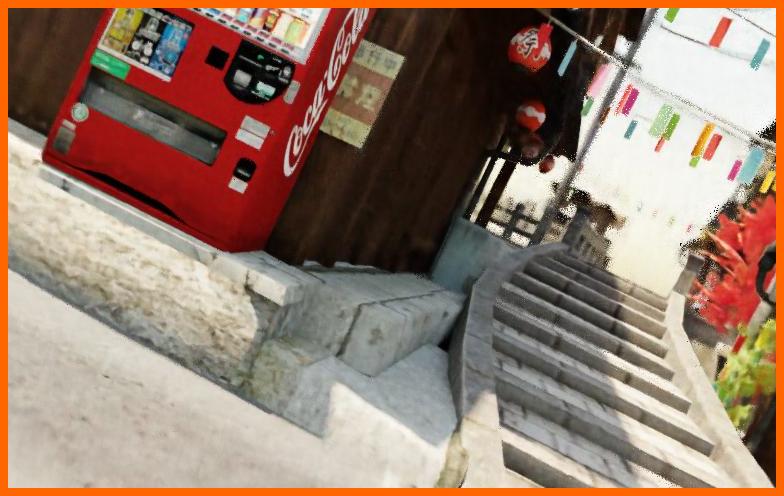} &
		\includegraphics[width=0.16\textwidth]{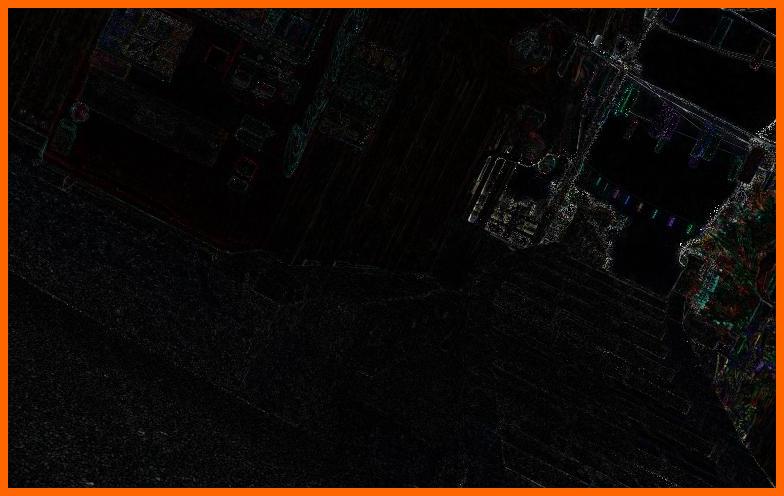}\\
		\specialrule{0em}{0.05pt}{0.05pt}
	\end{tabular}
%	\vspace{-0.5em}
	\captionsetup {font={small,stretch=0.5}}
	\caption{{\bf{Qualitative comparisons on synthetic datasets for rolling Shutter effect removal.}} The experimental results demonstrate that our method is superior than all prior works. The even columns present the error residual images between the rendered and ground-truth global shutter images. The darker the better. }
	\label{fig_RS_removal_longer}
	\vspace{0.em}
\end{figure}

\begin{table}
	\vspace{0em}
	\captionsetup {font={small,stretch=0.5}}
	\caption{{\textbf{Quantitative comparisons on TUM-RS datasets \citep{schubert2019RS-VIO} in terms of the Absolute Trajectory Error (ATE) metric (m).}} The experimental results demonstrate that our method performs much better than prior works in terms of the accuracy of motion trajectory estimation.} 
	\label{TUM_ATE}
	\setlength{\belowcaptionskip}{-10pt}
	\setlength{\tabcolsep}{2.8mm}
	\scriptsize
	\resizebox{\linewidth}{!}{
		%\begin{tabular}{c|c c c c c c}
		\begin{tabular}{c|c c c c c c}
			\toprule
			%			&{\scriptsize \quad COLMAP \citep{schonberger2016structure} } &{\scriptsize BARF\citep{lin2021barf} } &{\scriptsize RSBA \citep{hedborg2012RSBA}} & {\scriptsize \ NW-RSBA \citep{liao2023NW_RSBA} \ } & {\scriptsize \qquad USB-NeRF-linear \qquad} & {\scriptsize \qquad USB-NeRF-cubic \qquad} \\
			\makebox[0.08\textwidth][c]{} & \makebox[0.08\textwidth][c]{\scriptsize COLMAP} & \makebox[0.08\textwidth][c]{\scriptsize BARF} & \makebox[0.08\textwidth][c]{\scriptsize RSBA} & \makebox[0.08\textwidth][c]{\scriptsize NW-RSBA} & \makebox[0.09\textwidth][c]{\scriptsize USB-NeRF-linear} &\makebox[0.09\textwidth][c]{\scriptsize USB-NeRF-cubic} \\
			%			\makebox[0.08\textwidth][c]{} & \makebox[0.08\textwidth][c]{\scriptsize \citep{schonberger2016structure}} & \makebox[0.08\textwidth][c]{\scriptsize \citep{lin2021barf}} & \makebox[0.08\textwidth][c]{\scriptsize \citep{hedborg2012RSBA}} & \makebox[0.08\textwidth][c]{\scriptsize \citep{liao2023NW_RSBA}} & \makebox[0.09\textwidth][c]{} &\makebox[0.09\textwidth][c]{} \\
			\midrule
			seq-1 &{.0237}$\pm${.0233} &.0345$\pm$.0176 &.0136$\pm$.0074 &.1162$\pm$.0340 &.0047$\pm$.0029 &{\bf.0038}$\pm$.0021\\
			seq-2 &{.1432}$\pm${.0676} &.1847$\pm$.0635 &.3574$\pm$.1713 &.4349$\pm$.2047 &{.0592}$\pm$.0506 &{\bf.0560}$\pm$.0332\\
			seq-3 &{.0476}$\pm${.0340} &.0743$\pm$.0437 &{\bf.0106}$\pm$.0057 &.0123$\pm$.0065 &.0120$\pm$.0060 &.0111$\pm$.0060\\
			seq-4 &{.0180}$\pm${.0059} &.0294$\pm$.0113 &.0064$\pm$.0026 &.0473$\pm$.0289 &{.0064}$\pm$.0026 &{\bf.0050}$\pm$.0023\\
			seq-5 &{.0662}$\pm${.0275} &.0999$\pm$.0394 &.0144$\pm$.0048 &.0366$\pm$.0554 &.0119$\pm$.0043 &{\bf.0114}$\pm$.0041\\
			seq-6 &{.0349}$\pm${.0124} &.0719$\pm$.0451 &{.0137}$\pm$.0078 &.0626$\pm$.0327 &.0163$\pm$.0074 &{\bf.0035}$\pm$.0021\\
			seq-7 &{.0184}$\pm${.0065} &.0185$\pm$.0057 &.0056$\pm$.0034 &.0700$\pm$.0289 &.0036$\pm$.0015 &{\bf.0030}$\pm$.0012\\
			seq-8 &{.0417}$\pm${.0189} &.0638$\pm$.0340 &{.0102}$\pm$.0040 &.1787$\pm$.0863 &.0096$\pm$.0069 &{\bf.0095}$\pm$.0070\\
			seq-9 &{.0512}$\pm${.0191} &.1509$\pm$.0835 &.2432$\pm$.1061 &.2780$\pm$.1186 &{\bf.0128}$\pm$.0130 &{.0150}$\pm$.0166\\
			seq-10 &{.0417}$\pm${.0122} &.1450$\pm$.0475 &{\bf.0126}$\pm$.0090 &.1378$\pm$.0682 &.0135$\pm$.0085 &.0178$\pm$.0149\\
			\midrule
			Avg. &{.0486}$\pm${.0228} &{.0873}$\pm${.0391} &{.0688}$\pm${.0322} &{.1374}$\pm${.0664} &{.0150}$\pm${.0104} &{\bf.0136}$\pm${.0090}\\
			\bottomrule
		\end{tabular}
	}
	
\end{table}

\begin{table}[t]
	\begin{center}
		\captionsetup {font={small,stretch=0.5}}
		\caption{{\textbf{Quantitative comparisons on synthetic datasets in terms of the Relative Pose Error for rotation part (\textdegree/frame).}} The experimental results demonstrate that rolling shutter distortions affect the accuracy of motion trajectory estimations. Due to proper modeling, our method performs much better than state-of-the-art methods. It also demonstrates that cubic B-Spline interpolation is superior to linear interpolation. x denotes method failed on the corresponding sequence.}
		\label{rpe_rot_synthetic}
		\vspace{-0.5em}
		\scriptsize
		\setlength{\tabcolsep}{2.5mm}
		\setlength{\belowcaptionskip}{-10pt}
		\resizebox{\linewidth}{!}{
			\begin{tabular}{c|c c c c c c}
				\toprule
				\makebox[0.08\textwidth][c]{} & \makebox[0.08\textwidth][c]{\scriptsize COLMAP} & \makebox[0.08\textwidth][c]{\scriptsize BARF} & \makebox[0.08\textwidth][c]{\scriptsize RSBA} & \makebox[0.08\textwidth][c]{\scriptsize NW-RSBA} & \makebox[0.08\textwidth][c]{\scriptsize USB-NeRF-linear} &\makebox[0.08\textwidth][c]{\scriptsize USB-NeRF-cubic} \\
				%			&{\scriptsize \qquad COLMAP \qquad} &{\scriptsize BARF} &{\scriptsize \qquad RSBA \qquad} & {\scriptsize NW-RSBA} & {\scriptsize \qquad USB-NeRF-linear \qquad} & {\scriptsize \qquad USB-NeRF-cubic \qquad} \\
				\midrule
				Carla 		&2.357 $\pm$ 1.647 & 2.762 $\pm$ 2.041 & 1.844 $\pm$ 1.080 & 2.090 $\pm$ 1.492 & 0.586 $\pm$ 0.174 & \bf 0.528 $\pm$ 0.157 \\
				
				BlueRoom & 2.594 $\pm$ 1.411 & 8.123 $\pm$ 3.585 & 7.257 $\pm$ 3.413 & x            & 0.127 $\pm$ 0.074 & \bf 0.046 $\pm$ 0.041 \\
				LivingRoom & 2.145 $\pm$ 1.320 & 9.016 $\pm$ 6.243 & 7.405 $\pm$ 4.383 & x            & 0.381 $\pm$ 0.140 & \bf 0.108 $\pm$ 0.053 \\
				WhiteRoom & 1.810 $\pm$ 0.771 & 4.702 $\pm$ 2.077 & 4.212 $\pm$ 1.631 & x            & 0.194 $\pm$ 0.127 & \bf 0.089 $\pm$ 0.039 \\
				
				Adornment & 2.723 $\pm$ 1.438 & 10.390 $\pm$ 4.659 &       x     & x            & 0.581 $\pm$ 1.273 & \bf 0.234 $\pm$ 0.147 \\
				Factory & 2.194 $\pm$ 1.187 & 8.523 $\pm$ 4.028 &        x     & x            & 0.209 $\pm$ 0.112 & \bf 0.135 $\pm$ 0.080 \\
				Tanabata & 2.744 $\pm$ 1.662 & 15.187 $\pm$ 13.197 &        x     & x            & 1.042 $\pm$ 0.459 & \bf 1.008 $\pm$ 0.463 \\
				
				\midrule
                
                Avg. & 2.367$\pm$ 1.348 & 8.386$\pm$ 5.119 & 5.179$\pm$ 2.627 & 2.090$\pm$ 1.492 & 0.446$\pm$ 0.337 & \bf 0.307$\pm$ 0.140 \\

				\bottomrule
			\end{tabular}
		}
	\end{center}
\end{table}

\begin{table}[t]
	\begin{center}
		\vspace{-1em}
		\captionsetup {font={small,stretch=0.5}}
		\caption{{\textbf{Quantitative comparisons on real datasets in terms of the Relative Pose Error for rotation part (\textdegree/frame).}} The experimental results demonstrate that rolling shutter distortions affect the accuracy of motion trajectory estimations. Due to proper modeling, our method performs much better than state-of-the-art methods. It also demonstrates that cubic B-Spline interpolation is superior to linear interpolation. x denotes method failed on the corresponding sequence.}
		\label{rpe_rot_real}
		\vspace{-0.5em}
		\scriptsize
		\setlength{\tabcolsep}{2.5mm}
		\setlength{\belowcaptionskip}{-10pt}
		\resizebox{\linewidth}{!}{
			\begin{tabular}{c|c c c c c c}
				\toprule
				\makebox[0.08\textwidth][c]{} & \makebox[0.08\textwidth][c]{\scriptsize COLMAP} & \makebox[0.08\textwidth][c]{\scriptsize BARF} & \makebox[0.08\textwidth][c]{\scriptsize RSBA} & \makebox[0.08\textwidth][c]{\scriptsize NW-RSBA} & \makebox[0.08\textwidth][c]{\scriptsize USB-NeRF-linear} &\makebox[0.08\textwidth][c]{\scriptsize USB-NeRF-cubic} \\
				%			&{\scriptsize \qquad COLMAP \qquad} &{\scriptsize BARF} &{\scriptsize \qquad RSBA \qquad} & {\scriptsize NW-RSBA} & {\scriptsize \qquad USB-NeRF-linear \qquad} & {\scriptsize \qquad USB-NeRF-cubic \qquad} \\
				\midrule
				seq1  & 0.278 $\pm$ 0.284 & 0.283 $\pm$ 0.152 & 0.578 $\pm$ 0.470 & 0.671 $\pm$ 0.545 & 0.114 $\pm$ 0.072 & \bf 0.108 $\pm$ 0.071 \\
				seq2  & 0.794 $\pm$ 0.428 & \bf 0.737 $\pm$ 0.432 & 9.789 $\pm$ 14.344 & 28.325 $\pm$ 40.662 & 1.365 $\pm$ 1.733 & 0.777 $\pm$ 0.838 \\
				seq3  & 0.271 $\pm$ 0.145 & 0.272 $\pm$ 0.176 & 0.397 $\pm$ 0.208 & 0.250 $\pm$ 0.116 & 0.159 $\pm$ 0.115 & \bf 0.145 $\pm$ 0.108 \\
				seq4  & 0.225 $\pm$ 0.100 & 0.211 $\pm$ 0.110 & 0.247 $\pm$ 0.126 & 1.049 $\pm$ 0.571 & 0.160 $\pm$ 0.093 & \bf 0.155 $\pm$ 0.093 \\
				seq5  & 0.607 $\pm$ 0.286 & 0.545 $\pm$ 0.199 & 2.245 $\pm$ 1.784 & 3.131 $\pm$ 2.316 & 0.168 $\pm$ 0.087 & \bf 0.166 $\pm$ 0.082 \\
				seq6  & 0.345 $\pm$ 0.163 & 0.322 $\pm$ 0.247 & 0.575 $\pm$ 0.257 & 0.851 $\pm$ 0.874 & 0.180 $\pm$ 0.089 & \bf 0.149 $\pm$ 0.080 \\
				seq7  & 0.219 $\pm$ 0.123 & 0.224 $\pm$ 0.143 & 0.324 $\pm$ 0.201 & 1.063 $\pm$ 0.700 & 0.146 $\pm$ 0.100 & \bf 0.128 $\pm$ 0.097 \\
				seq8  & 0.520 $\pm$ 0.325 & 0.249 $\pm$ 0.103 & 0.502 $\pm$ 0.251 & 1.718 $\pm$ 1.011 & 0.151 $\pm$ 0.069 & \bf 0.135 $\pm$ 0.066 \\
				seq9  & 0.322 $\pm$ 0.165 & 0.445 $\pm$ 0.276 & 4.400 $\pm$ 7.312 & 41.983 $\pm$ 47.732 & \bf 0.193 $\pm$ 0.211 & 0.263 $\pm$ 0.524 \\
				seq10 & 0.351 $\pm$ 0.165 & 1.006 $\pm$ 2.460 & 1.889 $\pm$ 1.117 & 4.752 $\pm$ 4.748 & \bf 0.193 $\pm$ 0.277 & 0.250 $\pm$ 0.499 \\

				\midrule
				Avg.   & 0.393 $\pm$ 0.218 & 0.429 $\pm$ 0.430 & 2.095 $\pm$ 2.607 & 8.379 $\pm$ 9.928 & 0.283 $\pm$ 0.285 & \bf 0.227 $\pm$ 0.246 \\
				\bottomrule
			\end{tabular}
		}
	\end{center}
\end{table}

To further evaluate our method, we also train USB-NeRF with un-ordered images (\ie there is no pose dependency among input frames). As mentioned in our main paper and proved by \citet{albl2016degeneracies}, rolling shutter bundle adjustment with un-ordered images would have degenerated solutions if the input images are not properly captured. Thus, we follow the instructions suggested by \citet{albl2016degeneracies}, \ie the mutual angle between readout directions should be larger than 30 degrees, to generate the training dataset. We synthesized 3 sequences of un-ordered rolling shutter images with Unreal game engine ({i.e.} Unreal-RS-BlueRoom, Unreal-RS-LivingRoom, Unreal-RS-Roof) in total. During experiments, we represent the camera motion within individual frame readout time by cubic B-Spline with 4 control knots. \tabnref{table_unorganized} and \figrefer{fig_unordered} present the experimental results. Since prior rolling shutter effect removal networks usually rely on two consecutive frames to restore the global shutter image, they are not suitable for this dataset. We only evaluate our method against both NeRF \citep{mildenhall2020nerf} and BARF \citep{lin2021barf}. The experimental results demonstrate that our method also outperforms prior methods with un-ordered rolling shutter images. 

\begin{figure}[!ht]
	\setlength\tabcolsep{1.pt}
	\centering
	\begin{tabular}{cccc}

		\includegraphics[width=0.24\textwidth]{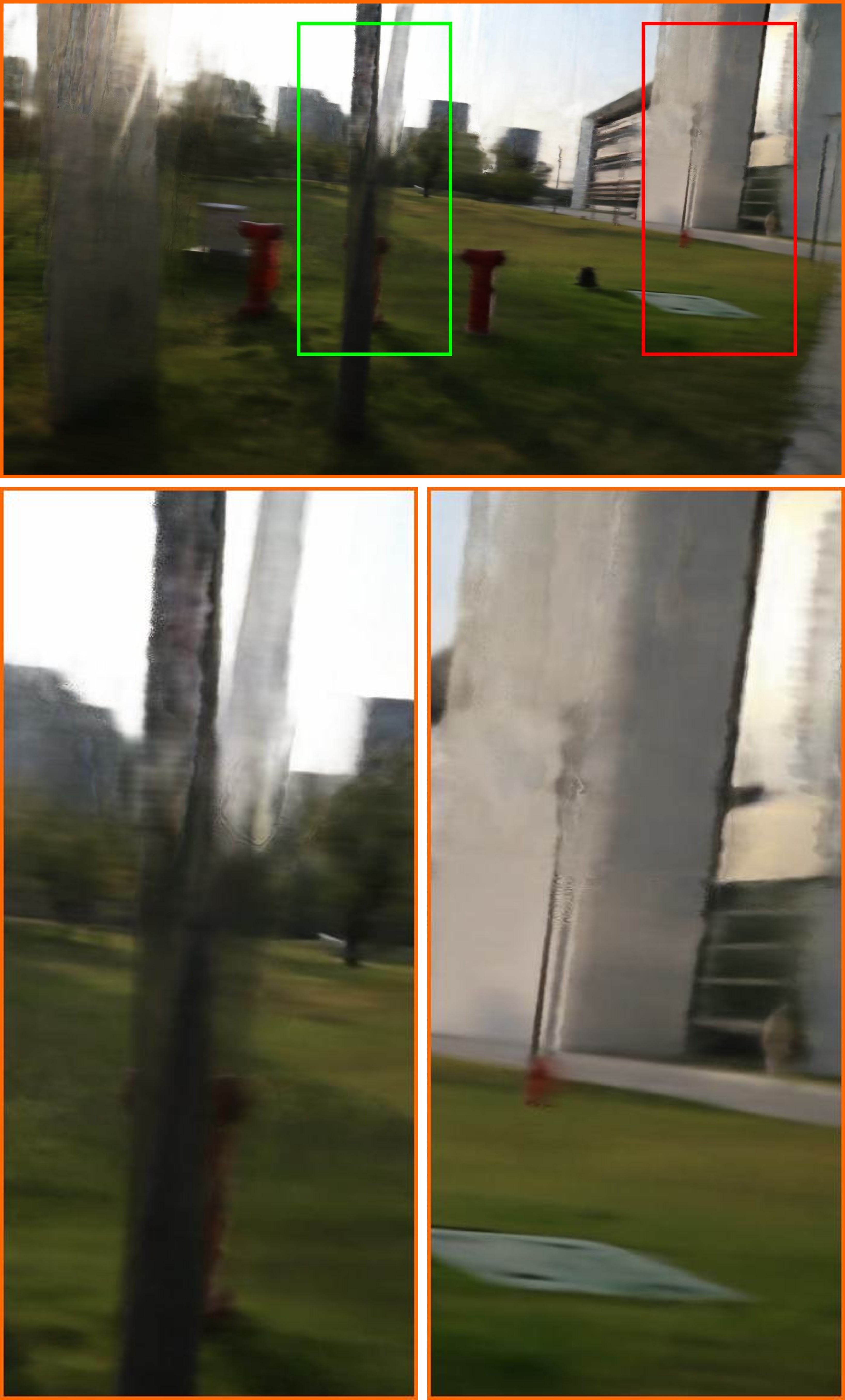} &
		\includegraphics[width=0.24\textwidth]{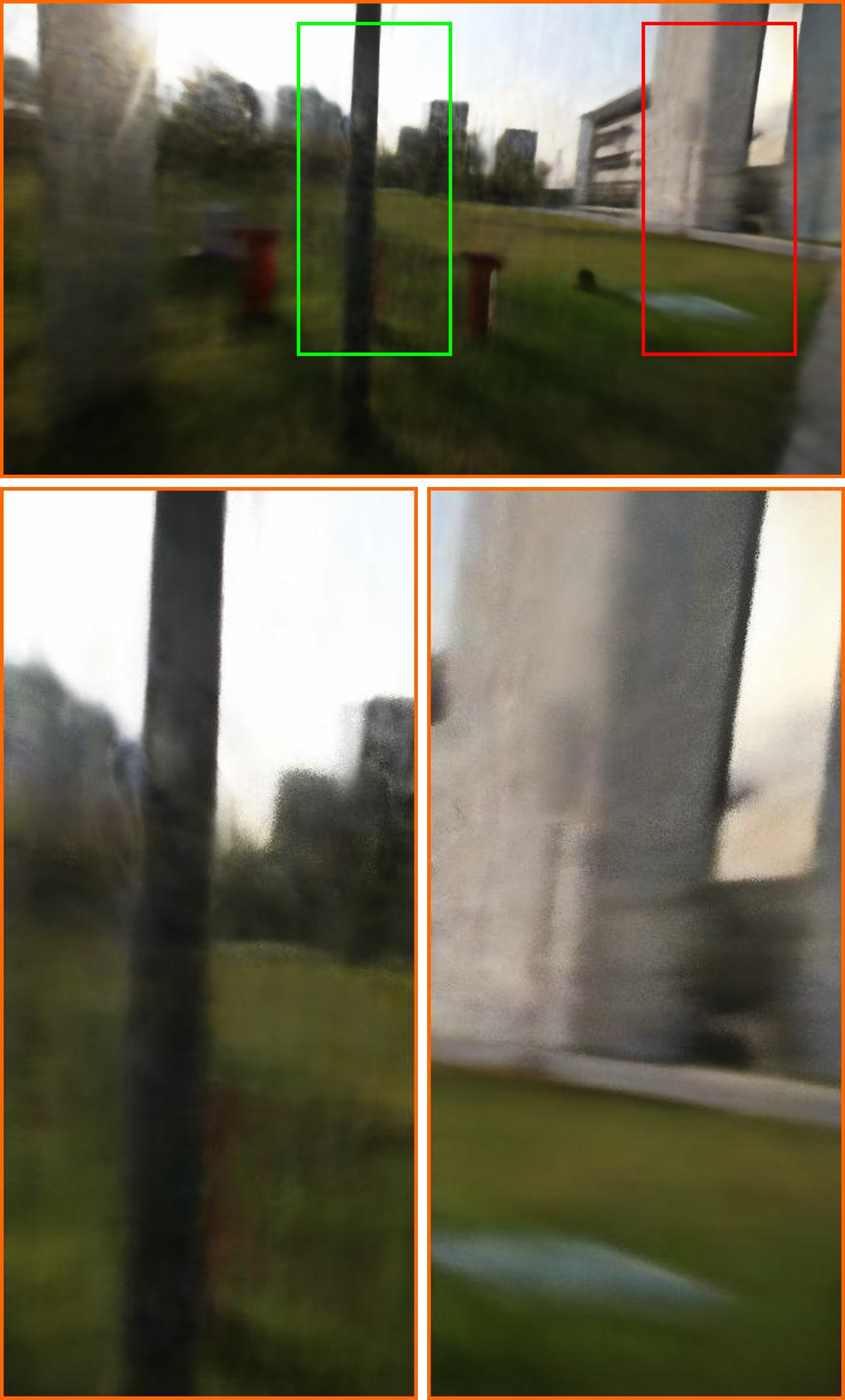} &
		\includegraphics[width=0.24\textwidth]{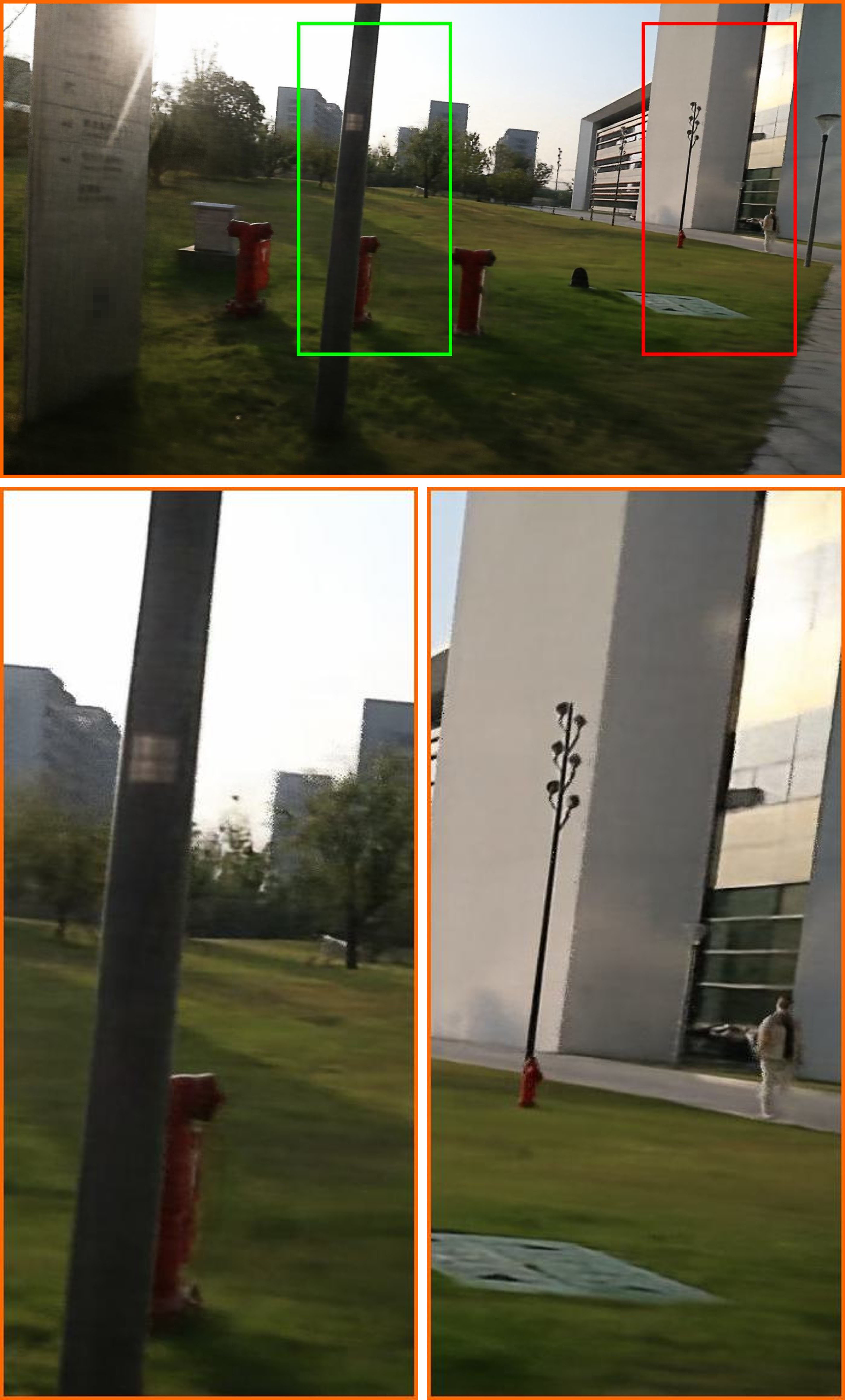} &
		\includegraphics[width=0.24\textwidth]{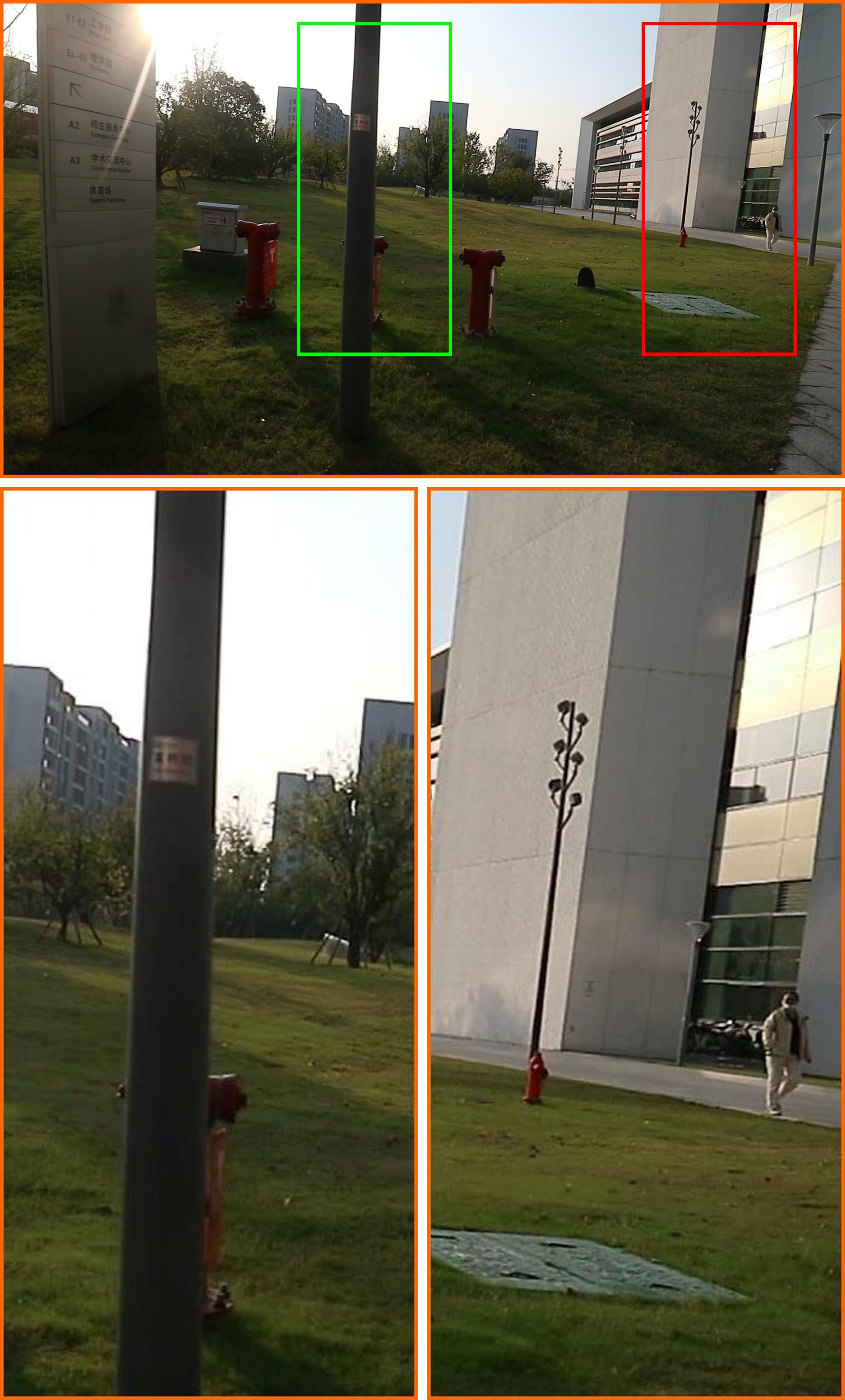} \\

		\includegraphics[width=0.24\textwidth]{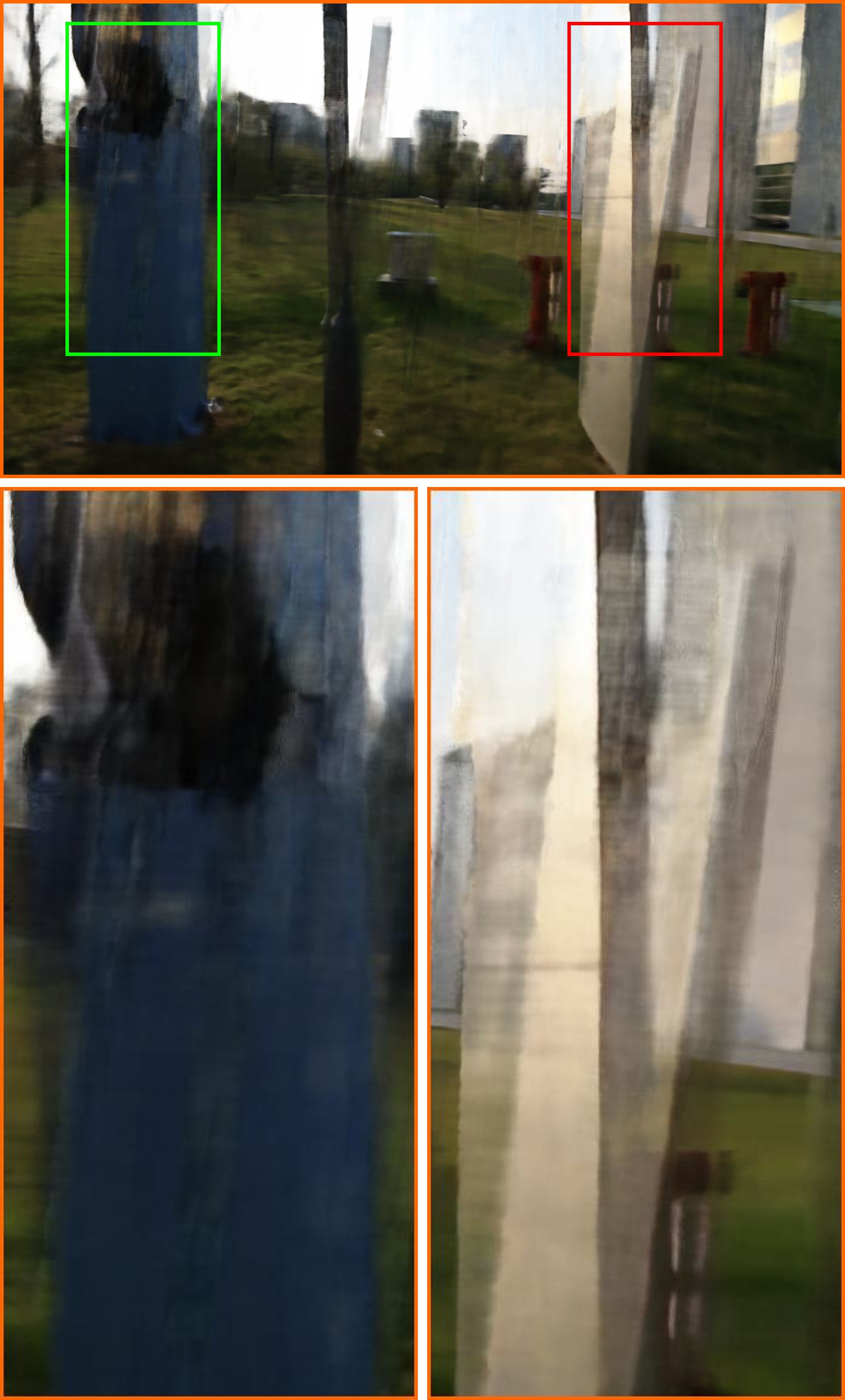} &
		\includegraphics[width=0.24\textwidth]{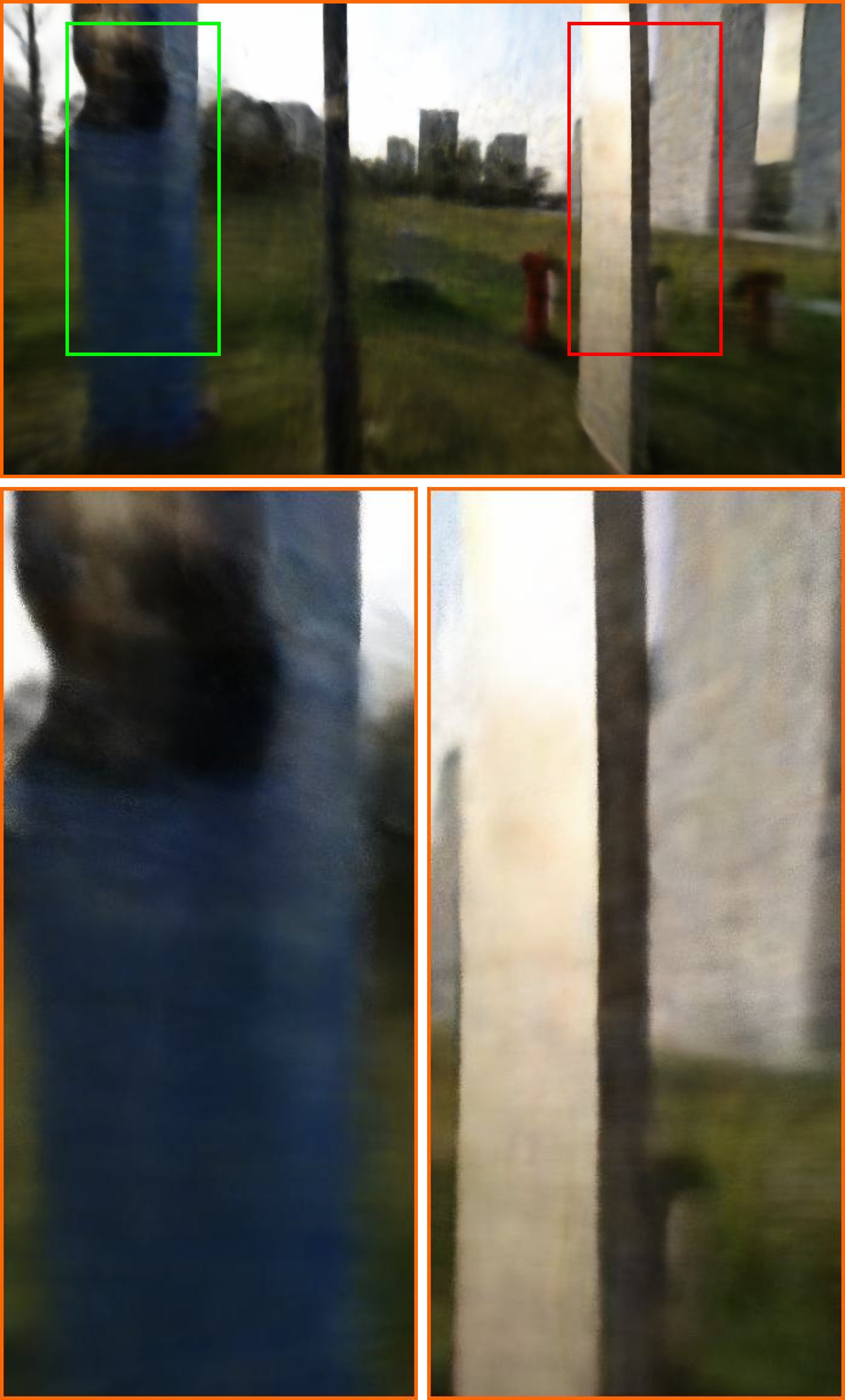} &
		\includegraphics[width=0.24\textwidth]{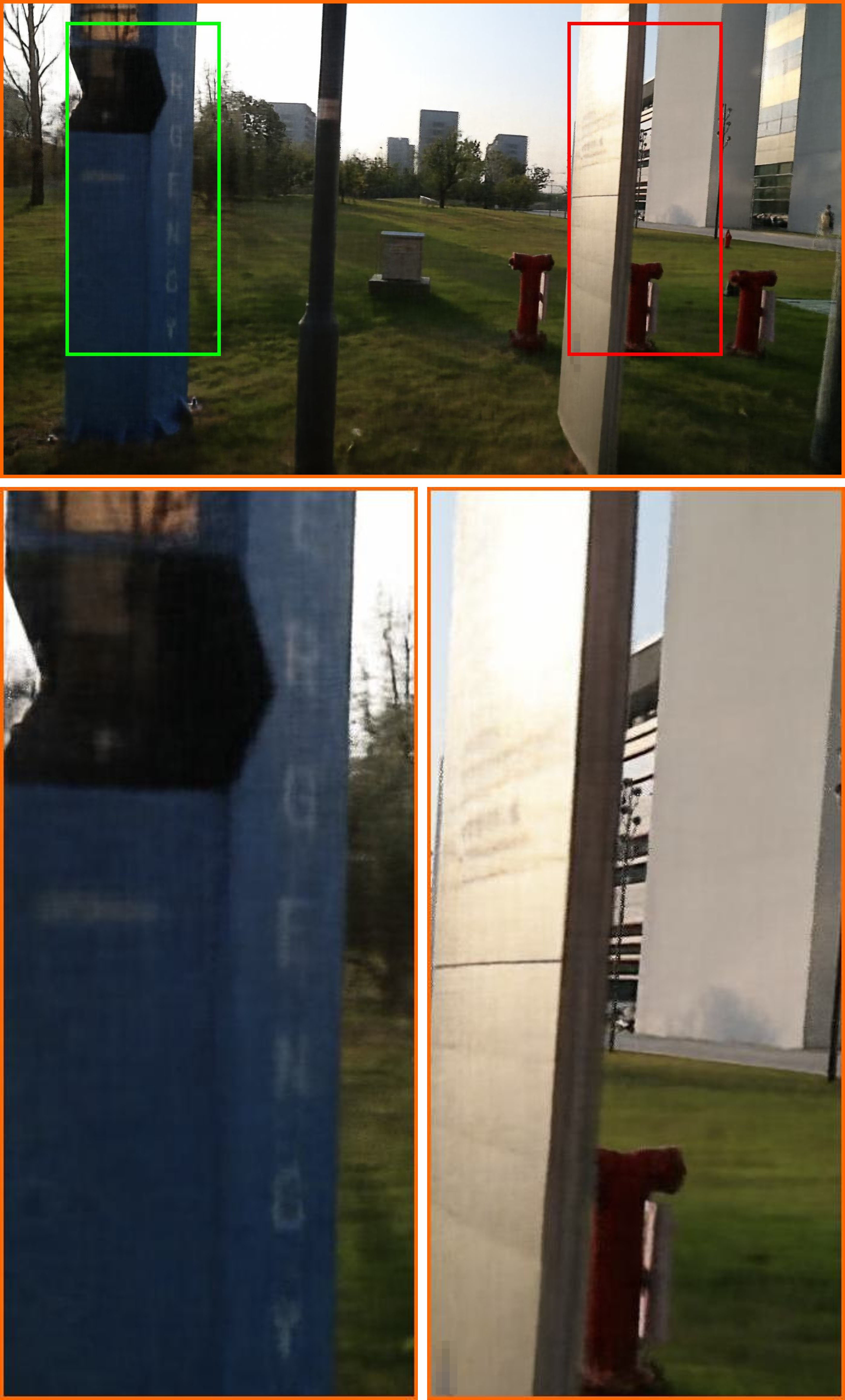} &
		\includegraphics[width=0.24\textwidth]{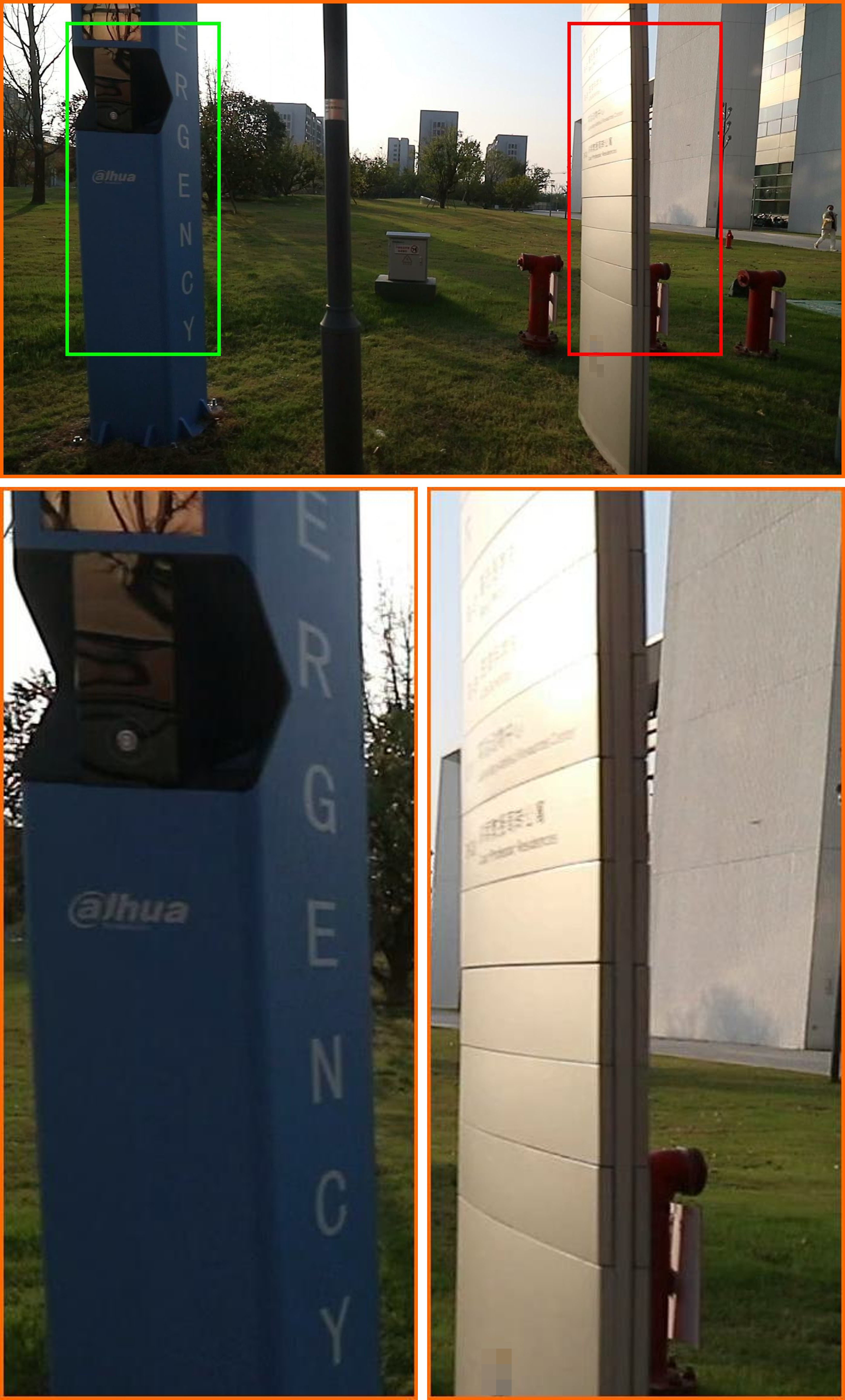} \\
		% \specialrule{-0.1em}{0.05pt}{0.05pt}
		% \specialrule{0em}{0.05pt}{0.05pt}
		\scriptsize{NeRF \citep{mildenhall2020nerf}} & \scriptsize{BARF \citep{lin2021barf}}  & \scriptsize{USB-NeRF} & \scriptsize{Input RS image}
	\end{tabular}
	\captionsetup {font={small,stretch=0.5}}
	\caption{\textbf{Qualitative comparisons on real-world datasets captured by a Canon camera.} The camera exhibits forward and backward motion, which challenges both NeRF and BARF to recover the true underlying 3D scene representation. Our method recovers the correct global shutter images. Note the camera is not parallel to the ground during capture.}
	\label{fig_rs_Canon}
\end{figure}

\begin{figure}[!ht]
	\vspace{1.em}
	\setlength\tabcolsep{1.pt}
	\centering
	\begin{tabular}{cccc}
		\includegraphics[width=0.24\textwidth]{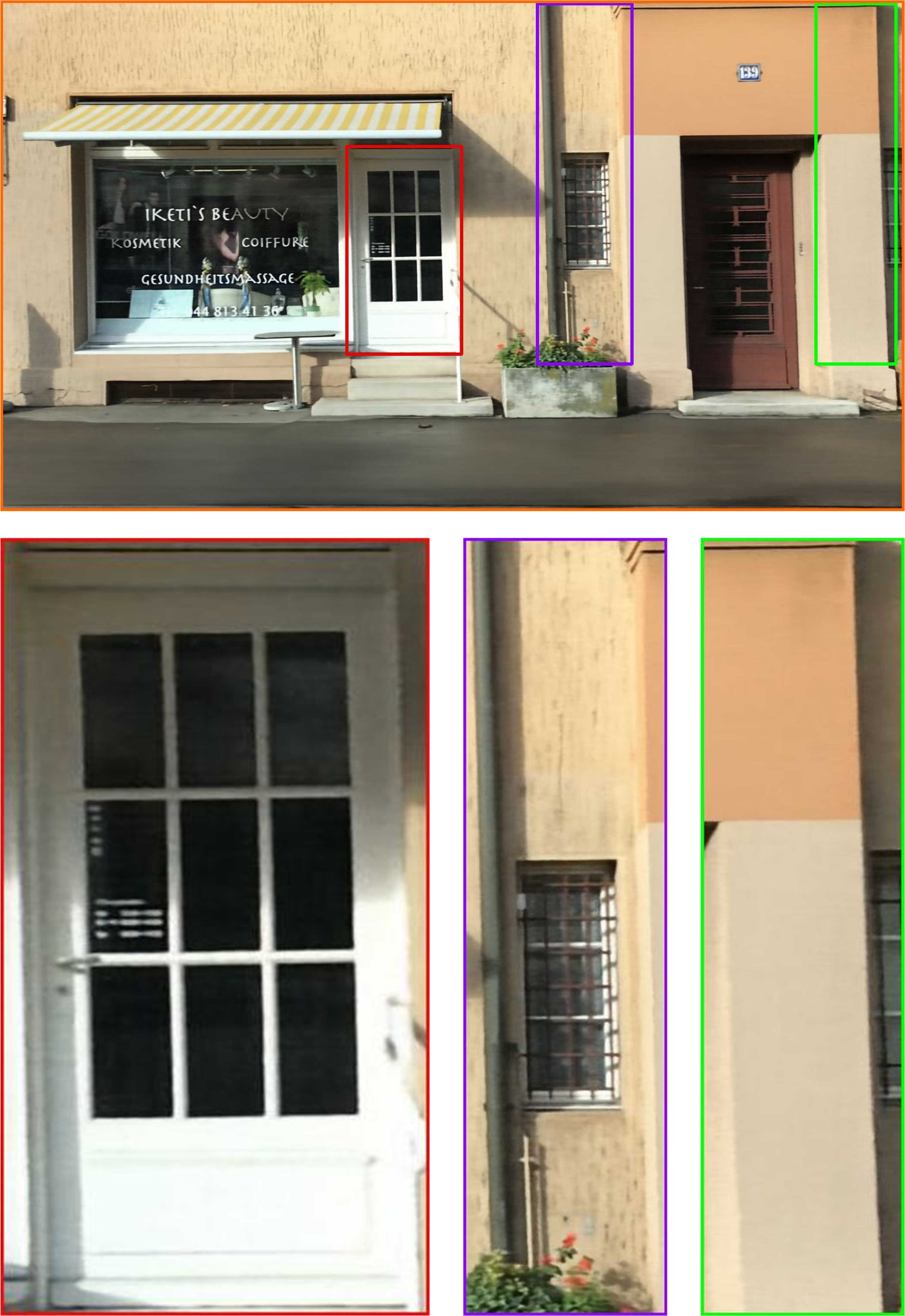} &
		\includegraphics[width=0.24\textwidth]{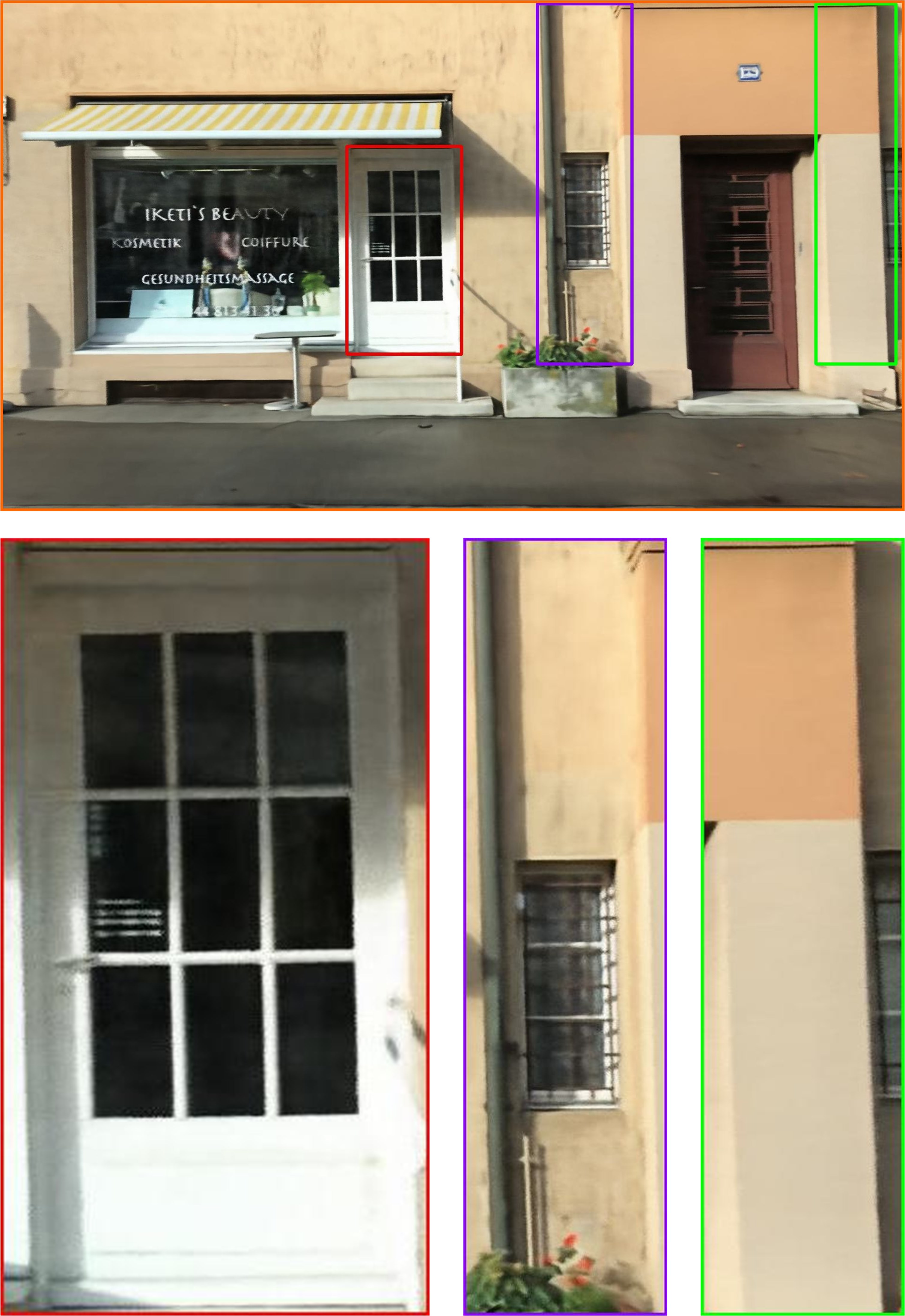} &
		\includegraphics[width=0.24\textwidth]{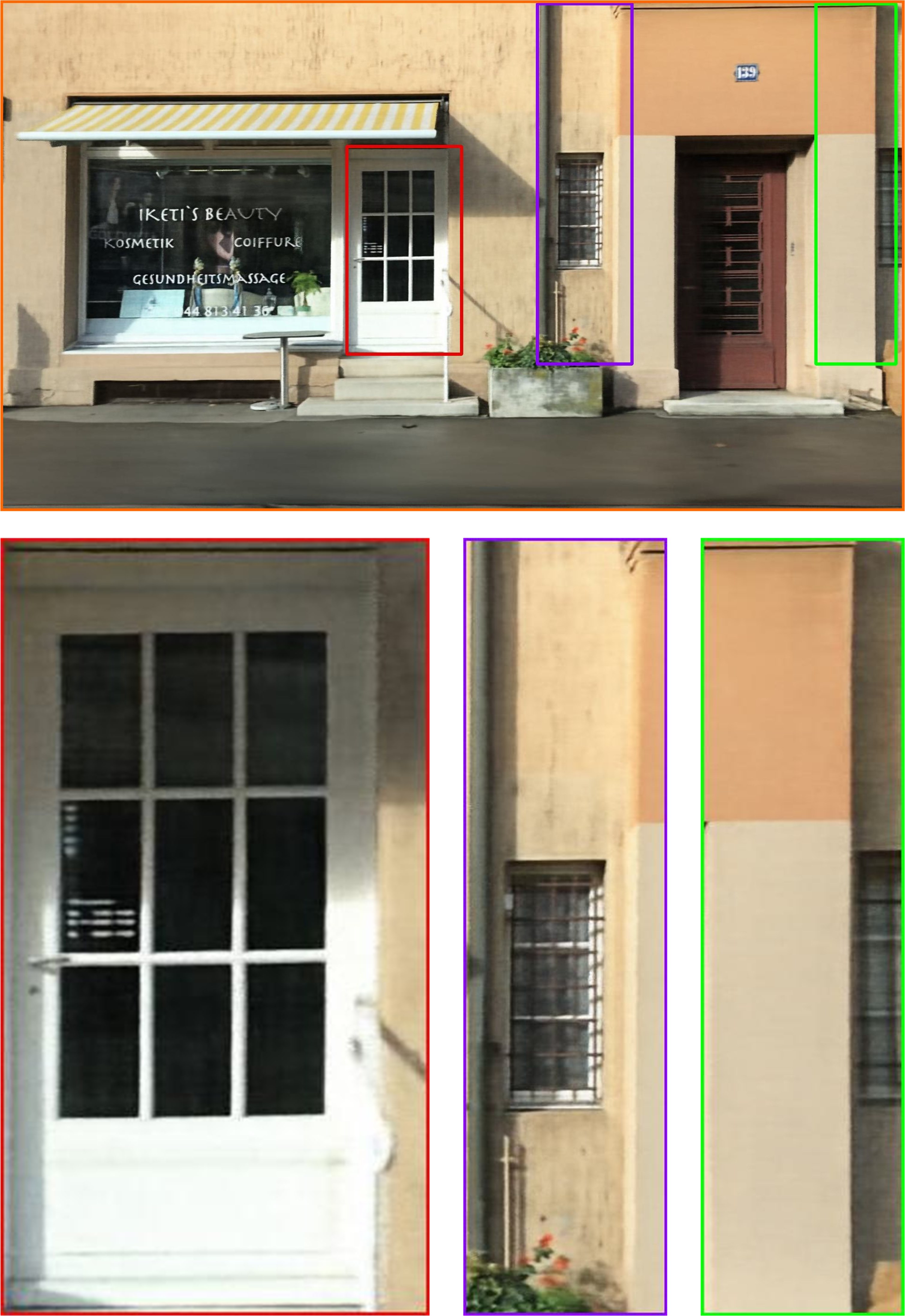} &
		\includegraphics[width=0.24\textwidth]{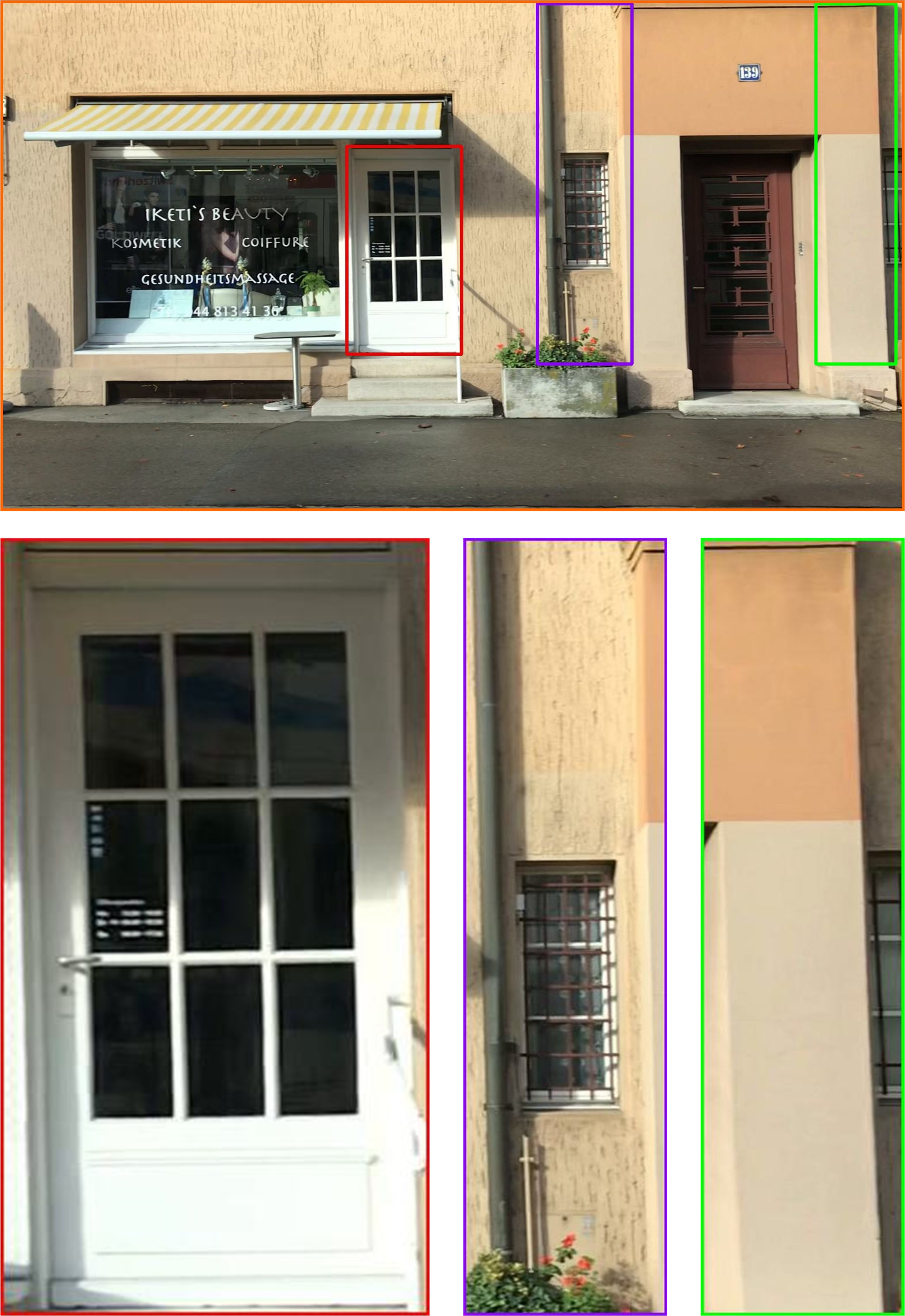} \\

        \scriptsize{NeRF \citep{mildenhall2020nerf}} & \scriptsize{BARF \citep{lin2021barf}}  & \scriptsize{USB-NeRF} & \scriptsize{Input RS image}
	\end{tabular}
%	\vspace{-0.5em}
	\captionsetup {font={small,stretch=0.5}}
	\caption{{\bf{Qualitative comparisons on real-world datasets captured by GoPro HERO6.}} The images are captured on a moving tram in constant direction. It demonstrates that even both NeRF and BARF can render clear images, the recovered scene is distorted. Our method can correctly un-distort it.}
	\label{fig_rs_GoPro}
\end{figure}

\begin{figure}[!ht]
	\setlength\tabcolsep{1.pt}
	\centering
	\begin{tabular}{cccc}
        \includegraphics[width=0.24\textwidth]{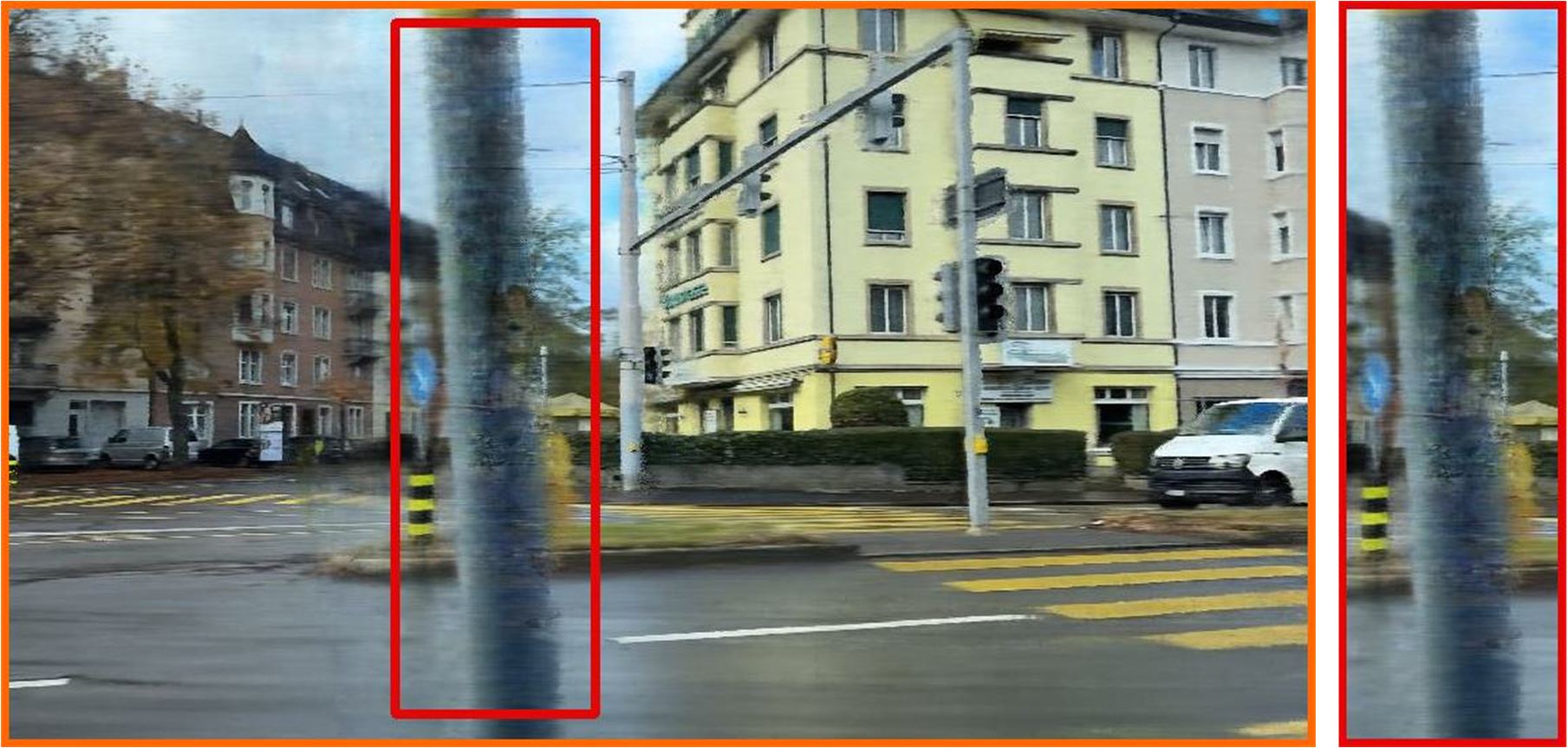} &
		\includegraphics[width=0.24\textwidth]{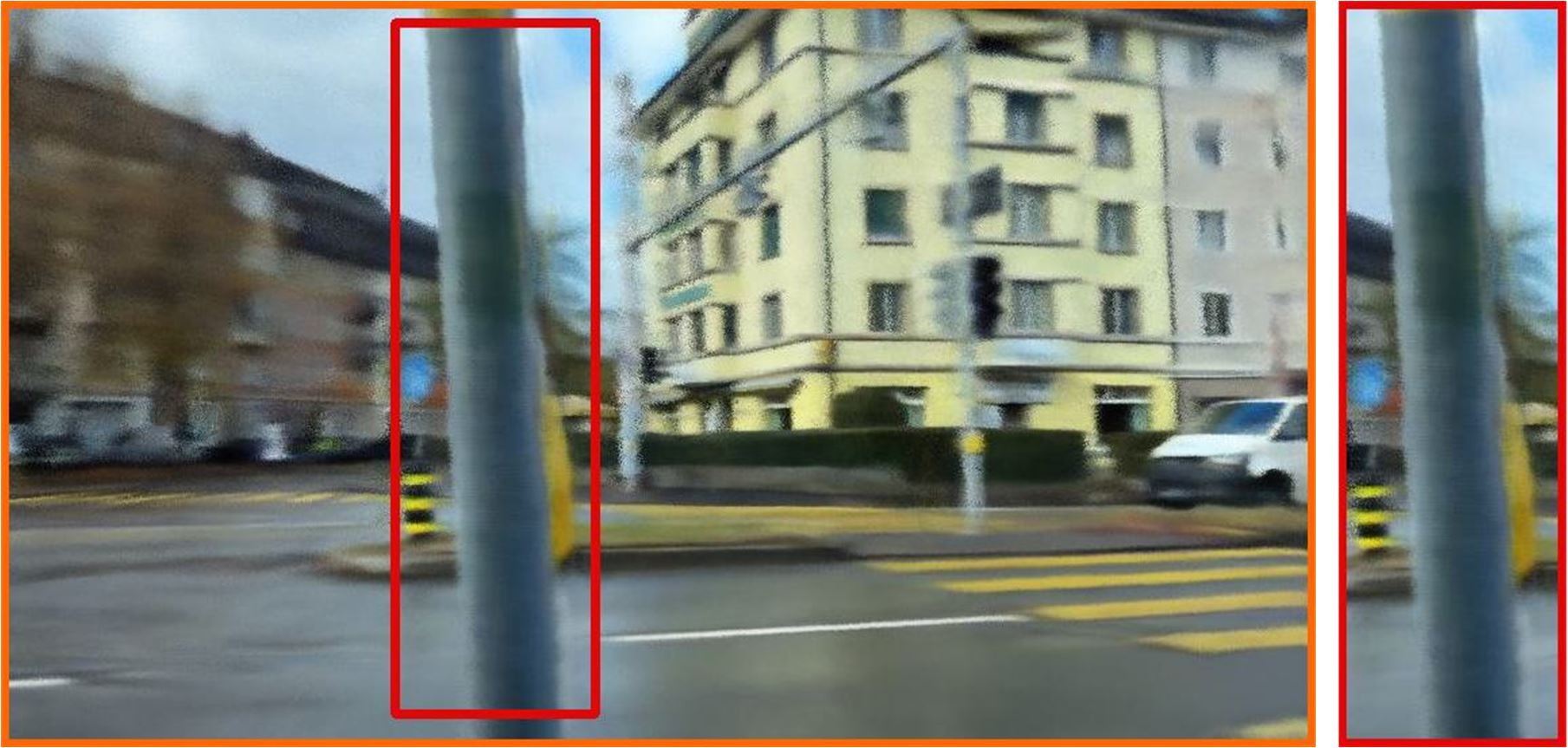} &
		\includegraphics[width=0.24\textwidth]{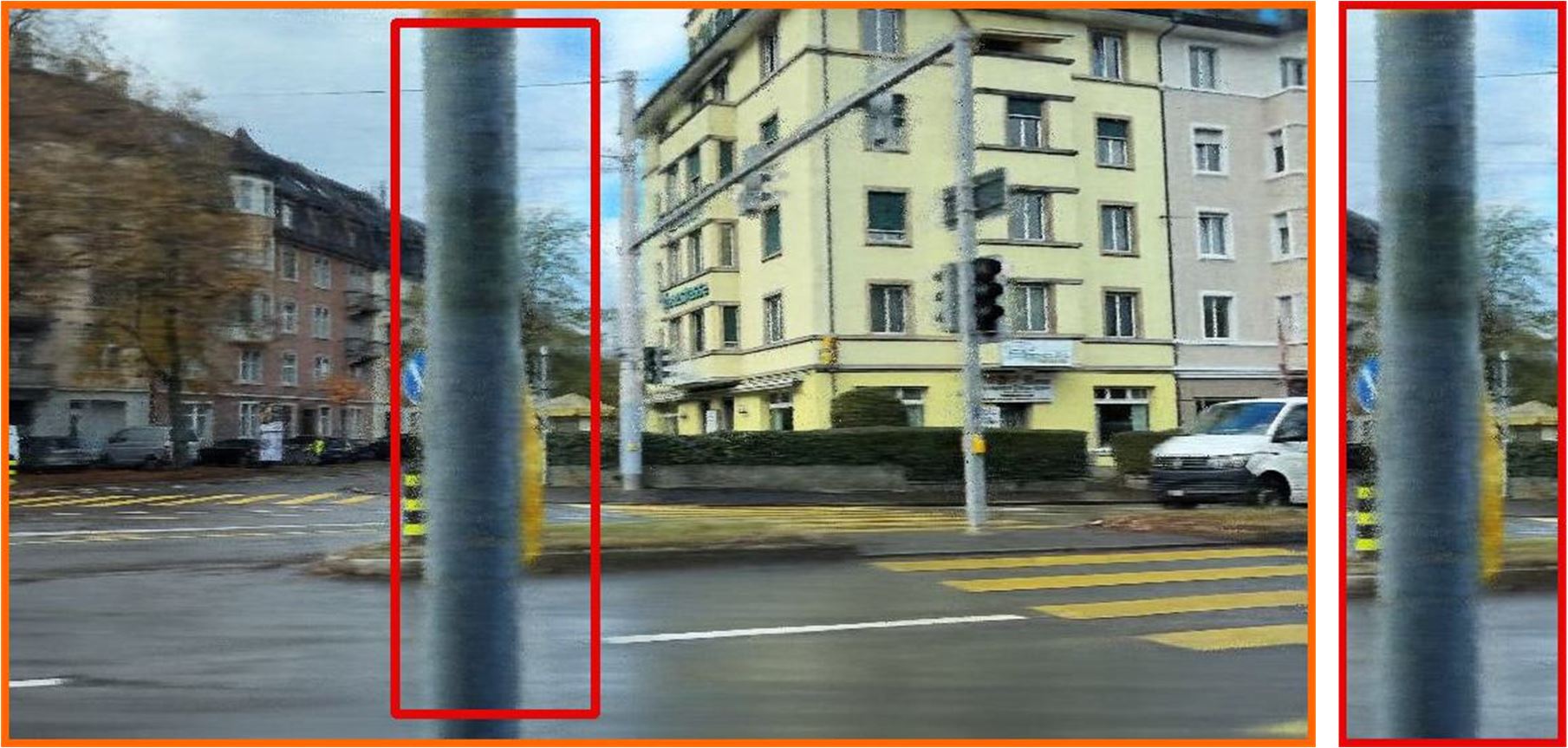} &
		\includegraphics[width=0.24\textwidth]{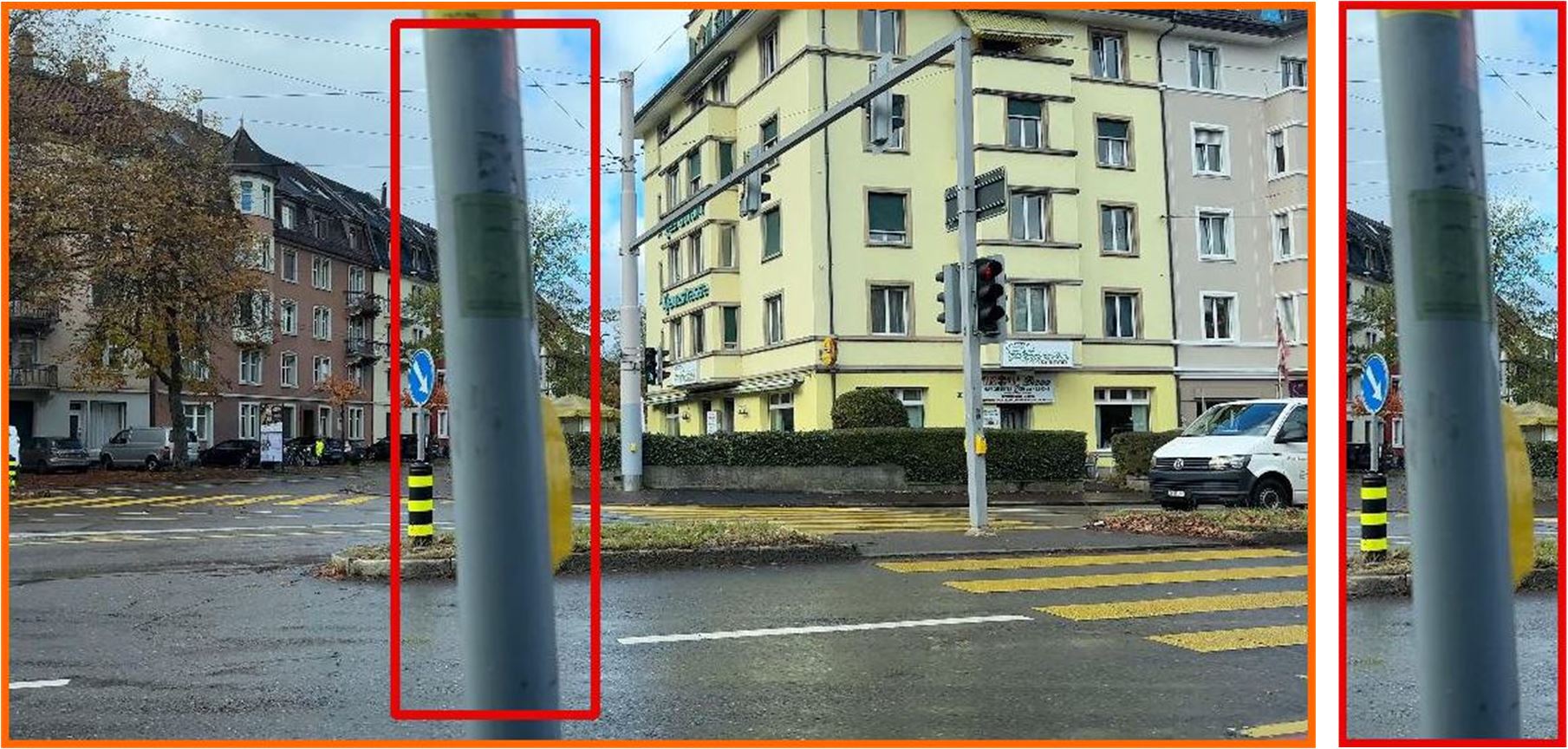} \\

        \includegraphics[width=0.24\textwidth]{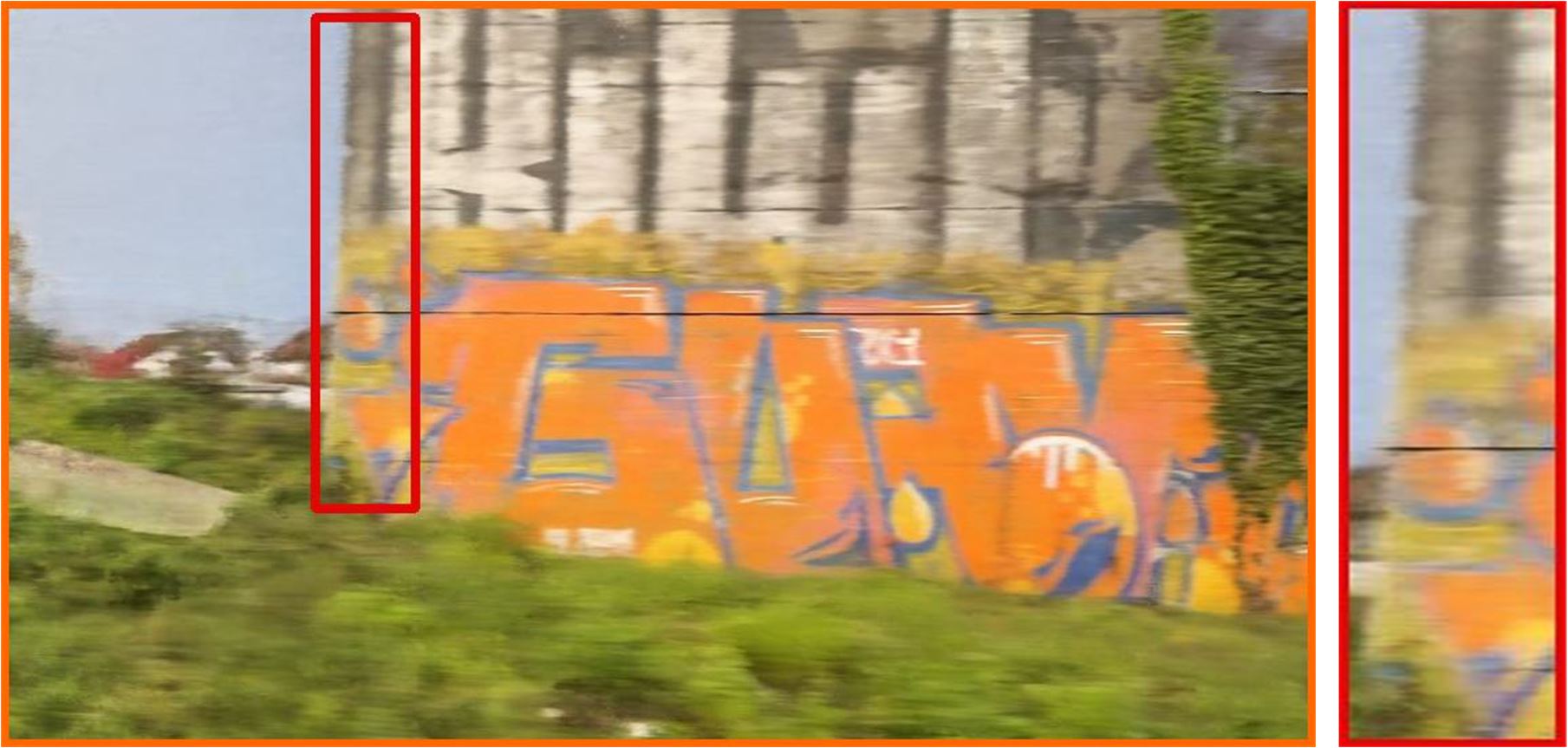} &
		\includegraphics[width=0.24\textwidth]{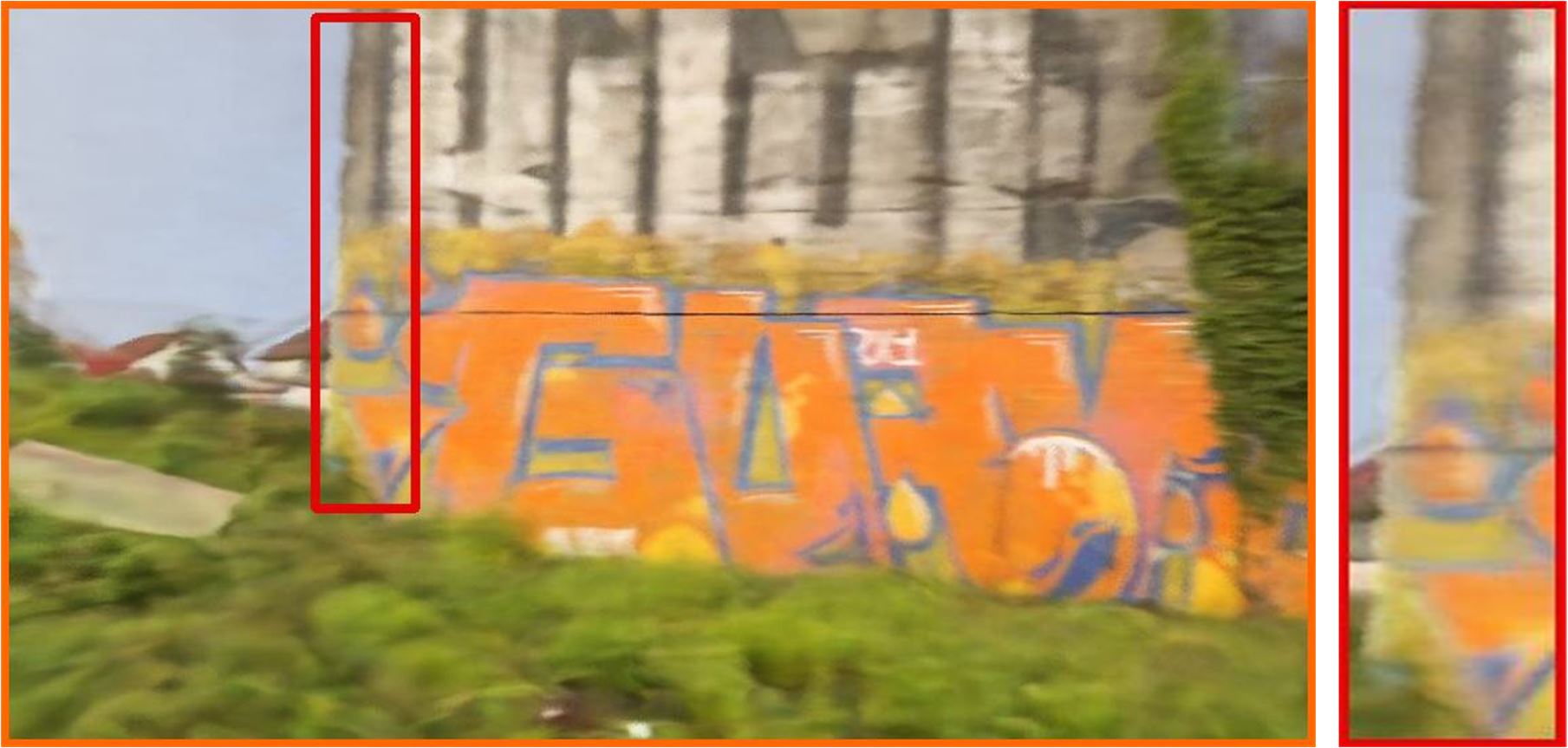} &
		\includegraphics[width=0.24\textwidth]{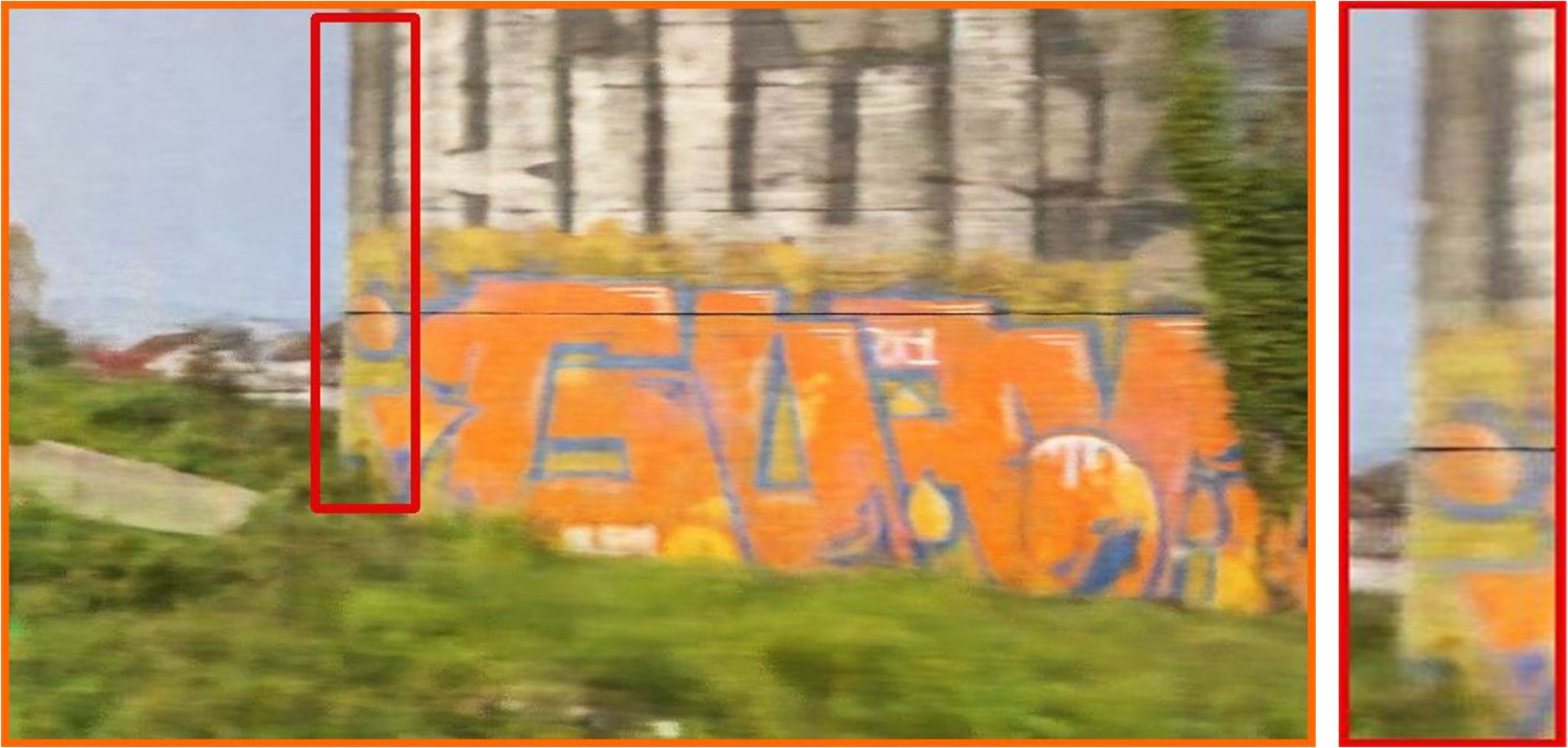} &
		\includegraphics[width=0.24\textwidth]{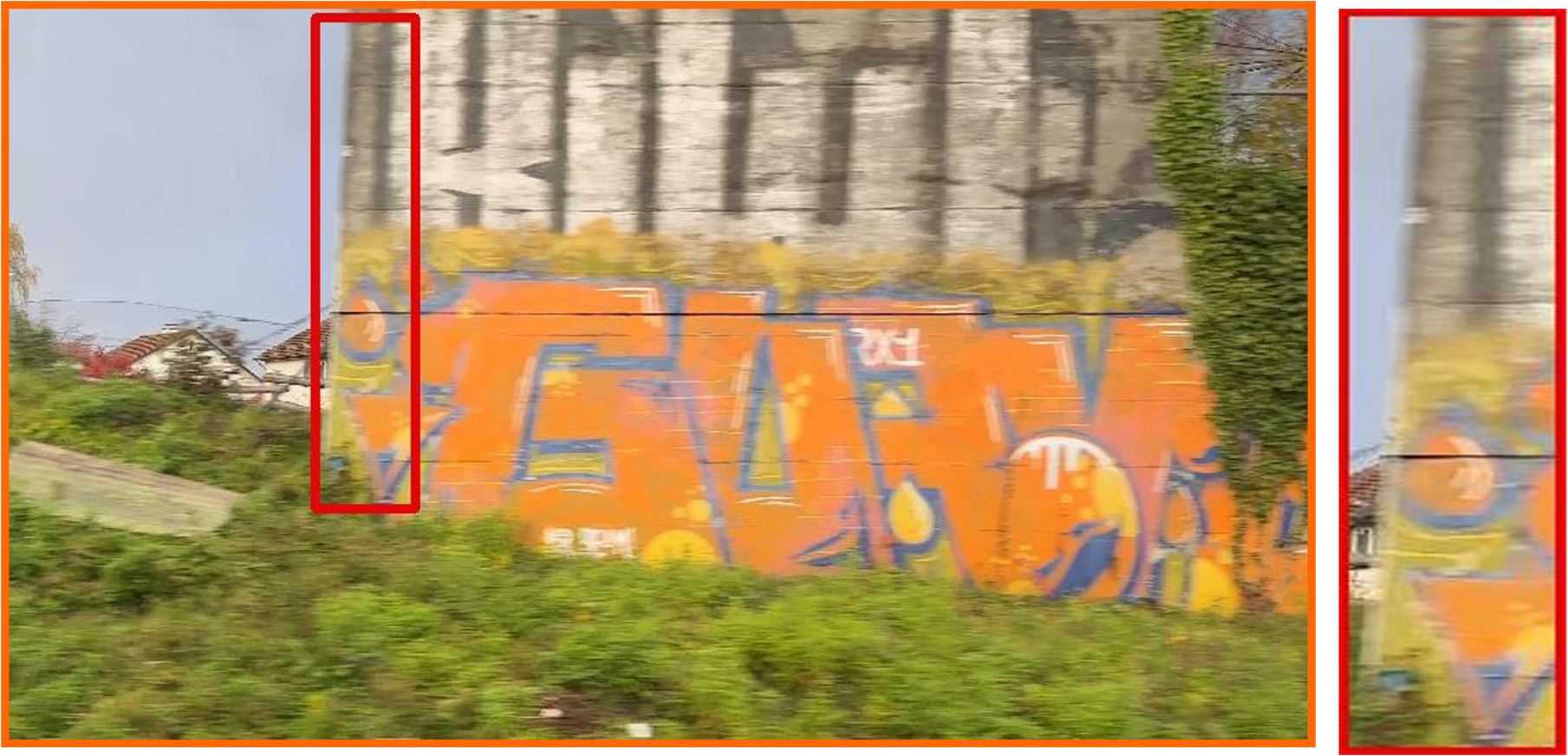} \\

        \includegraphics[width=0.24\textwidth]{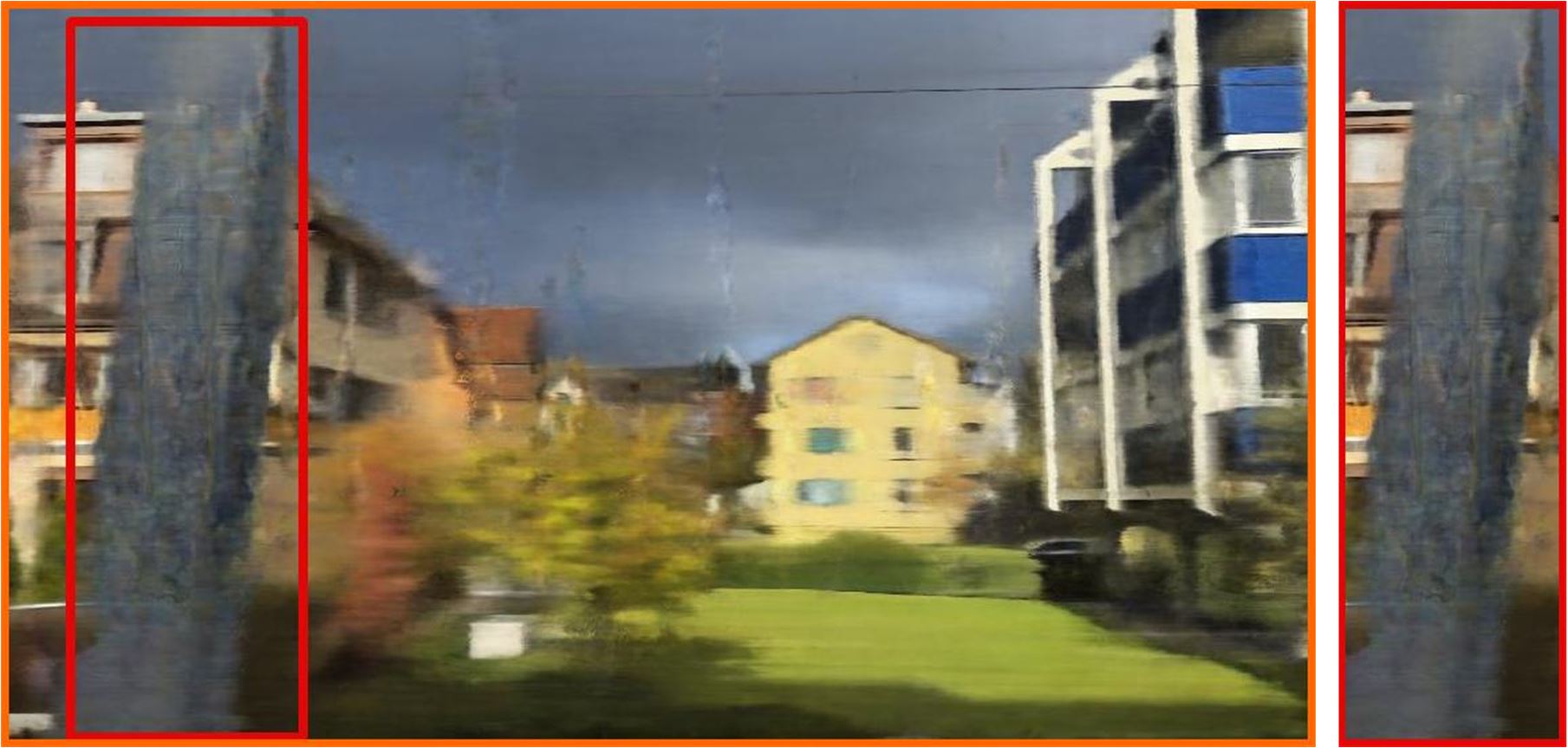} &
		\includegraphics[width=0.24\textwidth]{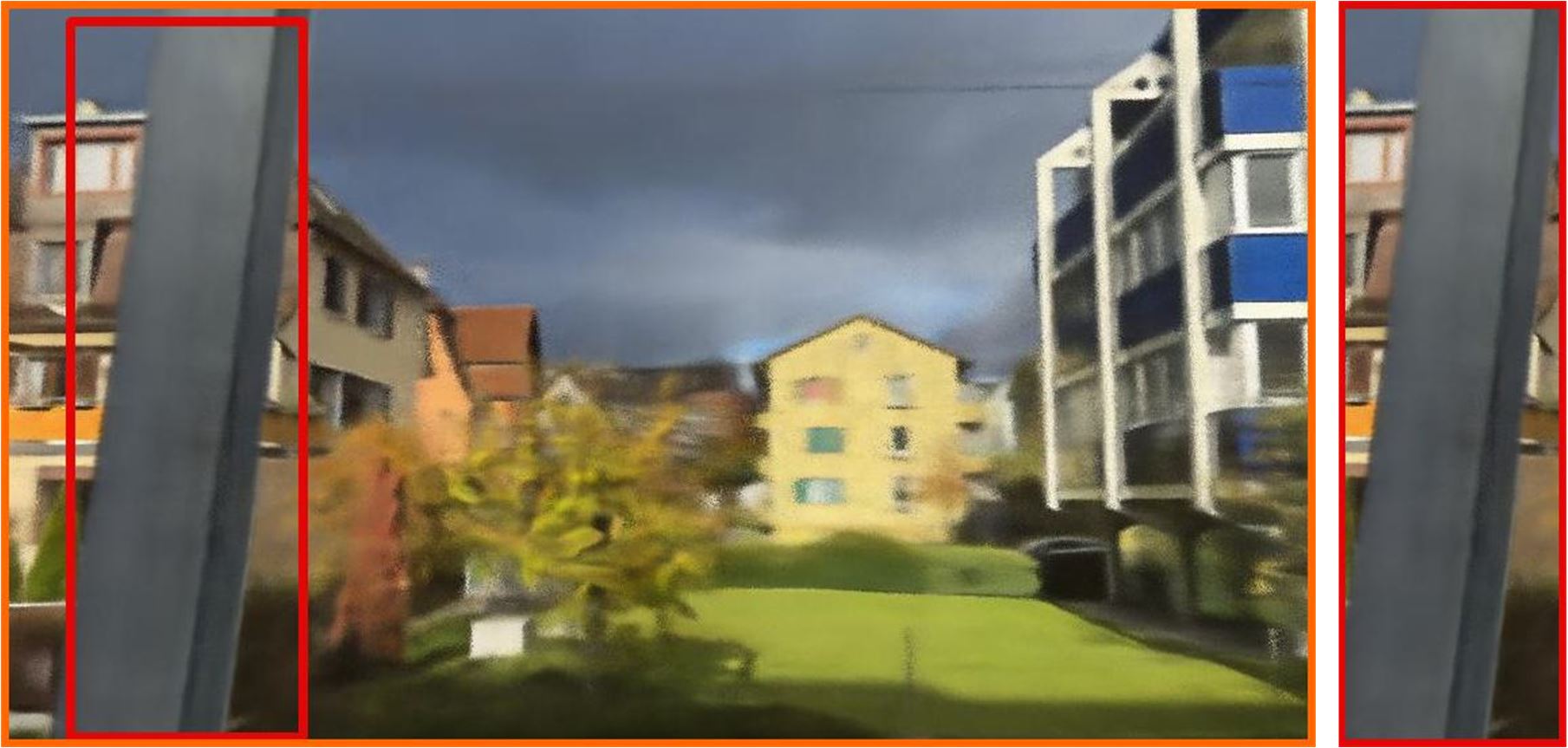} &
		\includegraphics[width=0.24\textwidth]{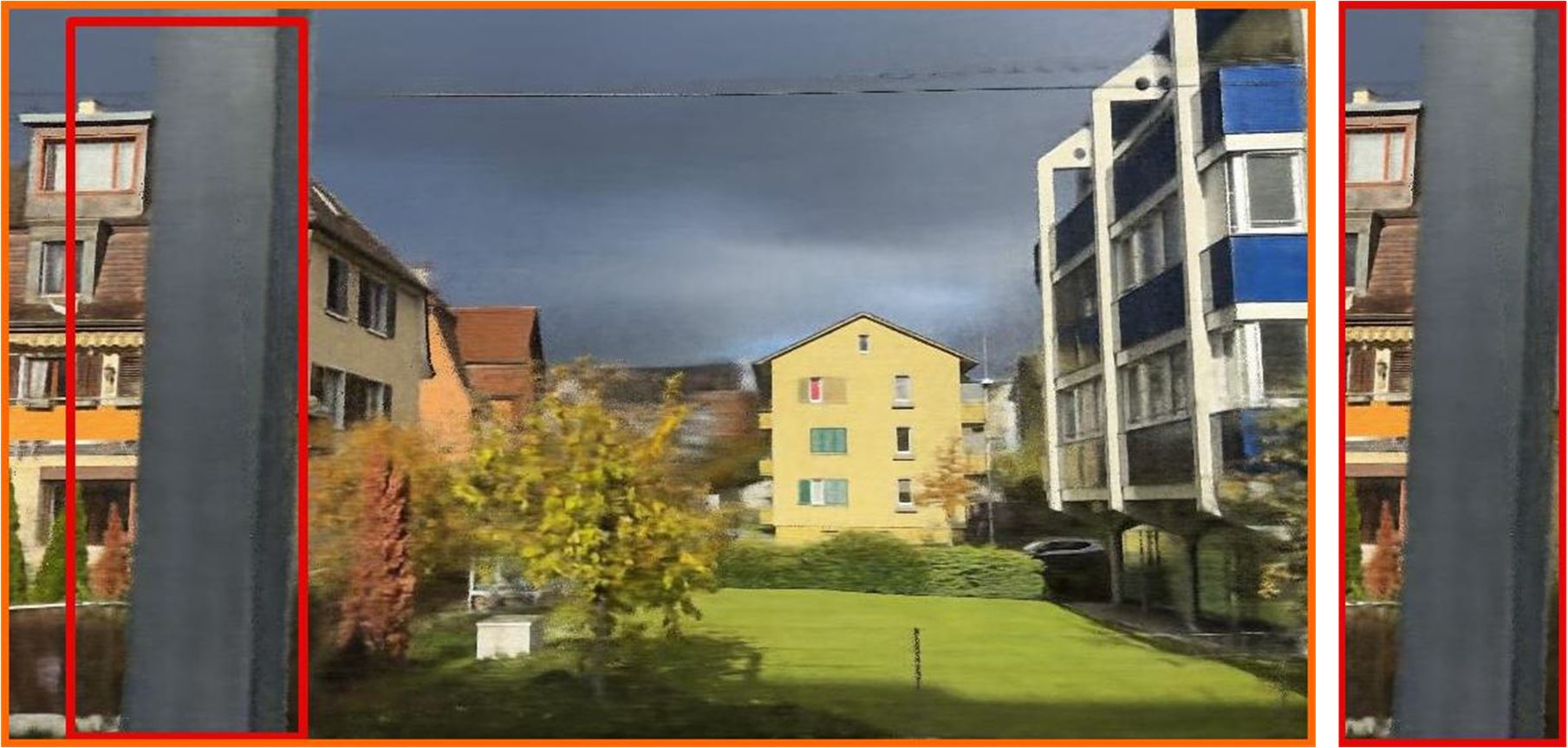} &
		\includegraphics[width=0.24\textwidth]{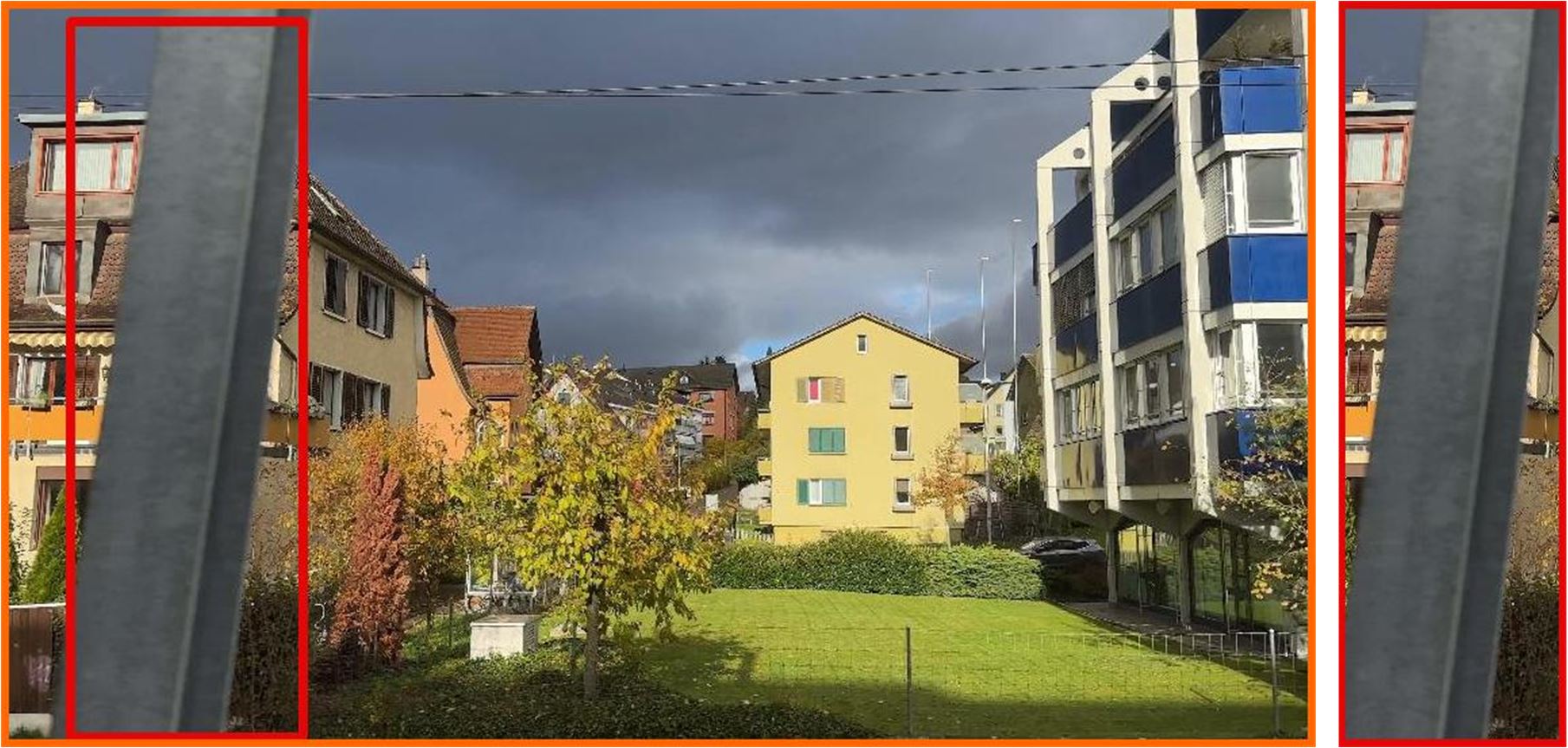} \\
        
        \scriptsize{NeRF \citep{mildenhall2020nerf}} & \scriptsize{BARF \citep{lin2021barf}}  & \scriptsize{USB-NeRF} & \scriptsize{Input RS image}
	\end{tabular}
	\captionsetup {font={small,stretch=0.5}}
	\caption{{\bf{Qualitative comparisons on real-world datasets captured by iPhone 14 Pro.}} The images are captured on a moving bus in constant direction. It demonstrates that both NeRF and BARF fail to correct the distortion and produce additional artifacts, while our method successfully restore the true underlying 3D scene representation.}
	\label{fig_rs_iphone_crop}
\end{figure}

Additional experimental results on more synthetic datasets in terms of the rolling shutter effect removal are also presented in \tabnref{table_our_synthesis_longer} and \figrefer{fig_RS_removal_longer}. It demonstrates that our method also performs better than prior state-of-the-art methods. Additional results on real-world datasets captured using  GoPro HERO6 Black, Canon camera (EOS M3), and iPhone 14 Pro are presented in \figrefer{fig_rs_Canon}, \figrefer{fig_rs_GoPro} and \figrefer{fig_rs_iphone_crop}. The results on real-world dataset demonstrates that both NeRF\citep{mildenhall2020nerf} and BARF\citep{lin2021barf} fail to correct the RS distortion, while our method renders correct global shutter images with no artifact.

\tabnref{TUM_ATE} presents the details on trajectory estimation in terms of the ATE metric for translation error with the TUM-RS dataset. \tabnref{rpe_rot_synthetic} and \tabnref{rpe_rot_real} presents the details of RPE metric for rotation error. The results show that our method performs consistently better than both COLMAP and BARF. It also demonstrates that our method is able to perform on-par against prior rolling shutter bundle adjustment methods, i.e. RSBA and NW-RSBA, and achieves better performance in terms of average ATE and RPE metrics over all sequences.

\subsection{Trajectory Visualization}
\label{sec:trajectory_visualization}
%
%\begin{figure}[!ht]
%	\setlength{\belowcaptionskip}{-6pt}
%	\setlength\tabcolsep{1pt}
%	\centering
%	\begin{tabular}{cc}
%		\raisebox{.9in}{\rotatebox[origin=t]{90}{\,}}
%		\includegraphics[width=0.49\textwidth]{supplementary_figure/trajectory/BlueRoom.pdf} &
%		\includegraphics[width=0.49\textwidth]{supplementary_figure/trajectory/LivingRoom.pdf} \\
%		Blue Room & Living Room  \\
%		\includegraphics[width=0.49\textwidth]{supplementary_figure/trajectory/WhiteRoom.pdf} &
%		\includegraphics[width=0.49\textwidth]{supplementary_figure/trajectory/Adornment.pdf} \\
%		White Room & Adornment \\
%		\includegraphics[width=0.49\textwidth]{supplementary_figure/trajectory/Factory.pdf} &
%		\includegraphics[width=0.49\textwidth]{supplementary_figure/trajectory/Tanabata.pdf} \\
%		Factory & Tanabata \\
%		\specialrule{0em}{0.05pt}{0.05pt}
%	\end{tabular}
%	\vspace{-0.0em}
%	\captionsetup {font={small,stretch=0.5}}
%	\caption{{\bf{Comparisons of estimated trajectories on synthetic RS dataset.}} The experimental results demonstrate that USB-NeRF is able to recover the motion trajectory accurately.}
%	\label{fig_syn_traj}
%	%	\vspace{-0.6em}
%\end{figure}

\begin{figure}[htbp]
	\setlength\tabcolsep{1.pt}
	\centering
	\begin{tabular}{cc}
		\includegraphics[width=0.49\textwidth]{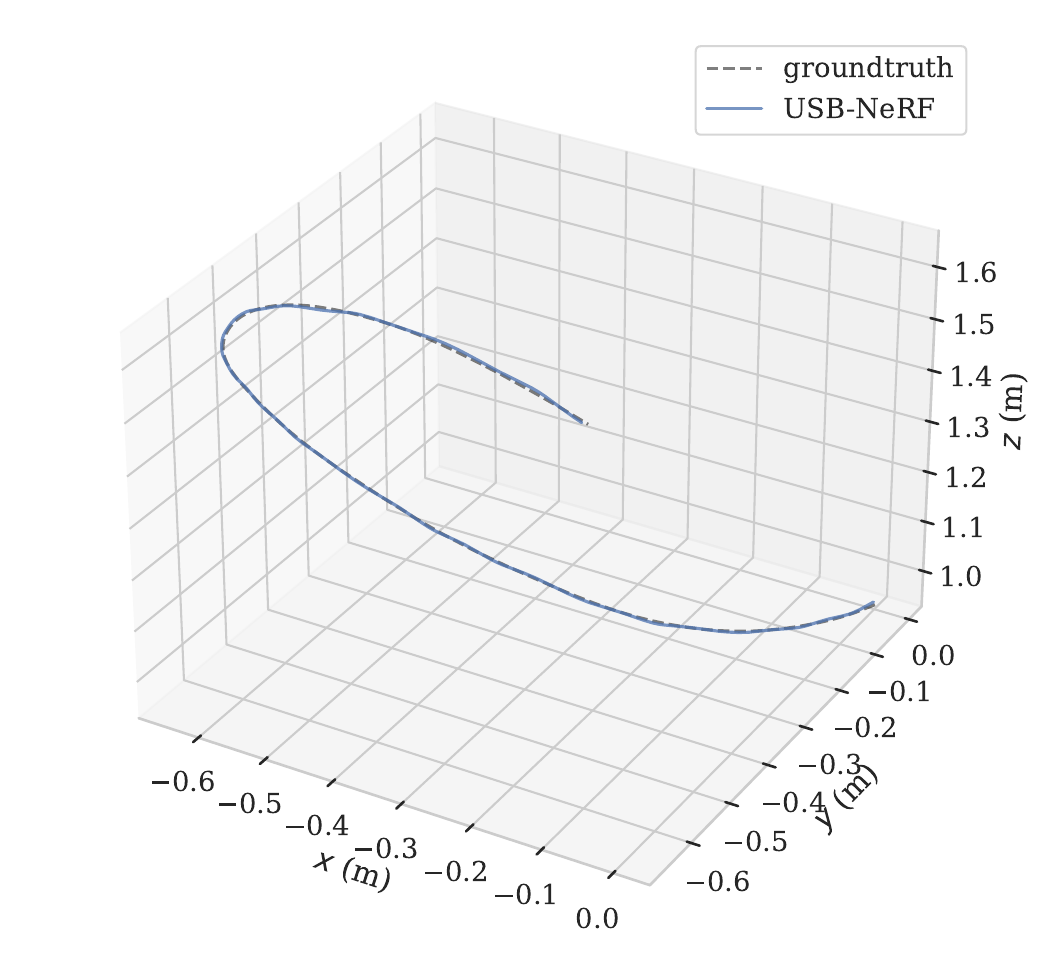} &
		\includegraphics[width=0.49\textwidth]{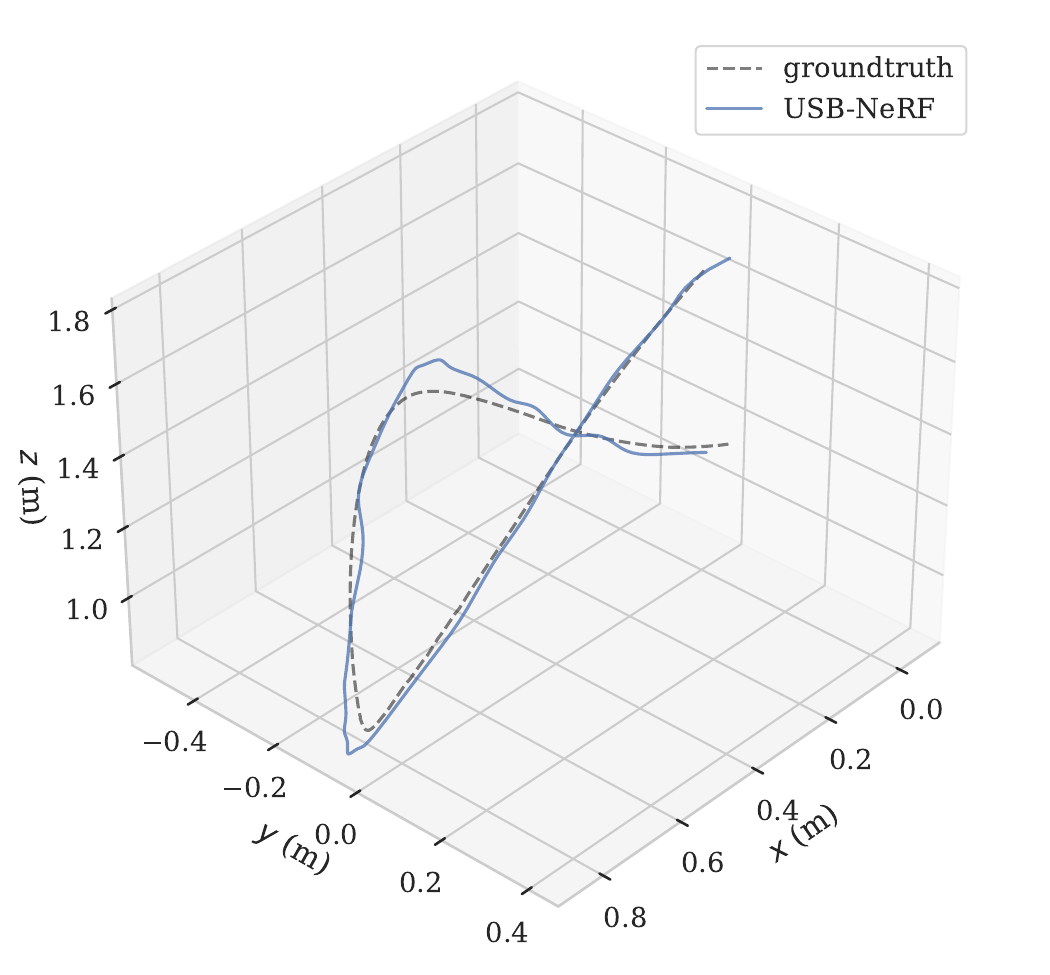} \\
		seq-1 & seq-2 \\
		\includegraphics[width=0.49\textwidth]{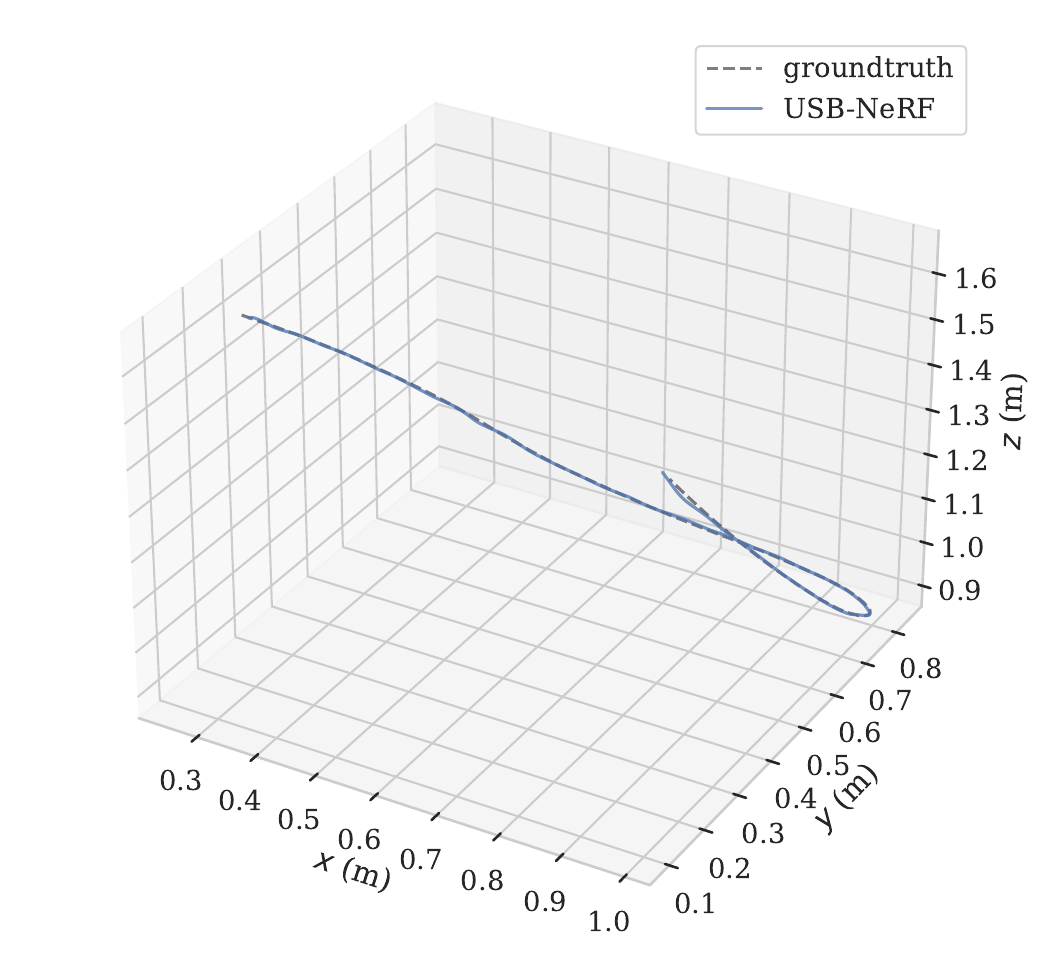} &
		\includegraphics[width=0.49\textwidth]{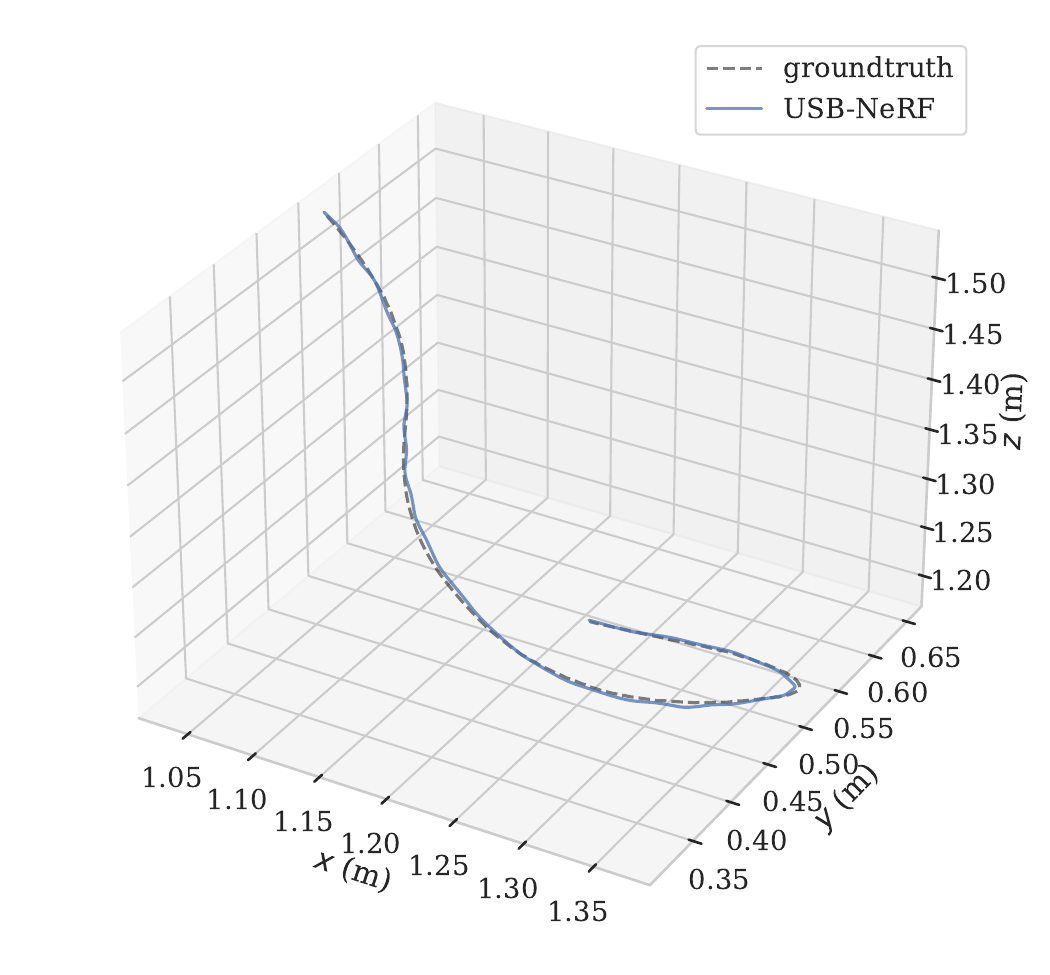} \\
		seq-3 & seq-4 \\
		\includegraphics[width=0.49\textwidth]{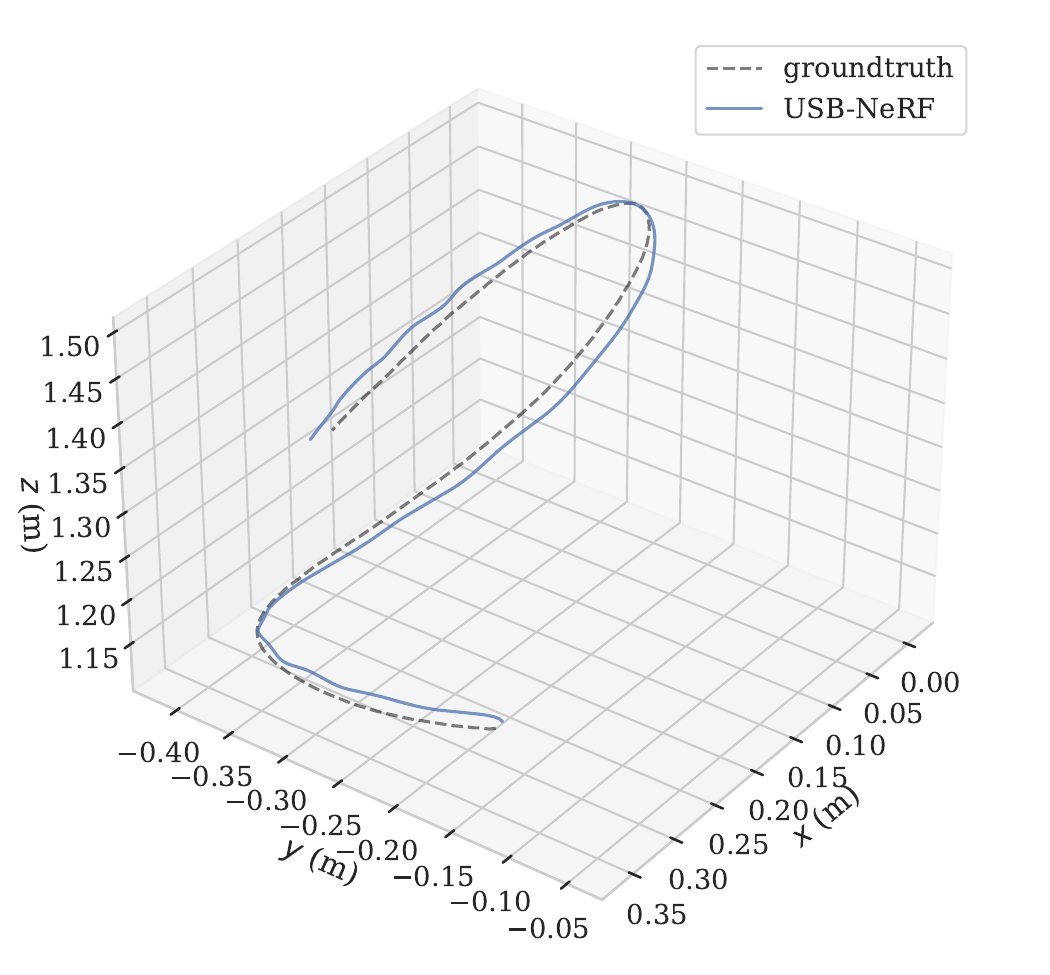} & 
		\includegraphics[width=0.49\textwidth]{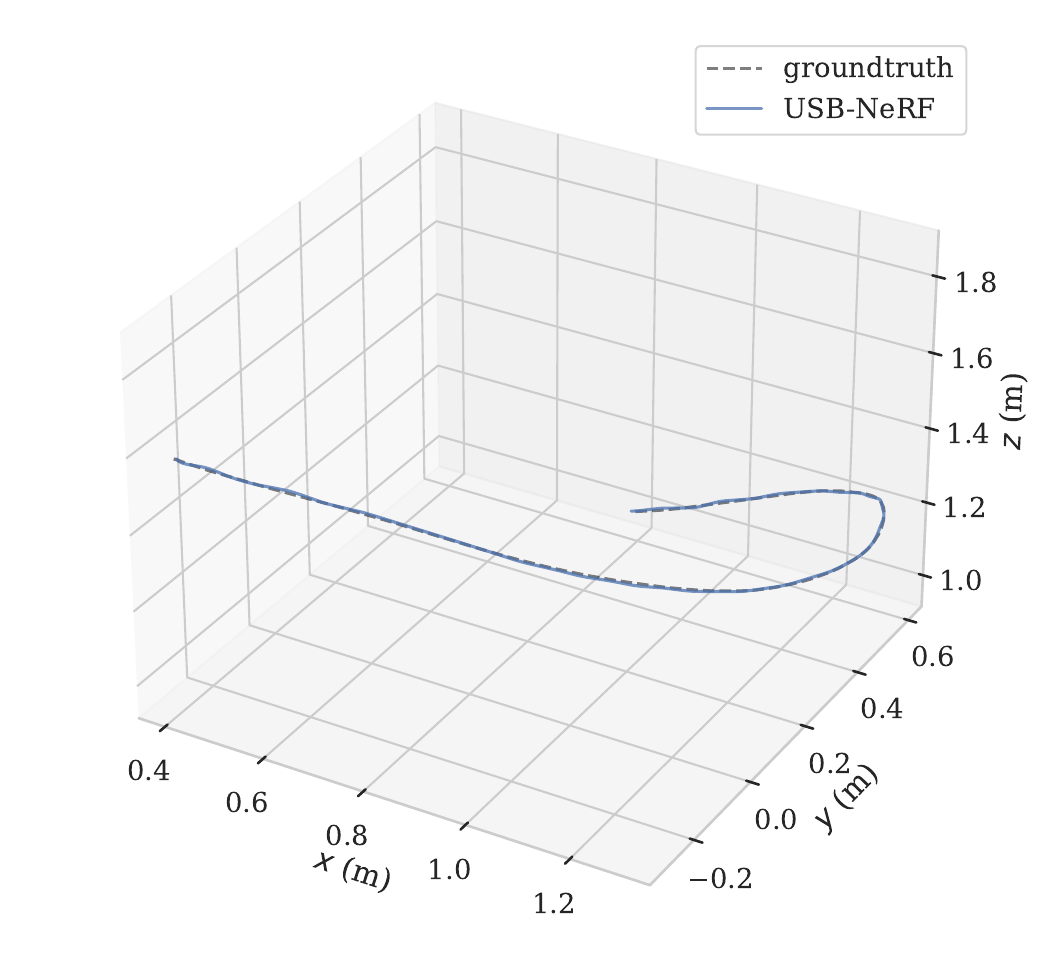} \\
		seq-5 & seq-6 \\
	\end{tabular}
	\captionsetup {font={small,stretch=0.5}}
	\caption{{\bf{Comparisons of estimated trajectories for sequence 1-6 of real TUM-RS datasets \citep{schubert2019RS-VIO}.}} The experimental results demonstrate that our method is able to estimate the motion trajectories with a sequence of rolling shutter images.}
	\label{fig_TUM_traj_1}
	\vspace{-0.6em}
\end{figure}

\begin{figure}[!t]
	\setlength\tabcolsep{1.pt}
	\centering
	\begin{tabular}{cc}
		\includegraphics[width=0.49\textwidth]{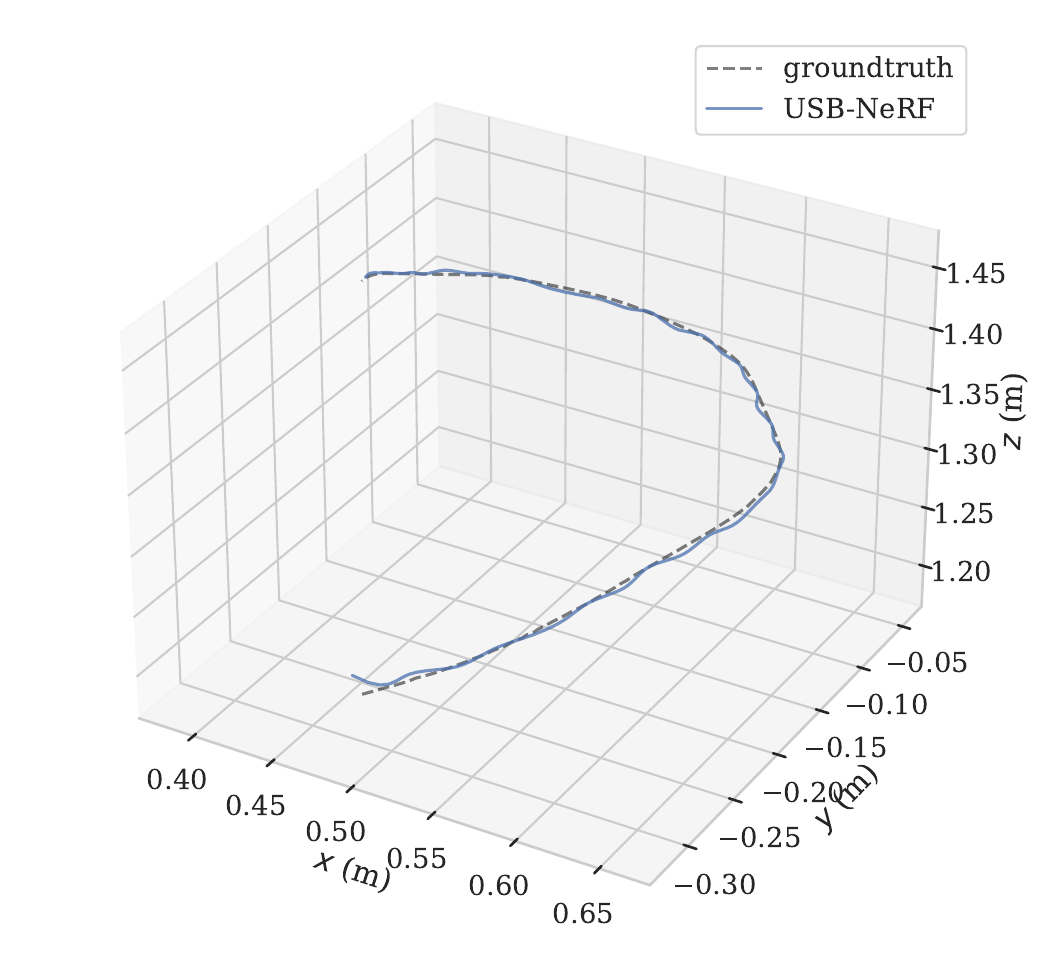} &
		\includegraphics[width=0.49\textwidth]{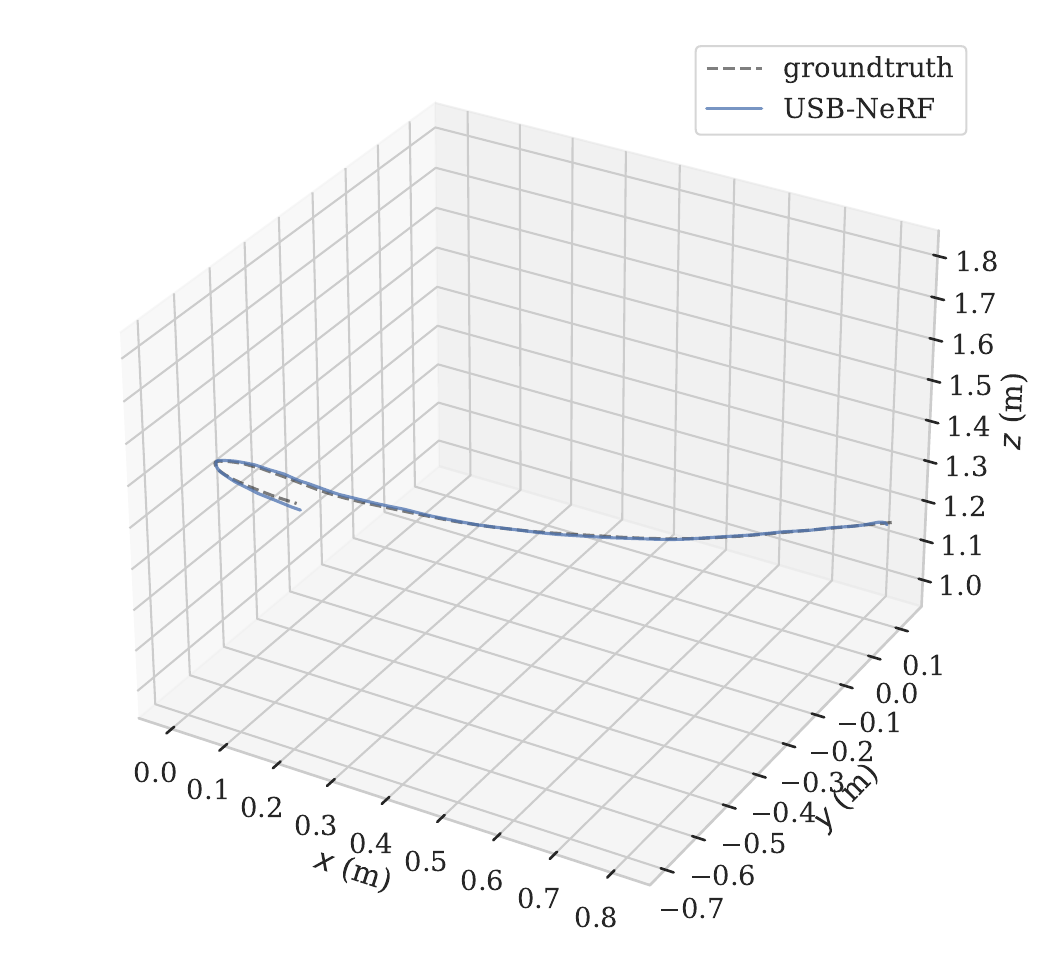} \\
		seq-7 & seq-8 \\
		\includegraphics[width=0.49\textwidth]{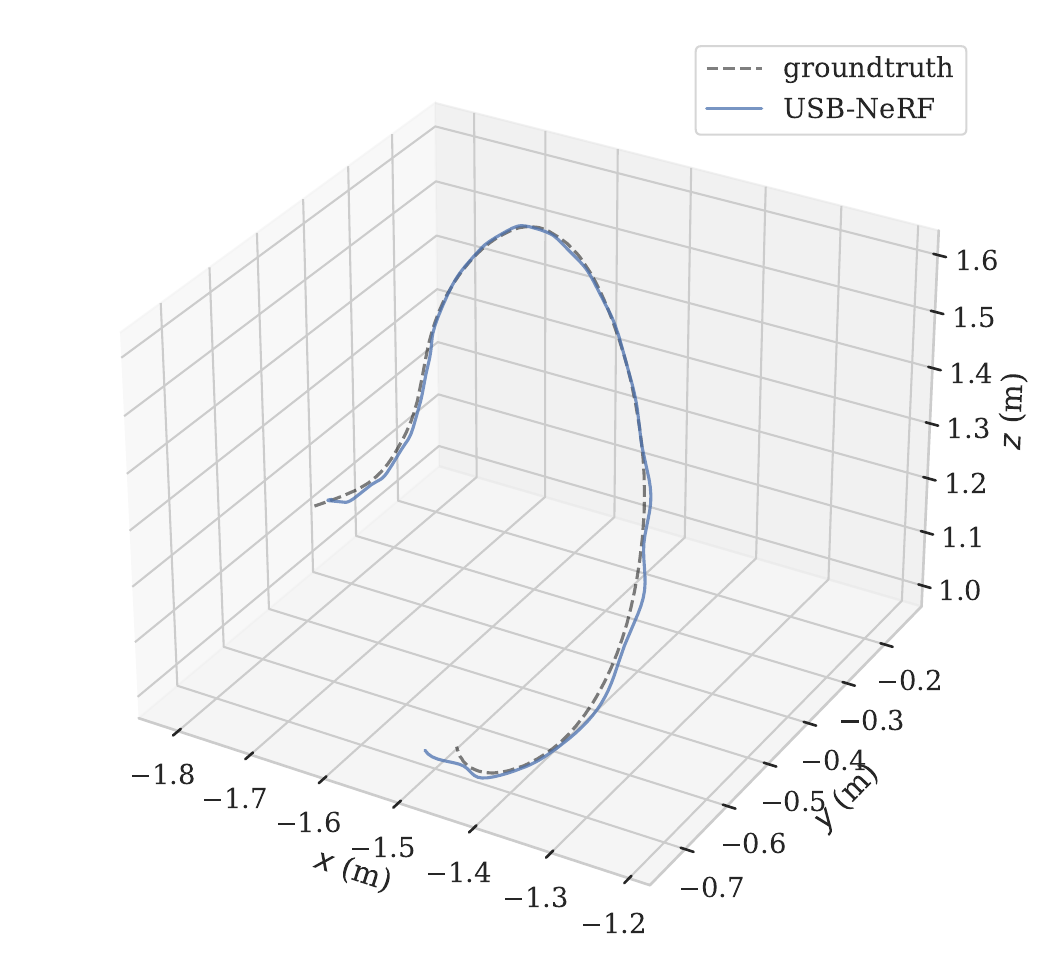} &
		\includegraphics[width=0.49\textwidth]{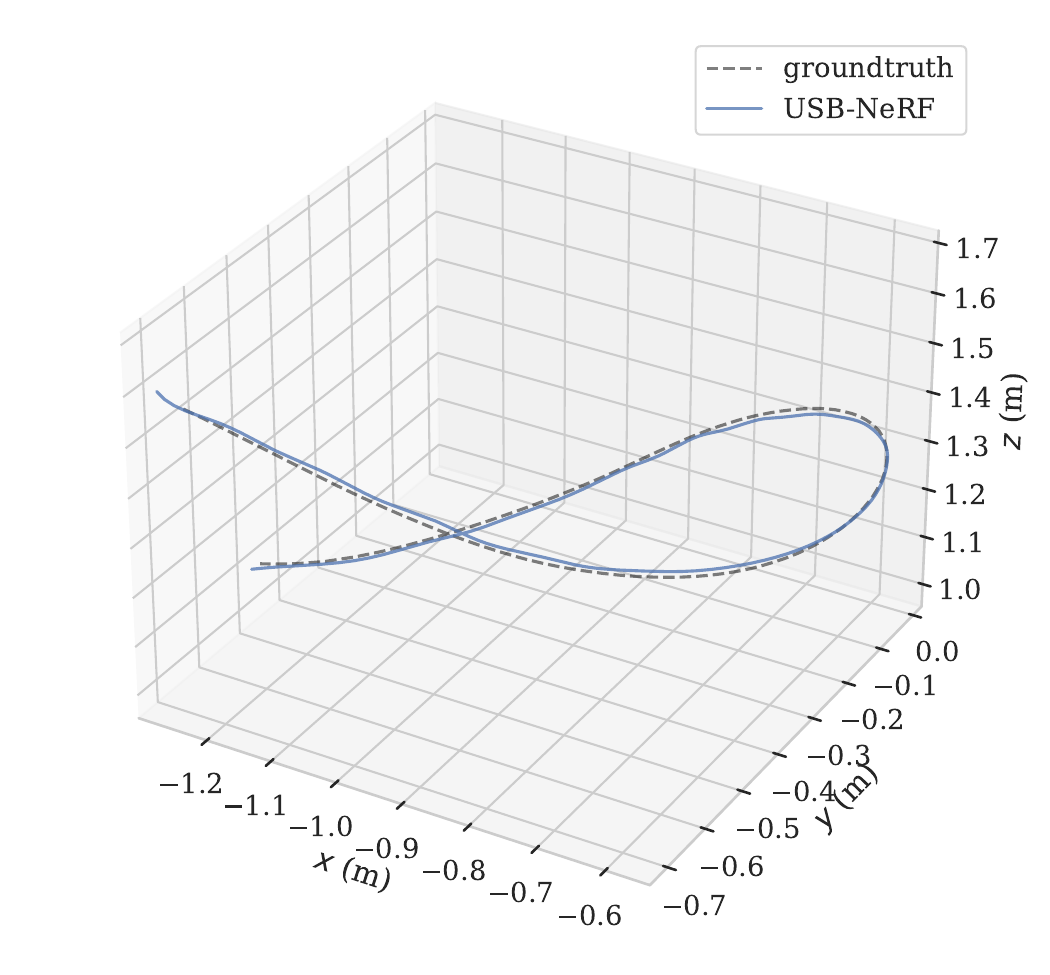} \\
		\specialrule{0em}{0.05pt}{0.05pt}
		seq-9 & seq-10
	\end{tabular}
	\vspace{-0.0em}
	\captionsetup {font={small,stretch=0.5}}
	\caption{{\bf{Comparisons of estimated trajectories for sequence 7-10 of real TUM-RS datasets \citep{schubert2019RS-VIO}.}} The experimental results demonstrate that our method is able to estimate the motion trajectories with a sequence of rolling shutter images.}
	\label{fig_TUM_traj_2}
	\vspace{-0.8em}
\end{figure}

%\begin{figure}[!t]
%	\setlength\tabcolsep{1.pt}
%	\centering
%	\begin{tabular}{c}
%		\includegraphics[width=0.95\textwidth]{supplementary_figure/full-seq/seq1.pdf} \\
%		\includegraphics[width=0.95\textwidth]{supplementary_figure/full-seq/seq2.pdf} \\
%		\includegraphics[width=0.95\textwidth]{supplementary_figure/full-seq/seq6.pdf} \\
%		\specialrule{0em}{0.05pt}{0.05pt}
%	\end{tabular}
%	\vspace{-0.0em}
%	\caption{{\bf{Comparisons of estimated trajectories for complete sequences of TUM-RS datasets \citep{schubert2019RS-VIO}.}} The experiment results show that BARF \citep{lin2021barf} totally failed in all trajectories.}
%	\label{fig_TUM_traj_long}
%	\vspace{-0.8em}
%\end{figure}

We also present additional qualitative results in terms of motion trajectory estimations. The experiments are conducted via the real TUM-RS datasets \citep{schubert2019RS-VIO}. The experimental results shown in \figrefer{fig_TUM_traj_1} and \figrefer{fig_TUM_traj_2} demonstrate that USB-NeRF is able to recover the motion trajectories on TUM-RS \citep{schubert2019RS-VIO} datasets.

\subsection{High-frame global shutter videos}
\label{sec:videos}

%\graphicspath{{./supplementary_figure/video/USB-NeRF}}
%\animategraphics[autoplay, loop, control, width=200pt]{10}{fig-}{0}{9}

To further demonstrate the advantage of our method, we also present a supplementary video which demonstrates the ability of our method to recover high quality high frame-rate global shutter images from a single rolling shutter image, which encodes rich temporal information. The video is attached as a separate file. The results also demonstrate the superior performance of our method against prior state-of-the-art methods.

%We attach several supplementary videos to present performance of high frame-rate global shutter video generation. Rolling shutter images include abundant temporal information, and this property helps predict a sequence of global shutter images. Our method employs cubic B-Spline interpolation, thus we can generate corresponding global shutter videos with arbitrary frame rates at any time. Our method use NeRF \cite{mildenhall2020nerf} to represent scenes, which produce better scene geometry and protect multi-view consistence, compared with CVR \cite{fan2022CVR}.

\end{document}